\documentclass{article}

\usepackage{microtype}
\usepackage{graphicx}
\usepackage{subfigure}
\usepackage{booktabs} 
\usepackage{caption}
\usepackage{bm}
\usepackage{amsmath,amssymb}
\usepackage{multirow}
\usepackage{tikz}
\usetikzlibrary {decorations.fractals,spy}

\usepackage{amsthm,amsmath,amsfonts,bm,xspace}
\usepackage{bbm}
\usepackage{color}
\usepackage{enumitem}

\def\1{\bm{1}}

\def\eps{{\epsilon}}

\def\vx{{\bm{x}}}
\def\veps{{\bm{\epsilon}}}

\def\vy{{\bm{y}}}
\def\vyt{{\bm{y_t}}}

\DeclareMathAlphabet{\mathsfit}{\encodingdefault}{\sfdefault}{m}{sl}
\SetMathAlphabet{\mathsfit}{bold}{\encodingdefault}{\sfdefault}{bx}{n}

\newcommand{\E}{\mathbb{E}}

\def\unet{Efficient U-Net}
\def\ours{DepthGen}
\def\stepunrolledheader{Step-unrolled Denoising Diffusion}
\def\stepunrolled{step-unrolled denoising diffusion}
\def\stepunrolledshort{SUD}
\def\first{\textbf}
\def\second{\underline}
\def\third{\textit}
\usepackage{hyperref}

\newcommand{\etal}{{\em{et al.}}}

\usepackage[accepted]{icml2023}

\usepackage{amsmath}
\usepackage{amssymb}
\usepackage{mathtools}
\usepackage{amsthm}

\usepackage[capitalize,noabbrev]{cleveref}

\theoremstyle{plain}

\theoremstyle{definition}

\theoremstyle{remark}

\usepackage[textsize=tiny]{todonotes}

\icmltitlerunning{Monocular Depth Estimation using Diffusion Models}

\begin{document}

\twocolumn[
\icmltitle{Monocular Depth Estimation using Diffusion Models}




\icmlsetsymbol{equal}{*}

\begin{icmlauthorlist}
\icmlauthor{Saurabh Saxena}{gr}
\icmlauthor{Abhishek Kar}{gr}
\icmlauthor{Mohammad Norouzi}{gr}
\icmlauthor{David J.\ Fleet}{gr}
\end{icmlauthorlist}

\icmlaffiliation{gr}{Google Research}


\icmlcorrespondingauthor{Saurabh Saxena}{srbs@google.com}

\icmlkeywords{Machine Learning, ICML}

\vskip 0.3in
]




\printAffiliations{} 


\begin{abstract}
\vspace*{-0.1cm}
We formulate monocular depth estimation using denoising diffusion models, inspired by their recent successes in high fidelity image generation.
To that end, we introduce innovations to address problems arising due to noisy, incomplete depth maps in training data,  including {\em \stepunrolled}, an $L_1$ loss, and depth infilling during training.
To cope with the limited availability of data for supervised training, we leverage pre-training on self-supervised image-to-image translation tasks.
Despite the simplicity of the approach, with a generic loss and architecture, our {\em \ours} model achieves SOTA performance on the indoor NYU dataset, and near SOTA results on the outdoor KITTI dataset. 
Further, with a multimodal posterior, \ours~ naturally represents depth ambiguity (e.g., from transparent  surfaces), and its zero-shot performance combined with depth imputation, enable a simple but effective text-to-3D pipeline. Project page: \href{https://depth-gen.github.io}{https://depth-gen.github.io} 
\vspace*{-0.1cm}

\end{abstract}

\section{Introduction}
\label{introduction}

Diffusion probabilistic models have emerged as a powerful family of generative models for high fidelity image synthesis, capturing remarkably rich knowledge about the visual world \cite{sohl2015deep,ho2020denoising,ramesh-dalle2,sahariac-imagen}.
Given their impressive generative capabilities, it is natural to ask to what extent are these models effective for image to image vision tasks like segmentation, optical flow or depth estimation?
Here, we adapt diffusion models to the problem of monocular depth estimation and investigate their effectiveness in the context of a large body of prior work.
We demonstrate state of the art performance on benchmarks while also enabling multi-modal inference to resolve depth ambiguities, and exploiting depth imputation for text to 3D generation.

Two key issues in training diffusion models for monocular depth inference concern the amount and quality of available training data.
First, much of the existing data is noisy and incomplete 
(e.g., see Figs.\ \ref{fig:multimodal_kitti_samples} and \ref{fig:nyu-train-missing-depth}).
This presents a challenge for the conventional training framework and   
iterative sampling in diffusion models, leading to a problematic distribution shift between training and testing.
To mitigate these issues we propose the use of an $L_1$ loss for robustness, infilling missing depth values during training, and the introduction of {\em \stepunrolled}.

Given the limited availability of labelled training data, we also consider the use of self-supervised pre-training.
This leverages the strong performance of diffusion models on tasks like colorization and inpainting \citep[e.g.,][]{sahariac-palette}, capturing rich image structure that may transfer to other tasks.
Accordingly, we propose a training pipeline comprising multi-task self-supervised pre-training followed by supervised fine-tuning. The model can then be used zero-shot or it can be further fine-tuned for a 
specific domain.

The resulting model, {\em \ours}, outperforms SOTA baselines on the indoor NYU dataset, and is competitive on KITTI.
Ablations show that unsupervised pre-training, depth infilling, the $L_1$ loss, and \stepunrolled~all significantly improve performance.
As a probabilistic model, \ours~has  other attractive properties:
With its ability to represent multi-modal distributions, we find that it can resolve depth ambiguities, e.g., due to reflective or transparent surfaces.
Given the ease of imputation with diffusion models, \ours~can also be used to infer missing depth values. 
We exploit this property, along with its zero-shot capability and existing text to image models to build a simple but effective framework for text to 3D scene generation and novel view synthesis.

In summary, our contributions are as follows:
\vspace*{-0.25cm}
\begin{enumerate}[nolistsep,leftmargin=0.5cm]
\itemsep 0.05cm
\item We introduce \ours, a diffusion model for monocular depth estimation,
comprising self-supervised pre-training and supervised fine-tuning.
Without specialized loss functions or architectures, it achieves SOTA relative error of 0.074 on the NYU benchmark.
\item To train diffusion models on noisy, incomplete depth data, we advocate the use of an $L_1$ loss, depth infilling, and \stepunrolled~(\stepunrolledshort) to reduce latent distribution shift between training and inference.
\item We show that \ours\  enables multimodal depth inference, and imputation of missing depths e.g, for 
text-to-3D generation and novel view synthesis.
\end{enumerate}


\begin{figure*}[t]
    \centering
    \begin{subfigure}
      \centering
      \includegraphics[width=\linewidth]{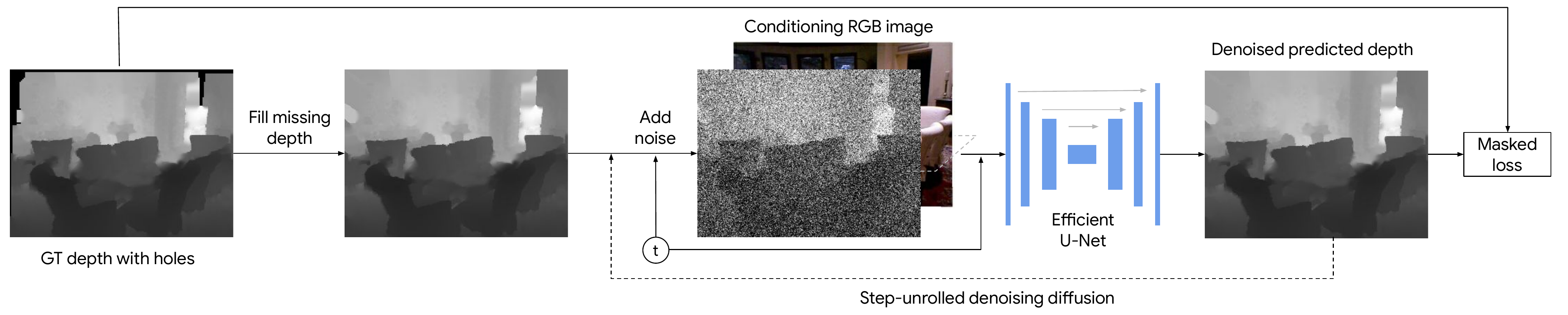}
    \end{subfigure}
    \vspace*{-0.75cm}
    \caption{
    \label{fig:arch} Training Architecture. Given a groundtruth depth map, we first infill missing depth using nearest neighbor interpolation. Then, following standard diffusion training, we add noise to the depth map and train a neural network to predict the noise given the RGB image and noisy depth map. During finetuning, we unroll one step of the forward pass and replace the groundtruth depth map with the prediction.}
    \vspace*{-0.25cm}
\end{figure*}

\section{Related Work}
\label{related}

\textbf{Monocular depth estimation} is essential for many vision applications \cite{jing2022depth,zhou2019does}.
And recent progress has been impressive, with the development of specialized loss functions and architectures
\citep[e.g.,][]{NIPS2005_17d8da81,4531745,eigen2014depth,eigen2014predicting,laina2016deeper,cao2016estimating,dorn,adabins,binsformer,pixelformer}.
We build on this rich literature, but with a simple, generic architecture, leveraging recent advances in generative models.

Prior work has shown that self-supervised tasks like colorization \cite{ZhangColorization2016,Shakhnarovich2016} serve as effective pre-training for downstream vision tasks. This motivates the choice to initialize our model with Palette \cite{sahariac-palette} style multi-task pre-training on the self-supervised image-to-image translation tasks.
Self-supervised training using masked prediction has also recently been found to be particularly effective \cite{xie2022revealing},
with subsequent work, concurrent to ours,  establishing the current SOTA \cite{AllInTokens}.
Our findings also support self-supervised pre-training, albeit with diffusion-based image-to-image translation, and we establish a new SOTA while also representing multi-modality and supporting zero-shot depth completion. 

Large-scale in-domain pre-training has also been effective for depth estimation \cite{ranftl2019robust,dpt,ren2017crossdomain}, which we find to be the case here as well.


\textbf{Diffusion models} have excelled at image generation, including unconditional and class-conditional generation \cite{dhariwal2021diffusion,ho2021cascaded}, image-to-image translation \cite{saharia2021image,sahariac-palette}, text-to-image synthesis \cite{rombach2022high,ramesh-dalle2,nichol-glide,sahariac-imagen}, and text-guided image editing
\cite{instructpix2pix,ImagenEditor,hertz2022prompttoprompt,meng2021sdedit}.
Despite this success, they have not been widely applied to vision tasks, except for recent work on image enhancement \cite{sahariac-palette}, used here for pre-training, and work on panoptic segmentation \cite{chen2022generalist}.
To the best of our knowledge, ours is the first to apply diffusion models to monocular depth estimation.

Also related to our work are diffusion models for view synthesis
from multi-view image data \cite{3dim}, 
generative models for point cloud data \cite{nichol2022pointe}, text-to-3D generative models
\cite{dreamfusion} and models for depth-aware novel view synthesis \cite{PixelSynth,infinite_nature_2020}.
While work on 3D generative models is exciting, our primary interest here is monocular depth estimation.






\section{\ours}
\label{method}

\subsection{Background}
Diffusion models are latent-variable generative models trained to transform a sample of a Gaussian noise  into a sample from a data distribution \cite{sohl2015deep,ho2020denoising}.
They comprise a {\it forward process} that gradually annihilates data by adding noise, as 'time' $t$ increases from 0 to 1, and a learned {\it generative process} that reverses the forward process, starting from a sample of random noise at $t=1$ and incrementally adding structure (attenuating noise) as $t$ decreases to 0. 
A conditional diffusion model conditions the steps of the reverse process. In the case of depth estimation, our conditioning signal is an RGB image, $\vx$, and the target is a conditional distribution over depth maps, $p(\vy \,|\, \vx)$.

Central to the model is a denoising network $f_{\theta}$ that is trained to take a noisy sample
at some time-step $t$, and predict a less noisy sample.
Using Gaussian noise in the forward process, one can express the training objective over the sequence of transitions (as $t$ slowly decreases) as a sum of non-linear regression losses, i.e.,
\begin{equation}
    \E_{(\vx,\, \vy)}\, \E_{(t,\, \veps)} \bigg\lVert f_\theta(\vx,\, \underbrace{\sqrt{\gamma_t} \,\vy + \sqrt{1\!-\!\gamma_t}\, \veps}_{\vyt}, \,t) - \veps\, \bigg\rVert^{2}_2
\label{eq:diffusion-loss}
\end{equation}
where $\veps \sim \mathcal{N}(0, I)$, $t \sim \mathcal{U}(0, 1)$, and where $\gamma_t > 0$ is computed with a pre-determined noise schedule.
For inference (i.e., sampling), one draws a random noise sample $\vy_1$, and then iteratively uses $f_{\theta}$
to estimate the noise, from which one can compute the next latent sample $\vy_s$, for $s<t$.

\subsection{Self-Supervised Pre-Training}

\ours~training comprises self-supervised pre-training,
then supervised training on RGB-D data.
The pre-trained model is a self-supervised multi-task diffusion model. 
Following \cite{sahariac-palette}, we train a Palette model from scratch on four image-to-image  translation tasks, i.e., colorization, inpainting, uncropping and JPEG artifact removal.

\subsection{Supervised training with noisy, incomplete depth}
\label{sec:handling-missing-depth}

Following pre-training, and with minor modifications to the architecture (see Sec.\ \ref{architecture}), training continues on paired RGB and depth data.
While straightforward conceptually, the training datasets available for depth estimation present  substantial challenges. The depth maps in particular are noisy and often contain regions with missing depth values.
The various causes for such \textit{holes} are due to highly reflective surfaces, light absorbing surfaces \cite{6664993} or regions outside the sensor's range of measurement. 
Holes are largely inconsequential for simple feed-forward nets or regression models, since one could only backpropagate the loss from the subset of pixels with known depth values, ignoring those with missing depth. For diffusion models, however, such corruption of the training data is problematic.

Diffusion models perform inference through iterative refinement -- in our case, of a depth map $\vy$ conditioned on an RGB image $\vx$. It starts with a sample of Gaussian noise $\vy_1$, and terminates with a sample from the predictive distribution $p(\vy_0 \,|\,\vx)$.
A refinement step from time $t$ to $s$, with $s\! <\!t$, proceeds by sampling from the parameterized distribution $p_\theta(\vy_s \,|\, \vy_t, \vx)$.
Simply put, during inference, each step operates on the output from the previous step.
In contrast, at training the different steps are somewhat decoupled (see Eqn.\ \ref{eq:diffusion-loss}), where the denoising network operates on a noisy version of the ground truth depth map instead of the output of the previous iteration (reminiscent of teaching forcing in training RNNs). 
This introduces a distribution shift between training and inference, since 
the marginal distribution over noisy training depth maps \textit{with holes} may differ significantly from the distribution of noisy depths at inference time, which should ideally (since we do not learn the distribution of holes in the loss) be a noisy version of the \textit{true}, complete depth maps (from a perfect, noiseless sensor for instance).
This has a significant negative impact on model performance.
This problem is further exacerbated by structured or heavy-tailed noise in training depth maps.

We find that these problems are effectively mitigated with the following modifications during training:

\textbf{Depth interpolation.}
To reduce distribution shift between training and inference we impute missing depth values.
We explored several ways to accomplish this, including various interpolation schemes, and using \ours\ (trained with nearest neighbor interpolation infilling) to infill missing depth. But, empirically, we found that two straightforward steps performed as well as more sophisticated approaches.
In particular, we find that nearest neighbor interpolation is sufficient to impute missing depths in indoor training data.
For outdoor data we continue to use nearest neighbor interpolation, 
except for sky regions, as they are often large and are much further from the camera 
than adjacent objects in the image.
We use an off-the-shelf sky segmenter \cite{Liba_2020_CVPR_Workshops}, and then set all
sky pixels to be the maximum modeled depth (here, 80m). 
Despite the imputation of missing depths, we note that the training loss is 
only computed at pixels with known (vs infilled) depth.



\begin{algorithm}[tb]
   \caption{Train step w/ infilling and \stepunrolledshort.}
   \label{alg:train_step}
\begin{algorithmic}
   \STATE {\bfseries Input:} rgb image $x$, depth map $y$ \\
   \STATE $t \xleftarrow{} U(0,1)$ \\
   \STATE $\eps \xleftarrow{} N(0,1)$ \\
   \STATE $valid\_mask = y > 0$
   \STATE $y = fill\_holes(y)$
   \STATE $y_t = \sqrt{\gamma_t} * y + \sqrt{1-\gamma_t} * \eps$ \\
   \IF{$unroll\_step$}
   \STATE $\eps_{pred} = stop\_grad(f_{\theta}(x, y_t, t))$
   \STATE $y_{pred} = (y_t - \sqrt{1-\gamma_t} * \eps_{pred}) / \sqrt{\gamma_t} $ \\
   \STATE $y_t = \sqrt{\gamma_t} * y_{pred} + \sqrt{1-\gamma_t} * \eps$
   \STATE $\eps = (y_t - \sqrt{\gamma_t} * y) / \sqrt{1-\gamma_t}$
   \ENDIF
   \STATE $\eps_{pred} = f_{\theta}(x, y_t, t)$
   \STATE $loss = reduce\_mean(\lvert \eps - \eps_{pred} \lvert[valid\_mask])$
\end{algorithmic}
\end{algorithm}

\textbf{\stepunrolledheader.}
\label{step_unrolled}
Another approach to tackling the distribution shift in the latent marginal distribution of $y_t$ between training and inference is to construct $y_t$ using the model's output instead of the ground truth depth. One can do this by slightly modifying the training procedure (Algorithm \ref{alg:train_step}) to run one forward pass of the model and build $y_t$ by adding noise to the model's output rather than the training depth map. We do not propagate gradients for this forward pass. We find that this slows down training by about 15\% on a TPU v4.
We refer to this as {\em \stepunrolled~(\stepunrolledshort)}.

We perform \stepunrolledshort~during fine-tuning only, not during supervised depth pre-training. Early in training the depth predictions are likely inaccurate. So the latent marginals over the noisy training depth maps would be much closer to the  desired \textit{true} marginals than those produced by adding noise to the model's outputs. 
Hence, doing \stepunrolledshort~early in supervised pre-training is not recommended.
One might consider the use of a curriculum for gradually introducing \stepunrolledshort~in the later stages of supervised pre-training, but this also introduces additional hyper-parameters, so we simply invoke \stepunrolledshort~during fine-tuning, and leave an exploration of curricula to future work.

This problem of training / inference distribution shift resembles that of \textit{exposure bias} \cite{DBLP:journals/corr/RanzatoCAZ15} in autoregressive models, for which the mismatch is caused by \textit{teacher forcing} during training \cite{6795228}. Several solutions have been proposed for this problem in the literature  \cite{lamb2016professor,yu2016seqgan,10.5555/2969239.2969370}. \stepunrolledshort~also closely resembles the approach in \cite{savinov2022stepunrolled} where they perform step-unrolling for training denoising autoencoders on text. 


Finally, we note that \cite{AllInTokens} faced a similar problem when training a vector-quantizer on depth data. They work around it by synthetically adding more holes following a carefully chosen masking ratio. In comparison, we prefer our approach since nearest neighbor infilling is hyper-parameter free and \stepunrolled~could be more generally applicable to other tasks with sparse data.

\textbf{$L_1$ Loss.}
While the $L_2$ loss in Eqn.\ \ref{eq:diffusion-loss} is appropriate for noise-free training data with additive Gaussian noise, good performance has been reported with an $L_1$ loss during training for image-to-image translation models \cite{saharia2021image}. Given the possibility of substantial noise in depth data, especially for large depths and near holes, we 
hypothesize that the robustness afforded by the $L_1$ loss may also be useful in training RGB-to-depth diffusion models.

\section{Experiments}
\label{sec:experiments}

\begin{table*}[t]
    \centering
    \small
    \caption{Comparison of performances on the NYU-Depth-v2 dataset. $\top$ indicates methods that use unsupervised pretraining, \dag indicates supervised pretraining and \ddag~indicates methods with supervised depth pretraining on auxiliary data. \textbf{Best} / \underline{second best} / \textit{third best} results are bolded / underlined / italicized respectively. $\downarrow$ denotes lower is better and $\uparrow$ denotes higher is better. Baselines: [1]~\citet{dorn}, [2]~\citet{yin2019enforcing}, [3] ~\citet{bts}, [4]~\citet{dav}, [5]~\citet{transdepth}, [6]~\citet{dpt}, [7]~\citet{adabins}, [8]~\citet{binsformer}, [9]~\citet{pixelformer}, [10]~\citet{xie2022revealing}, 
    [11]~\citet{AllInTokens}. $^{*}$ denotes concurrent work.}
    \vspace*{0.2cm}
    \begin{tabular}{ccccccccc}
        \toprule
        Method & Architecture & $\delta_1\uparrow$ & $\delta_2\uparrow$ & $\delta_3\uparrow$ & REL$\downarrow$ & RMS$\downarrow$ & $log_{10}\downarrow$ \\
        \midrule
        DORN [1] & ResNet-101$^\dag$ &0.828 & 0.965 & 0.992 & 0.115 & 0.509 & 0.051\\
        VNL [2]  & ResNeXt-101$^\dag$ & 0.875 & 0.976 & 0.994 & 0.108 & 0.416 & 0.048 \\
        BTS [3]  & DenseNet-161$^\dag$ &0.885 & 0.978 & 0.994 & 0.110 & 0.392 & 0.047\\
        DAV [4] & DRN-D-22$^\dag$ &0.882 & 0.980 & 0.996 & 0.108 & 0.412 & --\\
        TransDepth [5]  & Res-50+ViT-B$^\dag$ & 0.900 & 0.983 & 0.996 & 0.106 & 0.365 & 0.045\\
        DPT [6]  & Res-50+ViT-B$^{\dag\ddag}$ & 0.904 & 0.988 & \underline{0.998} & 0.110 & 0.357 & 0.045 \\
        AdaBins [7]  & E-B5+Mini-ViT$^{\dag}$ & 0.903 & 0.984 & \third{0.997} & 0.103 & 0.364 & 0.044 \\
        BinsFormer [8]  & Swin-Large$^\dag$ & 0.925 & 0.989 & \third{0.997} & 0.094 & 0.330 & 0.040 \\
        PixelFormer [9]  & Swin-Large$^\dag$ & 0.929 & \third{0.991} & \underline{0.998} & 0.090 & 0.322 & 0.039 \\
        MIM [10] & SwinV2-L$^\top$ & \underline{0.949} & \first{0.994} & \first{0.999} & 0.083 & \second{0.287} & \third{0.035} \\
        AiT-P [11]$^{*}$ & SwinV2-L$^\top$ &  \first{0.953} &  \second{0.993} & \first{0.999} & \third{0.076} & \first{0.279} & \second{0.033} \\
         \hline
        \ours\ (ours) &  &  &  &  &  & & \\
        samples=1 & \unet$^{\top\ddag}$ & 0.944 & 0.986 & 0.995 & \second{0.075} & 0.324 & \first{0.032} \\
        samples=2 & \unet$^{\top\ddag}$ & 0.944 & 0.987 & 0.996 & \first{0.074} & 0.319 & \first{0.032} \\
        samples=4 & \unet$^{\top\ddag}$ & \third{0.946} & 0.987 & 0.996 & \first{0.074} & 0.315 & \first{0.032} \\
        samples=8 & \unet$^{\top\ddag}$ & \third{0.946} & 0.987 & 0.996 & \first{0.074} & \third{0.314} & \first{0.032} \\
        \bottomrule
    \end{tabular} 
    \label{tab::results-nyu}
\end{table*}

\begin{table*}
    \centering
    \small
    \caption{Comparison of performances on the KITTI dataset. $\top$ indicates methods that use unsupervised pretraining, \dag~indicates supervised pretraining, and \ddag~indicates methods with supervised depth pretraining on auxiliary data. \textbf{Best} / \underline{second best} / \textit{third best} results are bolded / underlined / italicized respectively. $\downarrow$ denotes lower is better and $\uparrow$ denotes higher is better. E-B5: EfficientNet-B5~\cite{tan2019efficientnet}.
    Baselines: [11]~\cite{godard2017unsupervised}, [12]~\cite{johnston2020self}, [13]~\cite{gan2018monocular}, [1]~\citet{dorn}, [2]~\citet{yin2019enforcing}, [14]~\cite{xu2020probabilistic}, [3] ~\citet{bts}, [5]~\citet{transdepth}, [6]~\citet{dpt}, [7]~\citet{adabins}, [8]~\citet{binsformer}, [9]~\citet{pixelformer}, [10]~\citet{xie2022revealing}.}
    \label{tab::results-kitti}
    \vspace*{0.2cm}
    \begin{tabular}{ccccccccc}
       \toprule
        Method & Backbone & ~~~\textbf{$\delta_1$}$\uparrow$~~~ & ~~~\textbf{$\delta_2$}$\uparrow$~~~ & ~~~\textbf{$\delta_3$}$\uparrow$~~~ & ~REL~$\downarrow$~ & Sq-rel~$\downarrow$ & RMS~$\downarrow$ & RMS log~$\downarrow$ \\
        \midrule
      Godard~\etal [11] & ResNet-50 & 0.861 & 0.949 & 0.976 & 0.114   & 0.898  & 4.935 & 0.206 \\
      Johnston~\etal\ [12] & ResNet-101$^\dag$ & 0.889 & 0.962 & 0.982 & 0.106 & 0.861 & 4.699 & 0.185 \\
      Gan~\etal\ [13] & ResNet-101 & 0.890 & 0.964 & 0.985 & 0.098   & 0.666  & 3.933 & 0.173 \\
       DORN [1]  & ResNet-101$^\dag$ & 0.932 & 0.984 & 0.994 & 0.072   & 0.307  & 2.727 & 0.120\\
      VNL [2] & ResNext-101$^\dag$ & 0.938 & 0.990  & \third{0.998} & 0.072 & -- & 3.258 & 0.117 \\
      PGA-Net [14] & ResNet-50$^\dag$ & 0.952 & 0.992 & \third{0.998} & 0.063 & 0.267 & 2.634 & 0.101 \\
     BTS [3] & DenseNet-161$^\dag$ & 0.956 & 0.993 & \third{0.998} & 0.059   & 0.245  & 2.756 & 0.096 \\
      TransDepth [5] & ResNet-50+ViT-B$^\dag$ & 0.956 & 0.994 & \second{0.999} & 0.064 & 0.252  & 2.755 & 0.098 \\
    DPT [6]~ &ResNet-50+ViT-B$^{\dag\ddag}$ & 0.959 & \third{0.995} & \second{0.999} & 0.062 & --& 2.573 & 0.092 \\
    AdaBins [7] & E-B5+mini-ViT$^{\dag}$ & 0.964 & \third{0.995} & \second{0.999} & 0.058  & 0.190  & 2.360 & 0.088\\
BinsFormer [8] & Swin-Large$^{\dag}$ & \third{0.974} &  \second{0.997} & \second{0.999} & \third{0.052} & \third{0.151} & \second{2.098} & \third{0.079} \\
    PixelFormer [9] & Swin-Large$^{\dag}$ & \second{0.976} & \second{0.997}  & \second{0.999} & \second{0.051} & \second{0.149} & \second{2.081} & \second{0.077} \\
    MIM [10] & SwinV2-L$^\top$ & \first{0.977} & \first{0.998} & \first{1.000} & \first{0.050} & \first{0.139} & \first{1.966} & \first{0.075} \\

    \midrule
    \ours\ (ours) & & & & & & & & \\
    samples=1 & \unet$^{\top\ddag}$ & 0.951&   0.991&   0.997&   0.064&   0.389&   3.104&   0.103 \\
    samples=2 & \unet$^{\top\ddag}$ & 0.951&   0.991&   \third{0.998}&   0.064&   0.378&   3.07&   0.102 \\
    samples=4 & \unet$^{\top\ddag}$ &  0.951&   0.991&   \third{0.998}&   0.064&   0.373&   3.052&   0.102 \\
    samples=8 & \unet$^{\top\ddag}$ &  0.953&   0.991&   \third{0.998}&   0.064&   0.356&   2.985&   0.100 \\
    \bottomrule
\end{tabular}
\vspace{-0.25cm}
\end{table*} 

\subsection{Datasets}
\label{sec:datasets}


For unsupervised pre-training, we use the ImageNet-1K \cite{imagenet_cvpr09} and Places365 \cite{zhou2017places} datasets and train on the self-supervised tasks of colorization, inpainting, uncropping, and JPEG decompression, following \cite{sahariac-palette}.

\textbf{Indoor model.}
For supervised image-to-depth pre-training of the indoor model we use the following two datasets (with dataset mixing at the batch level):

\textit{ScanNet} \cite{scannet} is a dataset of 2.5M images captured using a Kinect v1-like sensor. It provides depth maps at $640\times480$ and RGB images at $1296\times968$.

\textit{SceneNet RGB-D} \cite{scenenet-rgbd} is a synthetic dataset of 5M images generated by rendering ShapeNet \cite{shapenet} objects in scenes from SceneNet \cite{https://doi.org/10.48550/arxiv.1511.07041} at a resolution of $320\times240$.

For indoor fine-tuning and evaluation we use 
\textit{NYU depth v2} \cite{nyu}, a commonly used dataset for evaluating indoor depth prediction models. It provides aligned image and depth maps at $640\times480$ resolution. We use the official split consisting of ~50k images for training and 654 images for evaluation. The predicted depth maps from our model are resized to the full resolution using bilinear up-sampling before evaluation. We evaluate on a cropped region proposed by \cite{eigen2014depth} following prior work.

\textbf{Outdoor model.}
For outdoor model training we use the 
\textit{Waymo Open Dataset} \cite{waymoOpenDataset},  a large-scale driving dataset consisting of about 200k frames. Each frame provides RGB images from 5 cameras and LiDAR maps. We use the RGB images from the FRONT, FRONT\_LEFT and FRONT\_RIGHT cameras and the TOP LiDAR only to build about 600k aligned RGB depth maps.

For subsequent fine-tuning and evaluation, we use 
\textit{KITTI}~\cite{kitti}, an outdoor driving dataset which provides RGB images and LiDAR scans at resolutions close to $1226\times370$. We use the training/test split proposed by \cite{eigen2014depth}, comprising  26k training images and 652 test images. The predicted depth from \ours\ is up-sampled to the full resolution using bilinear interpolation before evaluation. We evaluate on a cropped region proposed in \cite{garg2016unsupervised} following prior work.

\textbf{Data Augmentation and Preprocessing} We use random horizontal flip data augmentation for supervised depth training which is common in prior work. Where needed, images and depth maps are resized using bilinear interpolation to the model's resolution for training. Diffusion models expect inputs and generate outputs in the range $[-1., 1.]$. For the indoor model we use a max depth of 10 meters, and for the outdoor model we normalize the depth maps to the range with a maximum depth of 80 meters.

\subsection{Architecture}
\label{architecture}

The predominant architecture for diffusion models is the U-Net developed for the DDPM model \cite{ho2020denoising}, and 
later improved in several respects  \cite{nichol2021improved,song2021scorebased,dhariwal2021diffusion}. 
For \ours, we adapt the {\em \unet}~architecture that was developed for Imagen \cite{sahariac-imagen}.
The \unet~architecture is more efficient that the U-Nets used in prior work,
because it has fewer self-attention layers, fewer parameters and less computation at higher resolutions, along with other adjustments that make it well suited to training medium resolution diffusion models.

We make several minor changes to this architecture to adapt it for image-to-depth models. We drop the text cross-attention layers but keep the self-attention layer. \unet~has six input and three output channels,  since the target is a RGB image (input consists of a 3-channel source RGB image and a 3-channel noisy target image concatenated along the channel dimension). For depth models, since we have a scalar output image, we modify the architecture to have four input channels and one output channel. Note that this means we need to reinitialize the input and output convolutional kernels before the supervised depth pretraining stage. 

\textbf{Resolution.} Our re-trained Palette model was trained for images at a resolution of $256\times256$. For training depth models we choose resolutions that are close to this while preserving the aspect ratios of the original depth training datasets. The indoor model is trained at $320\times240$. For Waymo we use $384\times256$ and for KITTI $416\times128$. The model does not contain learned positional embeddings so it can be easily pretrained and finetuned at different resolutions.

\subsection{Hyper-parameters}
\label{sec:hparams}
The self-supervised model is trained for 2.8M steps with an $L_2$ loss and a mini-batch size of 512.  Other hyper-parameters are  similar to those in the original Palette paper. 

The depth models are trained with $L_1$ loss. We use a constant learning rate of 1e-4 during supervised depth pre-training but switch to a slightly lower learning rate of 3e-5 during fine-tuning which we find achieves slightly better results. We do learning rate warm-up over 10k steps for all models. All depth models are trained with a smaller mini-batch size of 64. The indoor depth model is trained on a mix of ScanNet and SceneNet RGBD for 2M steps and then fine-tuned on NYU for 50k steps. The outdoor depth model is trained on Waymo for 0.9M steps and fine-tuned on KITTI for 50k steps. Other details, like the optimizer and the use of EMA are similar to those outlined in \cite{sahariac-palette}.

\subsection{Sampler}
\label{sec:sampler}
We use the DDPM ancestral sampler \cite{ho2020denoising}  with 128 denoising steps. Increasing the number of denoising steps further did not greatly improve performance. We have not yet explored progressive  distillation \cite{salimans2022progressive} for faster sampling.
We believe the results on distillation for generative image models should transfer well to image-to-depth models, thereby,
reducing the gap between the speed of diffusion sampling and single-step depth estimation models. We leave this exploration to future work.
 
\subsection{Evaluation metrics}
\label{sec:metrics}
We follow the standard evaluation protocol used in prior work \cite{binsformer}. For both the NYU depth v2 and KITTI datasets we report the absolute relative error (REL), root mean squared error (RMS) and accuracy metrics ($\delta_i < 1.25^i$ for $i \in 1, 2, 3$). For NYU we additionally report absolute error of log depths ($log_{10}$). For KITTI we additionally report the squared relative error (Sq-rel) and root mean squared error of log depths (RMS log).

\subsection{Results}
\label{results}
Table \ref{tab::results-nyu} shows the results on NYU depth v2. We achieve a state-of-the-art absolute relative error of 0.074. Table \ref{tab::results-kitti} shows results on KITTI, where we perform competitively with prior work. We report results with averaging depth maps from one or more samples. Note that most prior works report average over two samples obtained by left-right reflection of the input image.

\label{ablations}
\begin{table}[t]
    \centering
    \small
    \caption{Ablations for pre-training (before fine-tuning), showing absolute relative error with a single sample for NYU and KITTI.}
    \vspace*{0.1cm}
    \begin{tabular}{l|c|c}
        \toprule
         & NYU~$\downarrow$ & KITTI~$\downarrow$ \\
        \midrule
        Training from scratch & 0.155 & 0.103 \\
        Self-supervised pre-training only & 0.116 & 0.086 \\
        Supervised pre-training only & 0.081 & 0.075 \\
        Self-supervised \& supervised pre-train & 0.075 & 0.064 \\
        \bottomrule
    \end{tabular} 
    
    \label{tab::pretraining_ablation}
    \vspace*{-0.5cm}
\end{table}

\begin{table}[t]
    \centering
    \small
    \caption{Ablation for handling incomplete depth showing absolute relative error with a single sample on NYU and KITTI.}
    \vspace*{0.1cm}
    \begin{tabular}{l|c|c}
        \toprule
        & NYU~$\downarrow$ & KITTI~$\downarrow$ \\
        \midrule
        Baseline & 0.079 & 0.115 \\
        \stepunrolledshort~only & 0.076 & 0.075 \\
        Infilling only & 0.077 & 0.066 \\
        \stepunrolledshort~and infilling & 0.075 & 0.064\\
        \bottomrule
    \end{tabular} 
    

    \vspace*{-0.5cm}
    \label{tab::missing_depth_ablation}
\end{table}
\begin{table}[t]
    \centering
    \small
    \caption{Ablation for $L_1$ vs $L_2$ loss showing absolute relative error with a single sample on NYU and KITTI.}
    \vspace*{0.1cm}
    \begin{tabular}{l|c|c}
        \toprule
        & NYU~$\downarrow$ & KITTI~$\downarrow$ \\
        \midrule
        $L_2$ & 0.085 & 0.071 \\
        $L_1$ & 0.075 & 0.064 \\
        \bottomrule
    \end{tabular} 
    \vspace*{-0.5cm}
    \label{tab::loss_ablation}
\end{table}
\def\nyuwidth{0.24}
\def\boxcolor{lightgray}

\def \nyuimagewithbox#1#2#3{\begin{tikzpicture} \node[inner sep=0] at (0, 0) {\includegraphics[width=\nyuwidth\textwidth]{#1}}; \draw[\boxcolor,ultra thin,dashed] #2 rectangle #3; \end{tikzpicture}}

\begin{figure*}
\begin{tabular}{cc|cc}
    \multicolumn{2}{c}{Groundtruth} &  \multicolumn{2}{c}{Predictions} \\
    \nyuimagewithbox{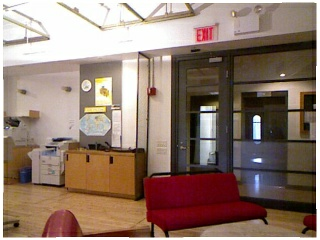}{(0.0,-0.5)}{(2,1.0)}
    &
    \nyuimagewithbox{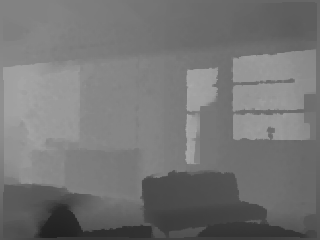}{(0.0,-0.5)}{(2,1.0)}
    &
    \nyuimagewithbox{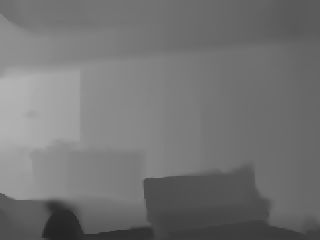}{(0.0,-0.5)}{(2,1.0)}
    &
    \nyuimagewithbox{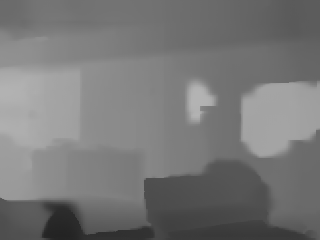}{(0.0,-0.5)}{(2,1.0)}
    
    \\
    
    \nyuimagewithbox{}{(1.4,0.0)}{(2.0,1.5)}
    &
    \nyuimagewithbox{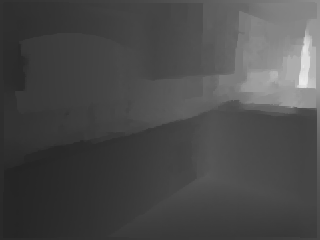}{(1.4,0.0)}{(2.0,1.5)}
    &
    \nyuimagewithbox{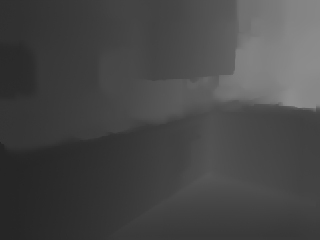}{(1.4,0.0)}{(2.0,1.5)}
    &
    \nyuimagewithbox{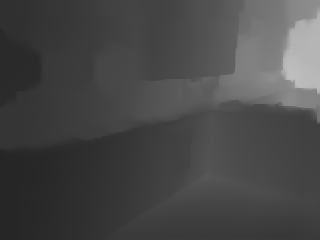}{(1.4,0.0)}{(2.0,1.5)}

    \\
    
    
    
    

    \nyuimagewithbox{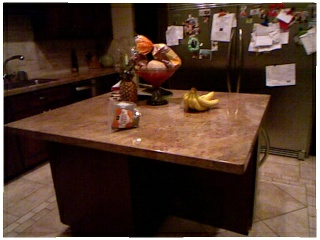}{(-0.45,1.0)}{(0.15,1.5)}
    &
    \nyuimagewithbox{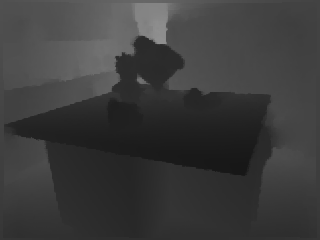}{(-0.45,1.0)}{(0.15,1.5)}
    &
    \nyuimagewithbox{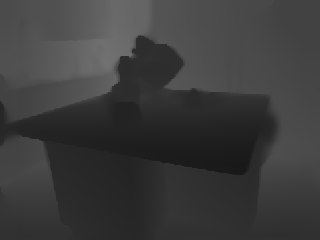}{(-0.45,1.0)}{(0.15,1.5)}
    &
    
    \nyuimagewithbox{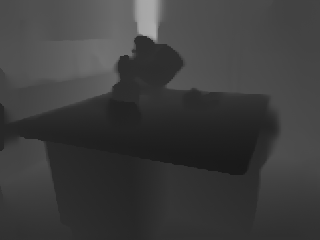}{(-0.45,1.0)}{(0.15,1.5)}
    
    \\

    \nyuimagewithbox{}{(-2.0,0.0)}{(-1.5,1.5)}
    
    &
    
    \nyuimagewithbox{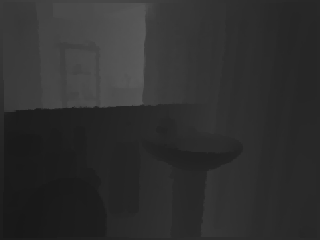}{(-2.0,0.0)}{(-1.5,1.5)}
    
    &
    
    \nyuimagewithbox{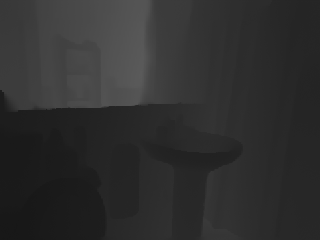}{(-2.0,0.0)}{(-1.5,1.5)}
    
    &
    
    \nyuimagewithbox{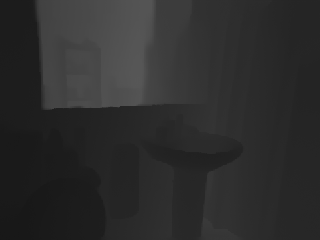}{(-2.0,0.0)}{(-1.5,1.5)}

    \\
\end{tabular}
\caption{Examples of multimodal predictions on the NYU Depth V2 val dataset. Rows 1-2 contain glass doors/windows where the model learns to predict the depth for either the glass surface or the surface behind it. Row 3 has a dark area next to the refrigerator for which the depth is unclear from RGB alone. In row 4 the model hallucinates the reflected door as a bath cabinet, which seems plausible from the RGB image.}
\label{fig:multimodal_nyu_samples}

\end{figure*}

\def\kittiwidth{0.24}
\def\boxcolor{lightgray}

\def \kittiimagewithbox#1#2#3{\begin{tikzpicture} \node[inner sep=0] at (0, 0) {\includegraphics[width=\kittiwidth\textwidth]{#1}}; \draw[\boxcolor,ultra thin,dashed] #2 rectangle #3; \end{tikzpicture}}

\begin{figure*}
\begin{tabular}{cc|cc}
    \multicolumn{2}{c}{Groundtruth} &  \multicolumn{2}{c}{Predictions} \\        
    
    

        \kittiimagewithbox{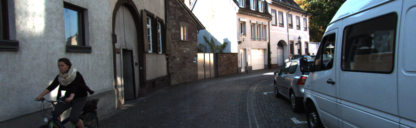}{(1.35,-0.15)}{(1.9, 0.4)}

    &
    \kittiimagewithbox{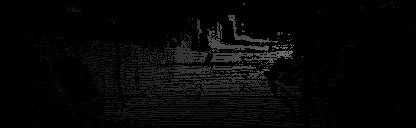}{(1.35,-0.15)}{(1.9, 0.4)}

    & 
    \kittiimagewithbox{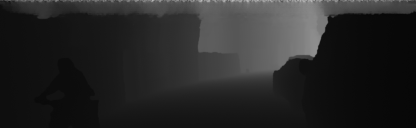}{(1.35,-0.15)}{(1.9, 0.4)}
    
    & 
    \kittiimagewithbox{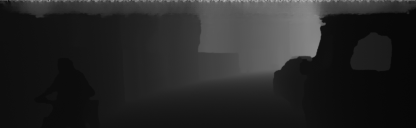}{(1.35,-0.15)}{(1.9, 0.4)}

    \\

    \kittiimagewithbox{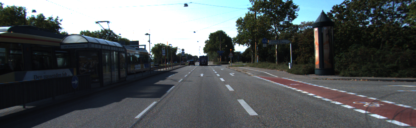}{(-2.05,-0.1)}{(-1.3, 0.2)}

    &
    \kittiimagewithbox{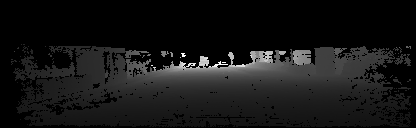}{(-2.05,-0.1)}{(-1.3, 0.2)}

    & 
    \kittiimagewithbox{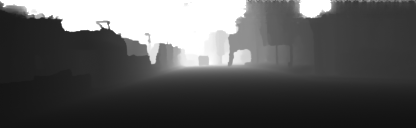}{(-2.05,-0.1)}{(-1.3, 0.2)}
    
    & 
    \kittiimagewithbox{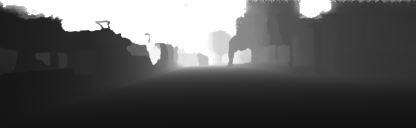}{(-2.05,-0.1)}{(-1.3, 0.2)}

        \\

    \kittiimagewithbox{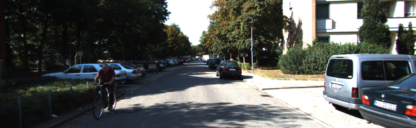}{(1.2,-0.15)}{(1.75, 0.1)}

    &
    \kittiimagewithbox{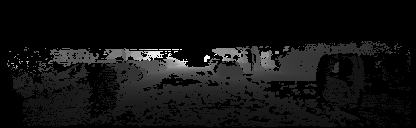}{(1.2,-0.15)}{(1.75, 0.1)}

    & 
    \kittiimagewithbox{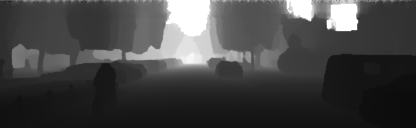}{(1.2,-0.15)}{(1.75, 0.1)}
    
    & 
    \kittiimagewithbox{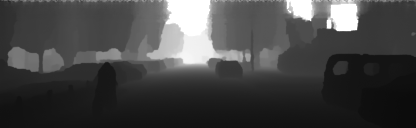}{(1.2,-0.15)}{(1.75, 0.1)}




    

            \\
    
    \kittiimagewithbox{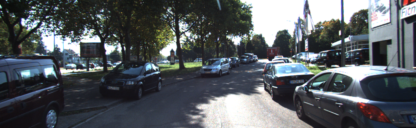}{(1.77,-0.07)}{(2.05, -0.3)}

    &
    \kittiimagewithbox{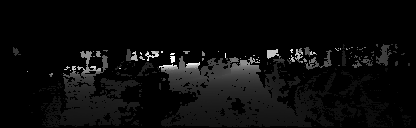}{(1.77,-0.07)}{(2.05, -0.3)}

    & 
    \kittiimagewithbox{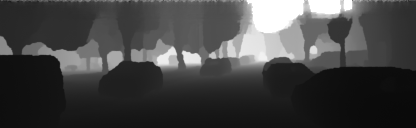}{(1.77,-0.07)}{(2.05, -0.3)}
    
    & 
    \kittiimagewithbox{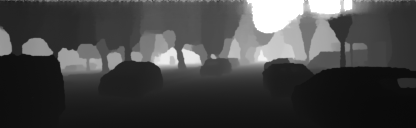}{(1.77,-0.07)}{(2.05, -0.3)}

\end{tabular}
\caption{Multimodal depth predictions on the KITTI val dataset.}
\label{fig:multimodal_kitti_samples}
\end{figure*}

\subsection{Ablations}

We find that both pre-training and accounting for missing depth are crucial to model performance.
Table  \ref{tab::pretraining_ablation} shows that both self-supervised pre-training and supervised depth pre-training are important, with supervised depth training having a larger impact, which is to be expected. 
Table \ref{tab::missing_depth_ablation} shows that depth infilling is extremely important for the outdoor KITTI dataset. It has less impact on NYU, which is understandable since KITTI has sparser depth maps. 
In the absence of filled depth maps, \stepunrolled~dramatically improves results especially on KITTI. Even with depth infilling, \stepunrolledshort~consistently improves performance for both indoor and outdoor datasets. Additionally, we ablate the choice of loss function in Table \ref{tab::loss_ablation}. We find that the $L_1$ loss yields better performance than $L_2$, likely because $L_1$ is more robust to noise at larger depths. 
(See Appendix for metrics other than absolute relative error.)

\def\imwidth{0.0975}
\def\imwidthlast{0.195}
\def\lastcoloffset{1.6cm}
\def\textpadding{3pt}
\def\ncols{9}

\begin{figure*}[t]
\centering
\renewcommand{\arraystretch}{0.2}
\addtolength{\tabcolsep}{-5pt}
\begin{tabular}{ccccccccc}
\includegraphics[width=\imwidth\linewidth]{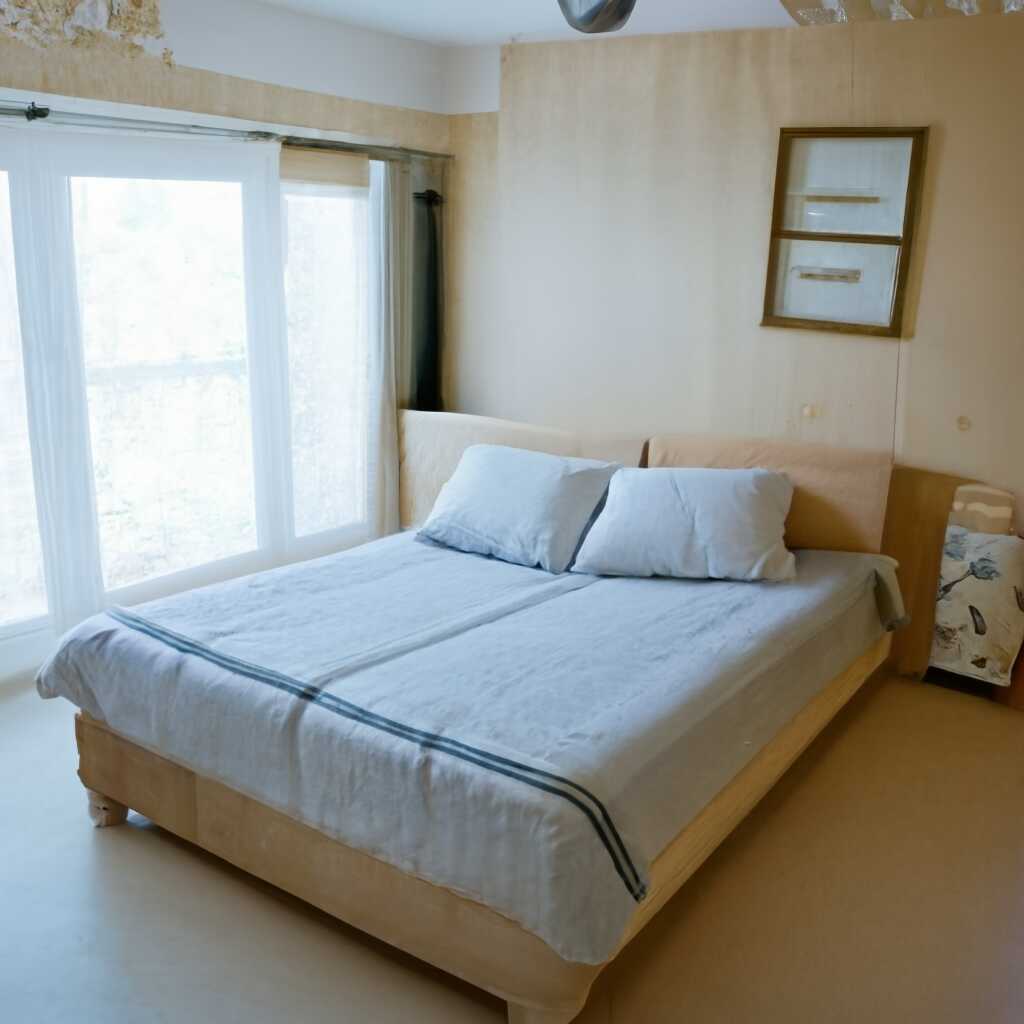} & \includegraphics[width=\imwidth\linewidth]{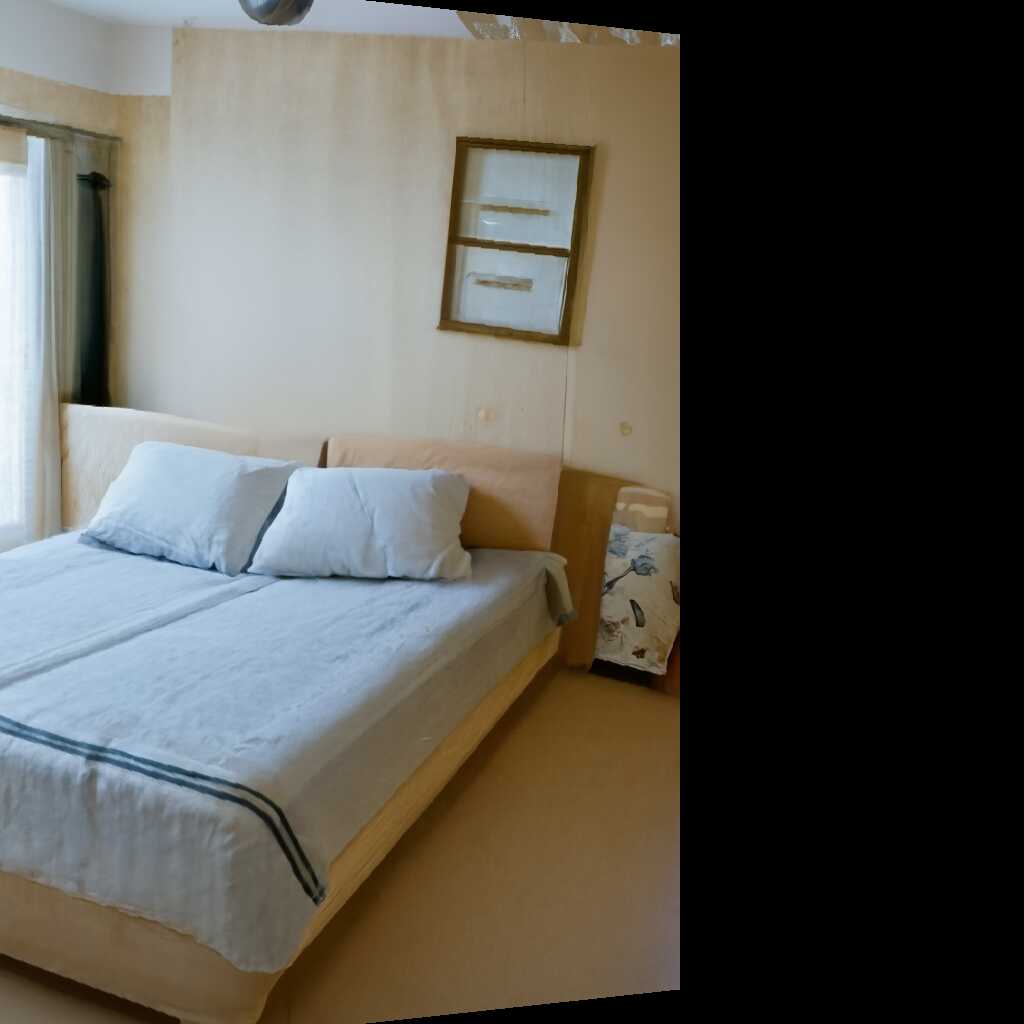} &\includegraphics[width=\imwidth\linewidth]{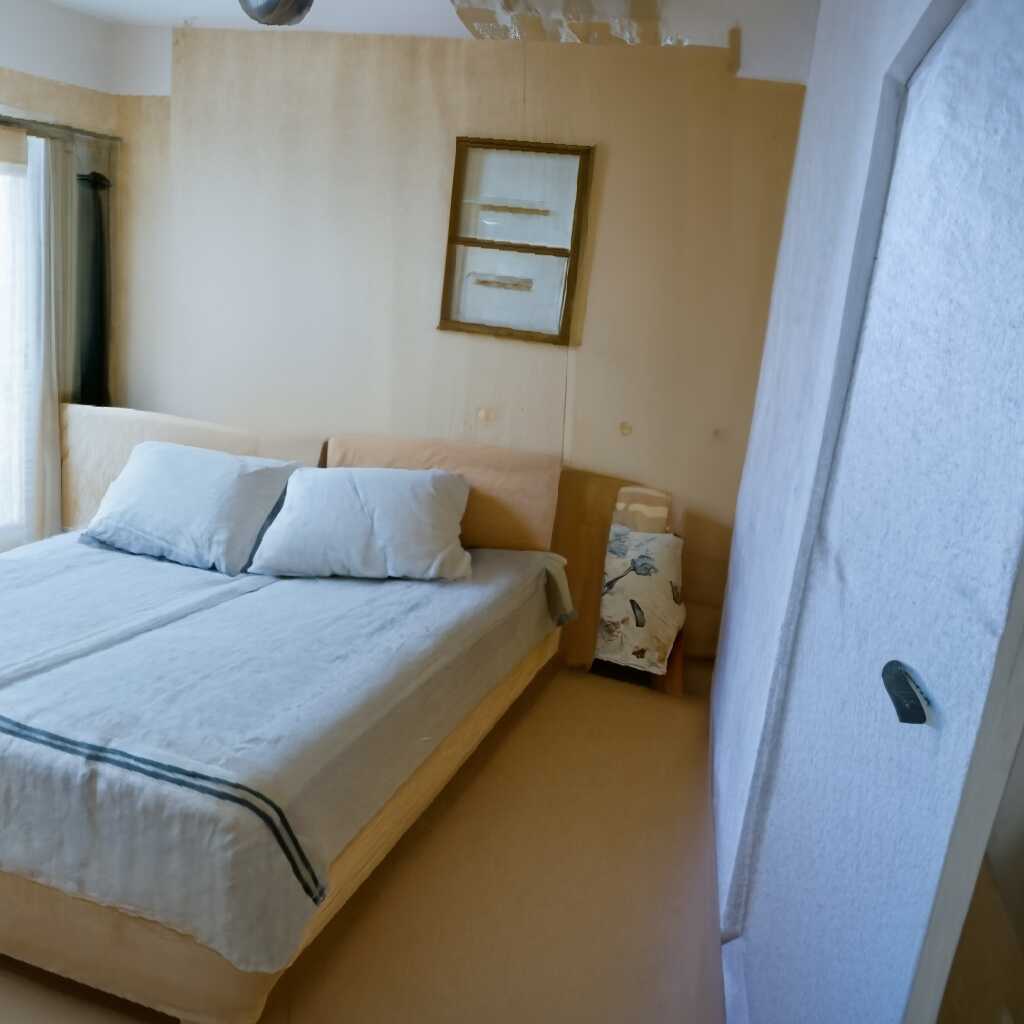} &\includegraphics[width=\imwidth\linewidth]{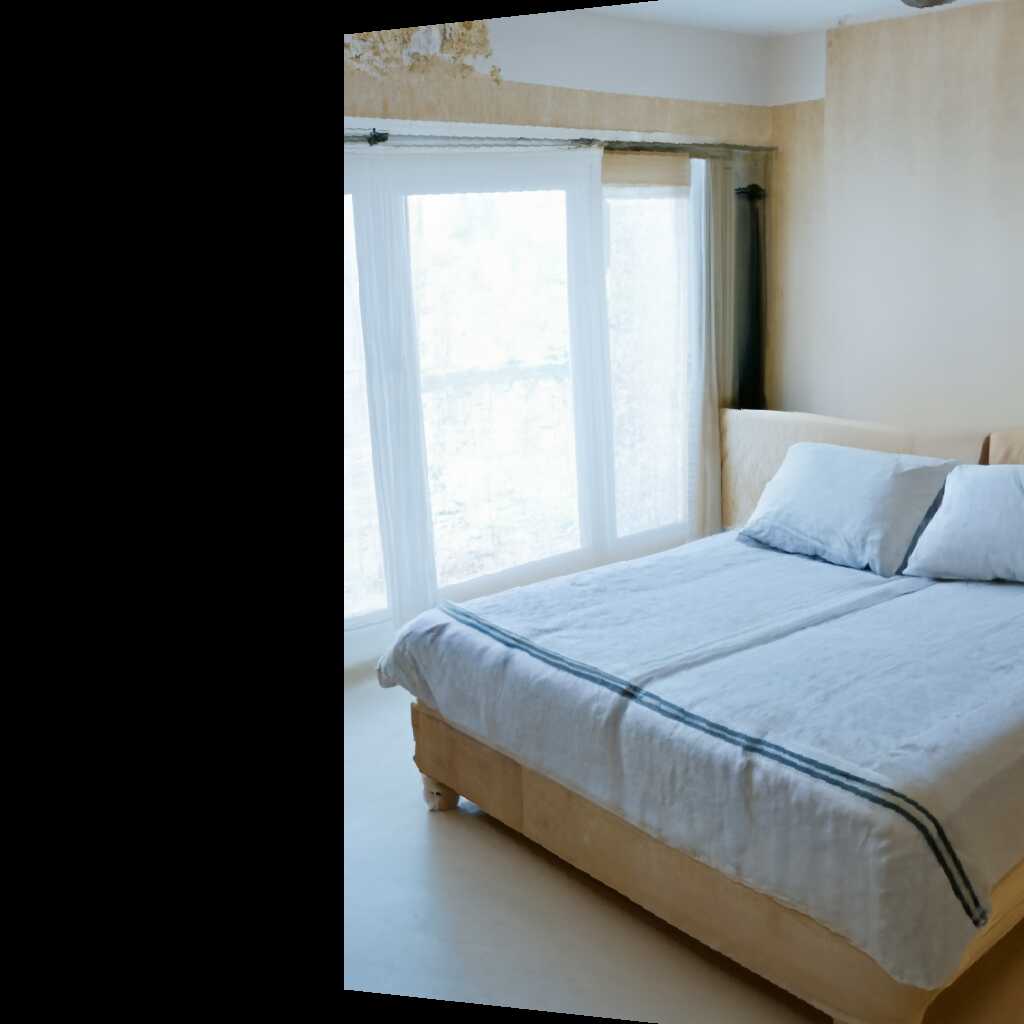} &\includegraphics[width=\imwidth\linewidth]{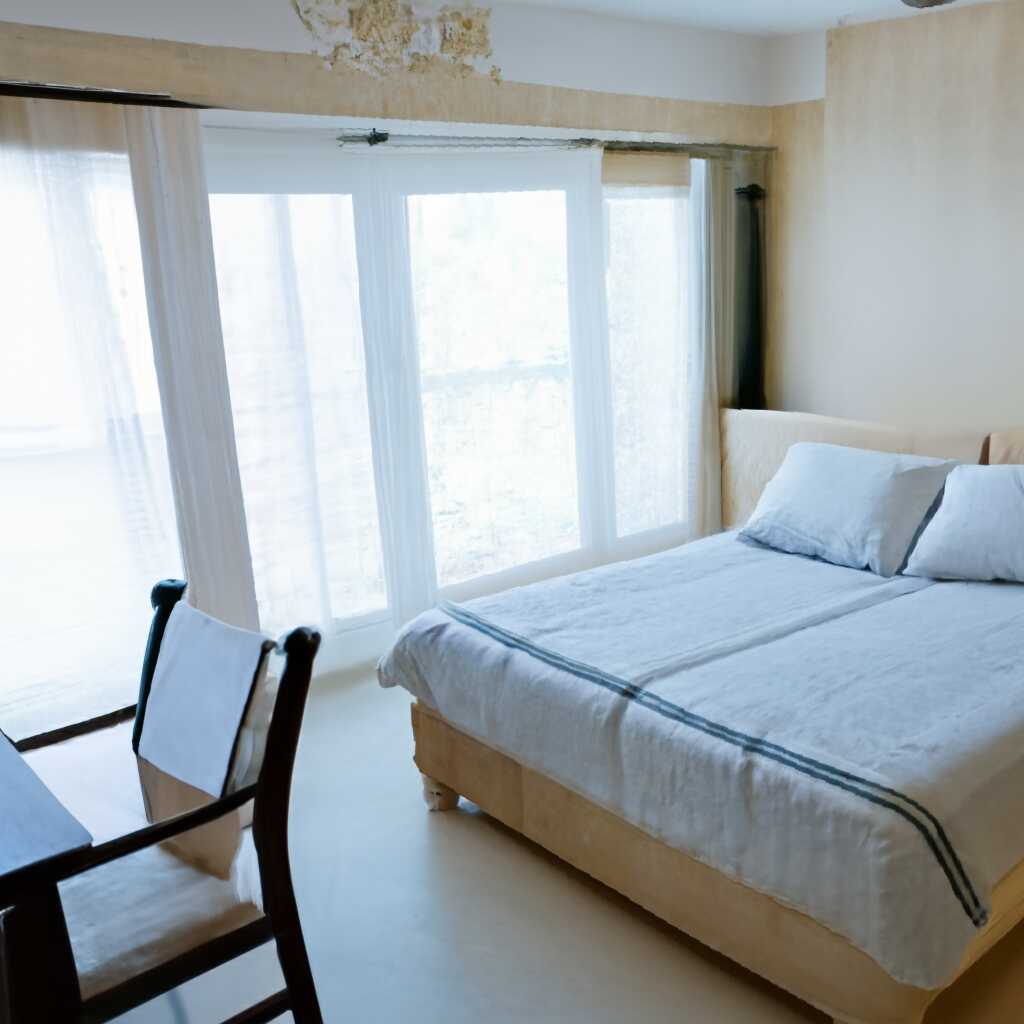} & \includegraphics[width=\imwidth\linewidth]{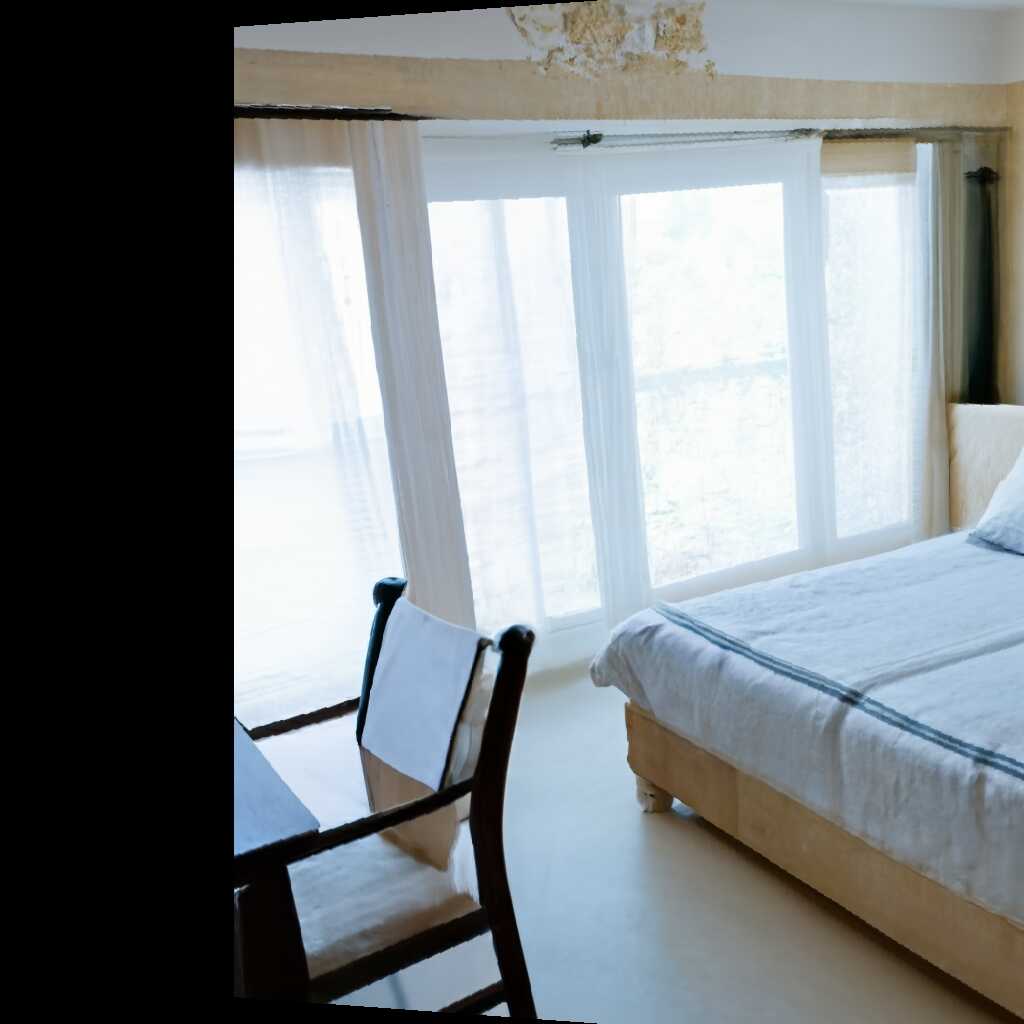} &\includegraphics[width=\imwidth\linewidth]{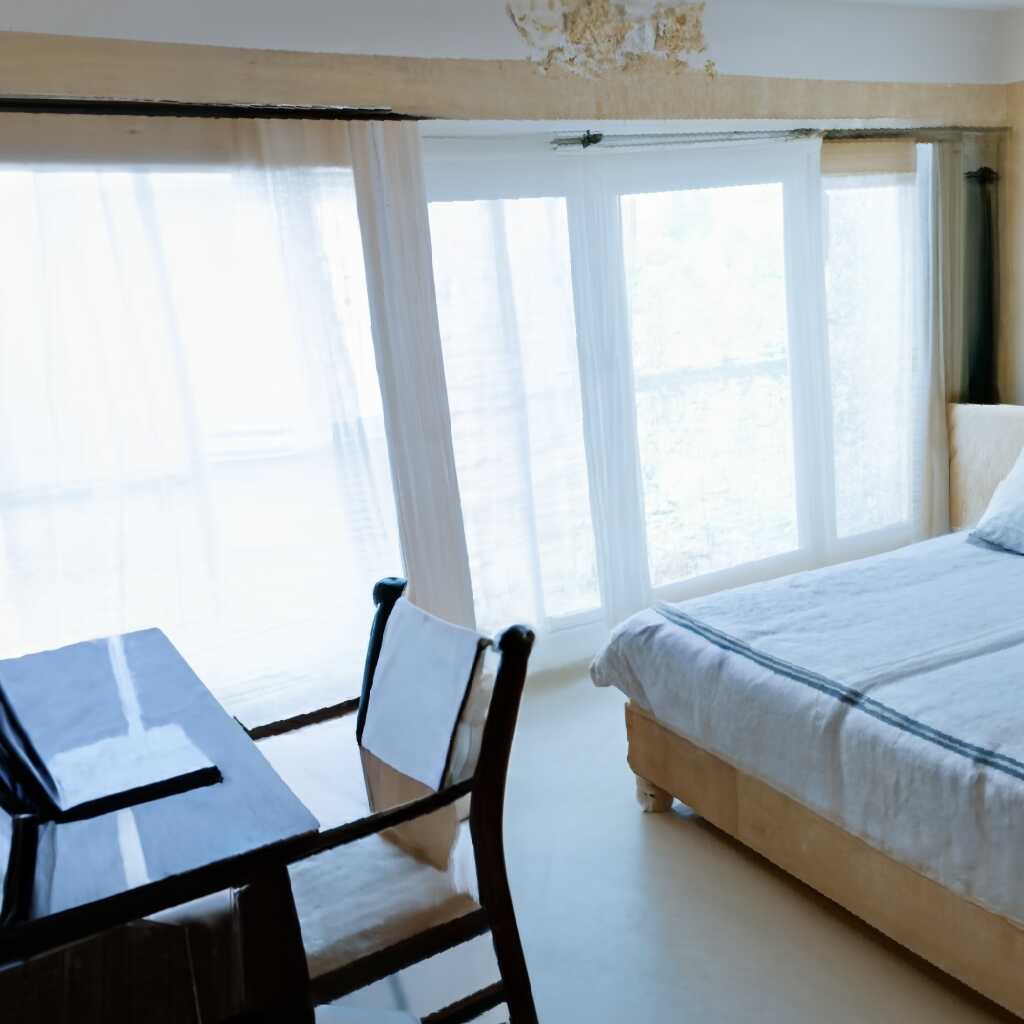} & \multicolumn{2}{c}{\multirow{2}{*}[\lastcoloffset]{\includegraphics[width=\imwidthlast\linewidth]{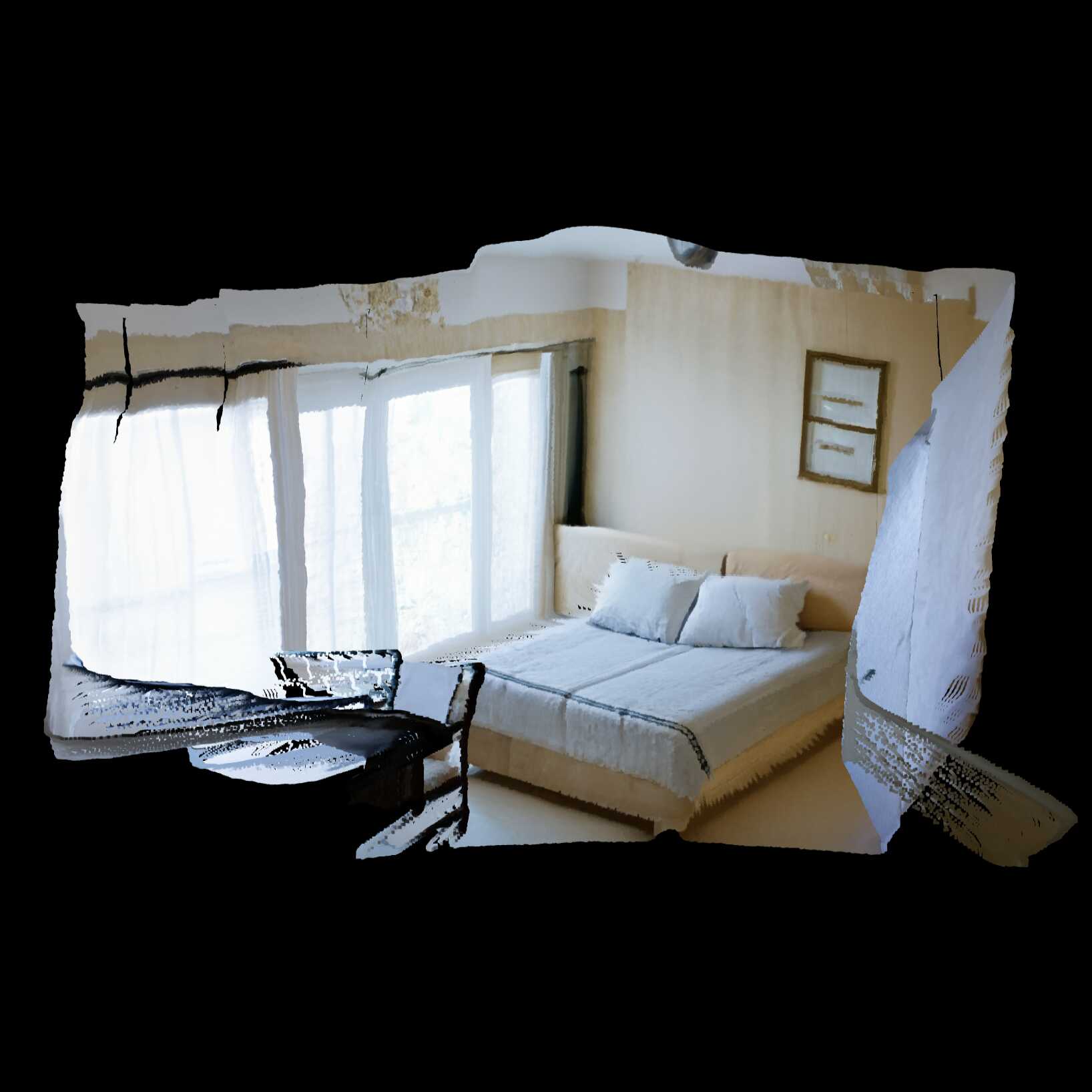}}} \\
\includegraphics[width=\imwidth\linewidth]{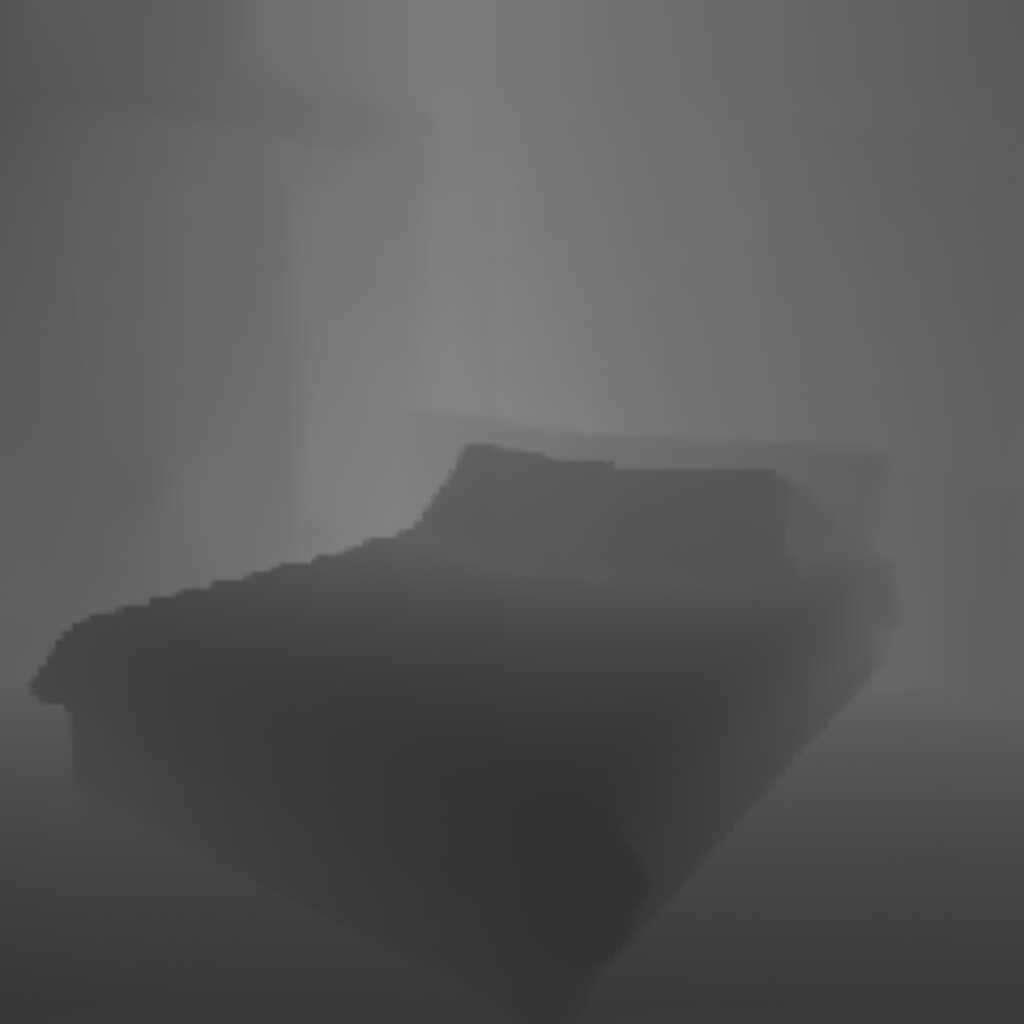} & \includegraphics[width=\imwidth\linewidth]{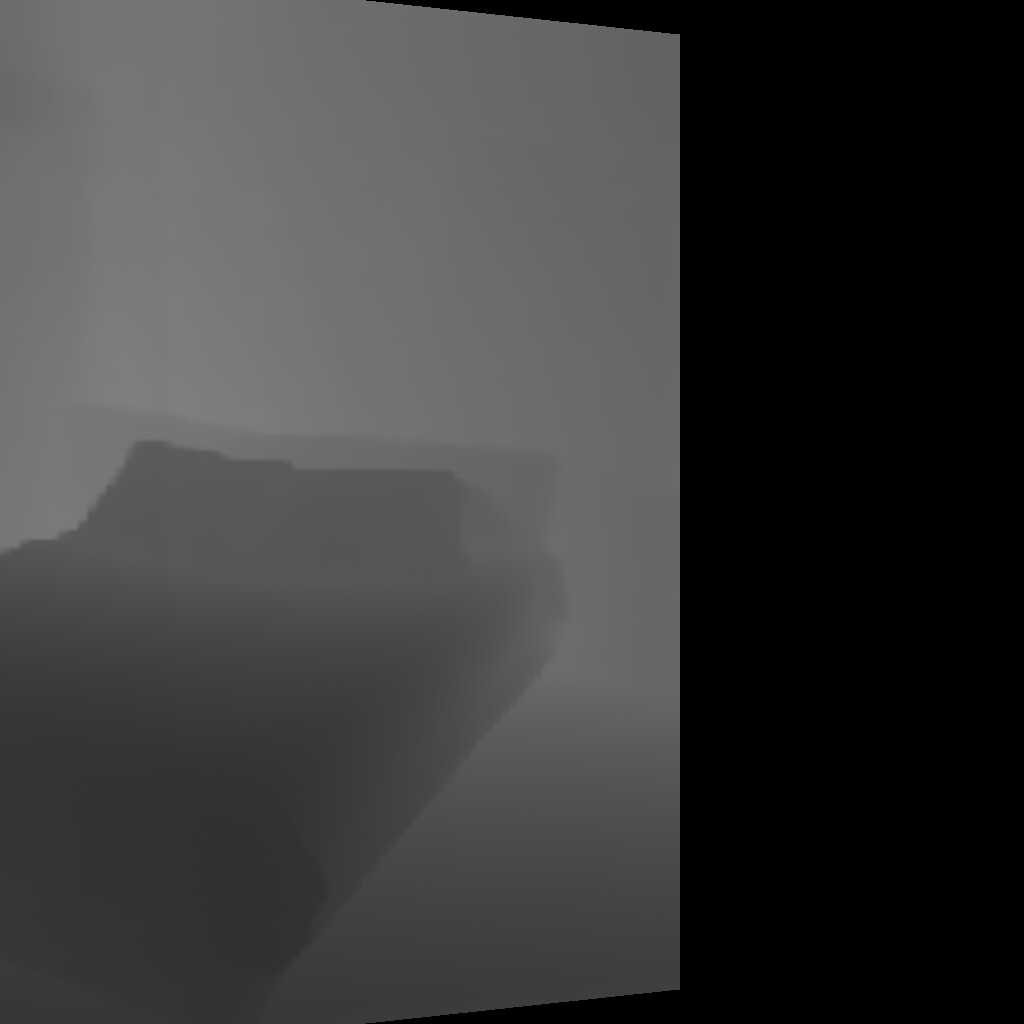} &\includegraphics[width=\imwidth\linewidth]{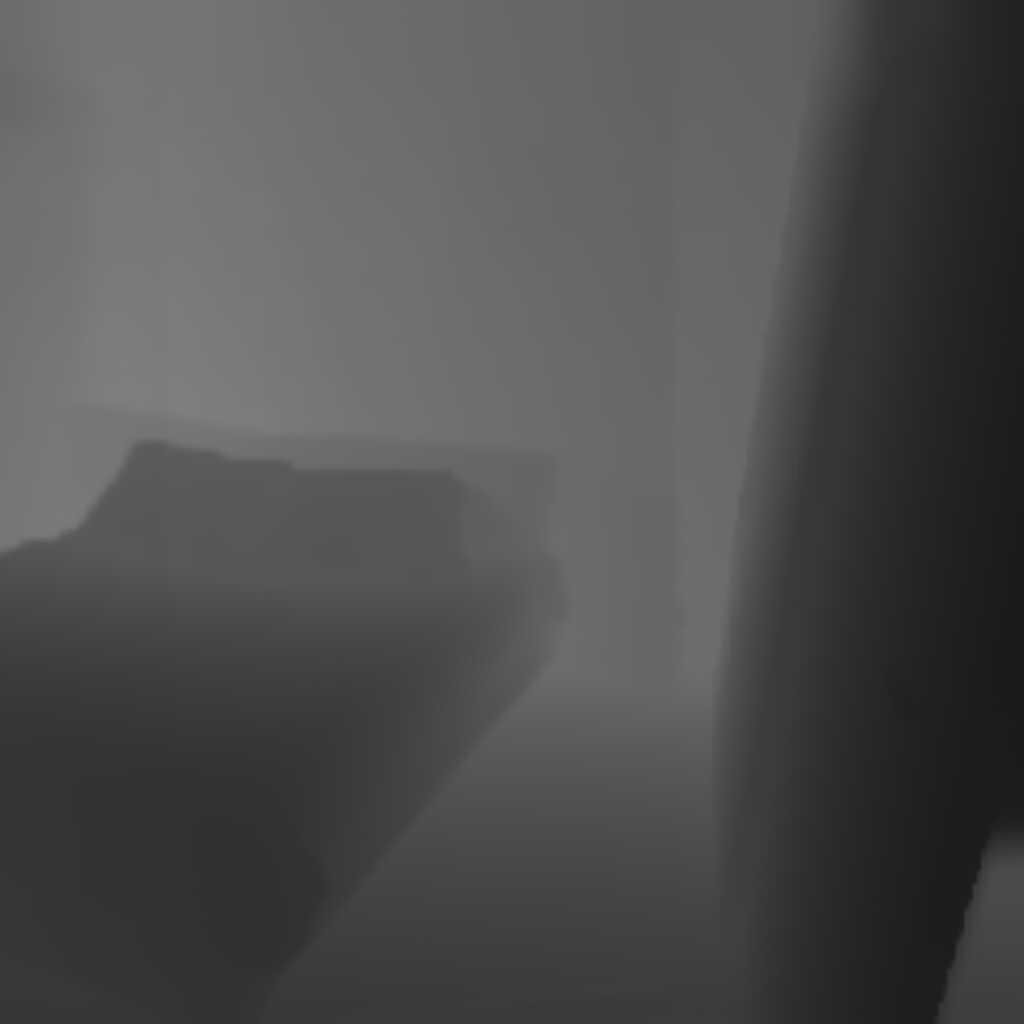} &  \includegraphics[width=\imwidth\linewidth]{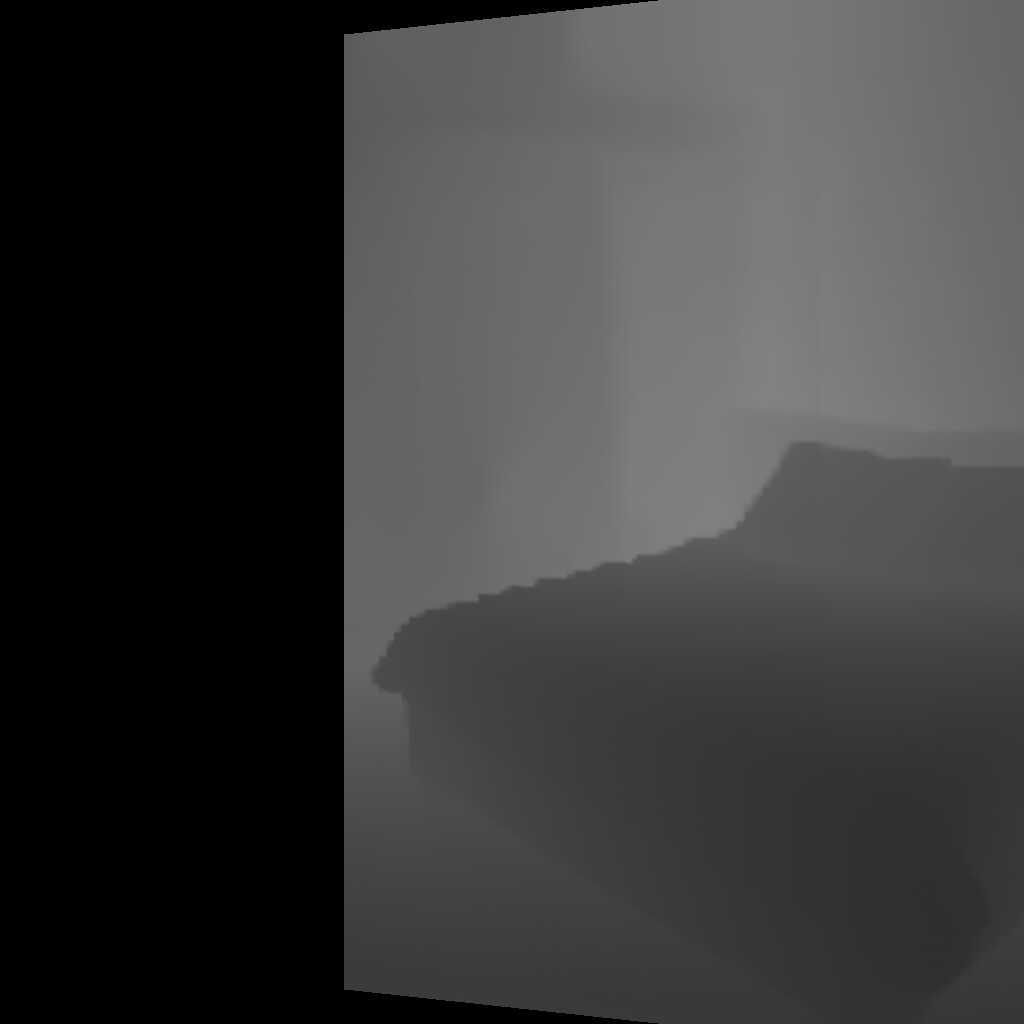} &\includegraphics[width=\imwidth\linewidth]{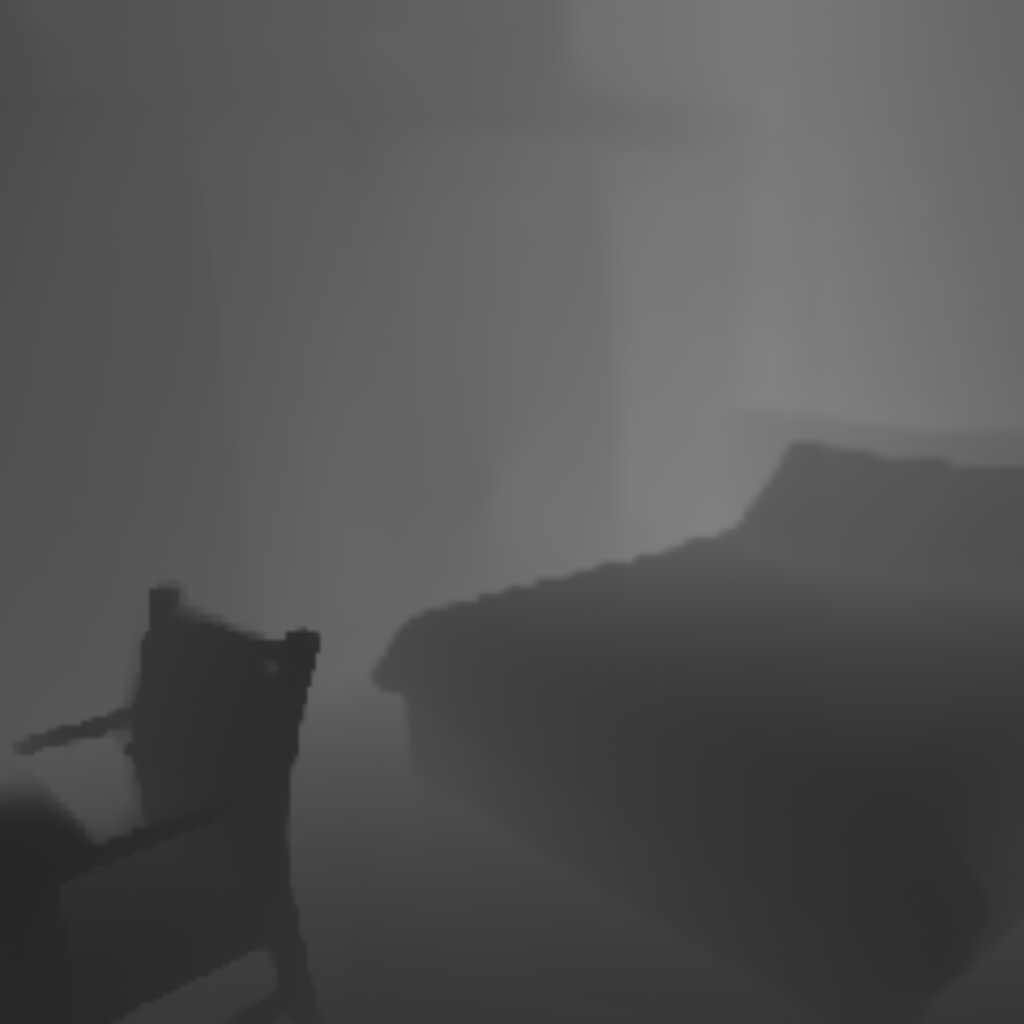} &\includegraphics[width=\imwidth\linewidth]{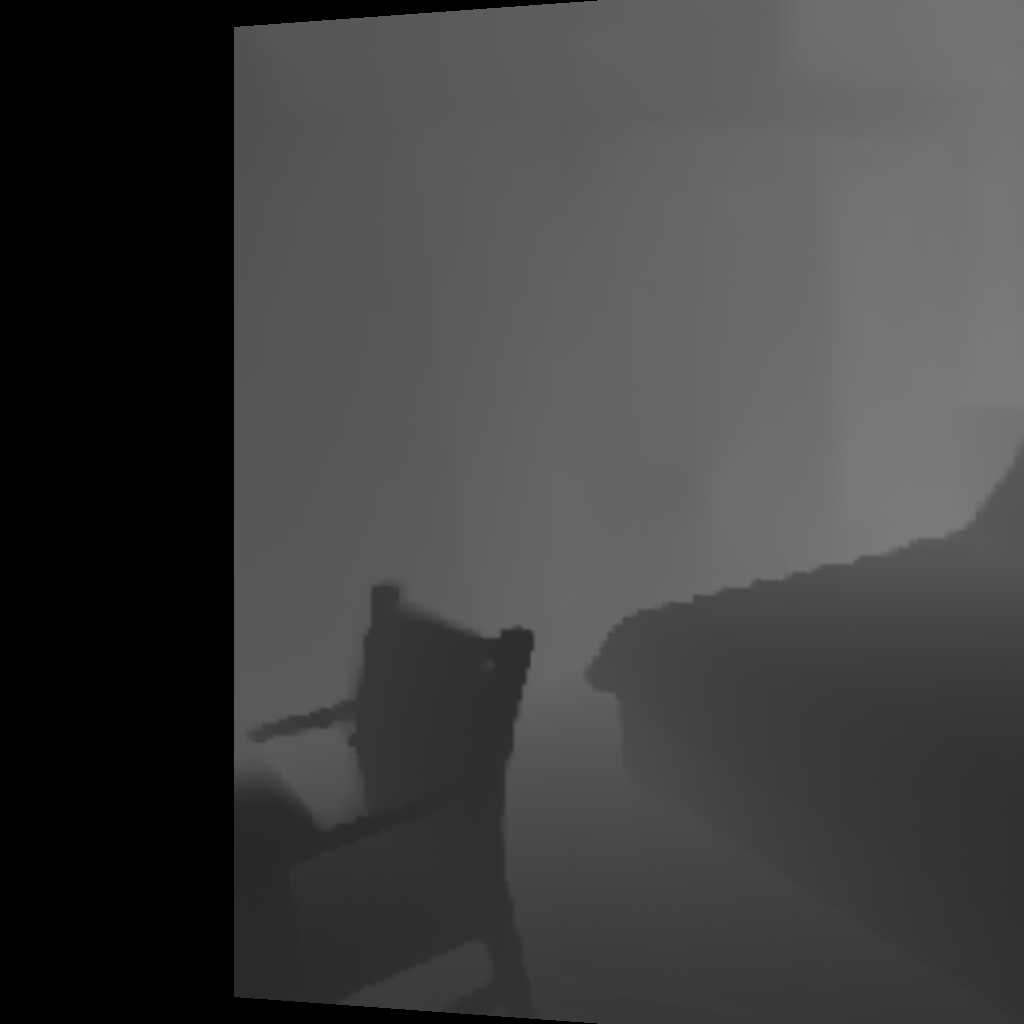} &\includegraphics[width=\imwidth\linewidth]{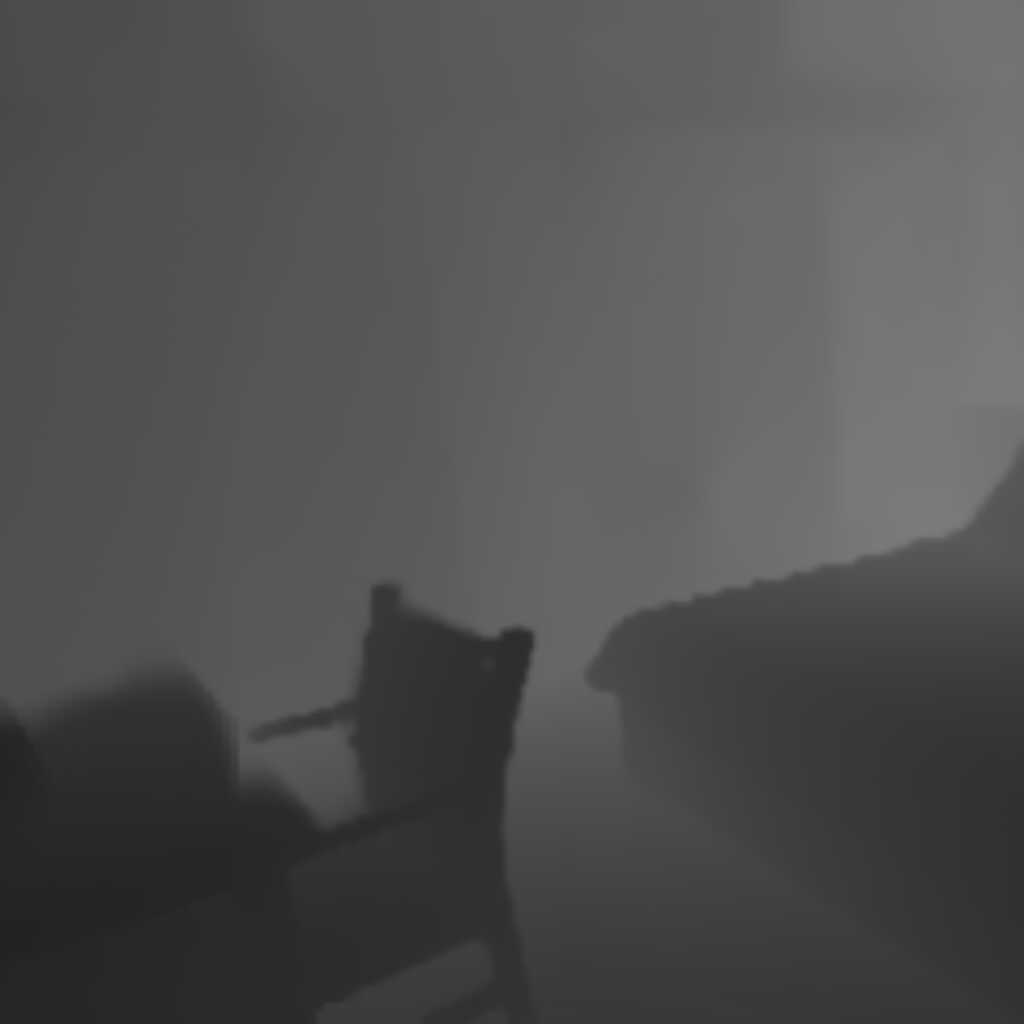} & \multicolumn{2}{c}{} \\
    \\[\textpadding]\multicolumn{\ncols}{c}{\footnotesize{Prompt: A bedroom.}}\\\\[\textpadding]

    \\

    \includegraphics[width=\imwidth\linewidth]{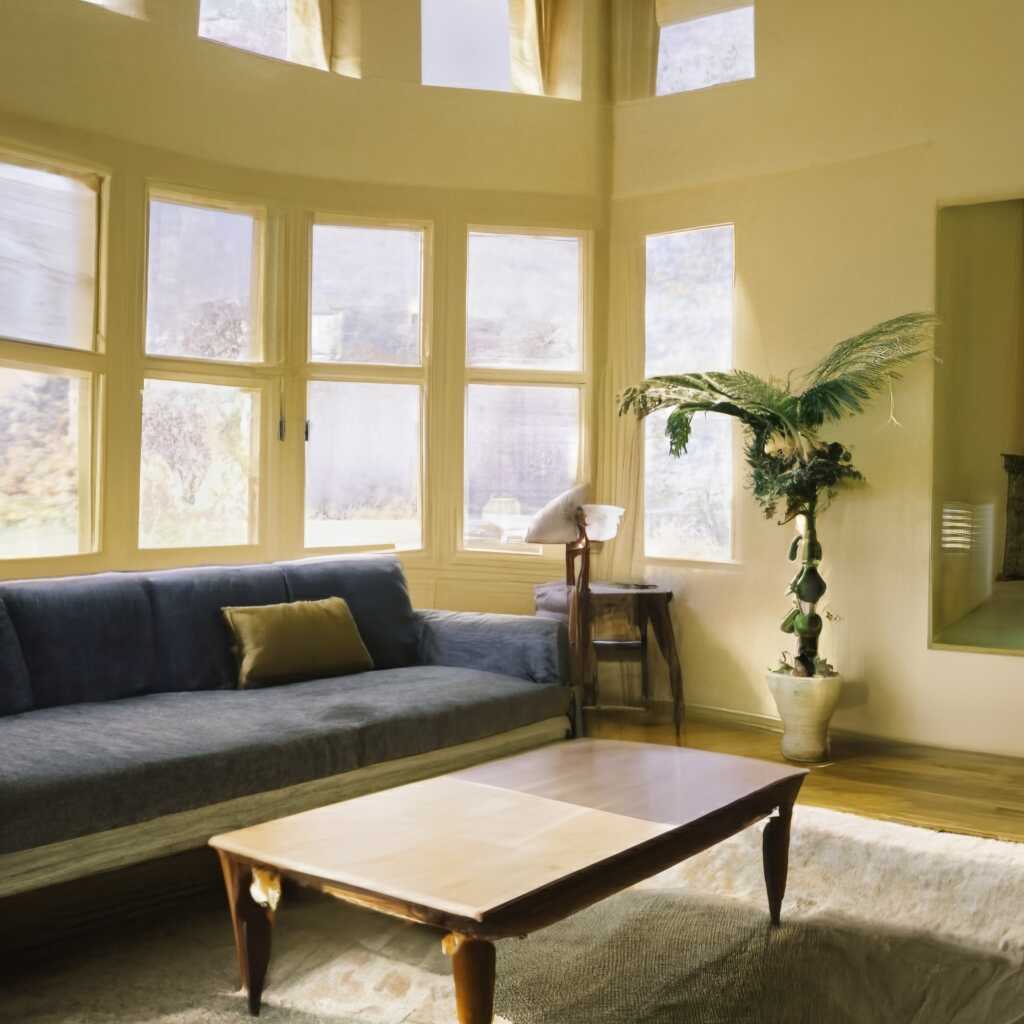} & \includegraphics[width=\imwidth\linewidth]{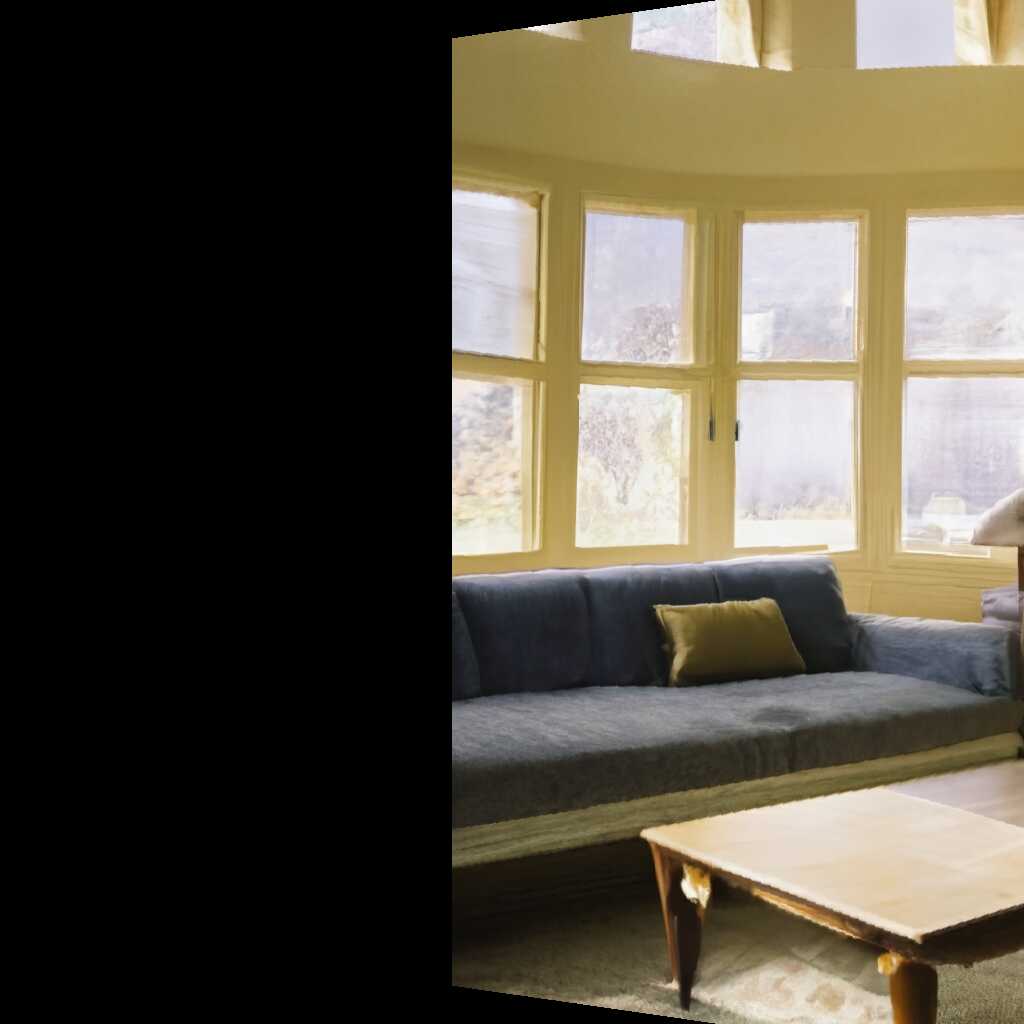} &\includegraphics[width=\imwidth\linewidth]{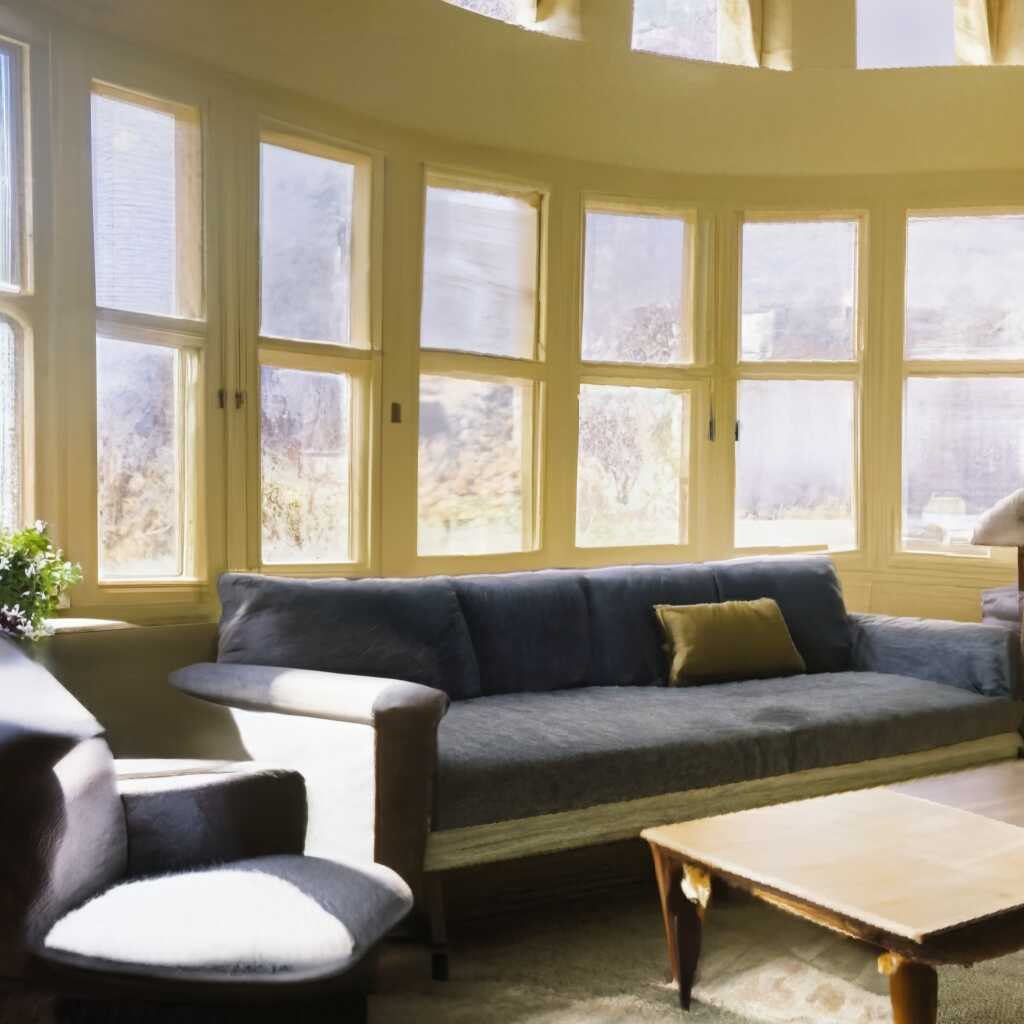} &\includegraphics[width=\imwidth\linewidth]{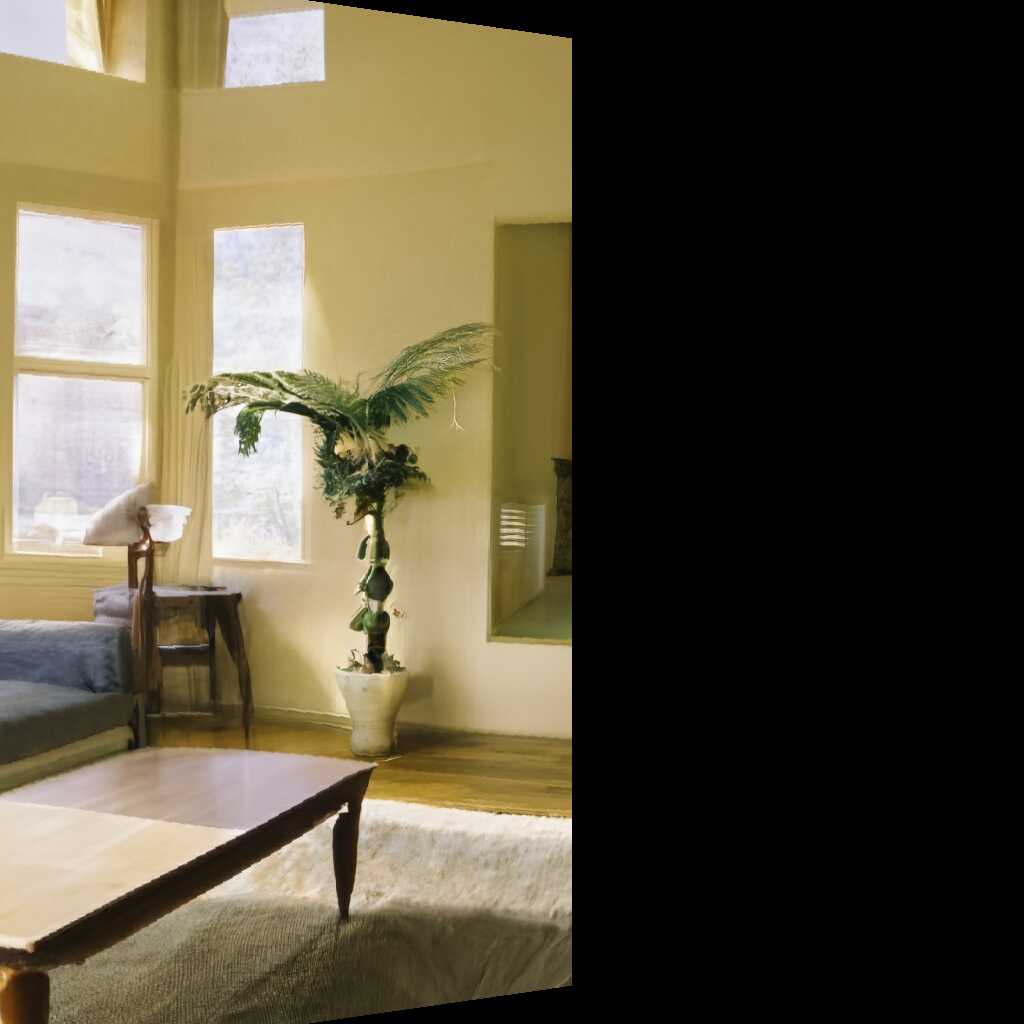} &\includegraphics[width=\imwidth\linewidth]{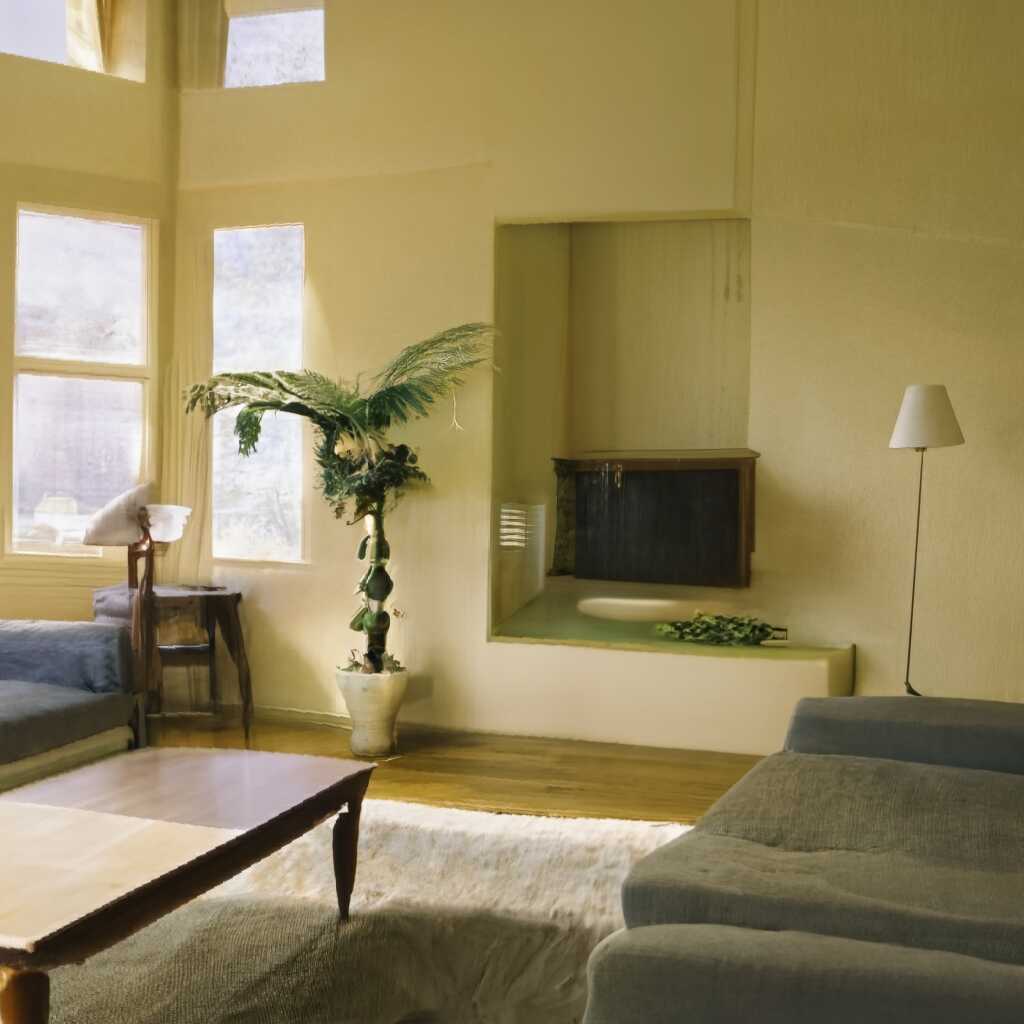} & \includegraphics[width=\imwidth\linewidth]{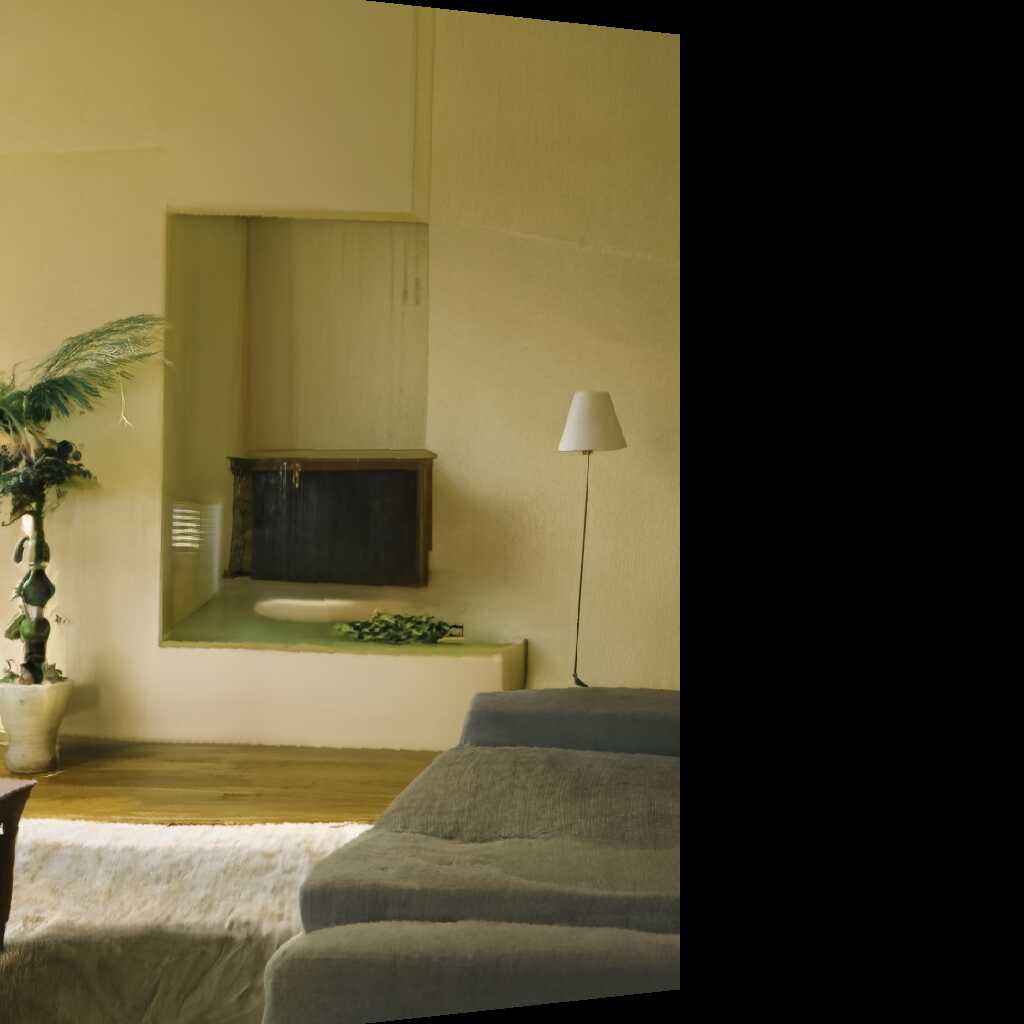} &\includegraphics[width=\imwidth\linewidth]{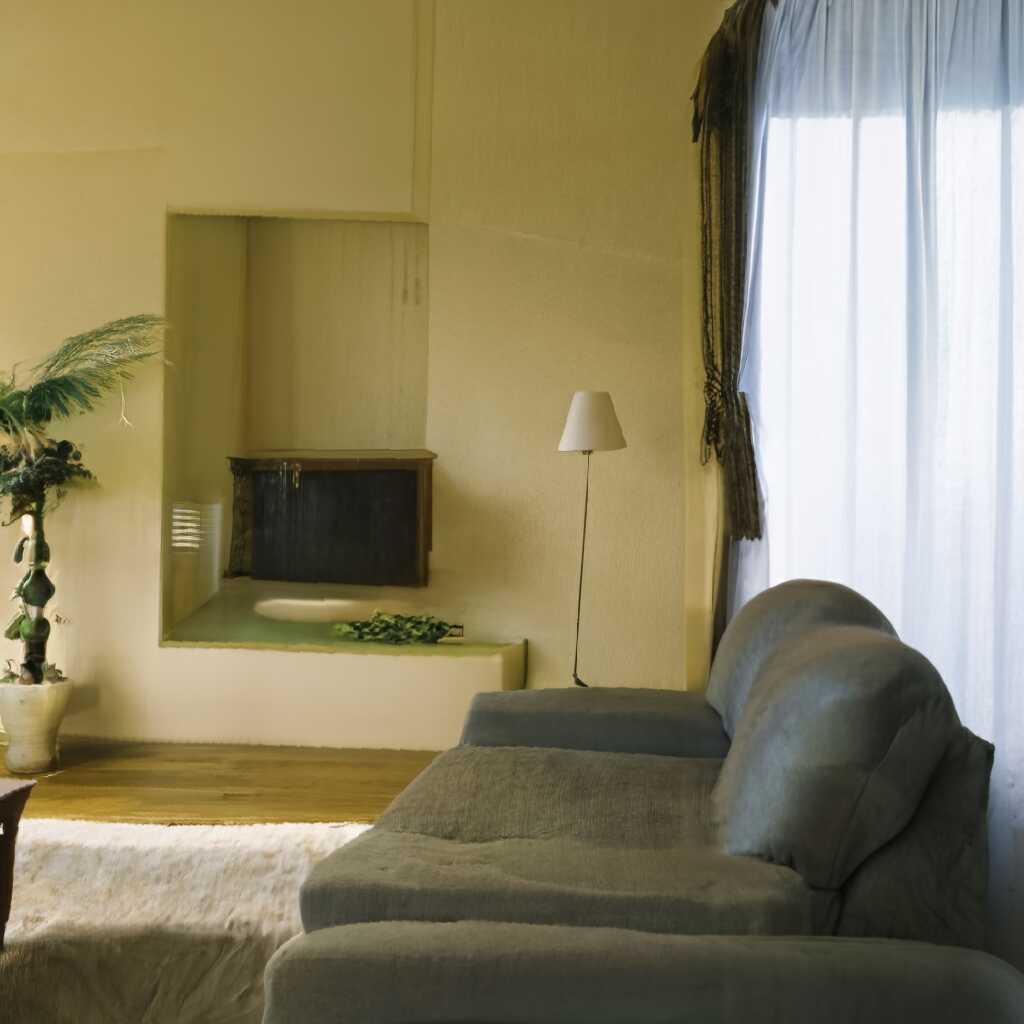} & \multicolumn{2}{c}{\multirow{2}{*}[\lastcoloffset]{\includegraphics[width=\imwidthlast\linewidth]{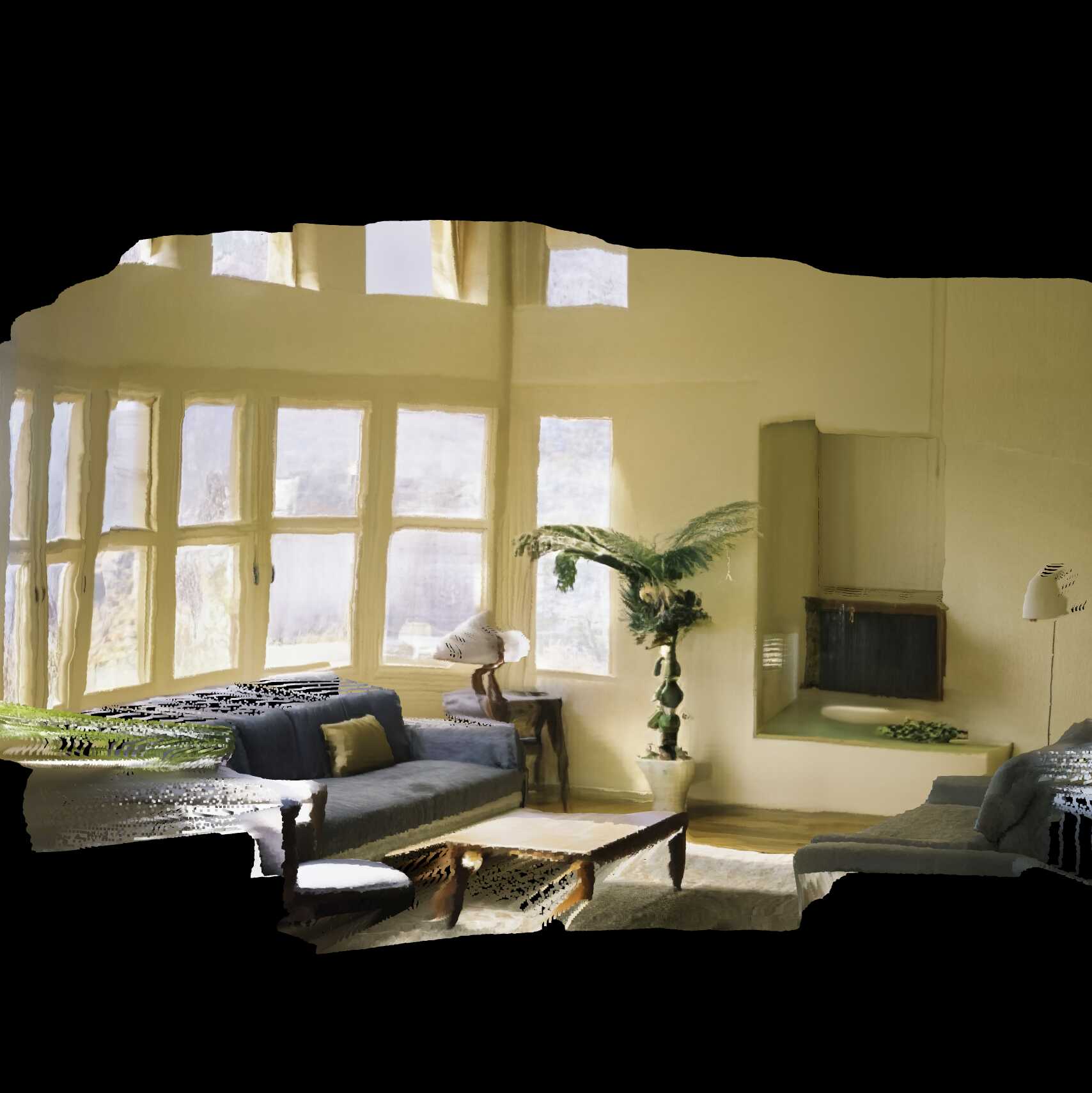}}} \\
    \includegraphics[width=\imwidth\linewidth]{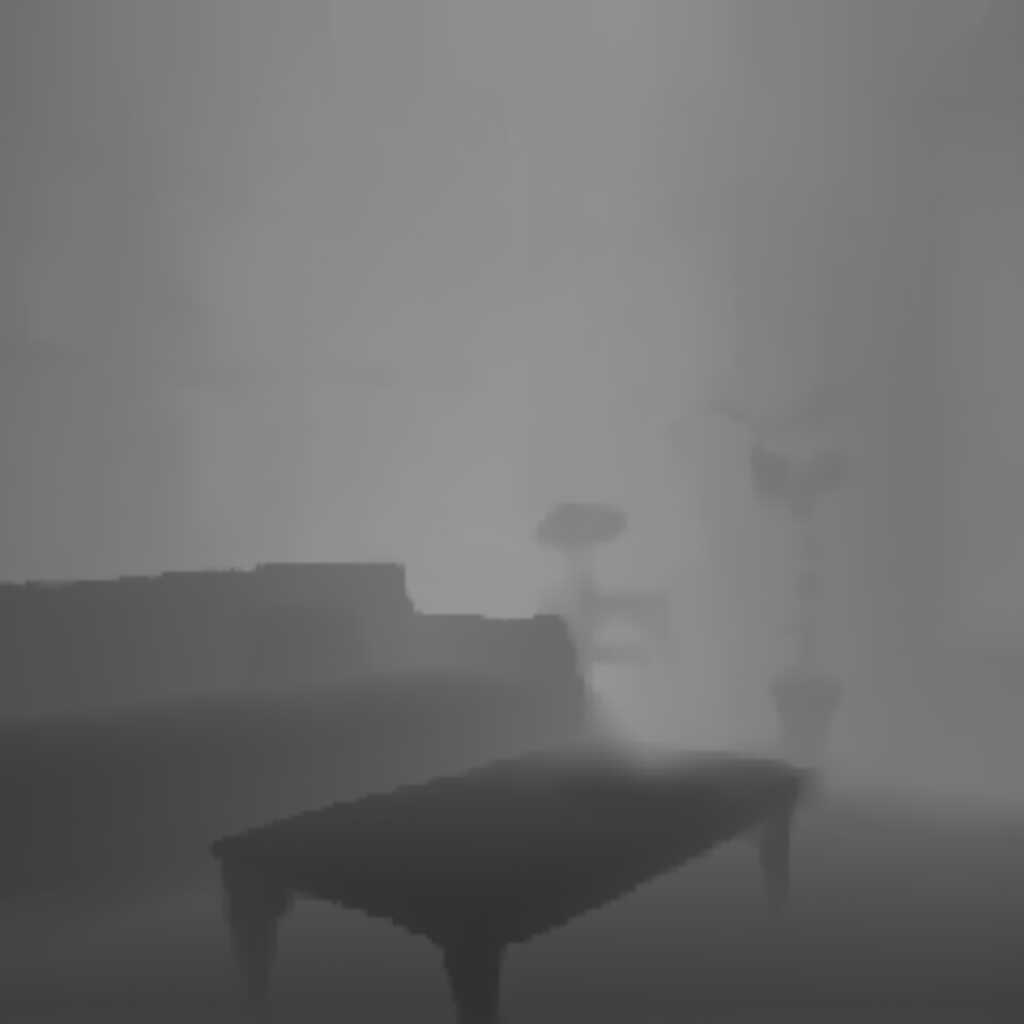} & \includegraphics[width=\imwidth\linewidth]{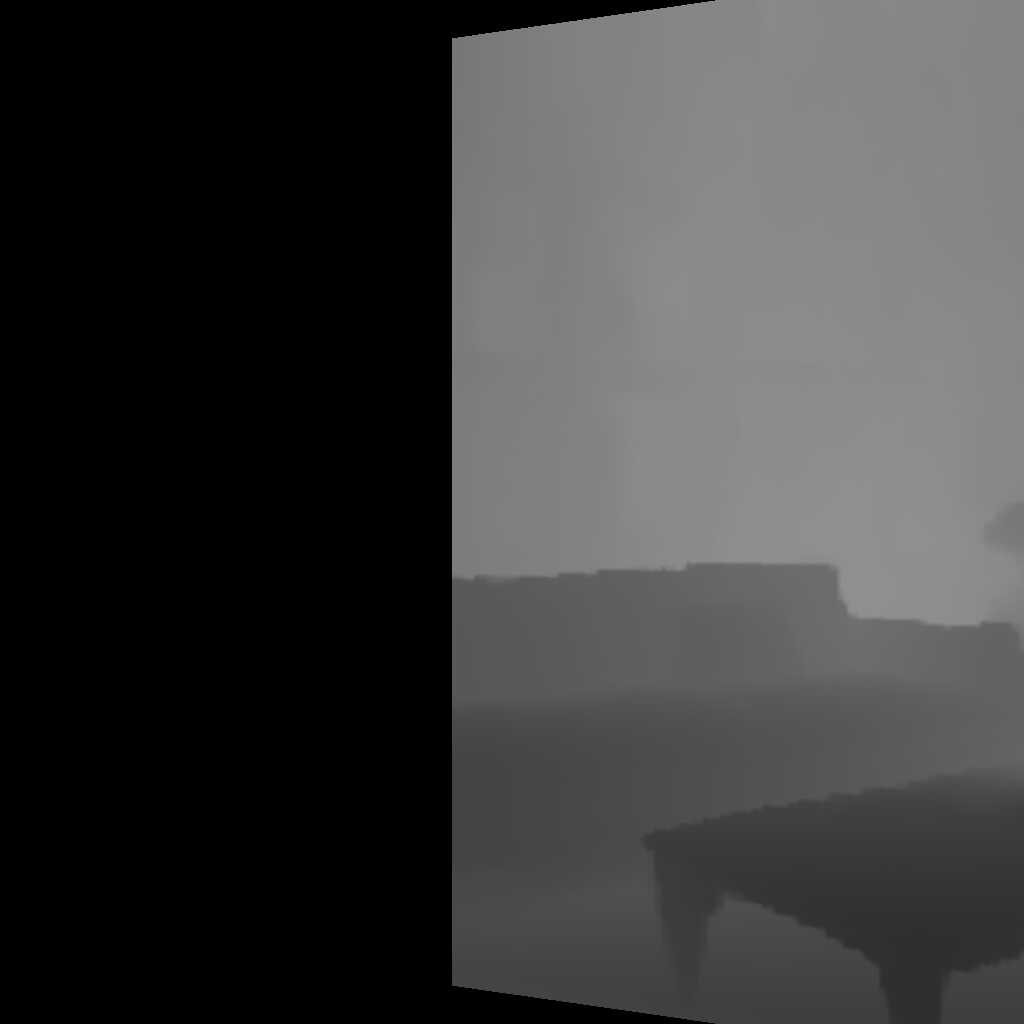} &\includegraphics[width=\imwidth\linewidth]{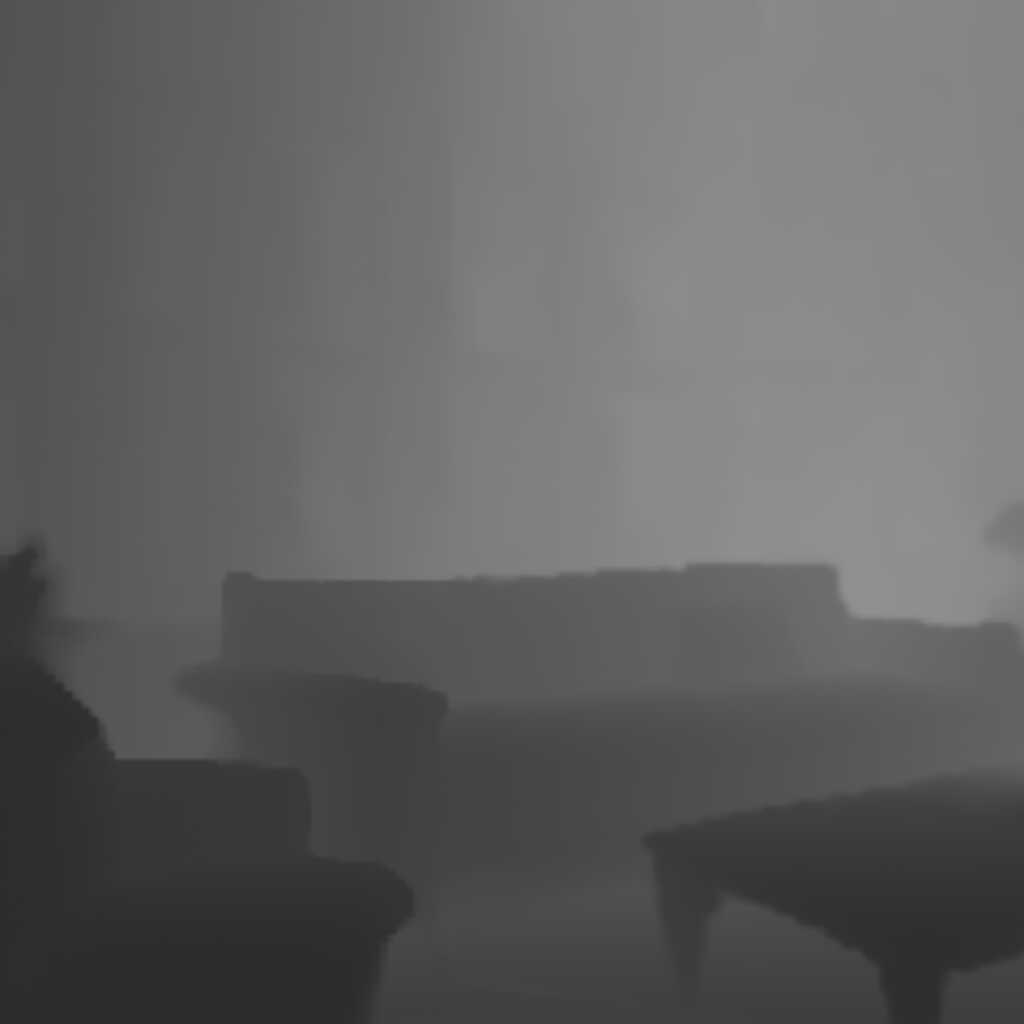} &  \includegraphics[width=\imwidth\linewidth]{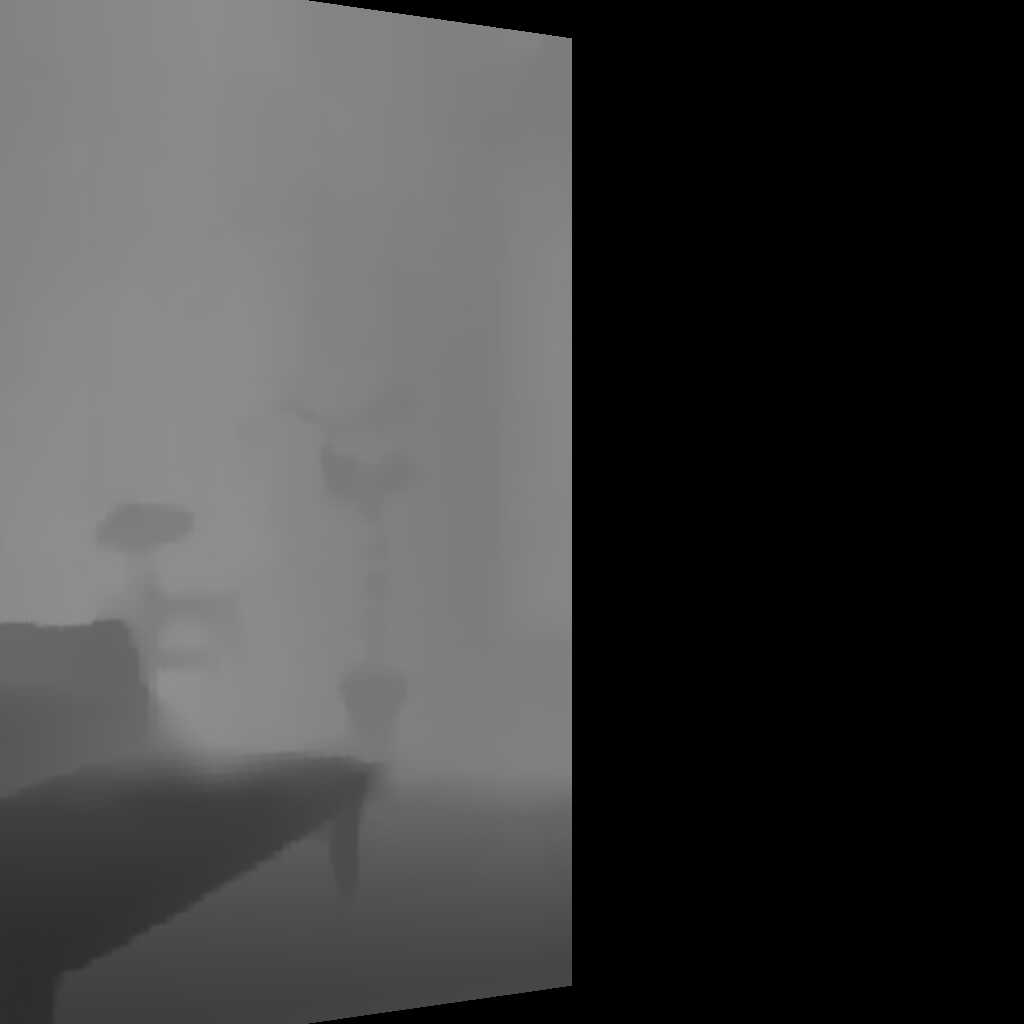} &\includegraphics[width=\imwidth\linewidth]{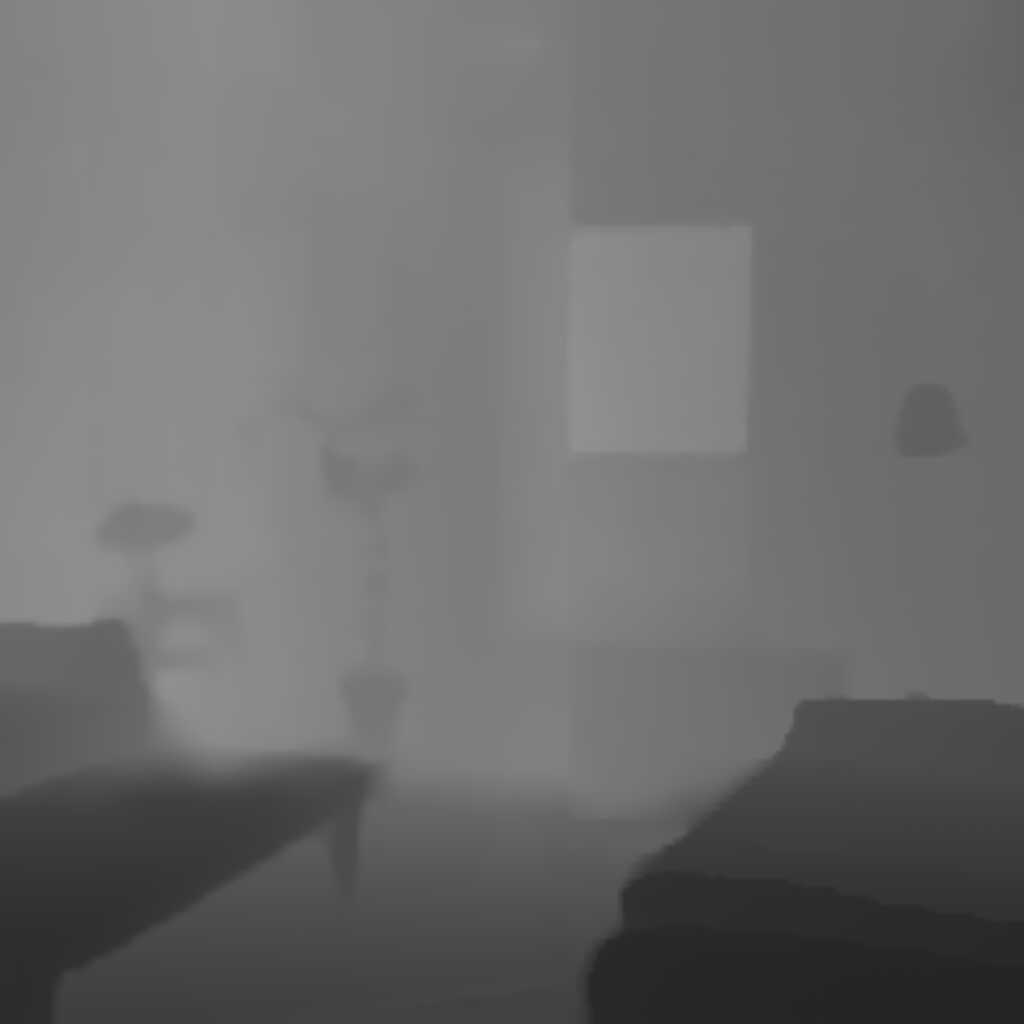} &\includegraphics[width=\imwidth\linewidth]{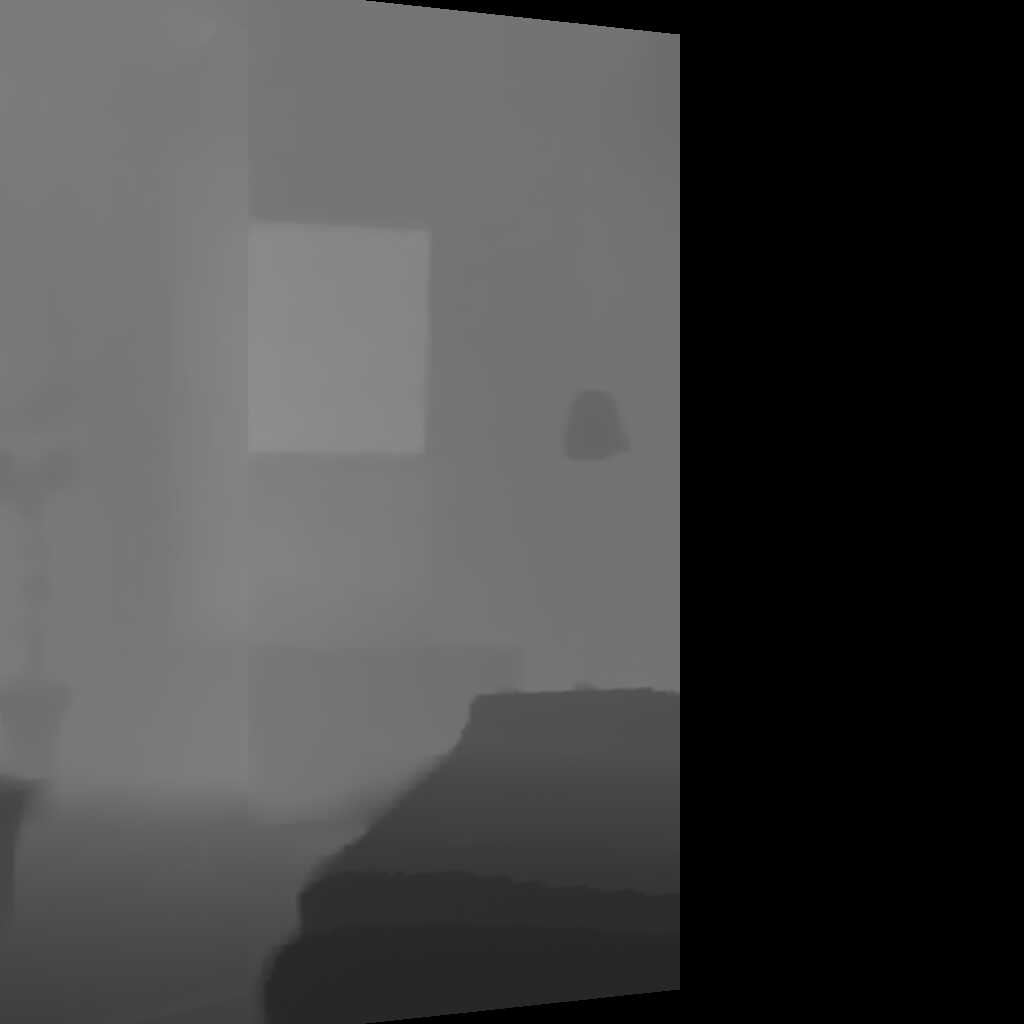} &\includegraphics[width=\imwidth\linewidth]{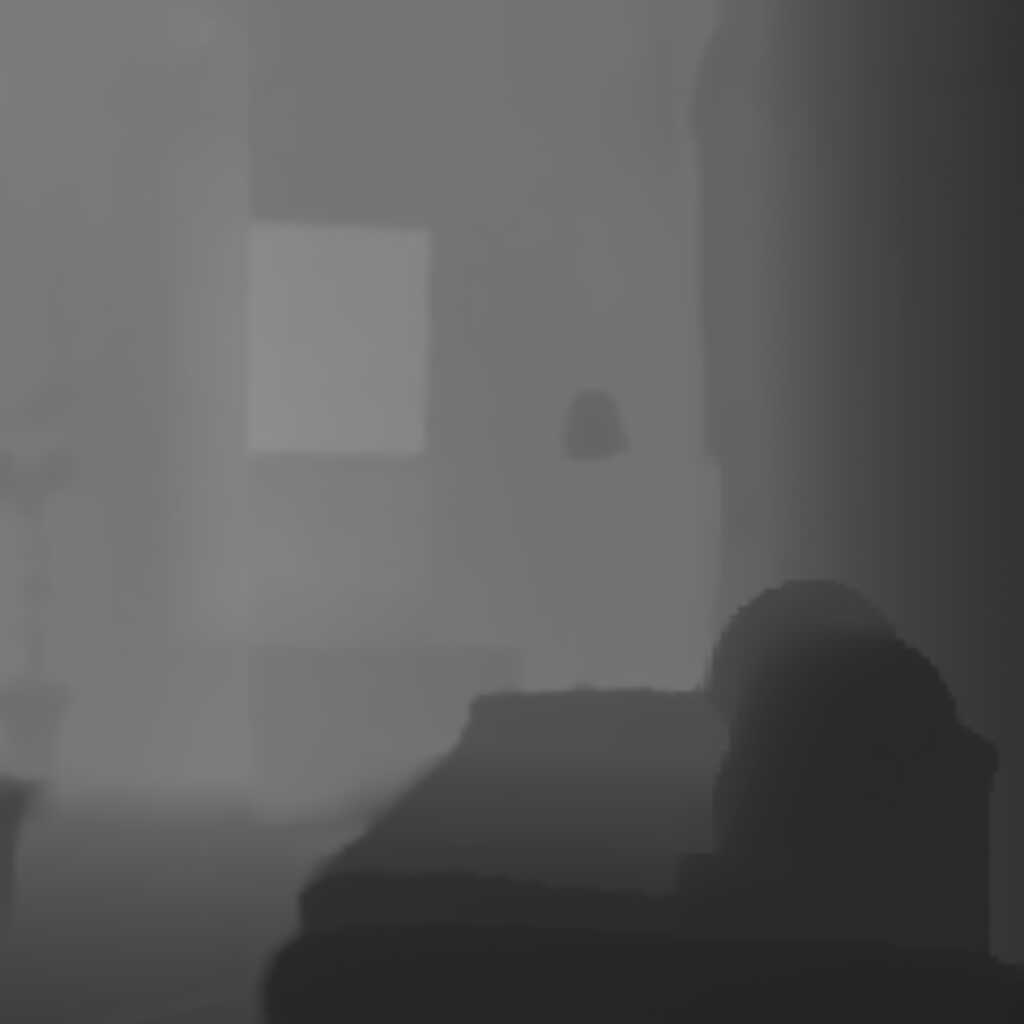} & \multicolumn{2}{c}{} \\
    \\[\textpadding]\multicolumn{\ncols}{c}{\footnotesize{Prompt: A living room.}}\\\\[\textpadding]


\end{tabular}
\vspace*{-0.3cm}
\caption{Text to 3D samples. Given a text prompt, an image is first generated using Imagen (first row of first column), after which depth is estimated (second row of first column). Subsequently the camera is moved to reveal new parts of the scene, which are then infilled using an image completion model and \ours~(which conditions on both the incomplete depth map and the filled image). At each step, newly generated RGBD points are added to a global point cloud which is visualized in the rightmost column. See \ref{tab:text_to_3d_samples_extras} for more samples.}
\vspace*{-0.2cm}
\label{tab:text_to_3d_samples}
\end{figure*}
\subsection{Multimodality}
One strength of diffusion models is their ability to capture complex multimodal distributions.  This can be effective in representing depth uncertainty, especially where there may exist natural depth ambiguities, e.g., in cases of transparency when looking through a window.
Figures \ref{fig:multimodal_nyu_samples} and \ref{fig:multimodal_kitti_samples} present multiple samples on the NYU and KITTI datasets which show that the model captures multimodality in sensor data and provides plausible samples in case of ambiguities.

\subsection{Novel View Synthesis}
\begin{figure}[h]
    \centering
    \includegraphics[width=.5\textwidth]{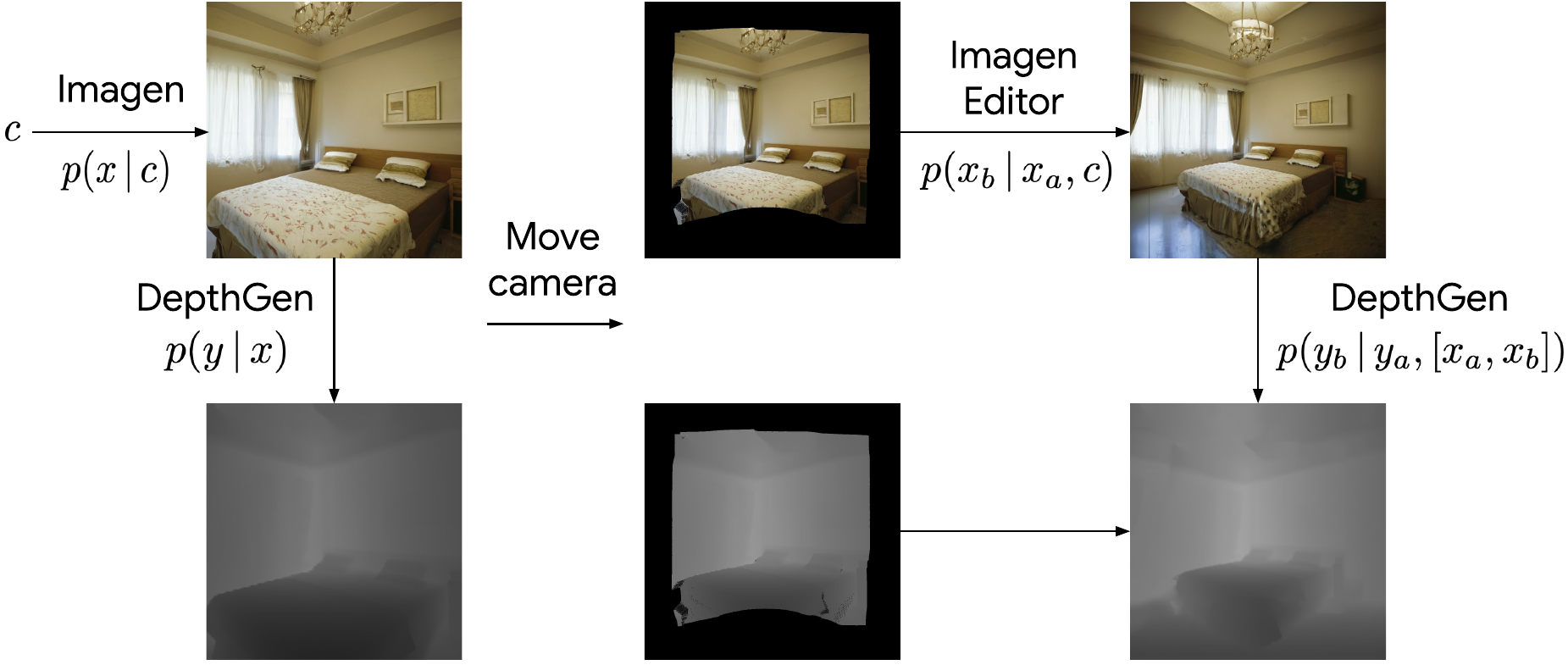}
    \vspace*{-0.4cm}
    \caption{Pipeline for iteratively generating a 3D scene conditioned on text $c=A\ bedroom.$
    See text for details.}
    \label{fig:text_to_3d}
\end{figure}

One advantage of diffusion models is the ease with which one can zero-shot impute one part of an image (or depth map) conditioned on the rest of the image (or depth map).
Here, we leverage this to build a limited but effective text-to-3D scene generation pipeline.
As depicted in Figure \ref{fig:text_to_3d}, we use the Imagen text-to-image model to generate an image, given text $c$, to which we apply \ours~(zero-shot) to sample a corresponding depth map. 
We then move the camera and, following  \cite{infinite_nature_2020}, render the RGBD point cloud from a new camera pose.
Of course this only provides RGB and depth values at a subset of pixels in the new frame since the fields of view are different.
Fortunately, the missing pixels are easily inferred 
using diffusion models (i.e., the Imagen Editor \cite{ImagenEditor} and \ours).

Let $x_a$ and $y_a$ be the RGB and depth values at pixel  locations rendered from the new camera pose respectively, and let $x_b$ and $y_b$ correspond to lines of sight not visible in the original frame.
We first infer the missing RGB values, i.e., $p(x_b | x_a, c)$, using
the uncropping/inpainting capability of the Imagen Editor.
We then use \ours~to impute the missing depth values, i.e., sampling from $p(y_b | y_a, [x_a, x_b])$.
There are several effective solutions to imputation with diffusion models, including the replacement method in \cite{song2021scorebased}, and the more sophisticated use of reconstruction guidance in \cite{ho-videodiffusion}.
For simplicity we use the replacement method to sample the unknown depths $y_b$ conditioned on existing depths $y_a$ and the image $x = [x_a, x_b]$.

\section{Conclusion}
\label{conclusion}
We propose a novel approach to monocular depth estimation using denoising diffusion models. We leverage self-supervised image-to-image pre-training, followed by subsequent training on supervised depth data to achieve SOTA results on challenging depth estimation benchmarks. We make several innovations that make it possible to effectively train diffusion models on imperfect training data that are commonplace for depth estimation.
We demonstrate the multimodal capability of diffusion models to represent depth uncertainty.
And we exploit the ease of imputation during iterative refinement in diffusion models  to show how \ours\ can be used for zero-shot depth completion. In combination with  text-to-image diffusion models, this enables a simple pipeline for novel view synthesis and text-to-3D.

\bibliography{main}
\bibliographystyle{icml2023}

\newpage

\appendix
\onecolumn
\section{Additional results}
\begin{table*}[!htbp]
    \centering
    \small
    \caption{Detailed ablation for handling incomplete depth on the NYU depth v2 dataset.}
    \vspace*{0.2cm}
    \begin{tabular}{lccccccc}
        \toprule
         & $\delta_1\uparrow$ & $\delta_2\uparrow$ & $\delta_3\uparrow$ & REL$\downarrow$ & RMS$\downarrow$ & $log_{10}\downarrow$ \\
        \midrule
        Baseline & 0.939 & 0.985 & 0.994 & 0.079 & 0.336 & 0.034  \\
        \stepunrolledshort~only & 0.944 & 0.986 & 0.995 & 0.076 & 0.325 & 0.033  \\
        Infilling only & 0.940 & 0.985 & 0.995 & 0.077 & 0.338 & 0.034  \\
        \stepunrolledshort~and infilling & 0.944 & 0.986 & 0.995 & 0.075 & 0.324 & 0.032 \\
        \bottomrule
    \end{tabular} 
    \label{tab::missing-depth-ablation-nyu}
\end{table*}
\begin{table*}[!htbp]
    \centering
    \small
    \caption{Detailed ablation for handling incomplete depth on the KITTI dataset.}
    \label{tab::missing-depth-ablation-kitti}
    \vspace*{0.2cm}
    \begin{tabular}{lcccccccc}
       \toprule
         & ~~~\textbf{$\delta_1$}$\uparrow$~~~ & ~~~\textbf{$\delta_2$}$\uparrow$~~~ & ~~~\textbf{$\delta_3$}$\uparrow$~~~ & ~REL~$\downarrow$~ & Sq-rel~$\downarrow$ & RMS~$\downarrow$ & RMS log~$\downarrow$ \\
        \midrule
        Baseline & 0.920 & 0.985 & 0.996 &  0.115 & 0.550 & 3.563 & 0.142 \\
        \stepunrolledshort~only & 0.938 & 0.986 & 0.996 &   0.075 & 0.468 & 3.437 & 0.117 \\
        Infilling only & 0.949 & 0.990 & 0.997 &  0.066 & 0.408 & 3.187 & 0.106  \\
        \stepunrolledshort~and infilling & 0.951&   0.991&   0.997&   0.064&   0.389&   3.104&   0.103 \\
    \bottomrule
\end{tabular}
\end{table*} 
\begin{table*}[!htbp]
    \centering
    \small
    \caption{Detailed ablation for pretraining on the NYU depth v2 dataset.}
    \vspace*{0.2cm}
    \begin{tabular}{lccccccc}
        \toprule
         & $\delta_1\uparrow$ & $\delta_2\uparrow$ & $\delta_3\uparrow$ & REL$\downarrow$ & RMS$\downarrow$ & $log_{10}\downarrow$ \\
        \midrule
        From scratch & 0.777 & 0.920 & 0.969 & 0.155 & 0.601 & 0.069  \\
        Unsupervised pretraining only & 0.877 & 0.964 & 0.988 & 0.116 & 0.450 & 0.049 \\
        Supervised pretraining only & 0.936 & 0.980 & 0.992 & 0.081 & 0.352 & 0.035  \\
        Unsupervised and supervised pretraining & 0.944 & 0.986 & 0.995 & 0.075 & 0.324 & 0.032 \\
        \bottomrule
    \end{tabular} 
    \label{tab::pretraining-ablation-nyu}
\end{table*}
\begin{table*}[!htbp]
    \centering
    \small
    \caption{Detailed ablation for pretraining on the KITTI dataset.}
    \vspace*{0.2cm}
    \begin{tabular}{lcccccccc}
       \toprule
         & ~~~\textbf{$\delta_1$}$\uparrow$~~~ & ~~~\textbf{$\delta_2$}$\uparrow$~~~ & ~~~\textbf{$\delta_3$}$\uparrow$~~~ & ~REL~$\downarrow$~ & Sq-rel~$\downarrow$ & RMS~$\downarrow$ & RMS log~$\downarrow$ \\
        \midrule
        From scratch & 0.880 &   0.959 &   0.986 &   0.103 &   0.801 &   4.643 &   0.172 \\
        Unsupervised pretraining only & 0.912 &   0.978 &   0.993 &   0.086 &   0.604 &   3.901 &   0.137 \\
        Supervised pretraining only & 0.936 & 0.985 & 0.996 &  0.075 & 0.470 & 3.448 & 0.118 \\
        Unsupervised and supervised pretraining & 0.951&   0.991&   0.997&   0.064&   0.389&   3.104&   0.103 \\
    \bottomrule
\end{tabular}
\label{tab::pretraining-ablation-kitti}
\end{table*} 
\begin{table*}[!htbp]
    \centering
    \small
    \caption{Detailed ablation for the choice of loss function on the NYU depth v2 dataset.}
    \vspace*{0.2cm}
    \begin{tabular}{lccccccc}
        \toprule
         & $\delta_1\uparrow$ & $\delta_2\uparrow$ & $\delta_3\uparrow$ & REL$\downarrow$ & RMS$\downarrow$ & $log_{10}\downarrow$ \\
        \midrule
        $L_2$ & 0.932 & 0.981 & 0.994 & 0.085 & 0.349 & 0.037  \\
        $L_1$ & 0.944 & 0.986 & 0.995 & 0.075 & 0.324 & 0.032 \\
        \bottomrule
    \end{tabular} 
    \label{tab::loss-ablation-nyu}
\end{table*}
\begin{table*}[!htbp]
    \centering
    \small
    \caption{Detailed ablation for the choice of loss function on the KITTI dataset.}
    \label{tab::loss-ablation-kitti}
    \vspace*{0.2cm}
    \begin{tabular}{lcccccccc}
       \toprule
         & ~~~\textbf{$\delta_1$}$\uparrow$~~~ & ~~~\textbf{$\delta_2$}$\uparrow$~~~ & ~~~\textbf{$\delta_3$}$\uparrow$~~~ & ~REL~$\downarrow$~ & Sq-rel~$\downarrow$ & RMS~$\downarrow$ & RMS log~$\downarrow$ \\
        \midrule
        $L_2$ & 0.943 &   0.990 &   0.997 &   0.071 &   0.416 &   3.221 &   0.110   \\
        $L_1$ & 0.951&   0.991&   0.997&   0.064&   0.389&   3.104&   0.103 \\
    \bottomrule
\end{tabular}
\end{table*} 

\clearpage
\def\imwidth{0.095}
\def\imwidthlast{0.19}
\def\lastcoloffset{1.56cm}
\def\textpadding{3pt}
\def\ncols{9}

\begin{figure*}[t]
\centering
\renewcommand{\arraystretch}{0.2}
\addtolength{\tabcolsep}{-5pt}
\begin{tabular}{ccccccccc}
    \includegraphics[width=\imwidth\linewidth]{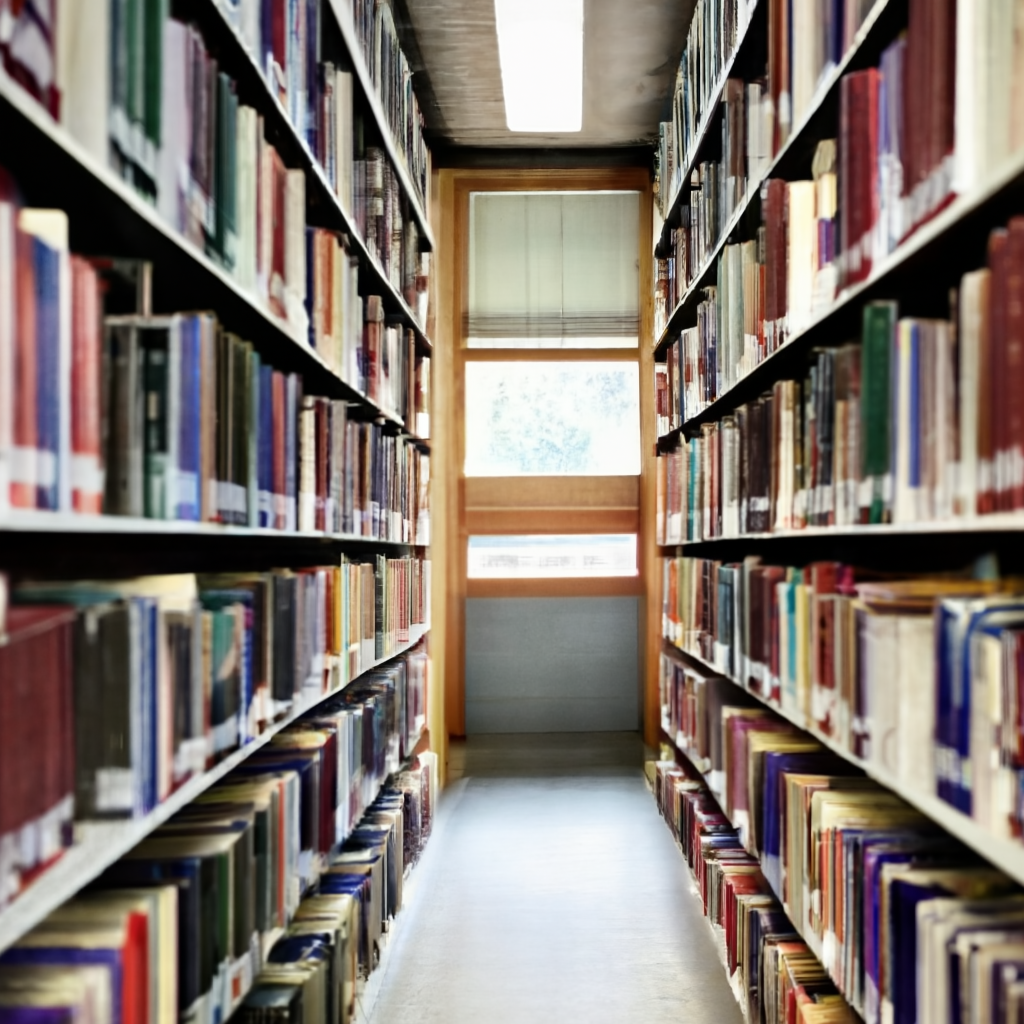} & \includegraphics[width=\imwidth\linewidth]{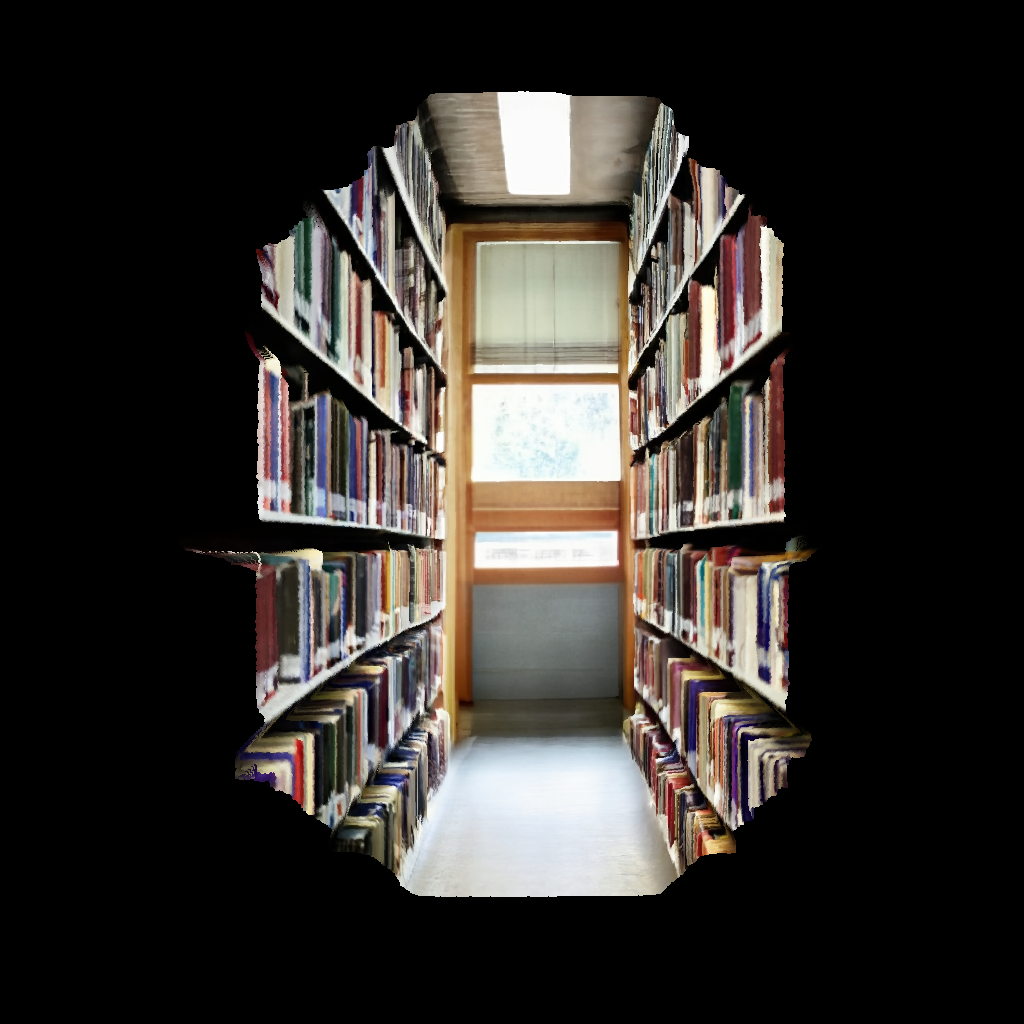} &\includegraphics[width=\imwidth\linewidth]{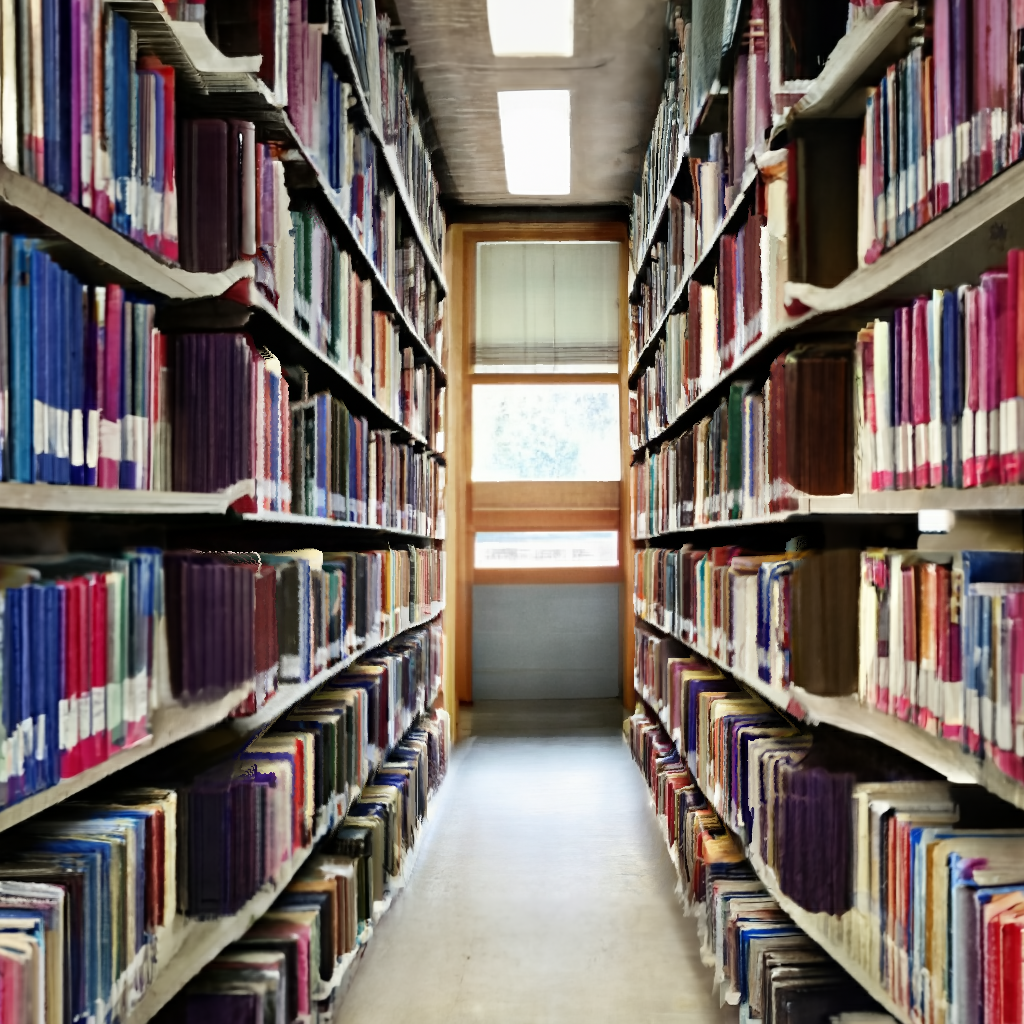} &\includegraphics[width=\imwidth\linewidth]{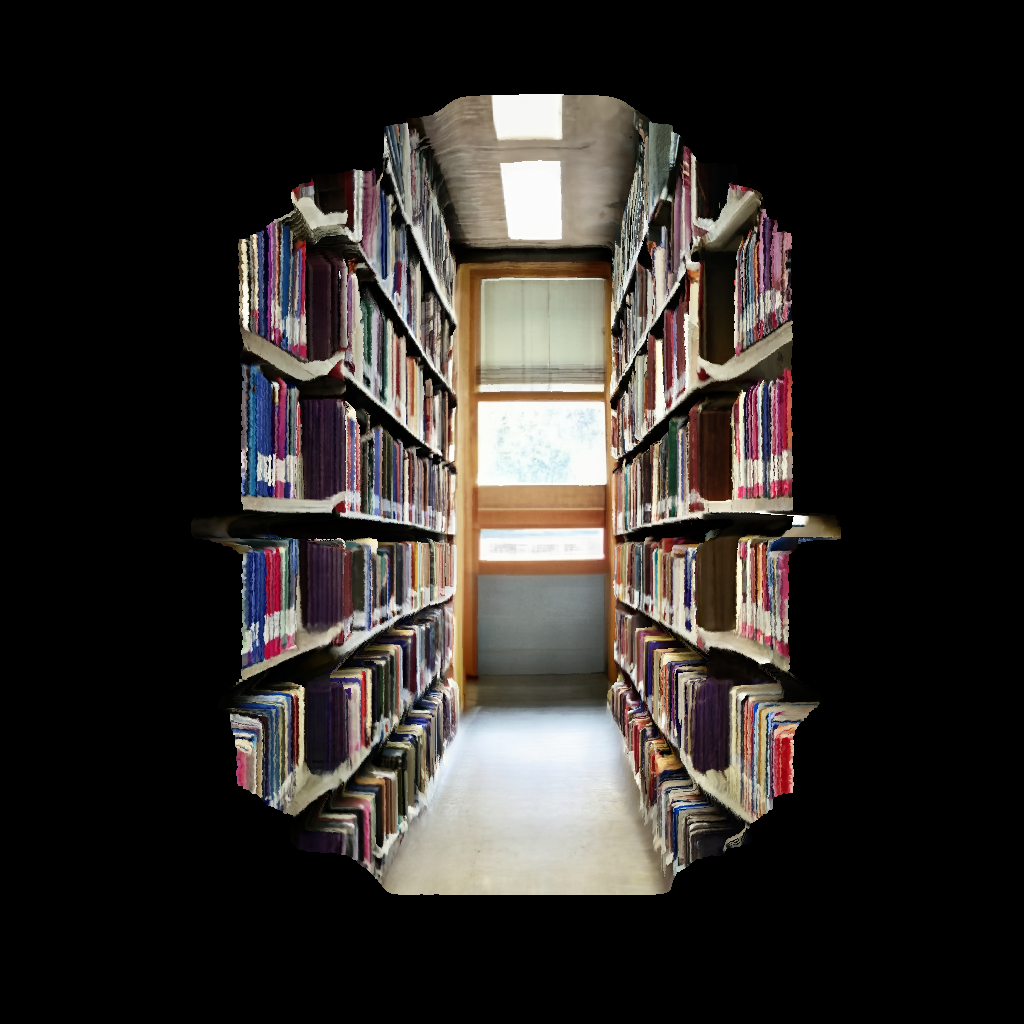} &\includegraphics[width=\imwidth\linewidth]{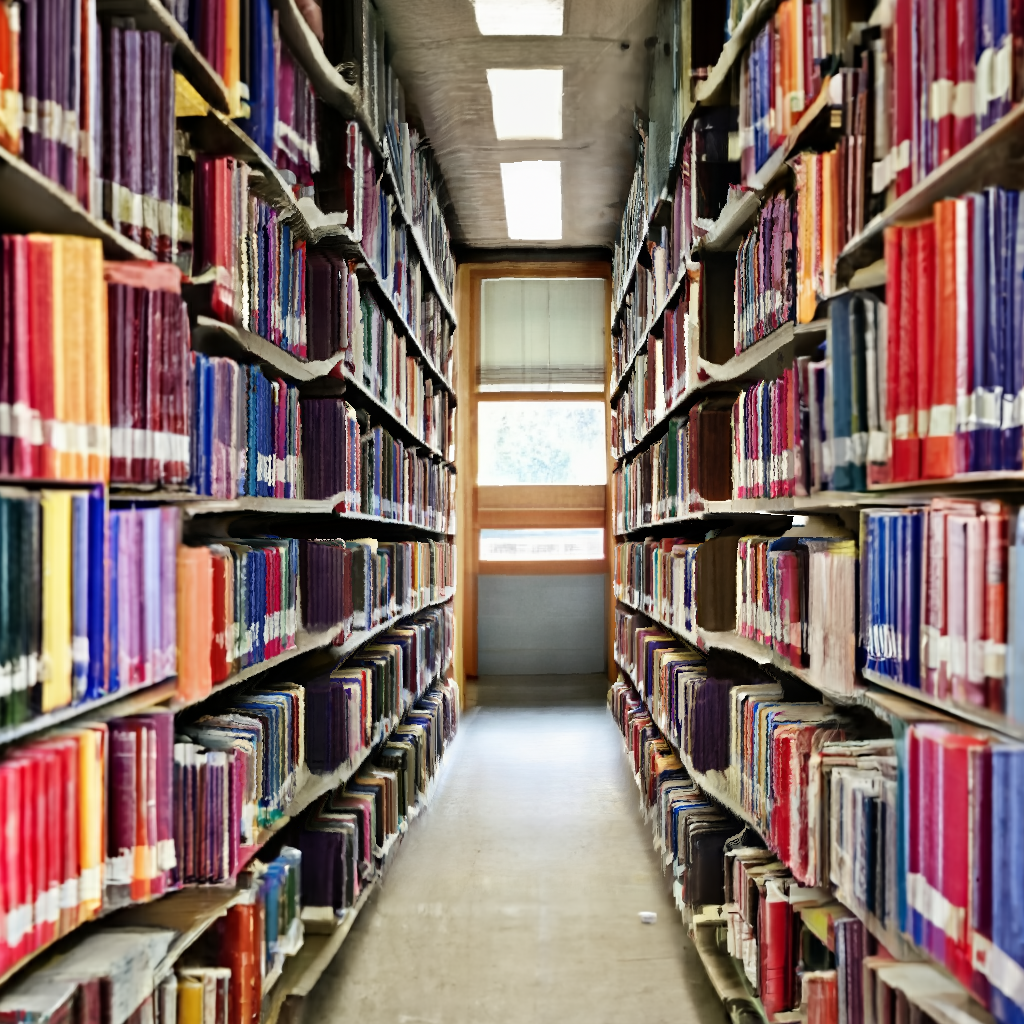} & \includegraphics[width=\imwidth\linewidth]{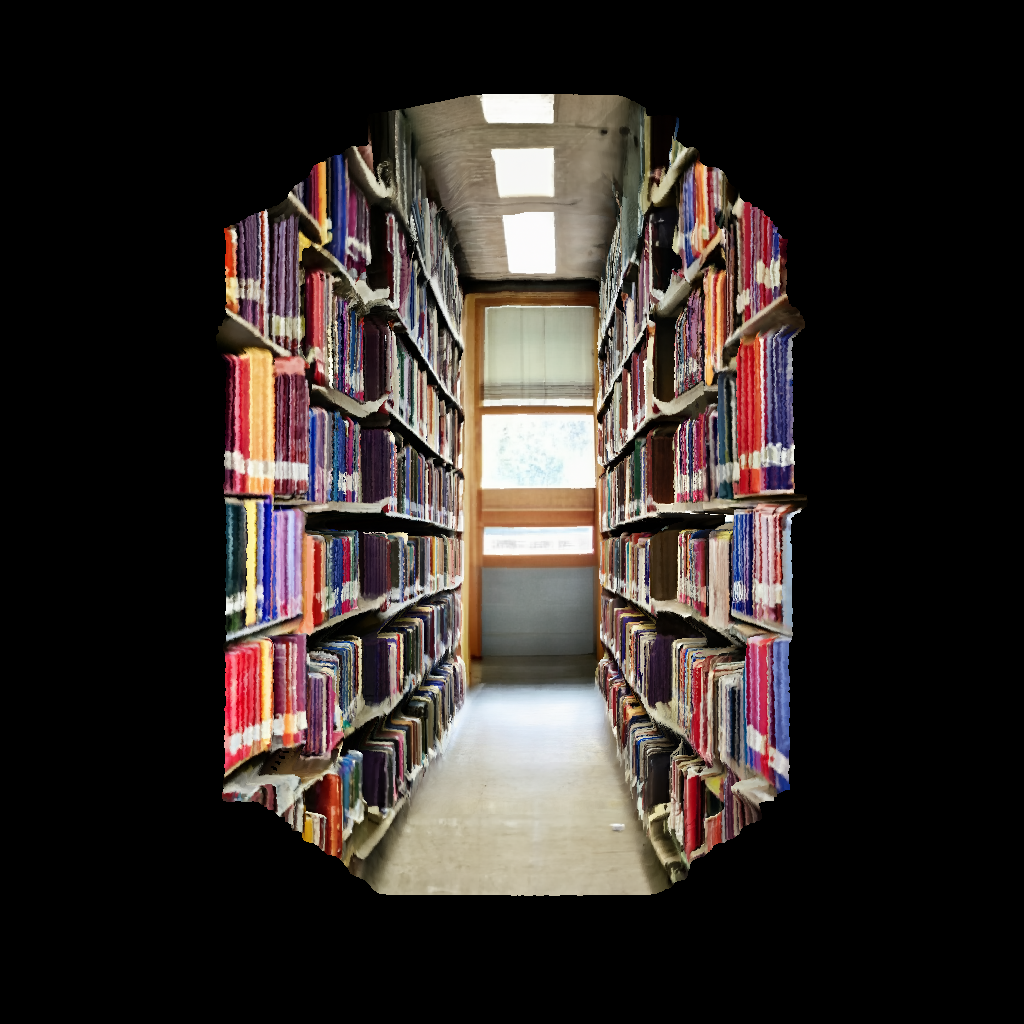} &\includegraphics[width=\imwidth\linewidth]{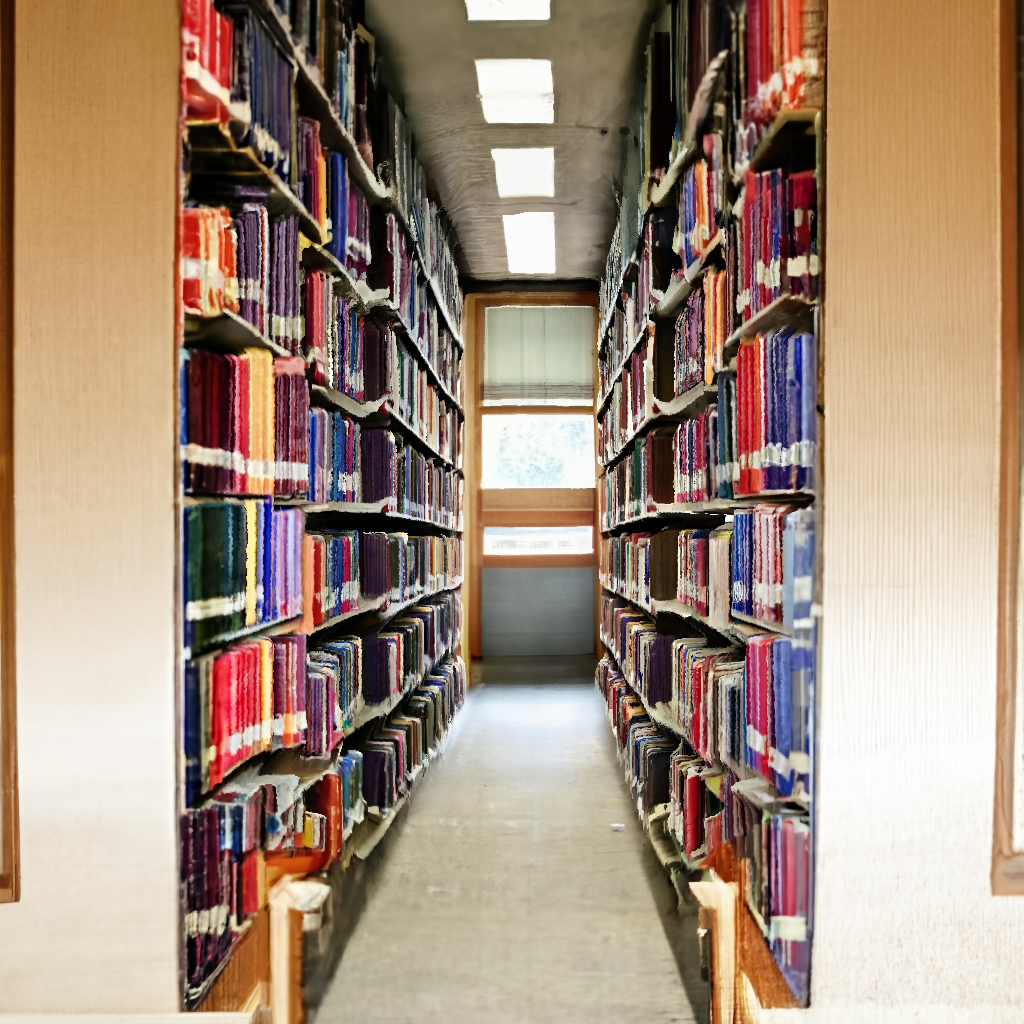} & \multicolumn{2}{c}{\multirow{2}{*}[\lastcoloffset]{\includegraphics[width=\imwidthlast\linewidth]{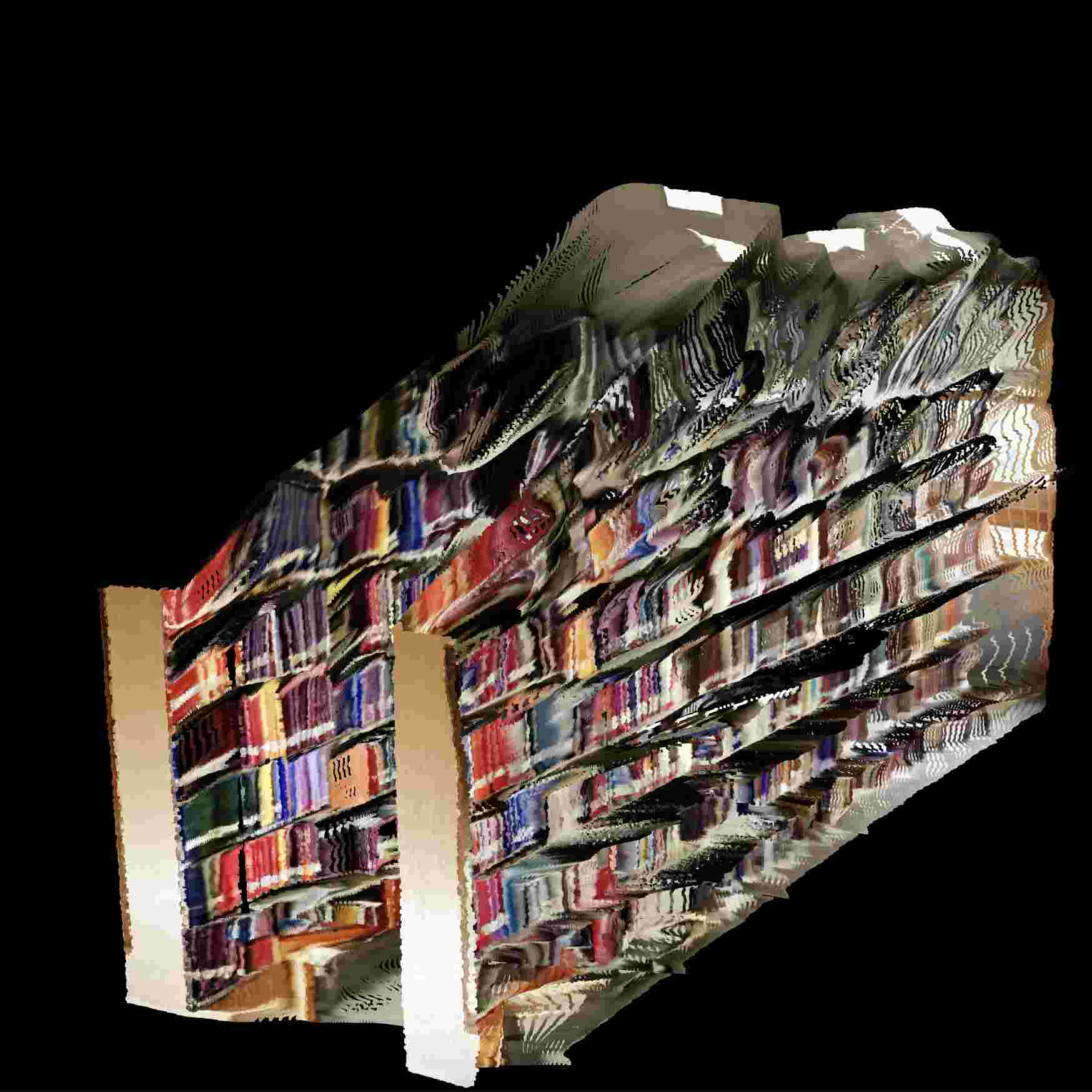}}} \\
    \includegraphics[width=\imwidth\linewidth]{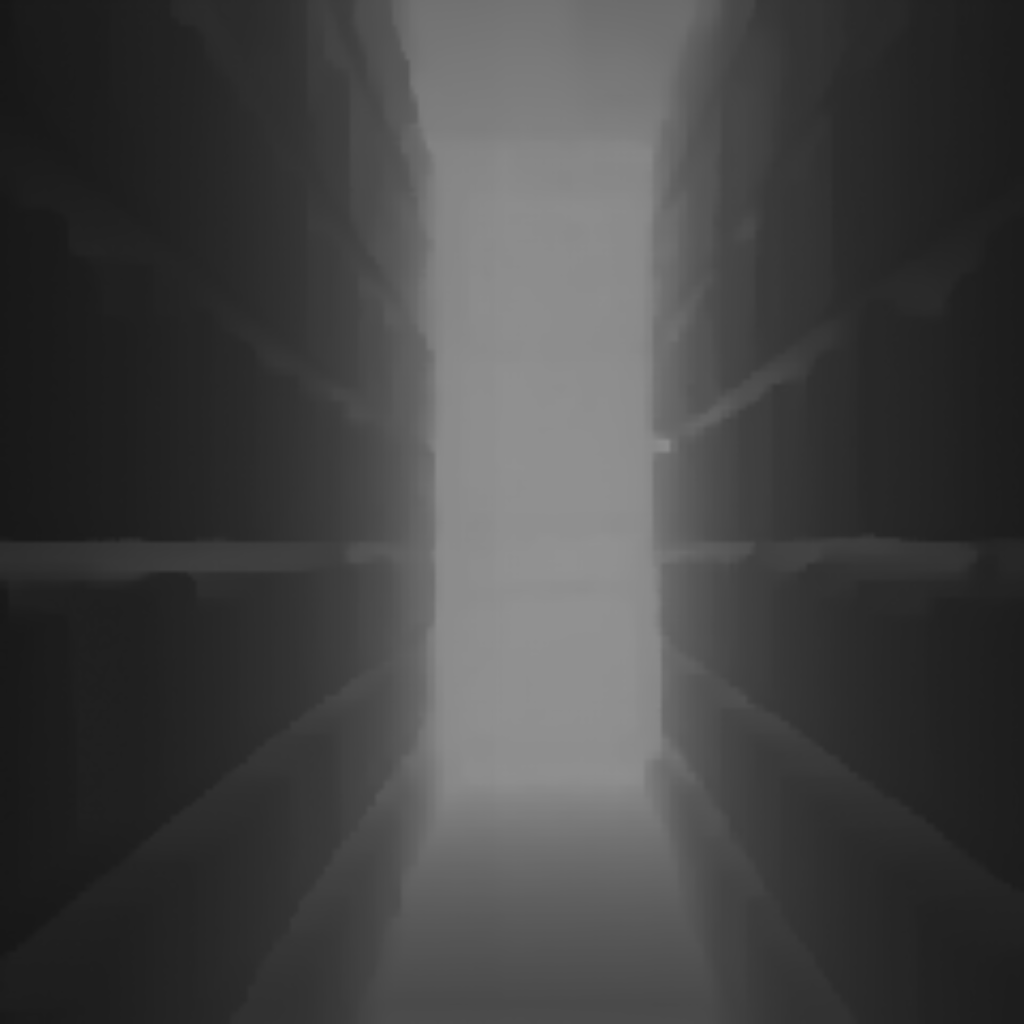} & \includegraphics[width=\imwidth\linewidth]{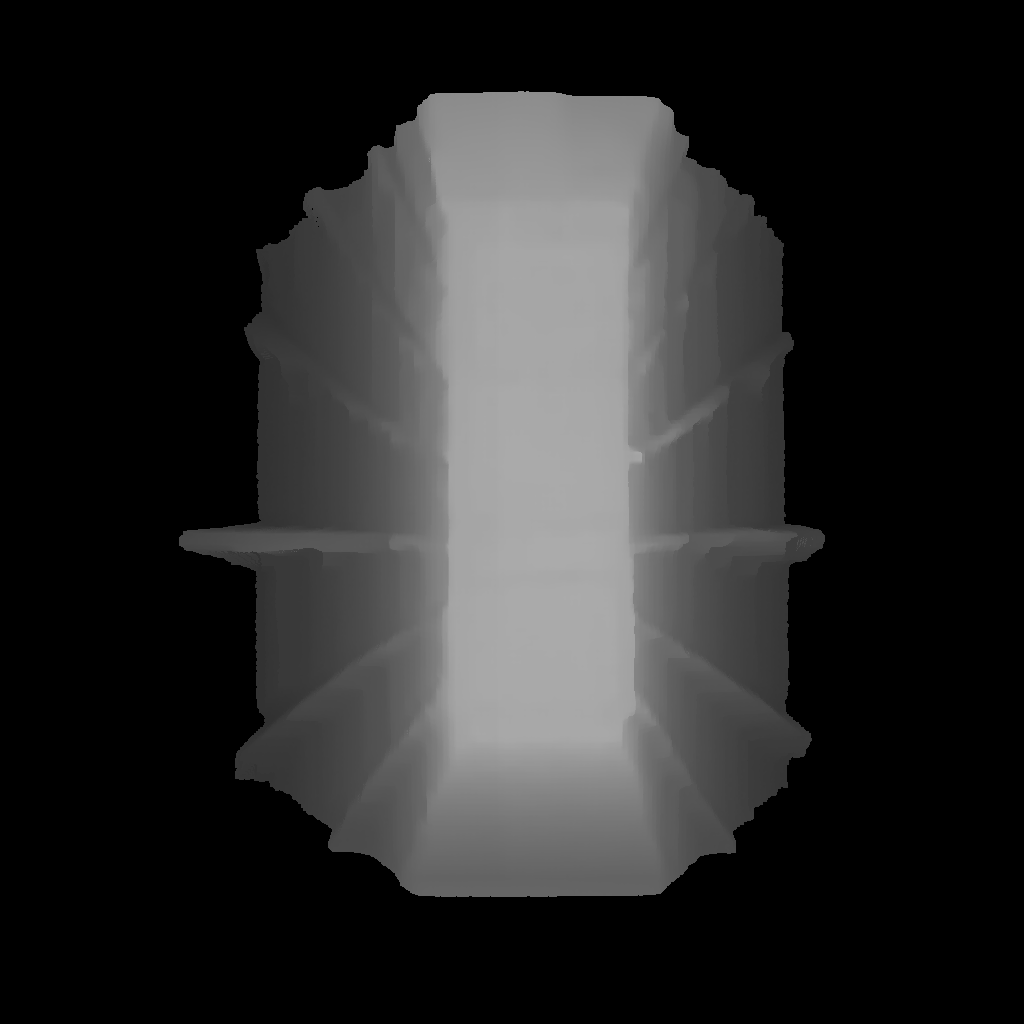} &\includegraphics[width=\imwidth\linewidth]{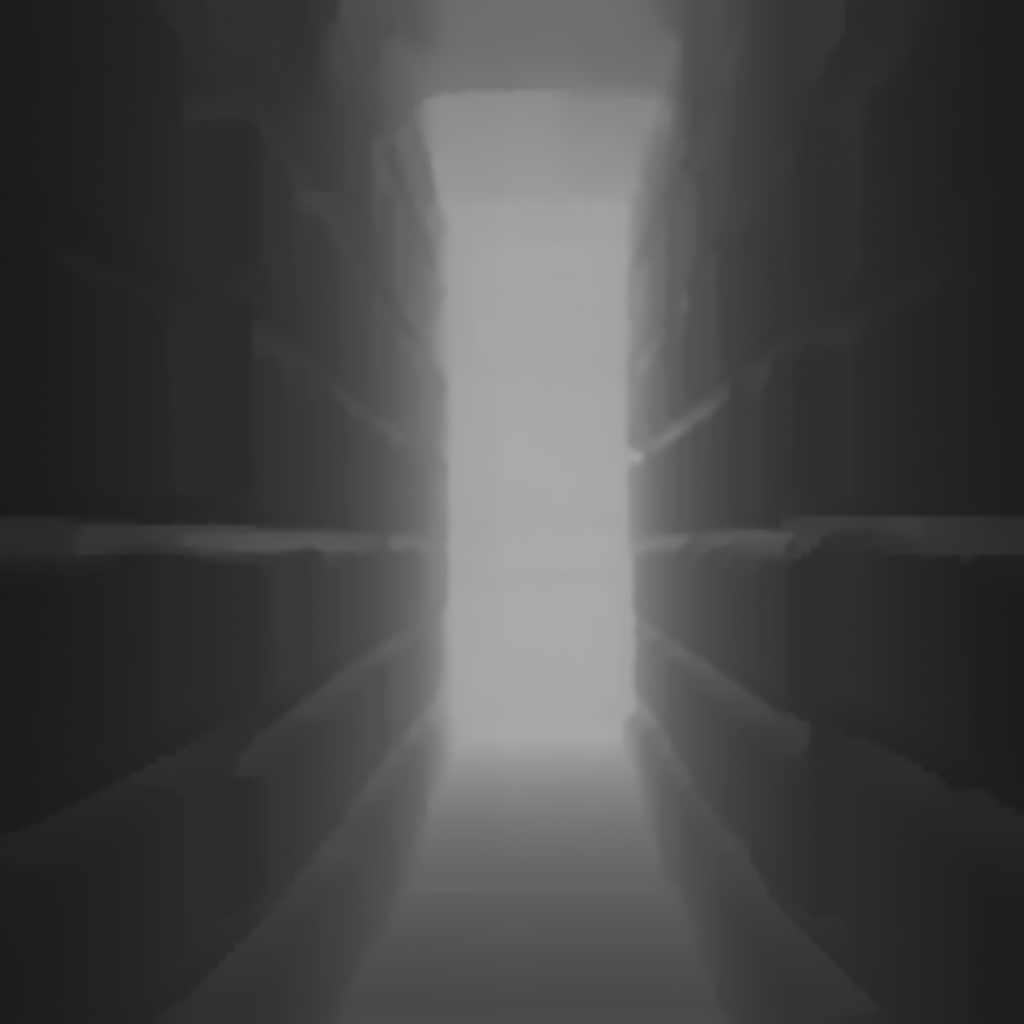} &  \includegraphics[width=\imwidth\linewidth]{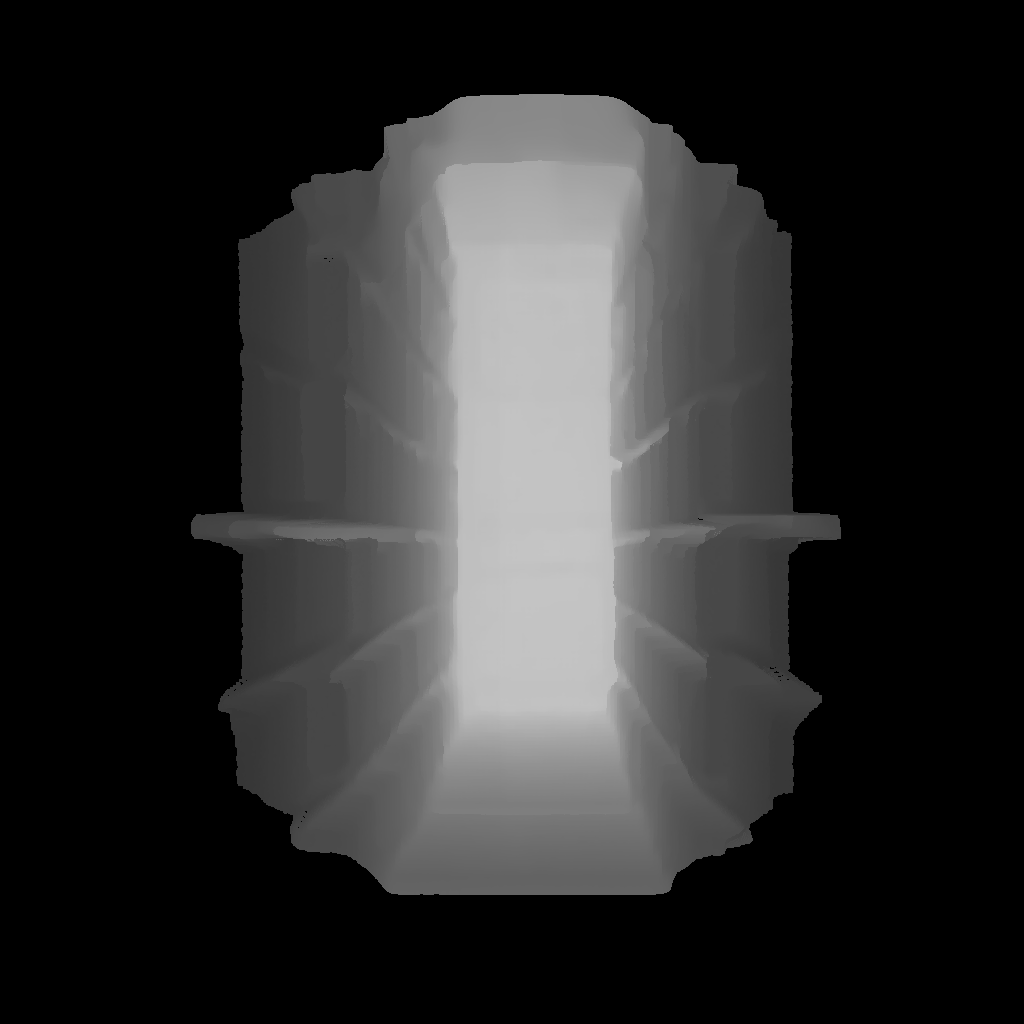} &\includegraphics[width=\imwidth\linewidth]{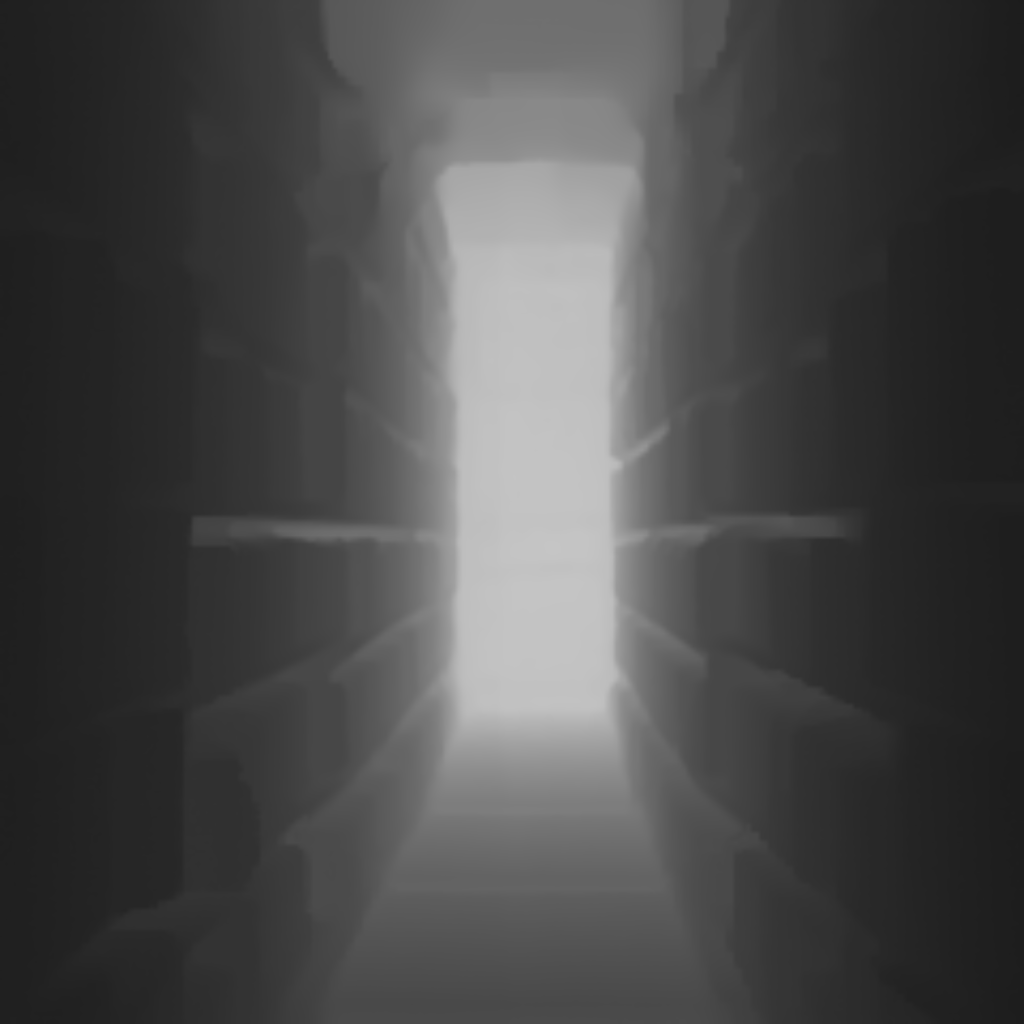} &\includegraphics[width=\imwidth\linewidth]{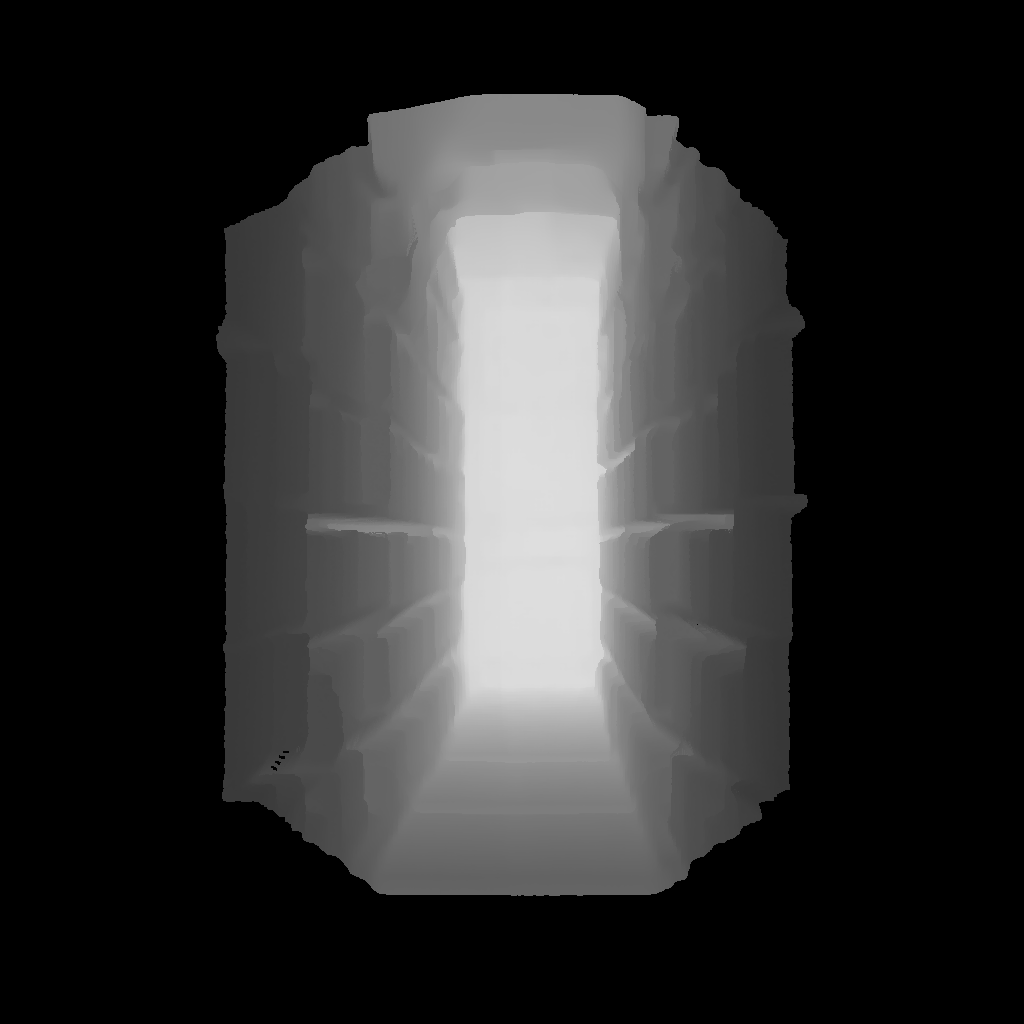} &\includegraphics[width=\imwidth\linewidth]{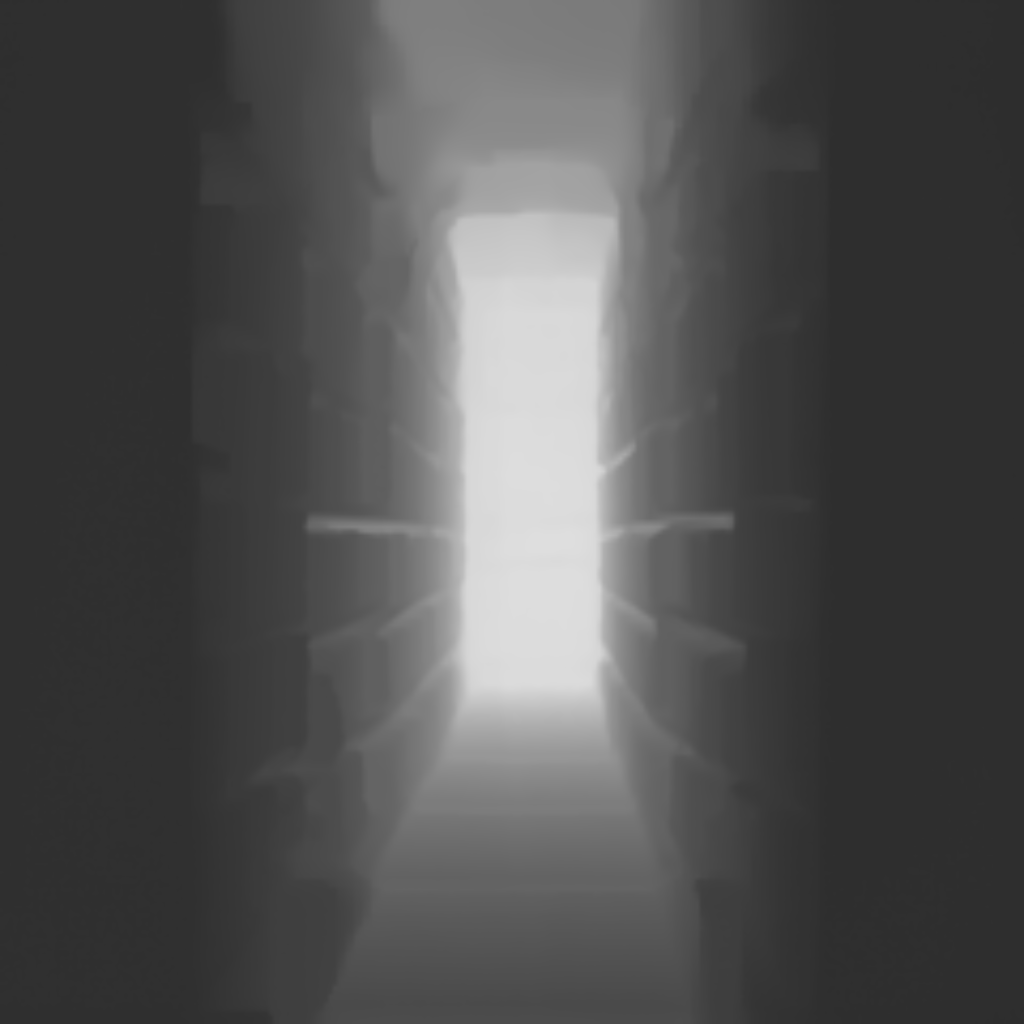} & \multicolumn{2}{c}{} \\
    \\[\textpadding]\multicolumn{\ncols}{c}{Prompt: A library.}\\\\[\textpadding]

    \includegraphics[width=\imwidth\linewidth]{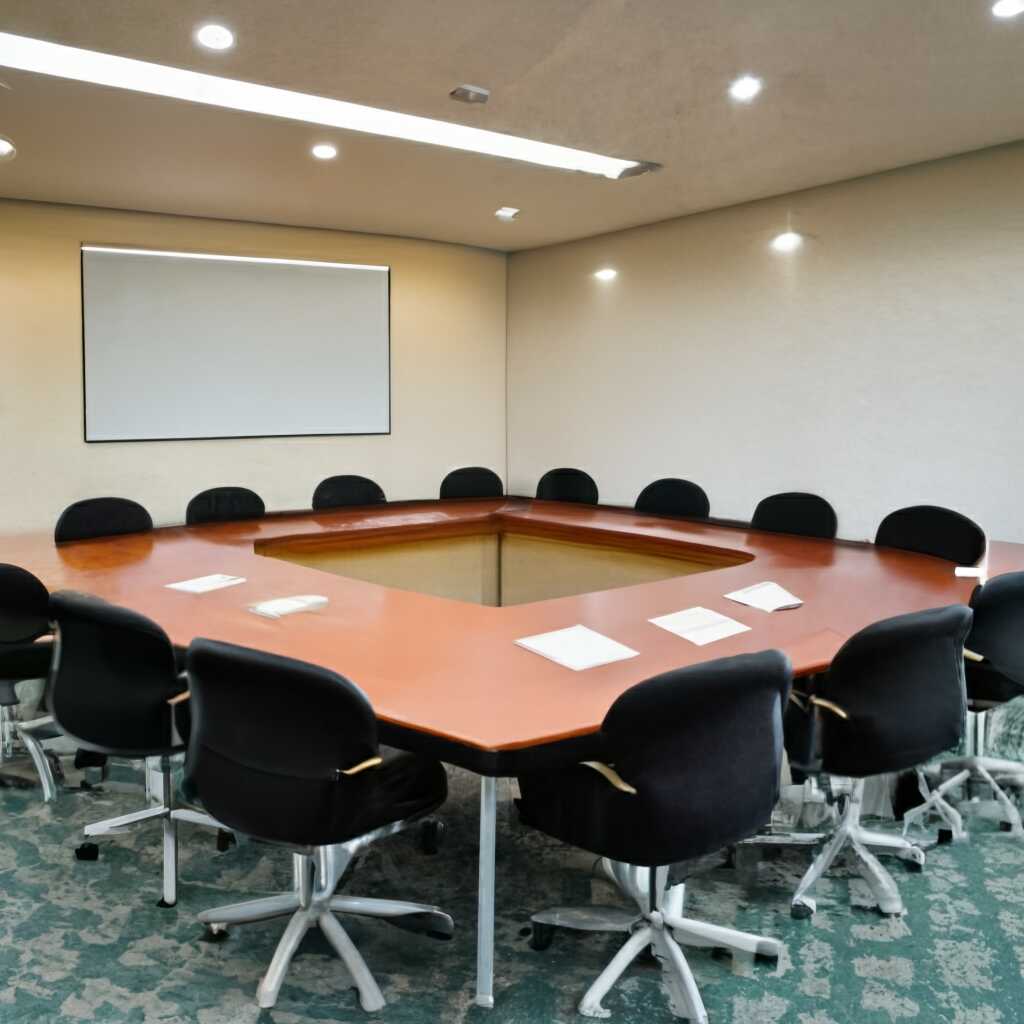} & \includegraphics[width=\imwidth\linewidth]{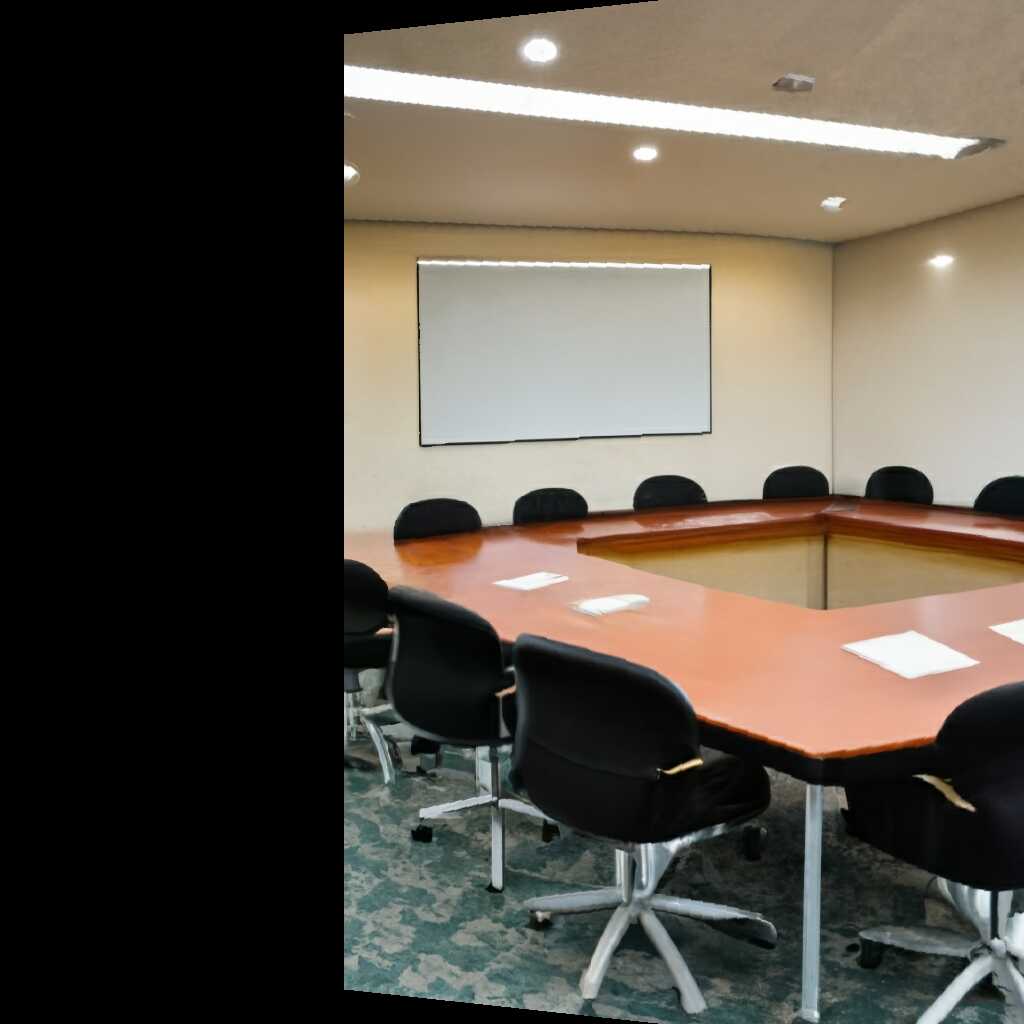} &\includegraphics[width=\imwidth\linewidth]{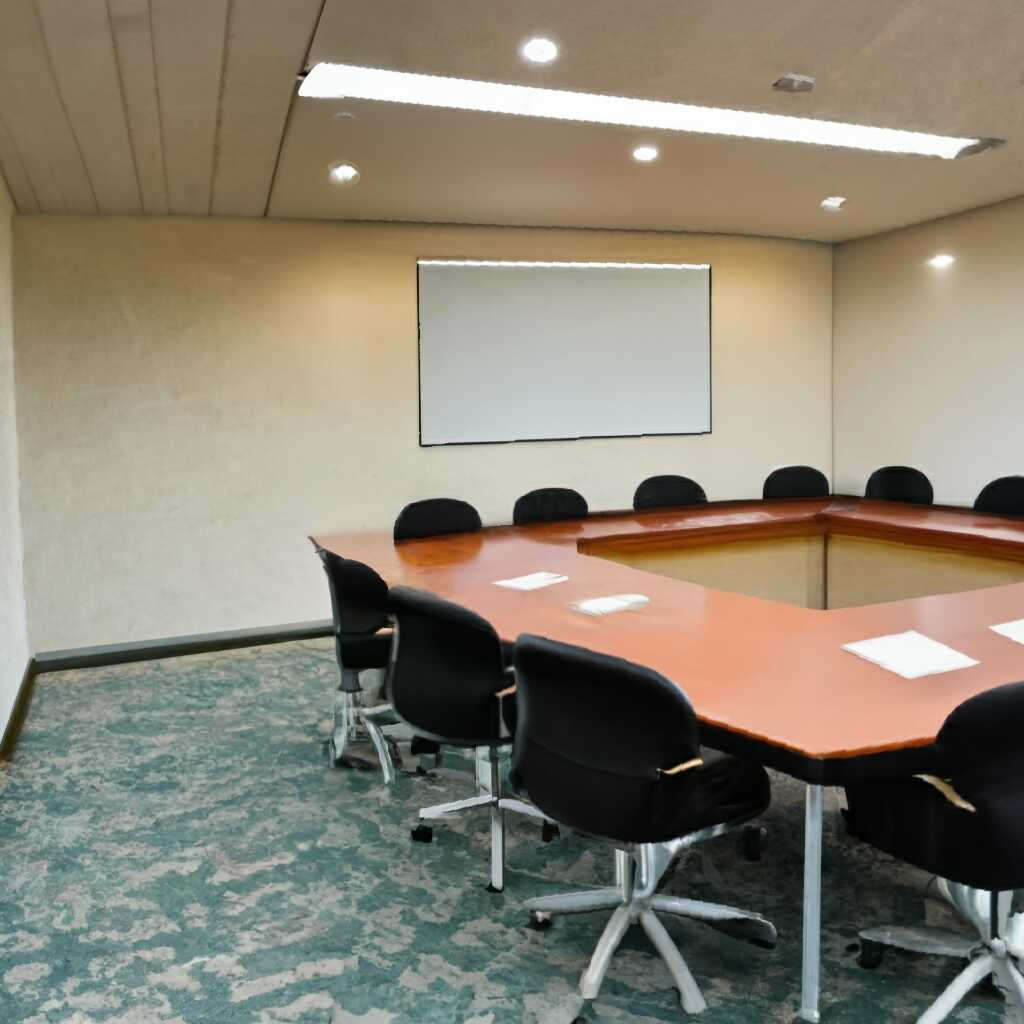} &\includegraphics[width=\imwidth\linewidth]{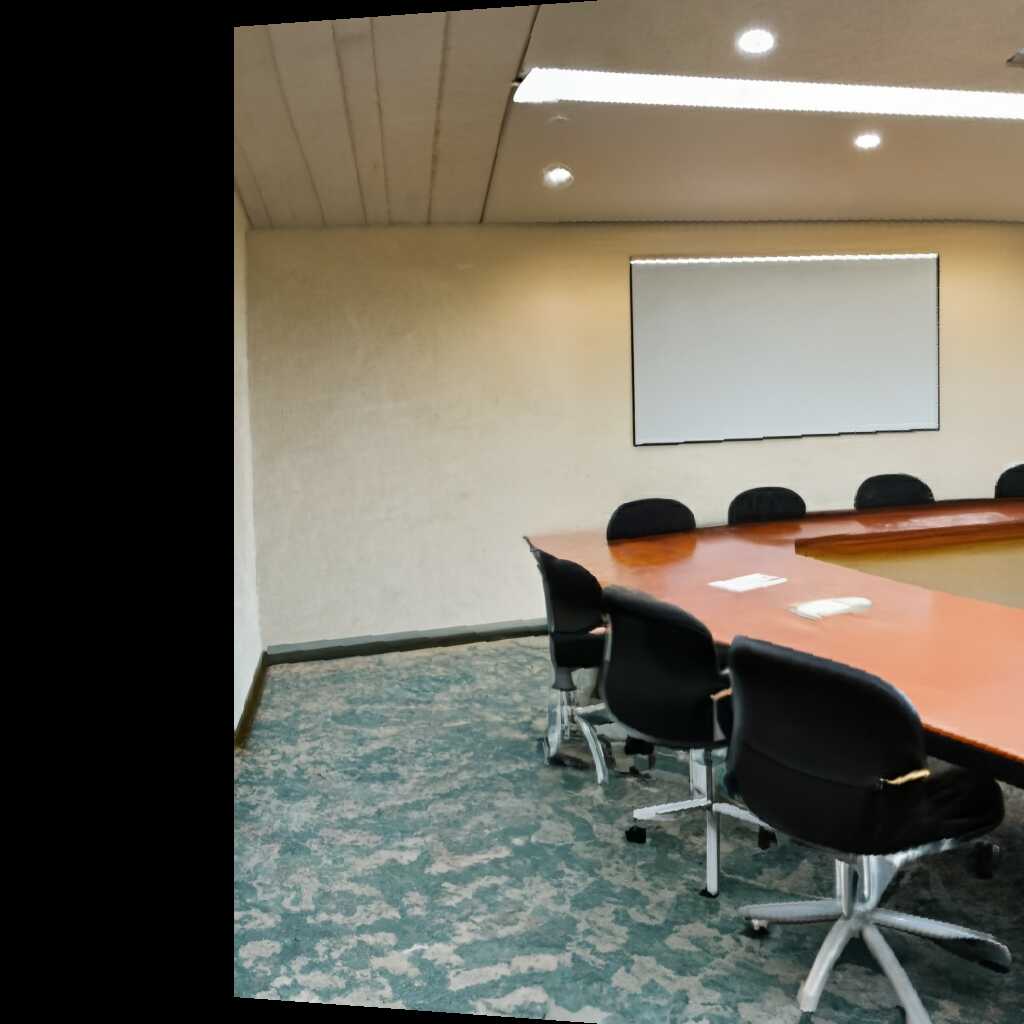} &\includegraphics[width=\imwidth\linewidth]{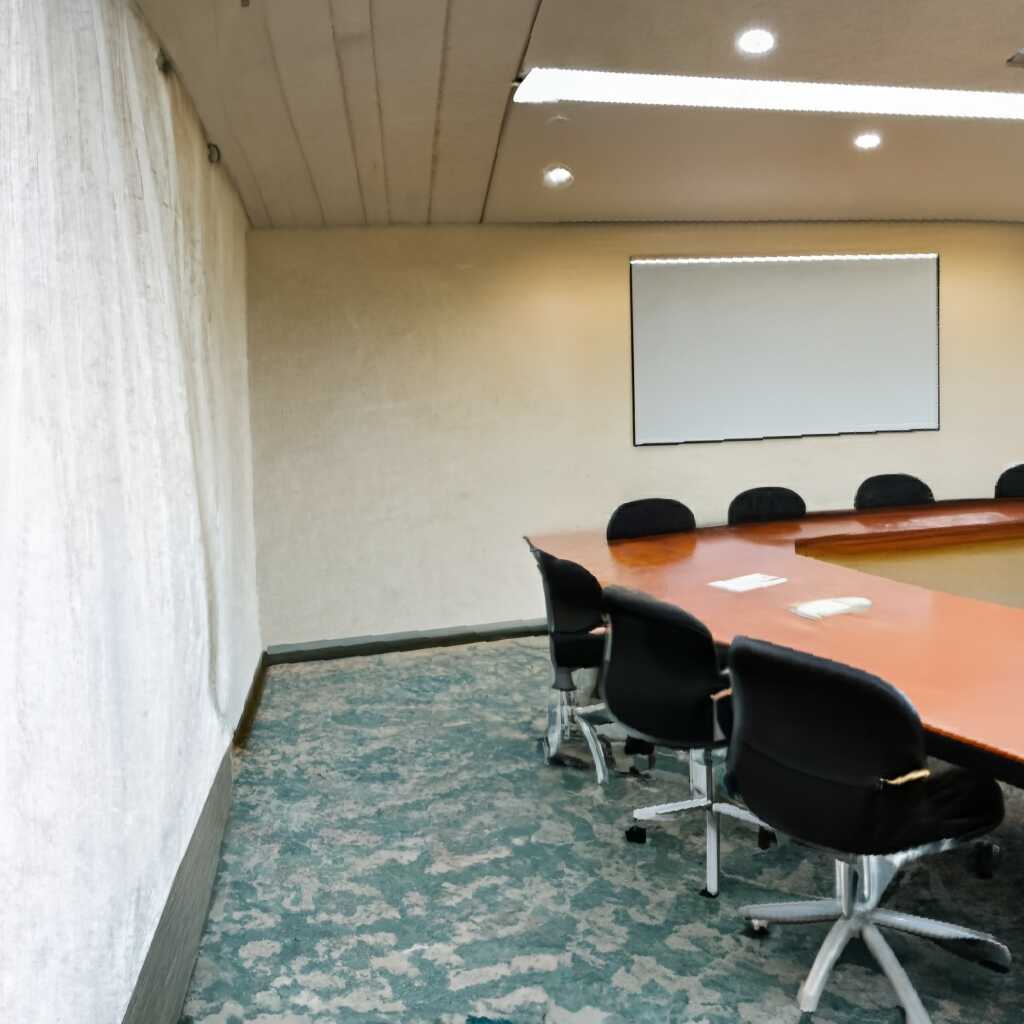} & \includegraphics[width=\imwidth\linewidth]{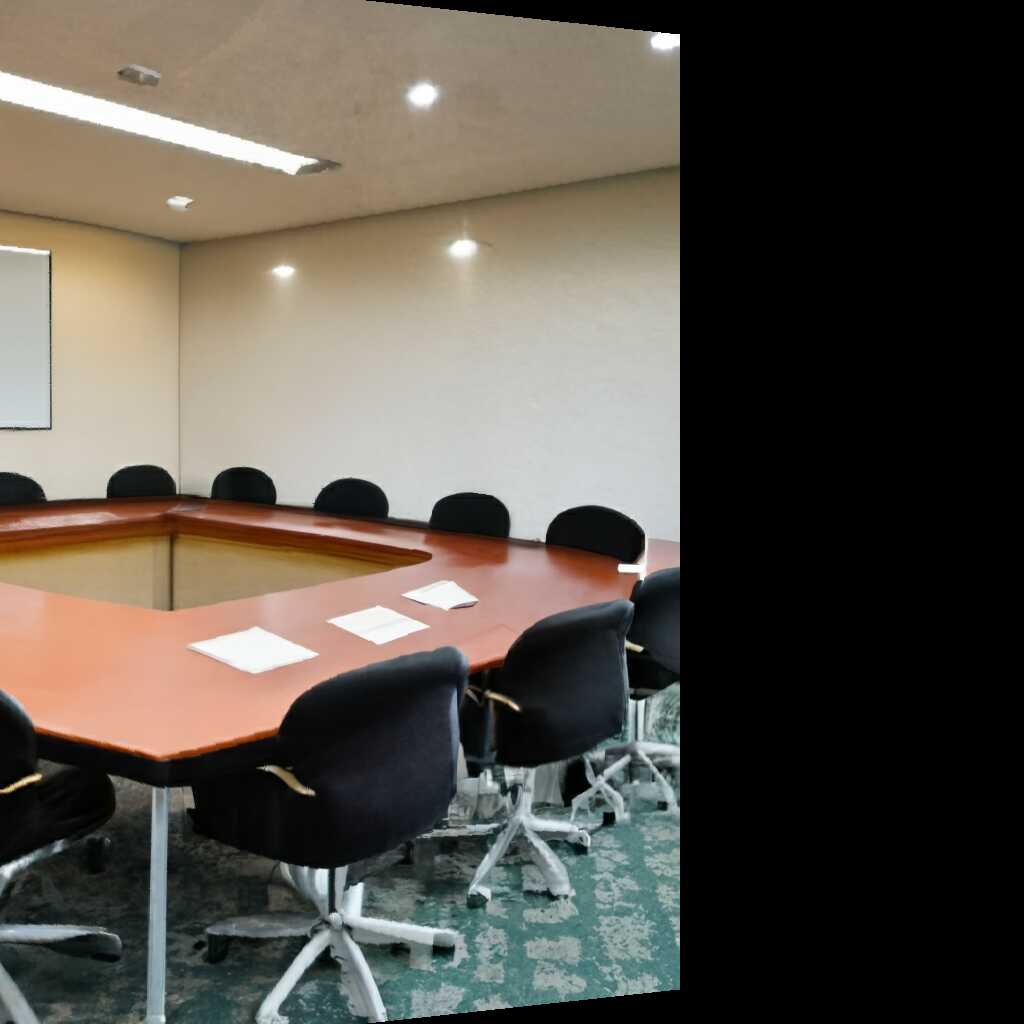} &\includegraphics[width=\imwidth\linewidth]{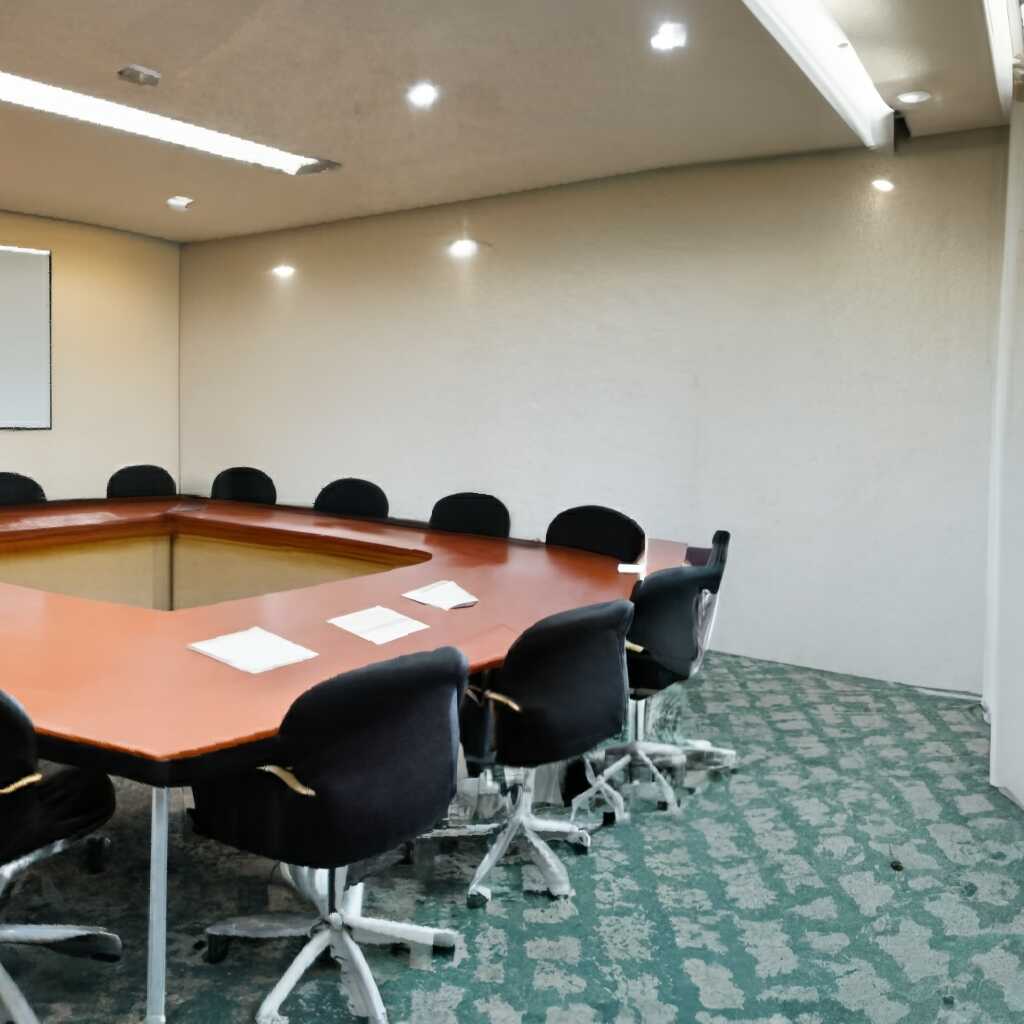} & \multicolumn{2}{c}{\multirow{2}{*}[\lastcoloffset]{\includegraphics[width=\imwidthlast\linewidth]{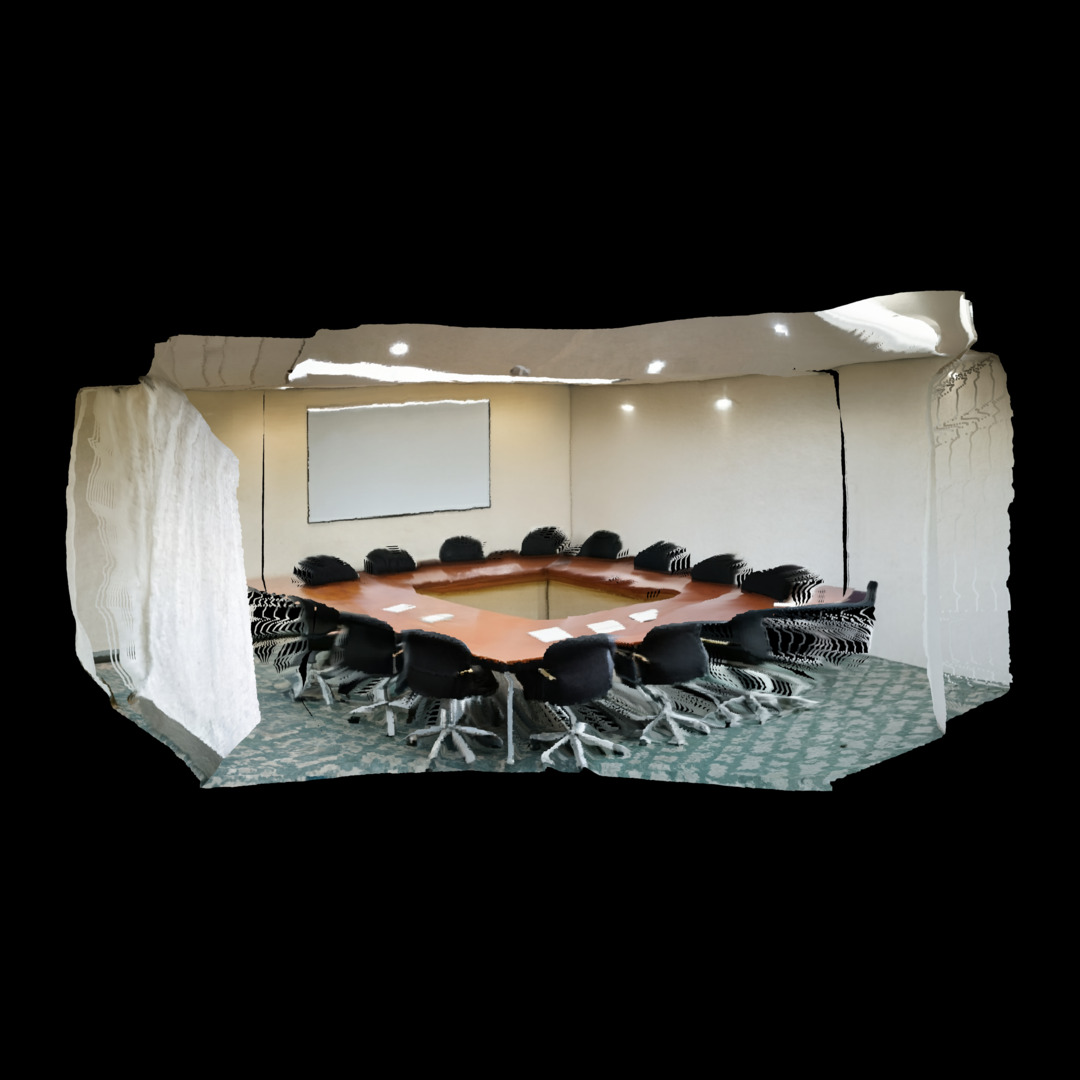}}} \\
    \includegraphics[width=\imwidth\linewidth]{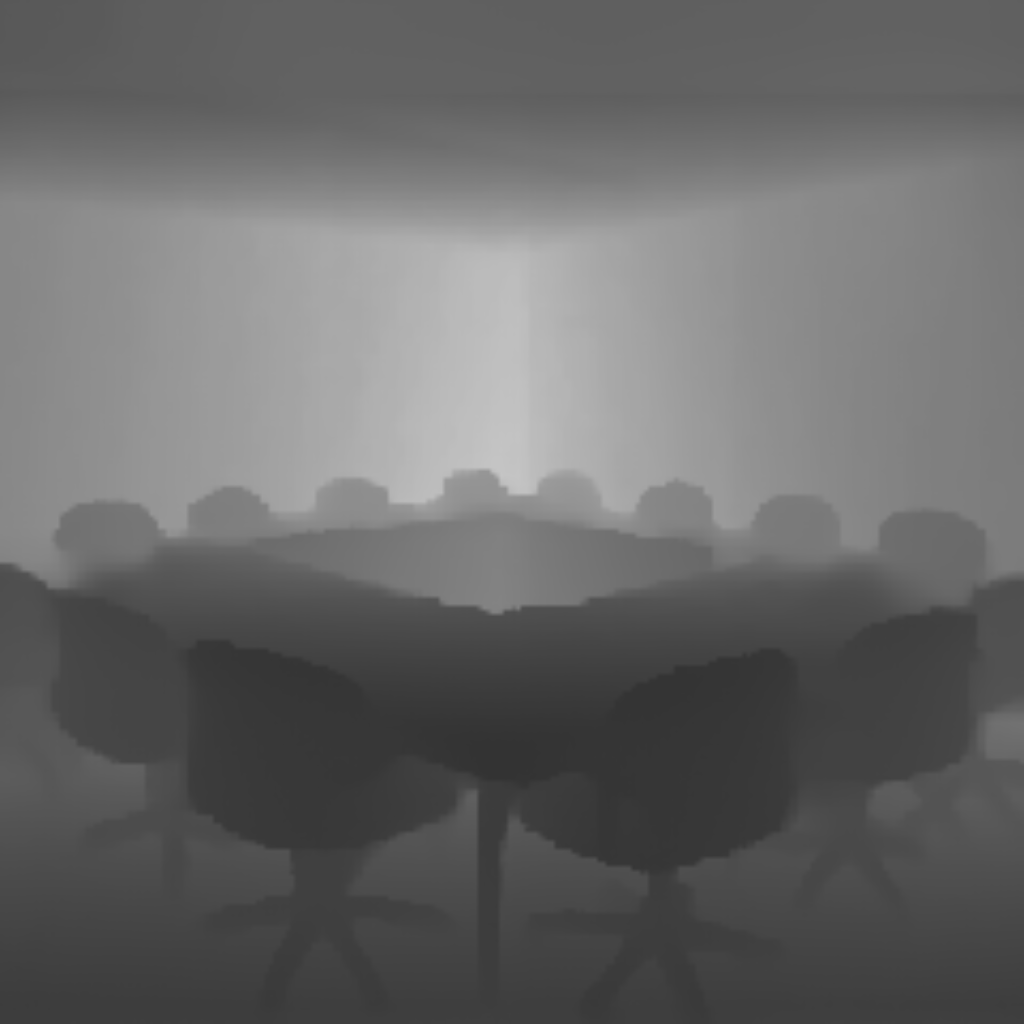} & \includegraphics[width=\imwidth\linewidth]{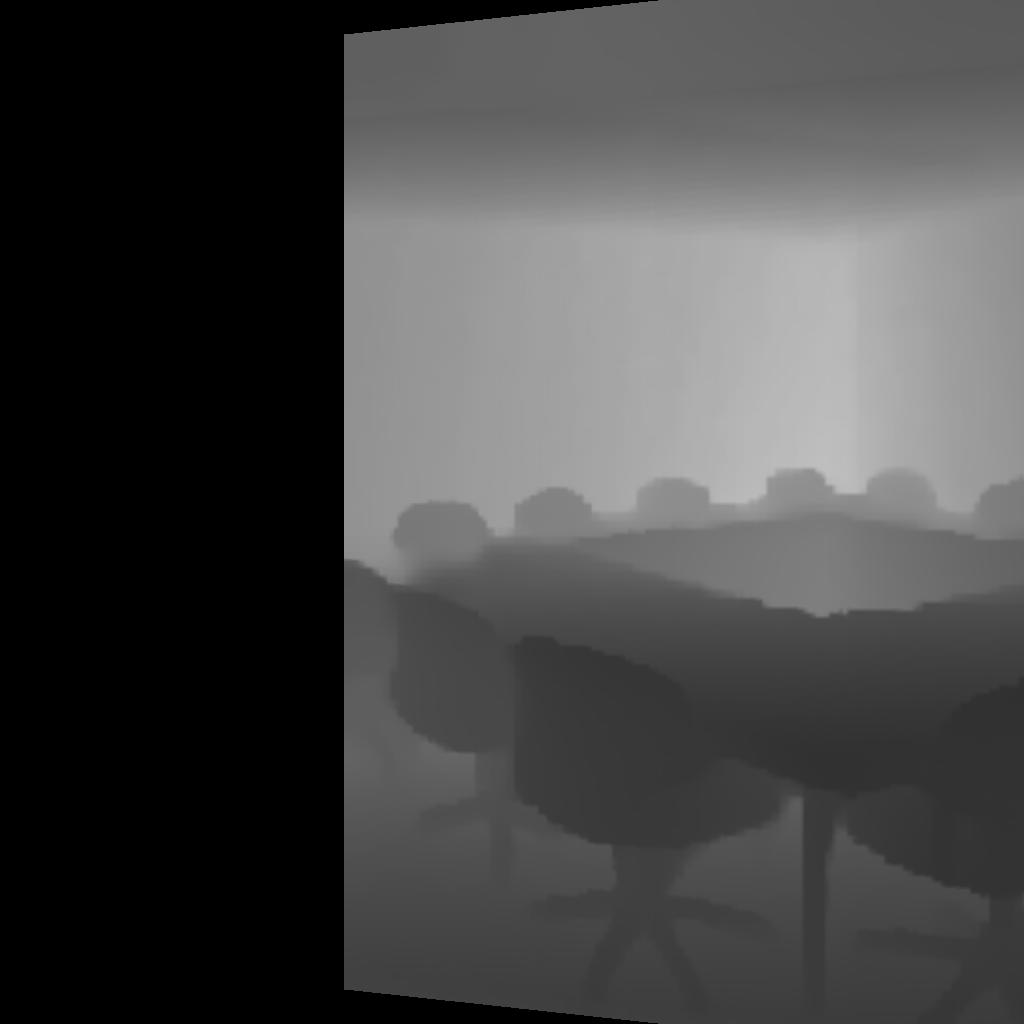} &\includegraphics[width=\imwidth\linewidth]{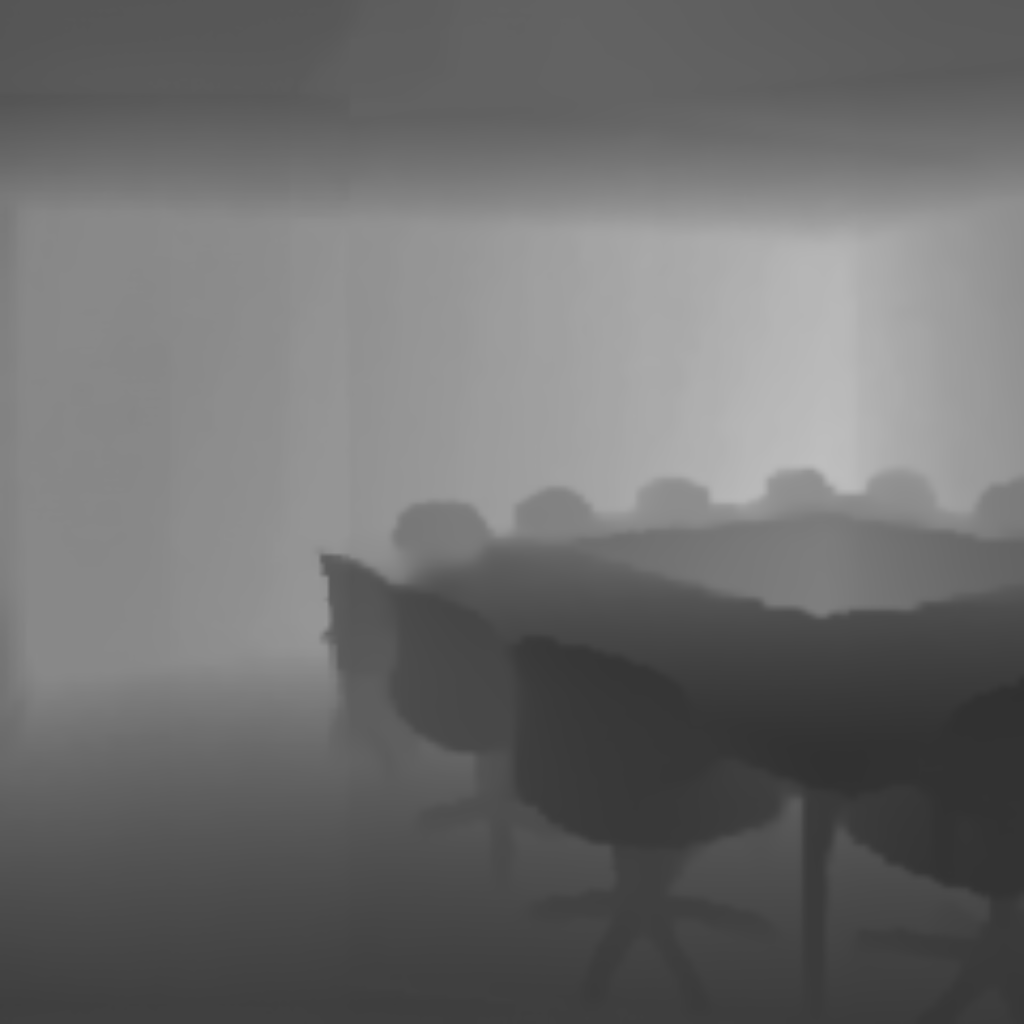} &  \includegraphics[width=\imwidth\linewidth]{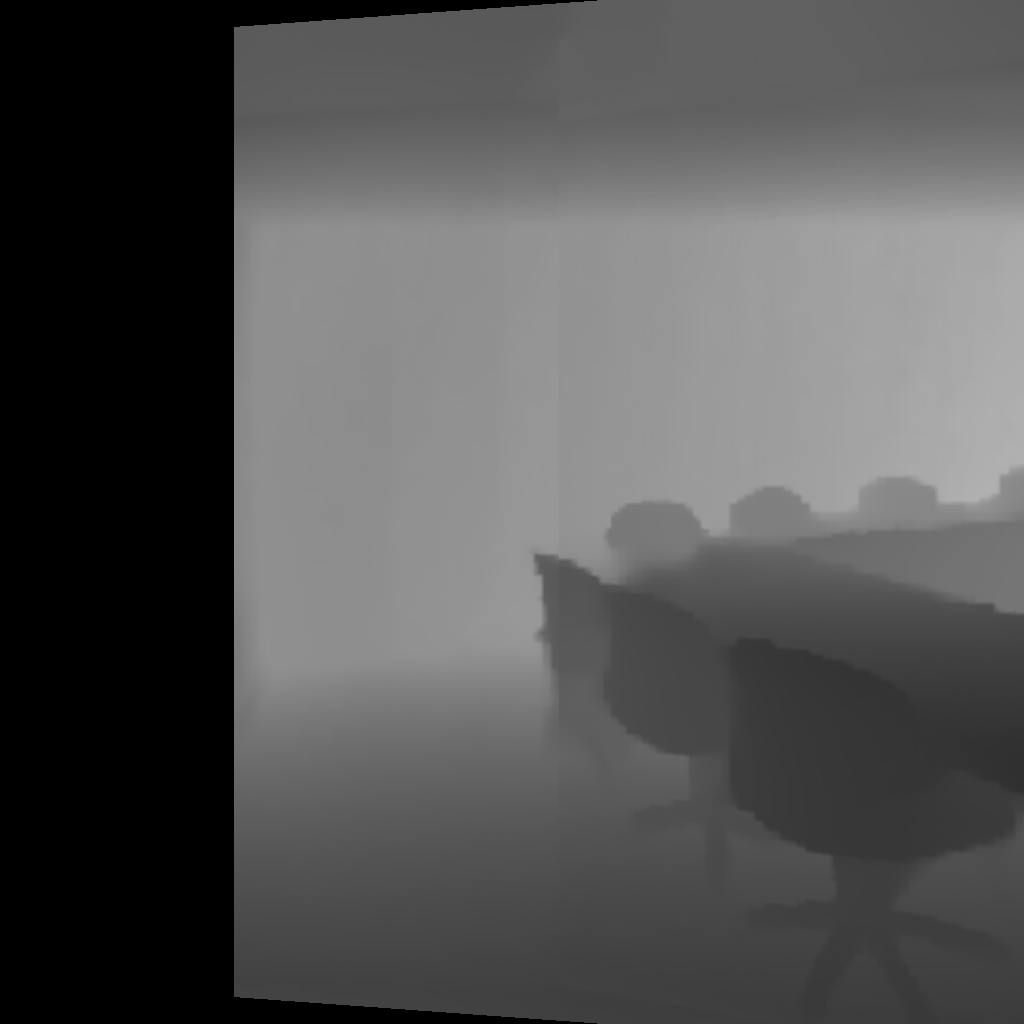} &\includegraphics[width=\imwidth\linewidth]{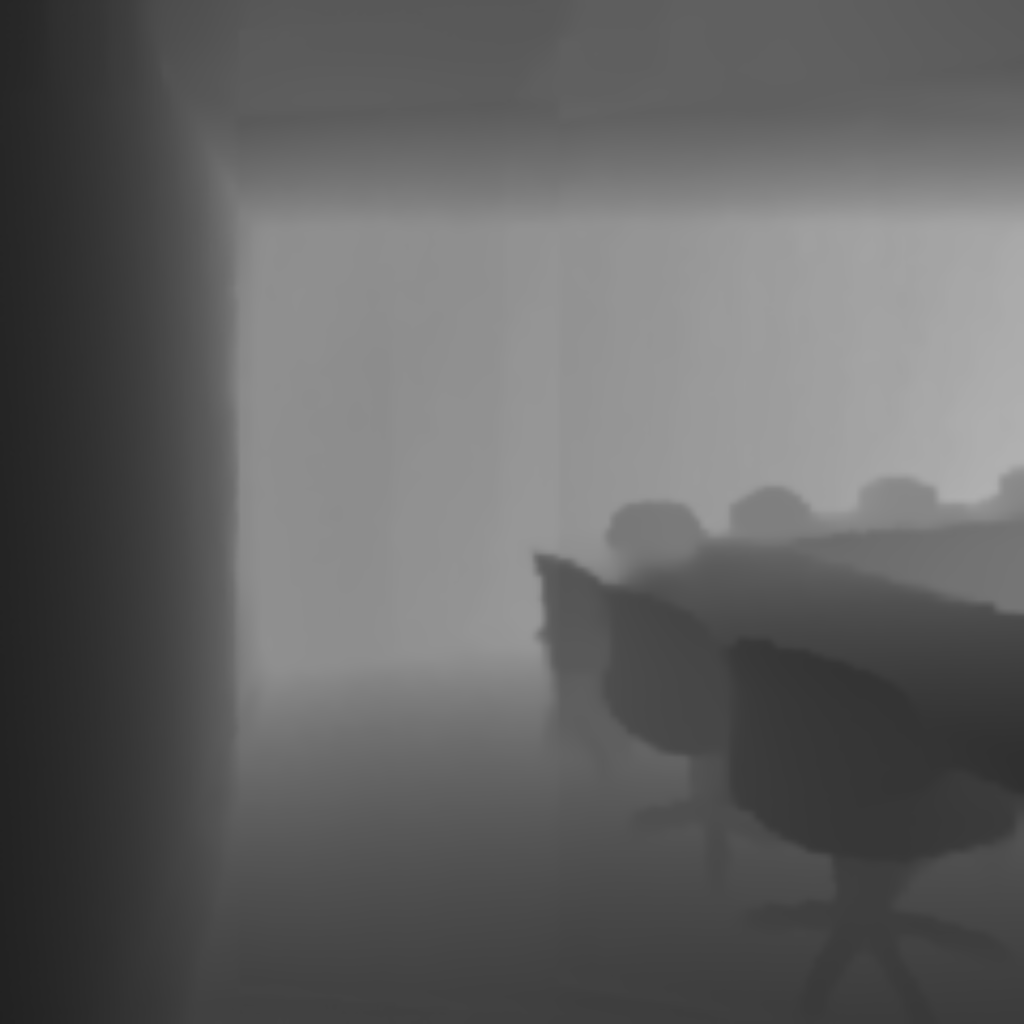} &\includegraphics[width=\imwidth\linewidth]{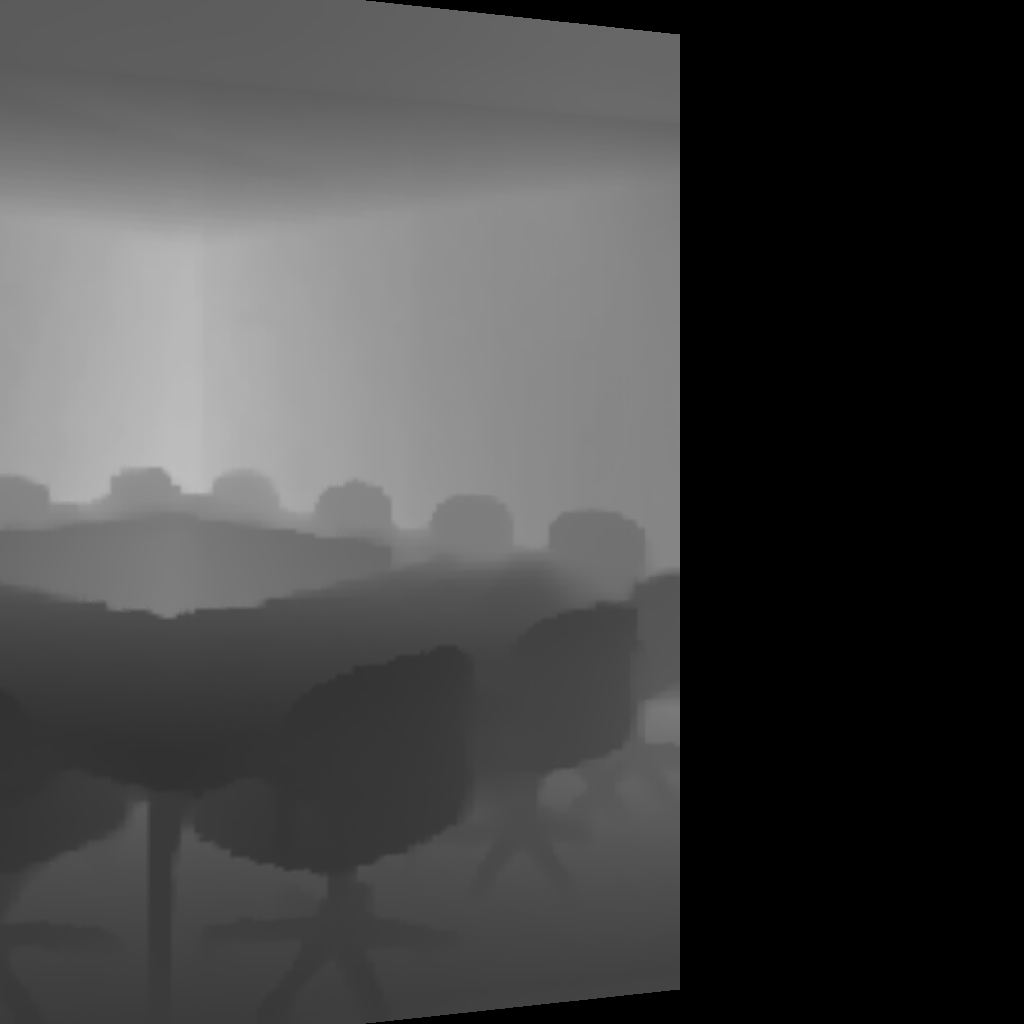} &\includegraphics[width=\imwidth\linewidth]{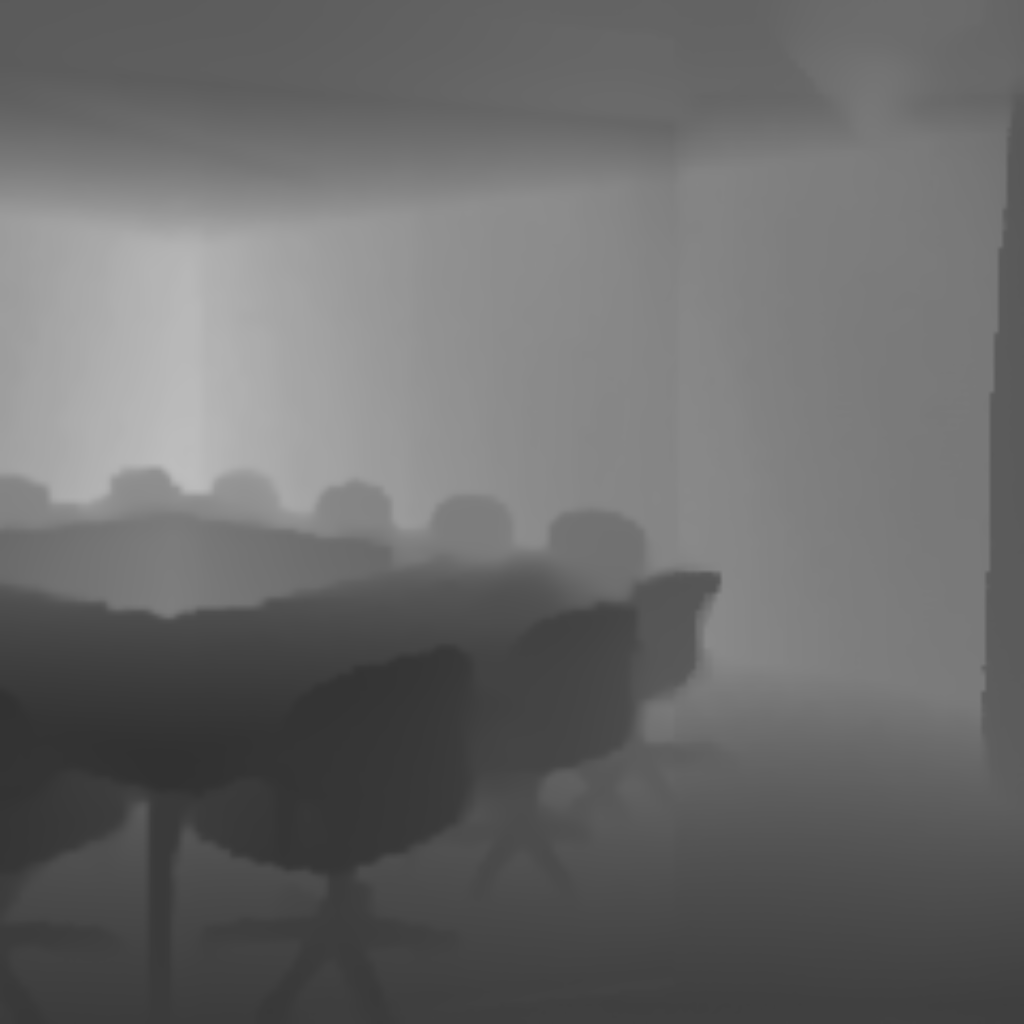} & \multicolumn{2}{c}{} \\
    \\[\textpadding]\multicolumn{\ncols}{c}{Prompt: A meeting room.}\\\\[\textpadding]

    \includegraphics[width=\imwidth\linewidth]{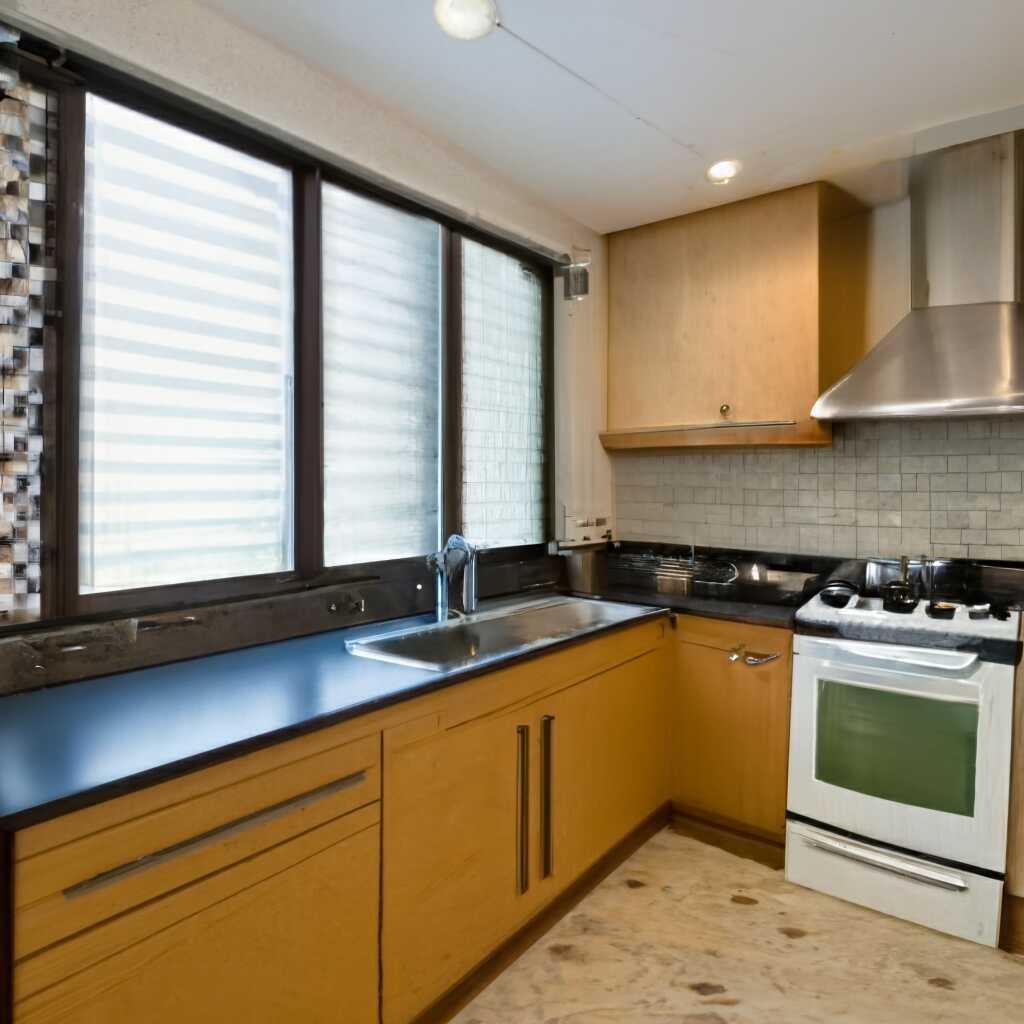} & \includegraphics[width=\imwidth\linewidth]{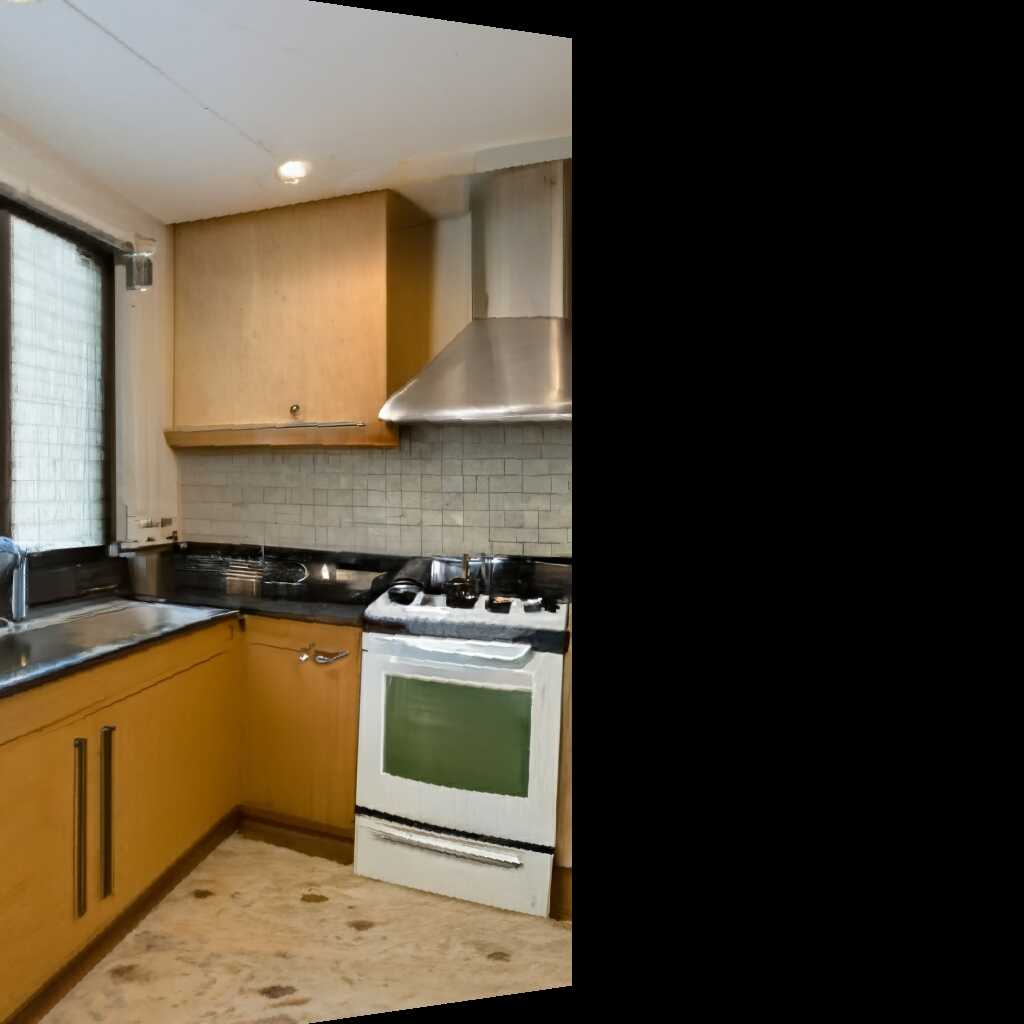} &\includegraphics[width=\imwidth\linewidth]{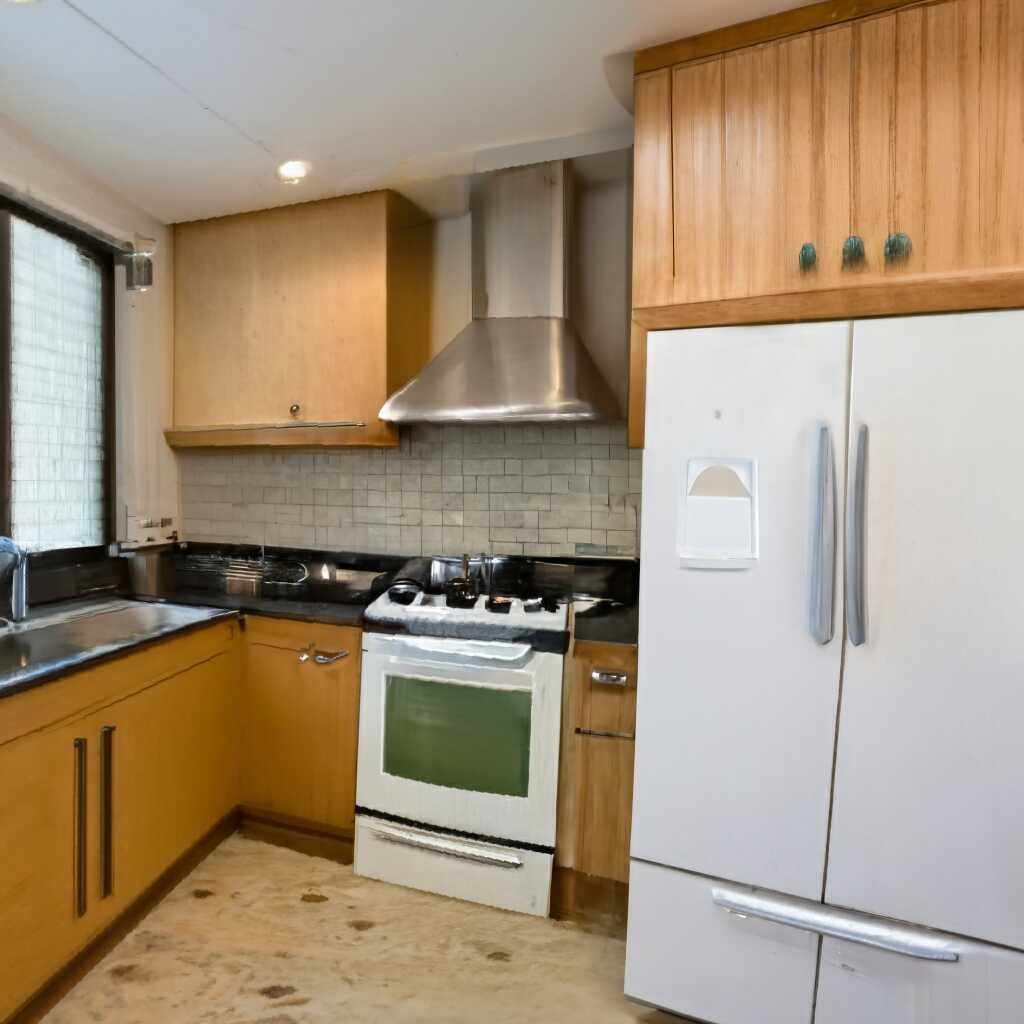} &\includegraphics[width=\imwidth\linewidth]{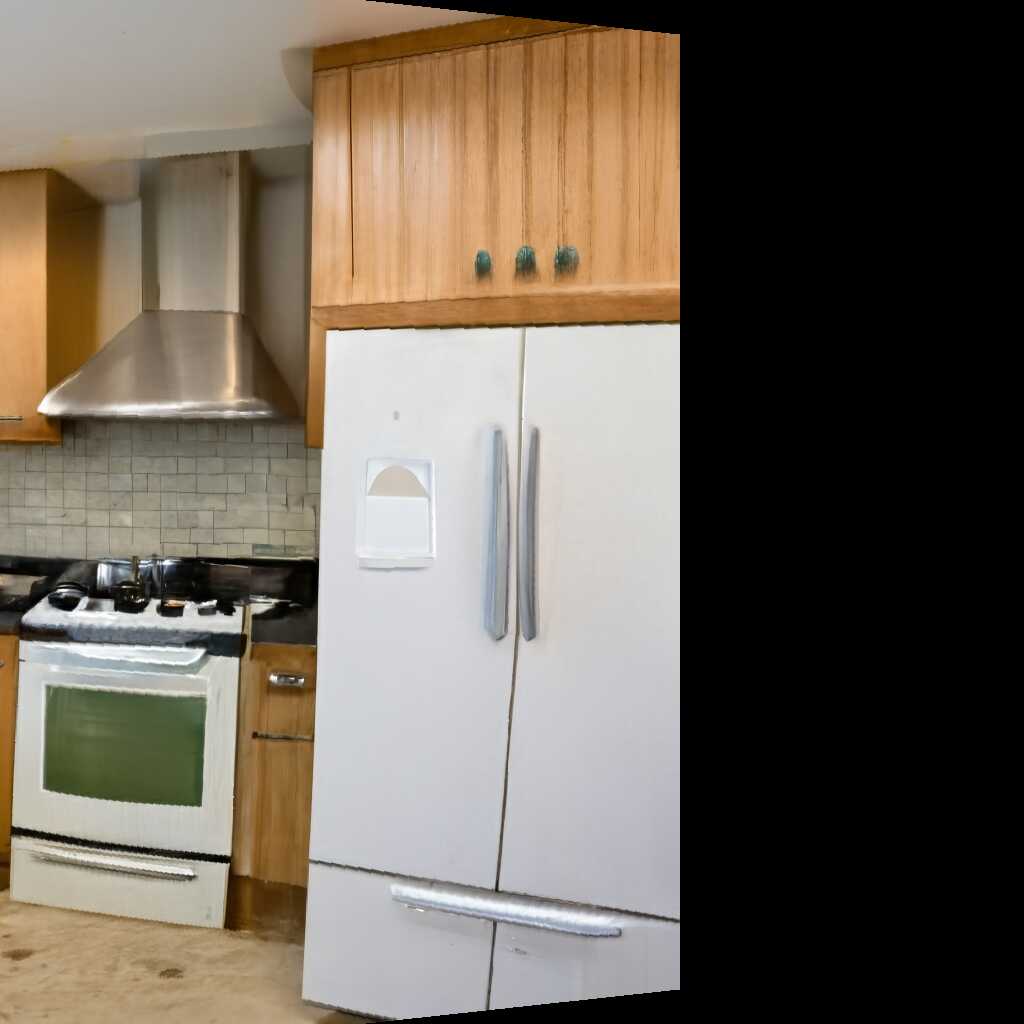} &\includegraphics[width=\imwidth\linewidth]{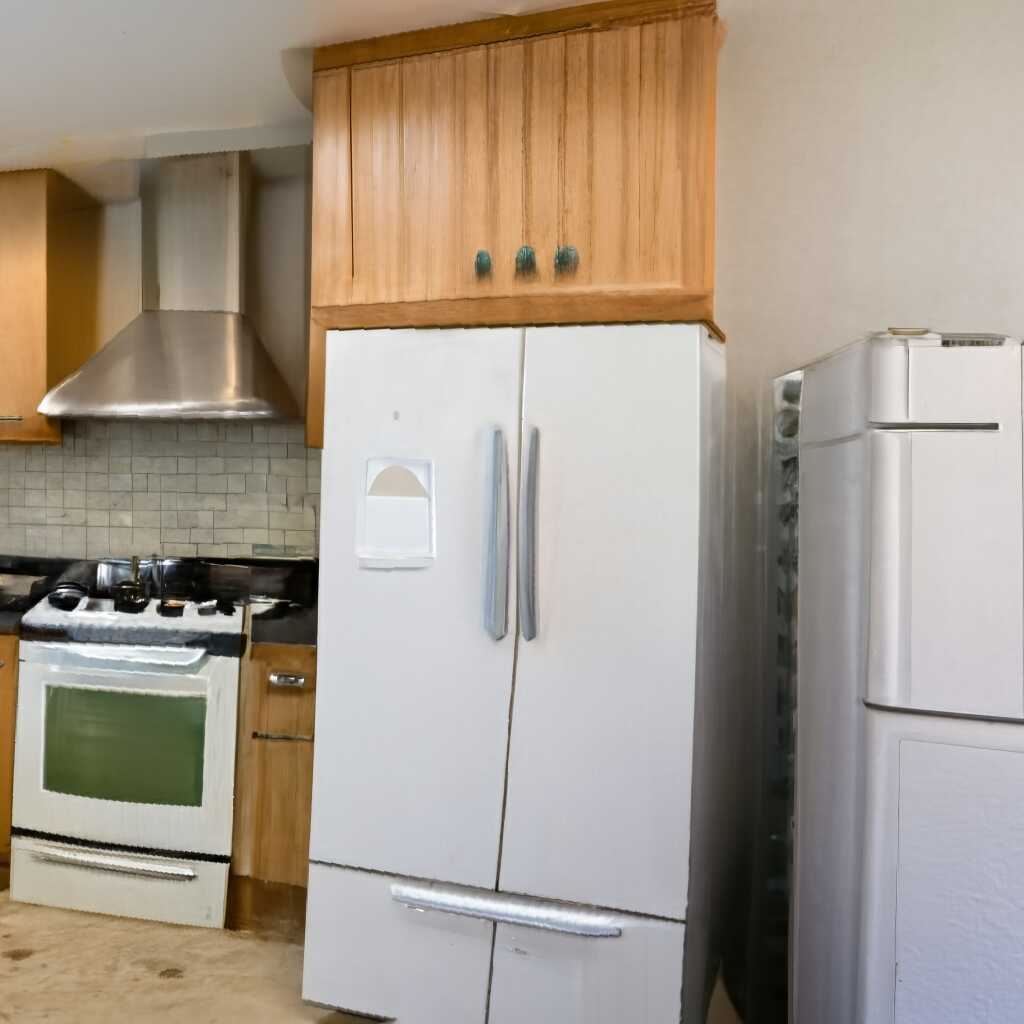} & \includegraphics[width=\imwidth\linewidth]{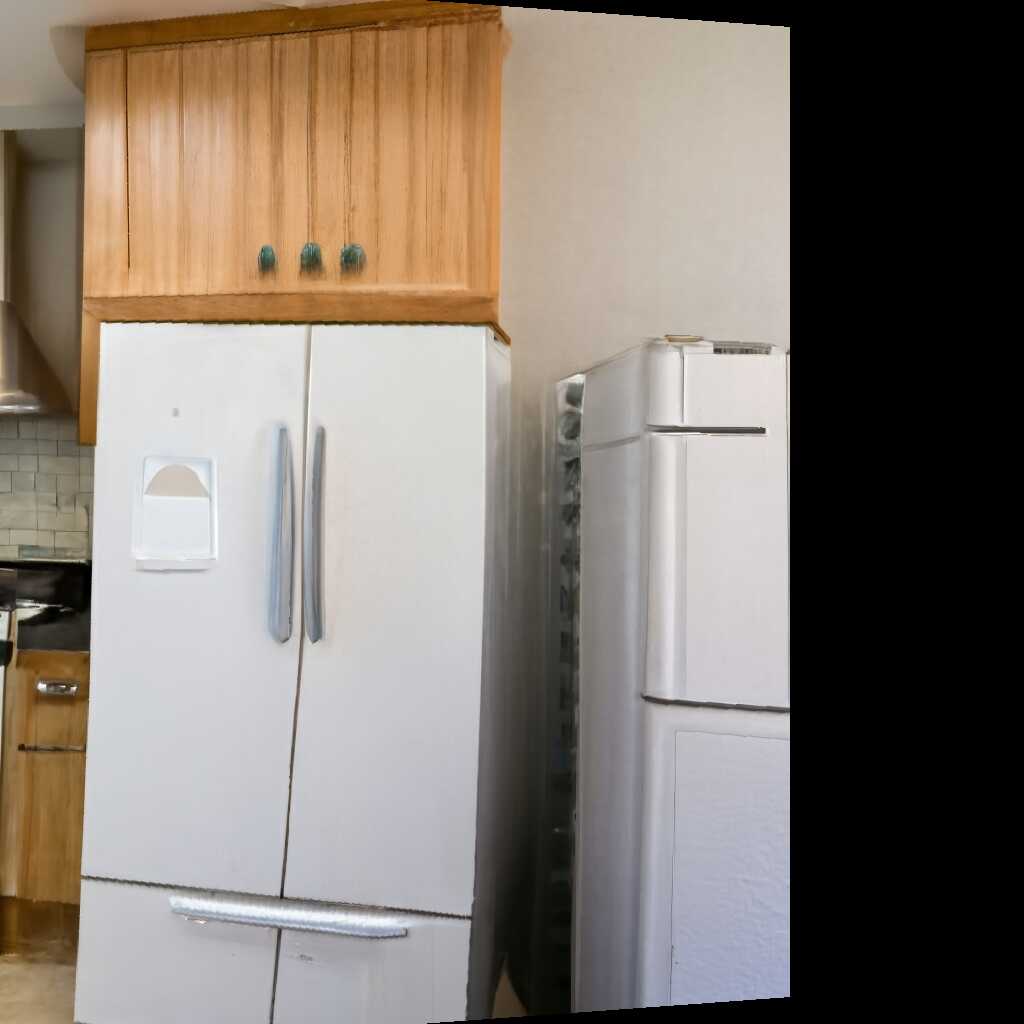} &\includegraphics[width=\imwidth\linewidth]{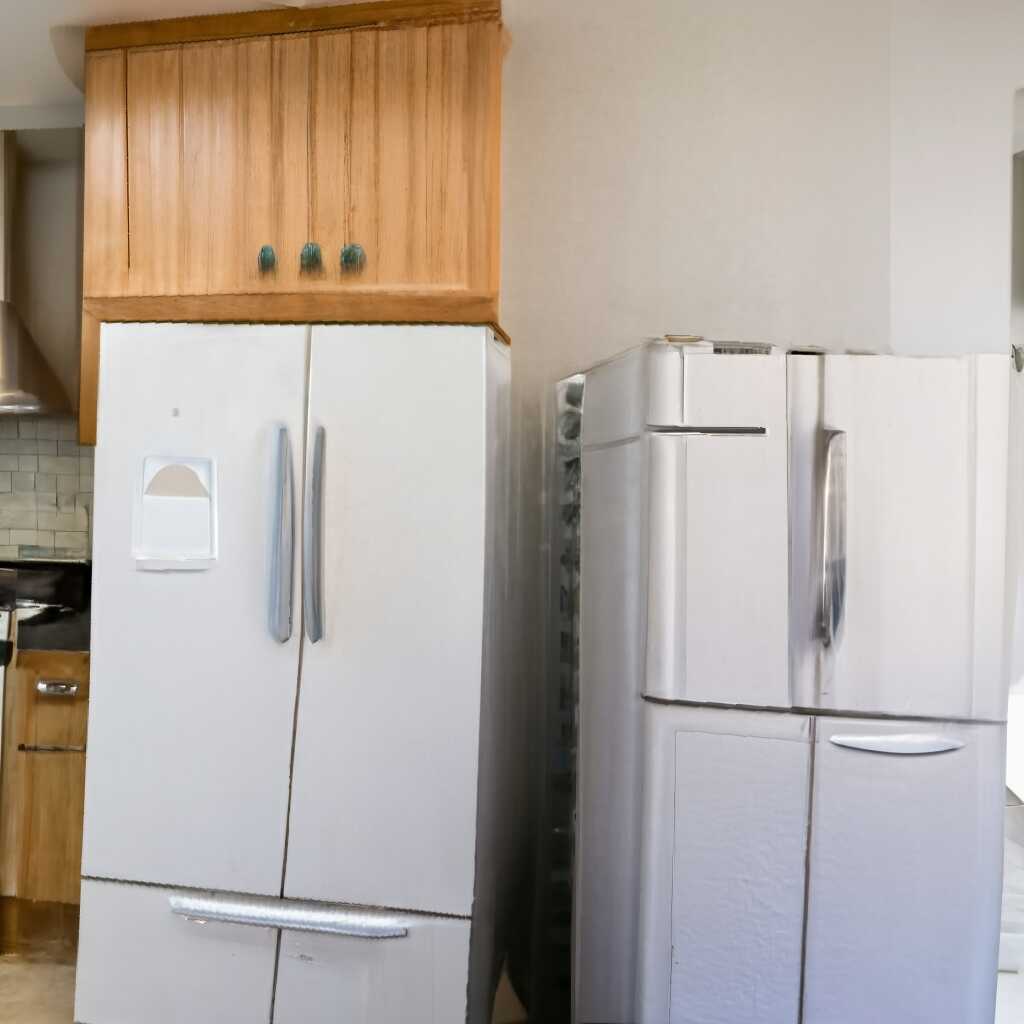} & \multicolumn{2}{c}{\multirow{2}{*}[\lastcoloffset]{\includegraphics[width=\imwidthlast\linewidth]{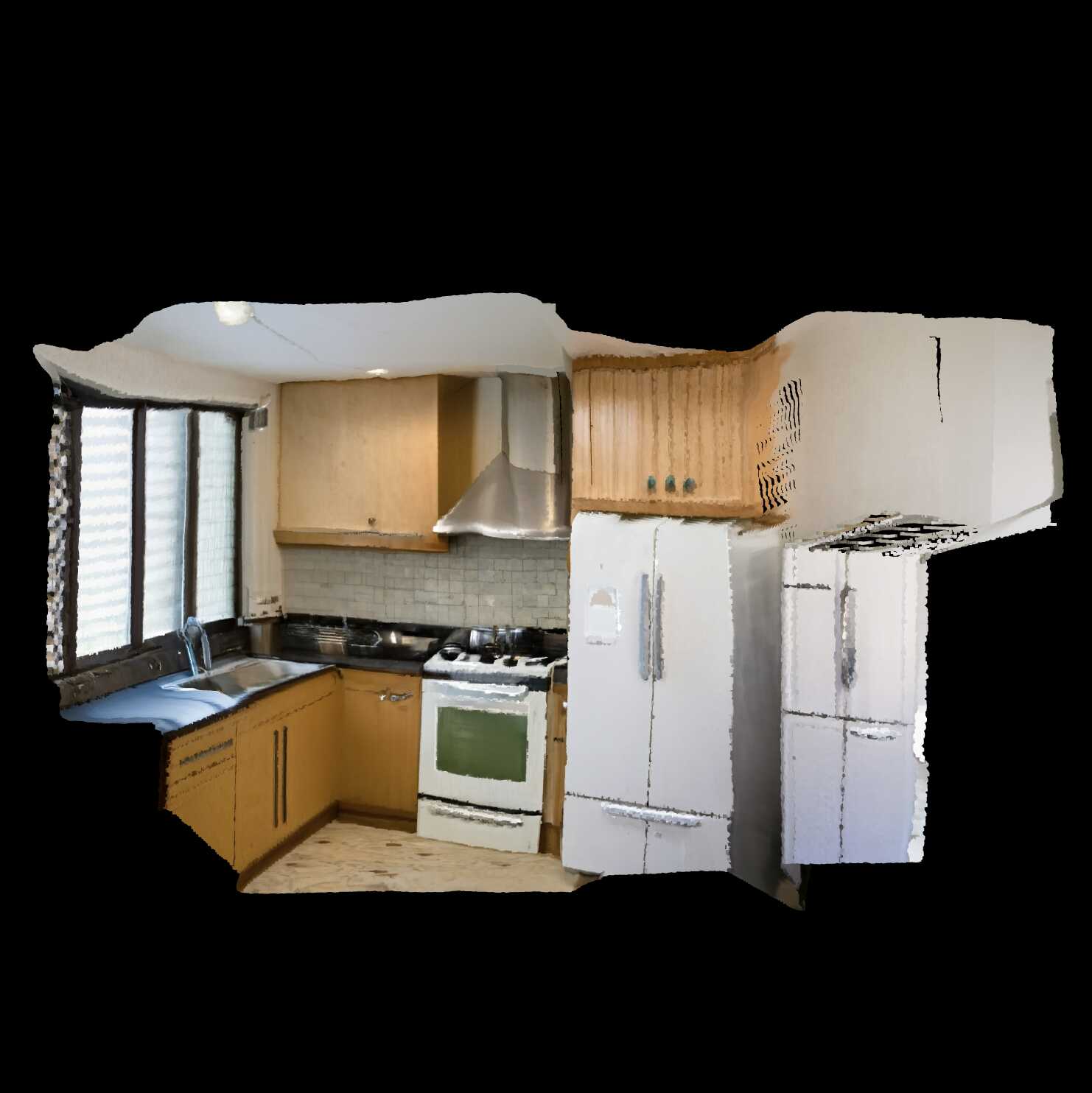}}} \\
    \includegraphics[width=\imwidth\linewidth]{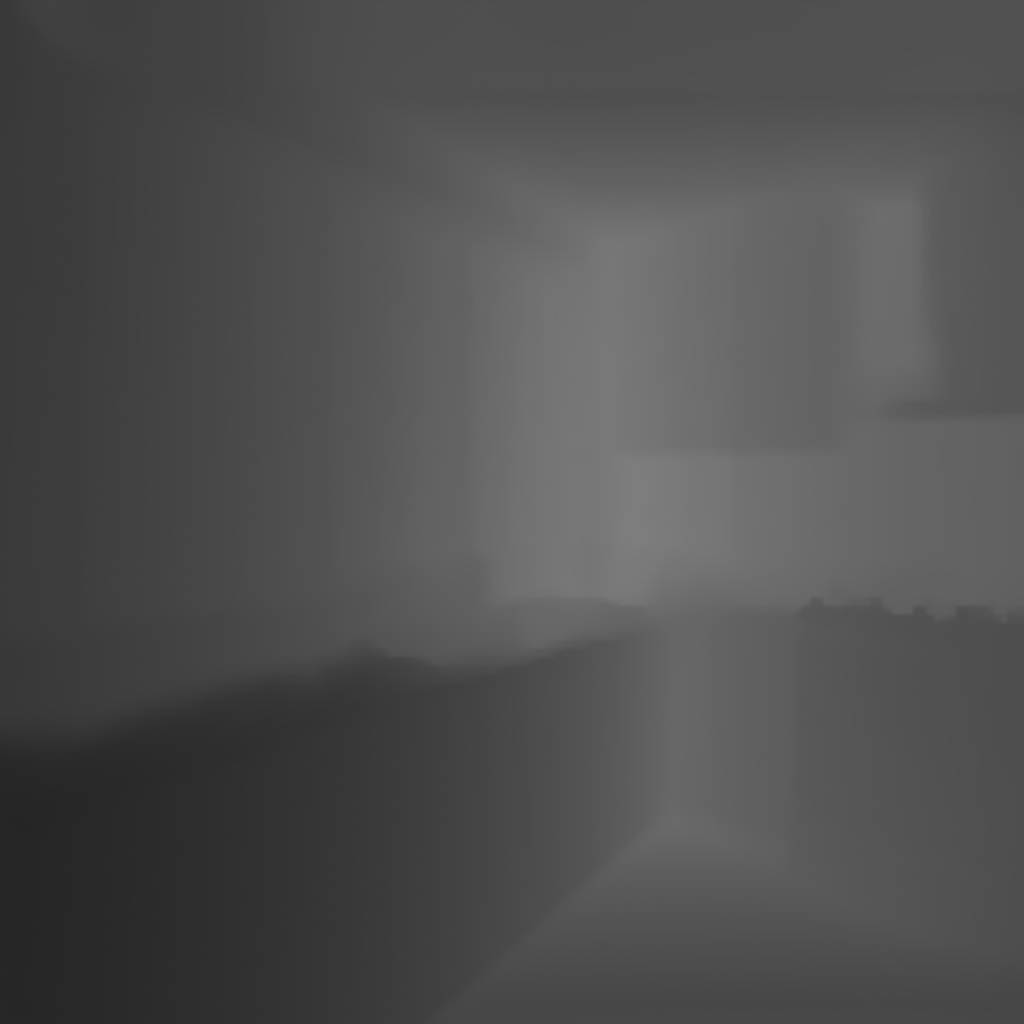} & \includegraphics[width=\imwidth\linewidth]{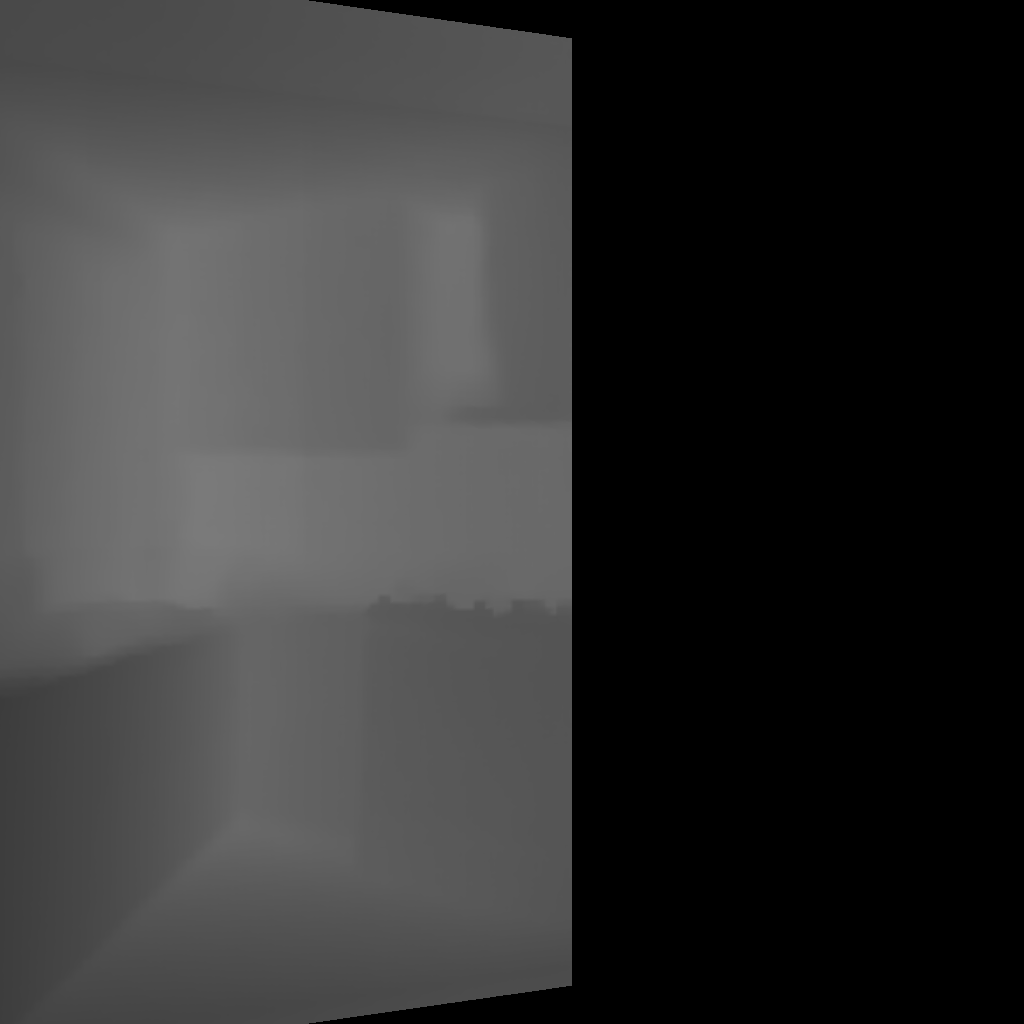} &\includegraphics[width=\imwidth\linewidth]{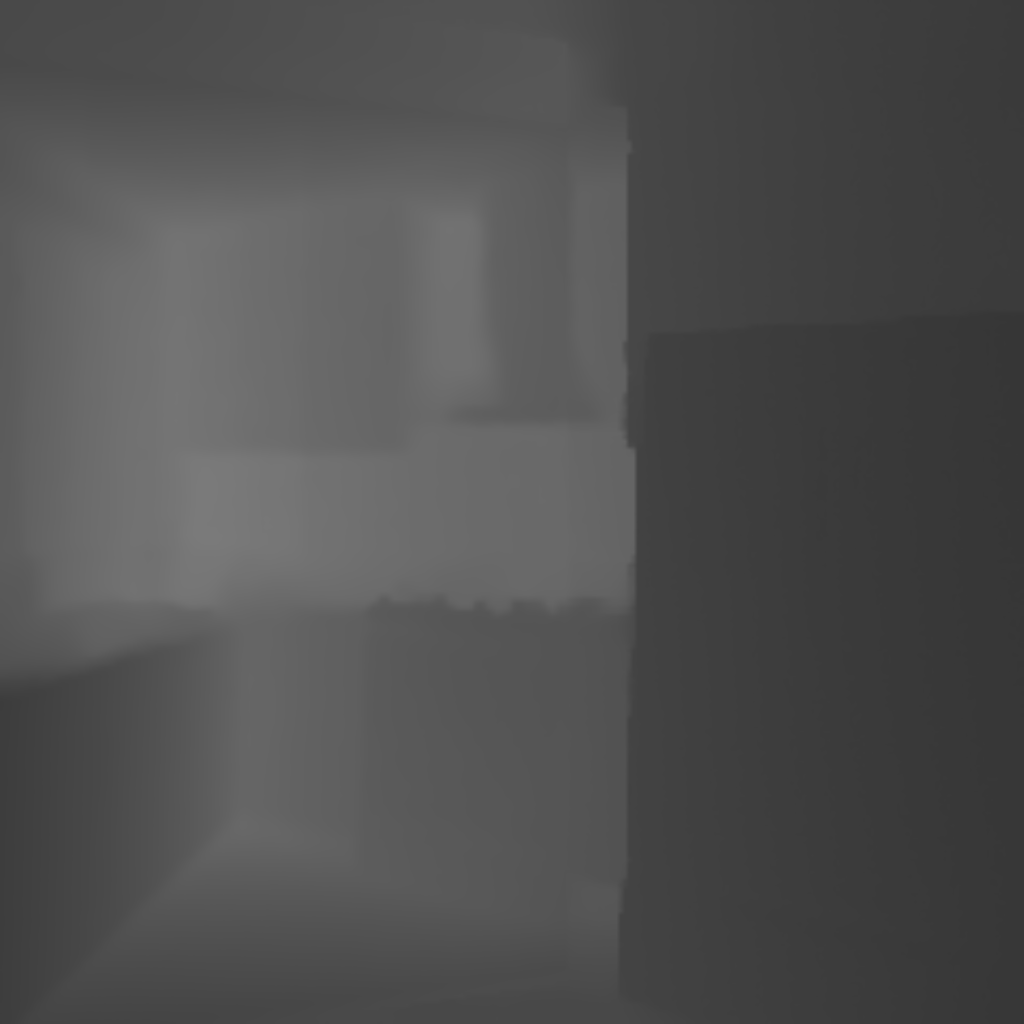} &  \includegraphics[width=\imwidth\linewidth]{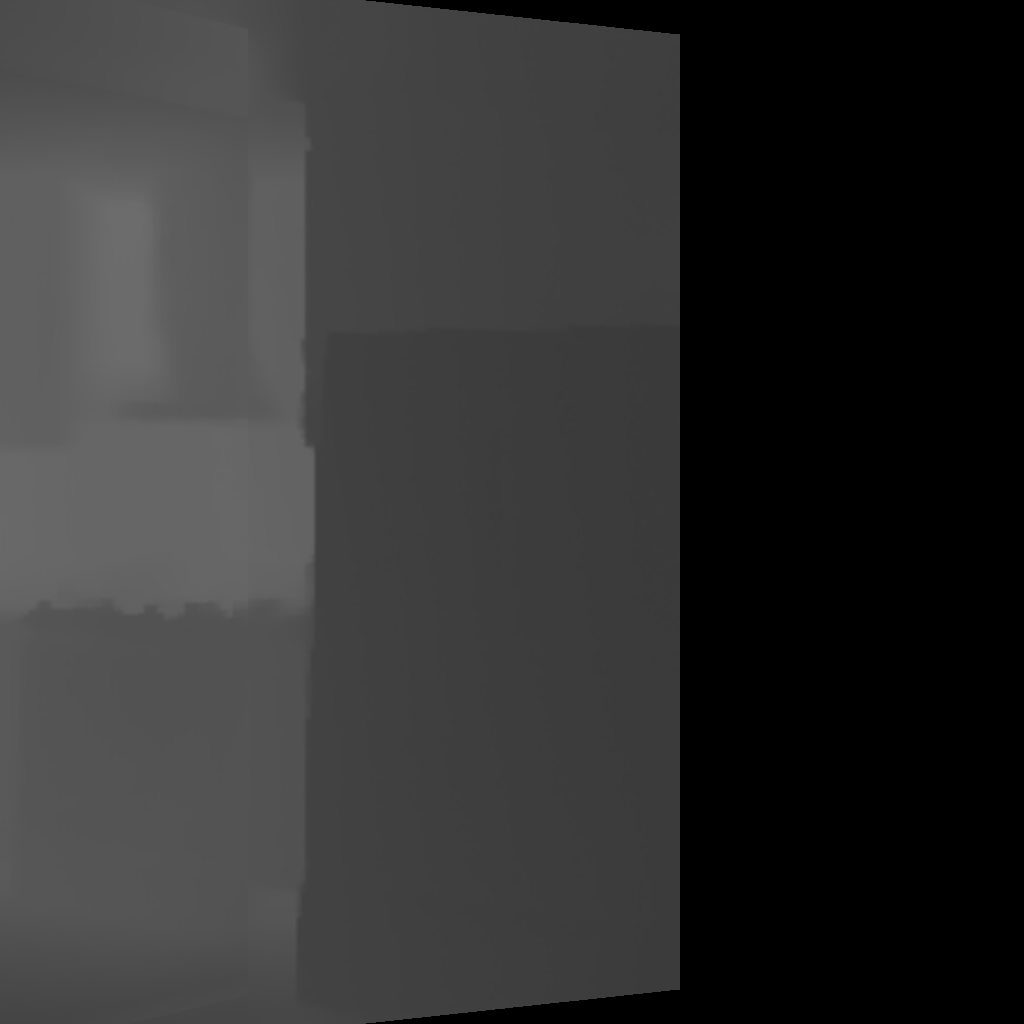} &\includegraphics[width=\imwidth\linewidth]{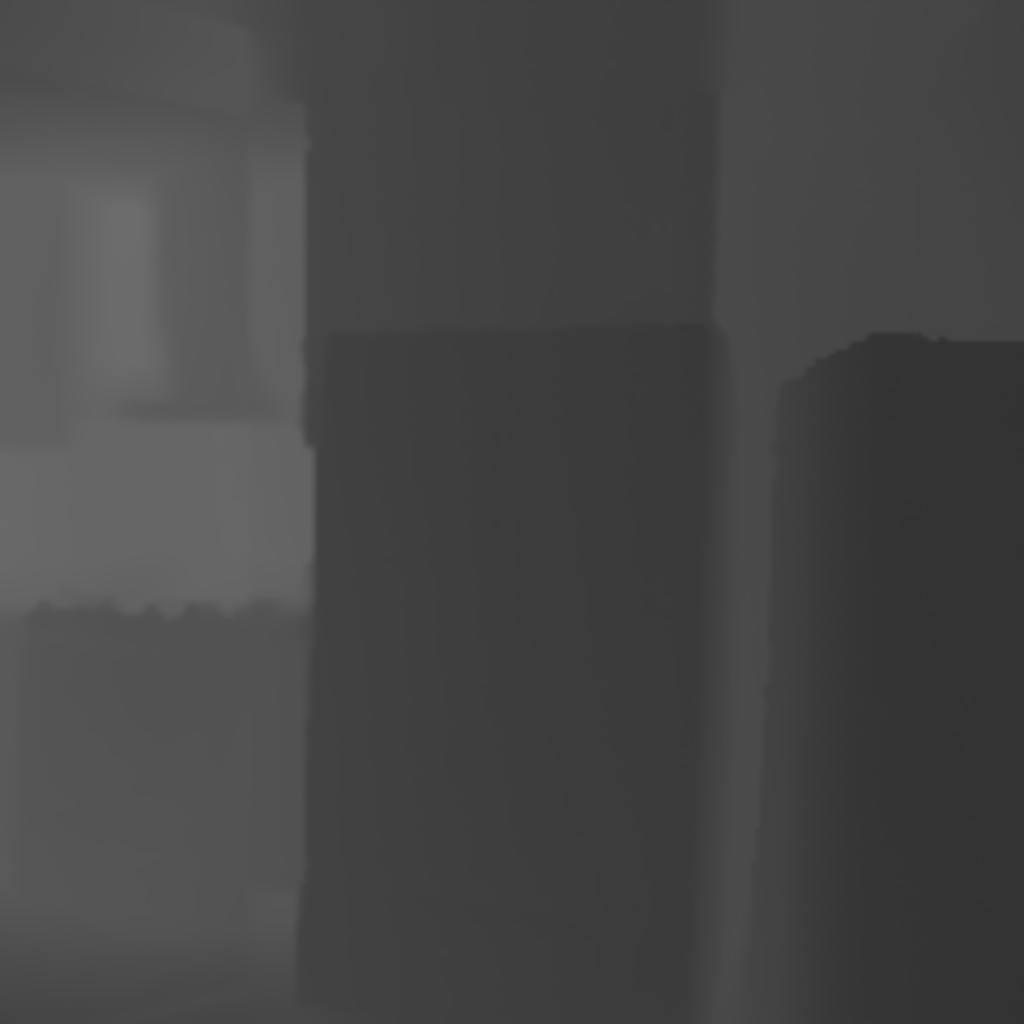} &\includegraphics[width=\imwidth\linewidth]{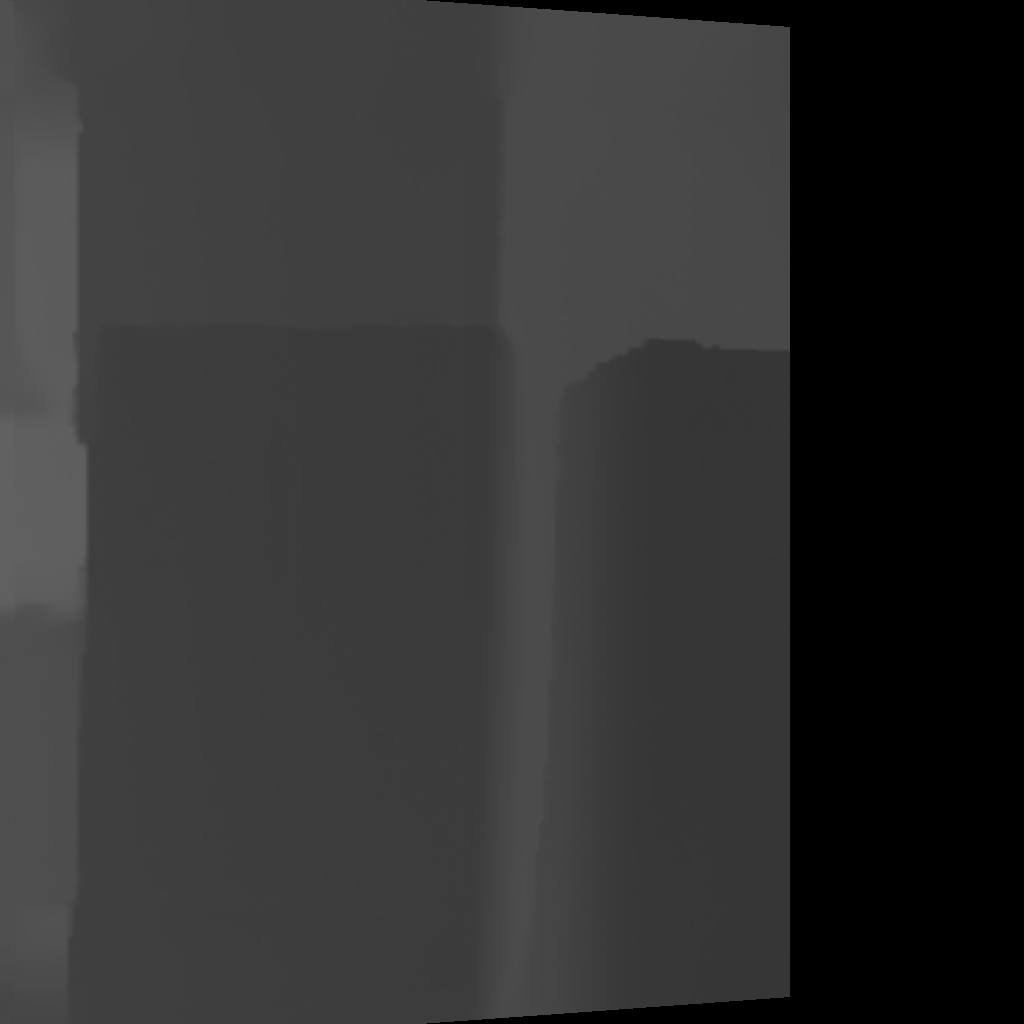} &\includegraphics[width=\imwidth\linewidth]{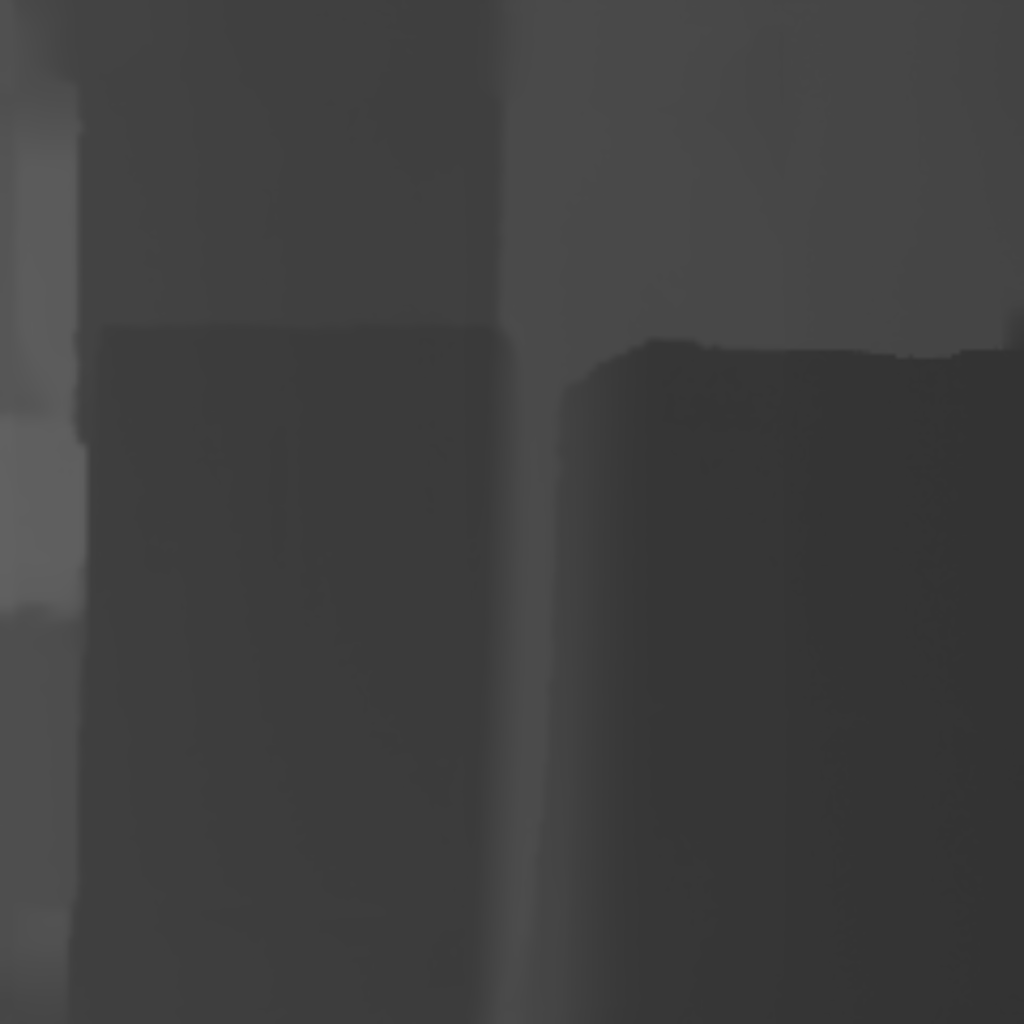} & \multicolumn{2}{c}{} \\
    \\[\textpadding]\multicolumn{\ncols}{c}{Prompt: A kitchen.}\\\\[\textpadding]

    \includegraphics[width=\imwidth\linewidth]{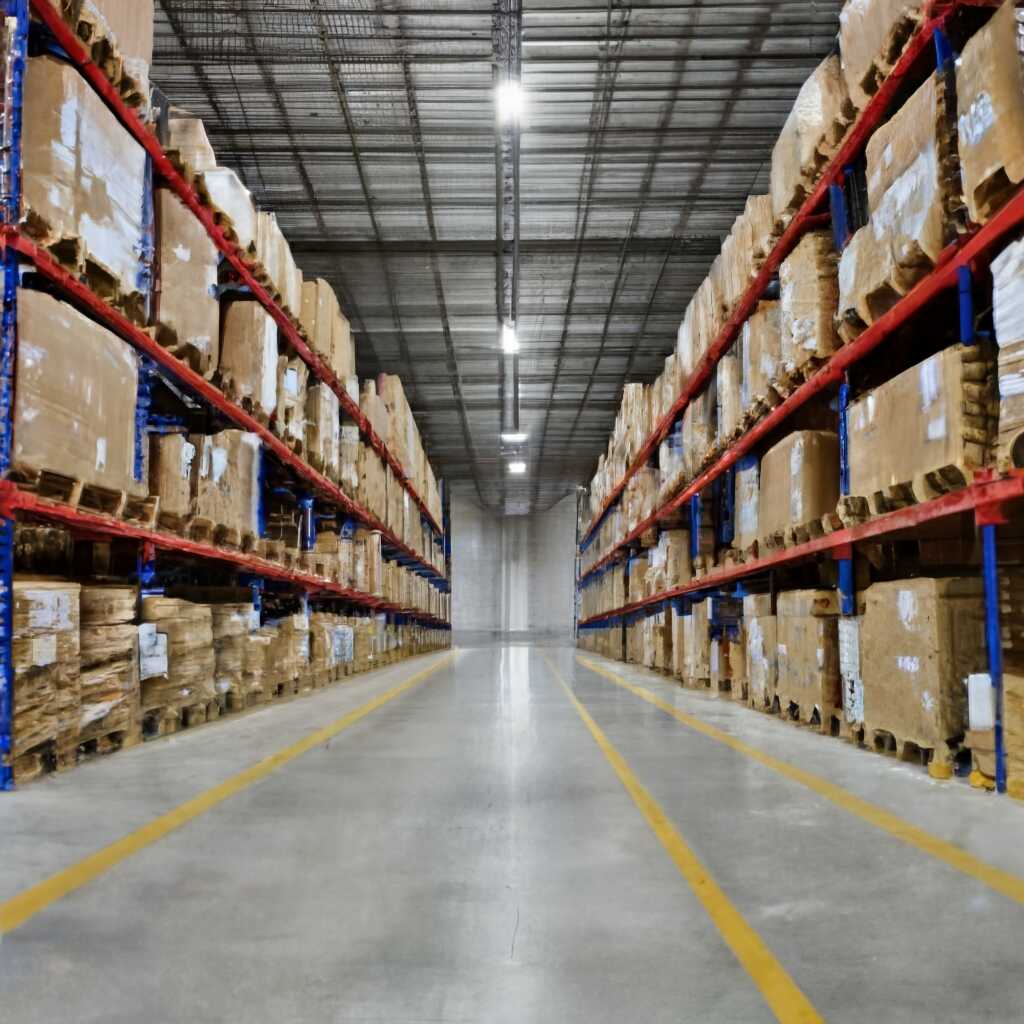} & \includegraphics[width=\imwidth\linewidth]{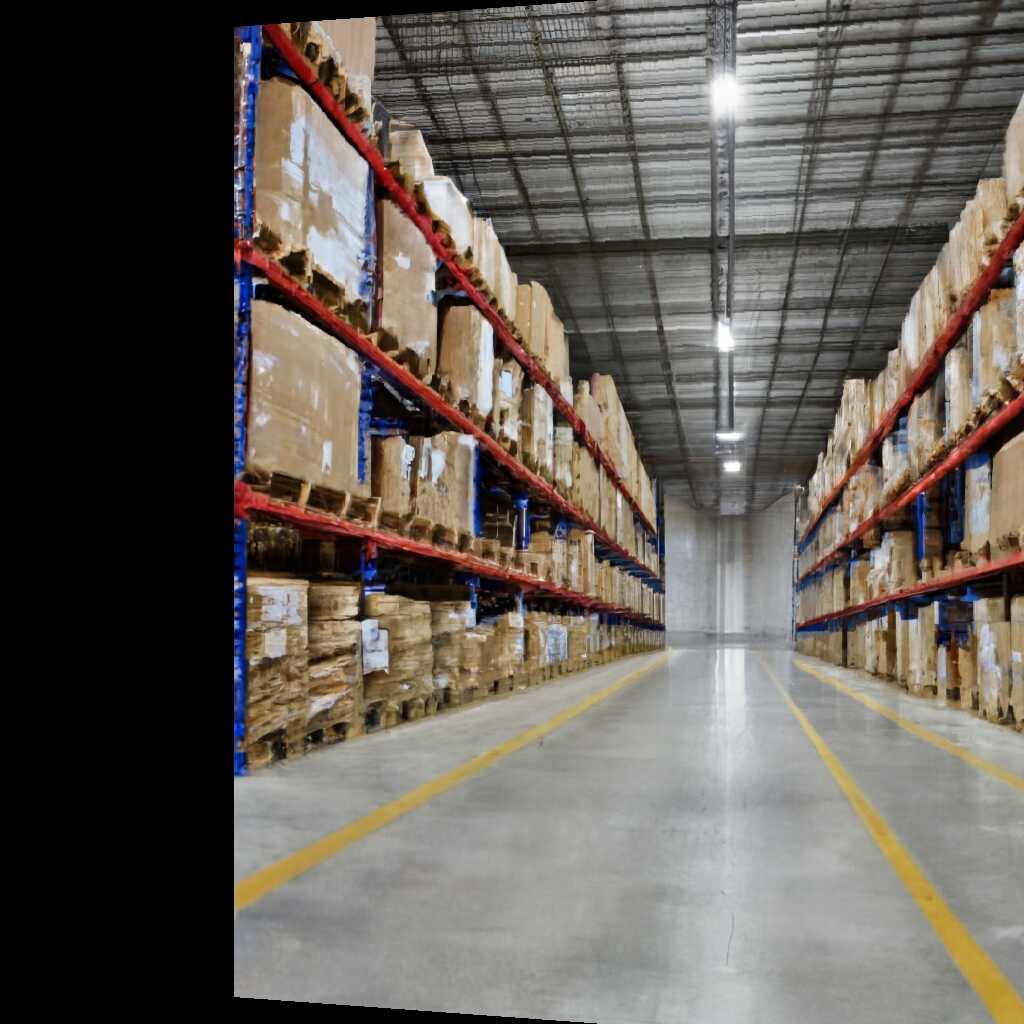} &\includegraphics[width=\imwidth\linewidth]{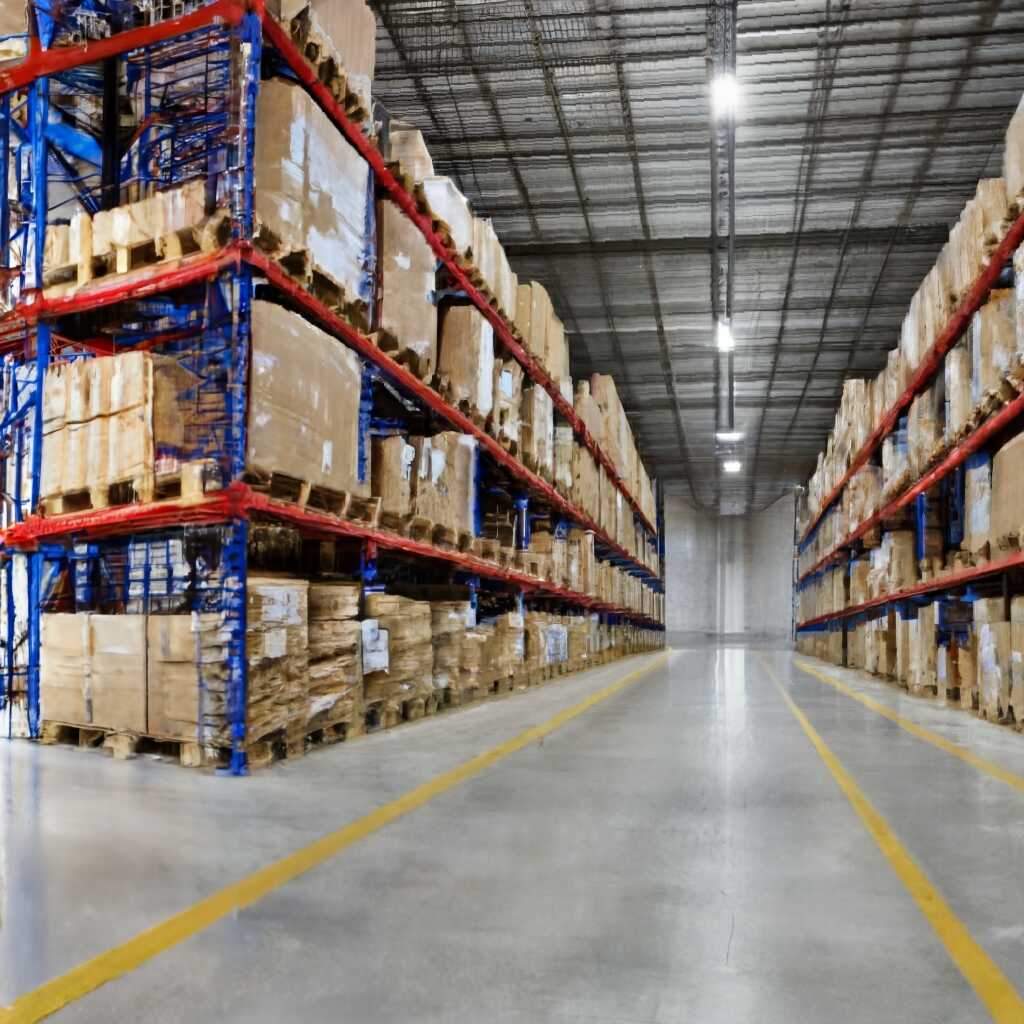} &\includegraphics[width=\imwidth\linewidth]{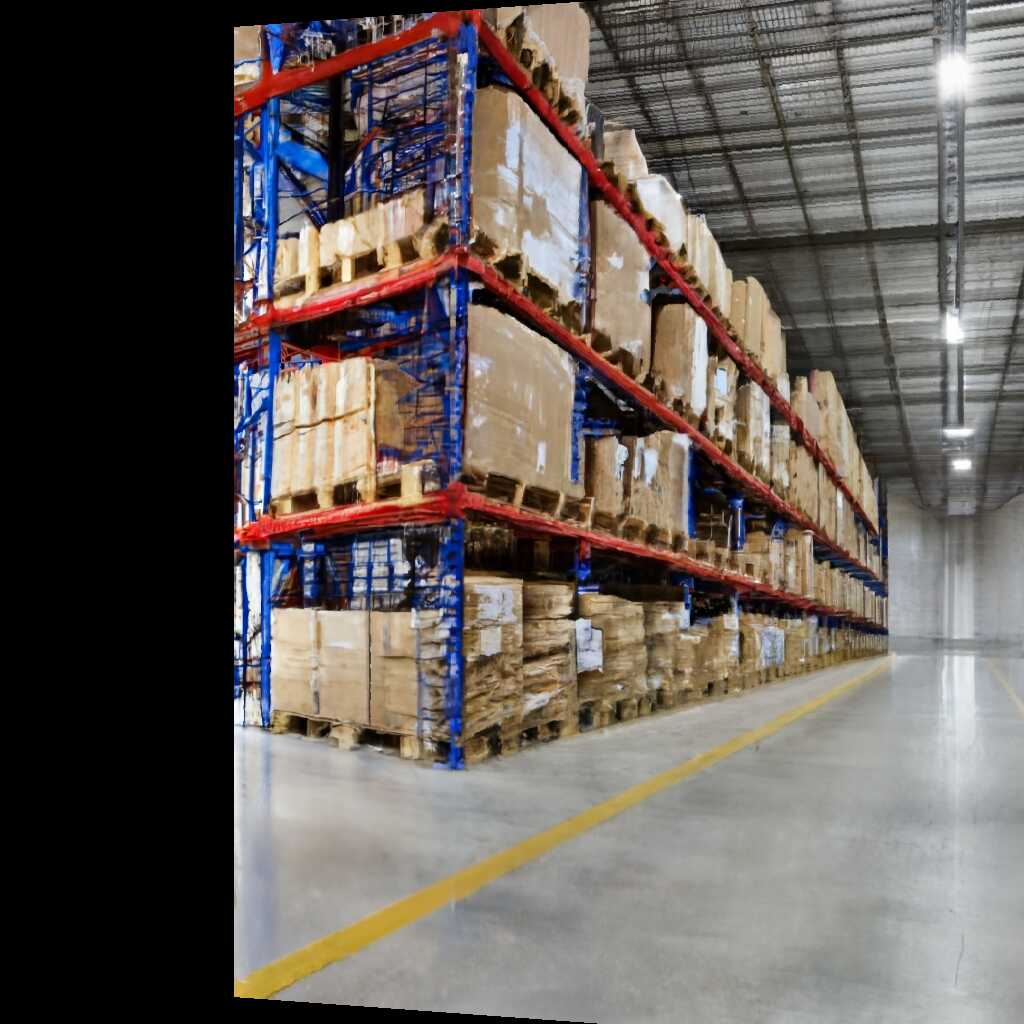} &\includegraphics[width=\imwidth\linewidth]{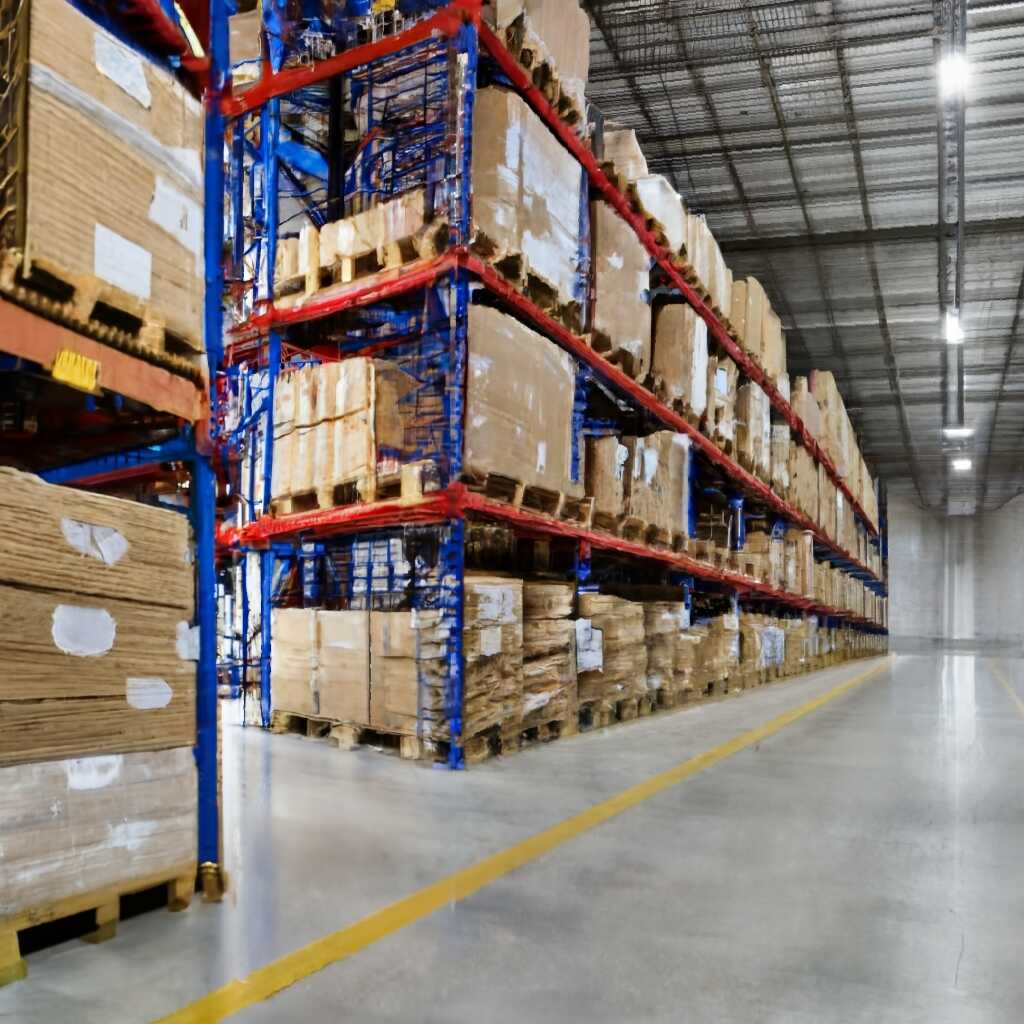} & \includegraphics[width=\imwidth\linewidth]{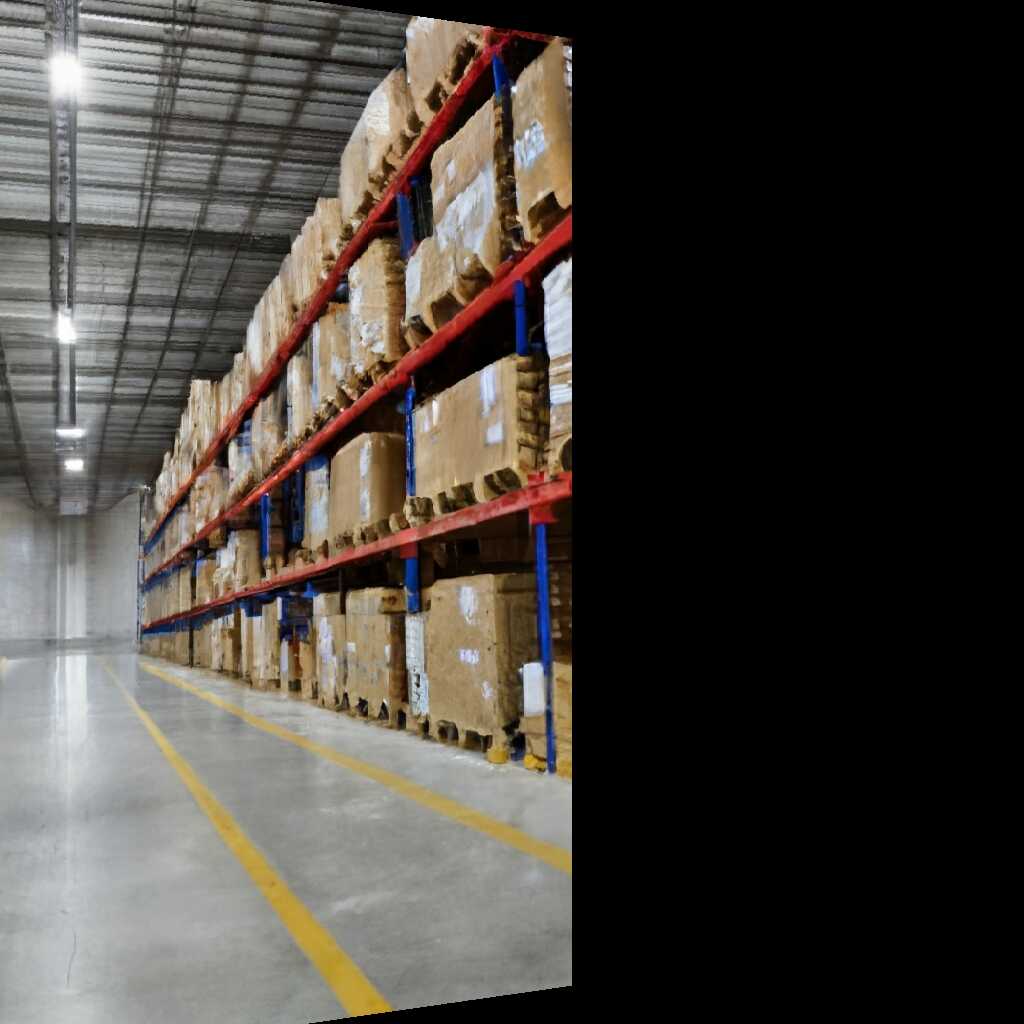} &\includegraphics[width=\imwidth\linewidth]{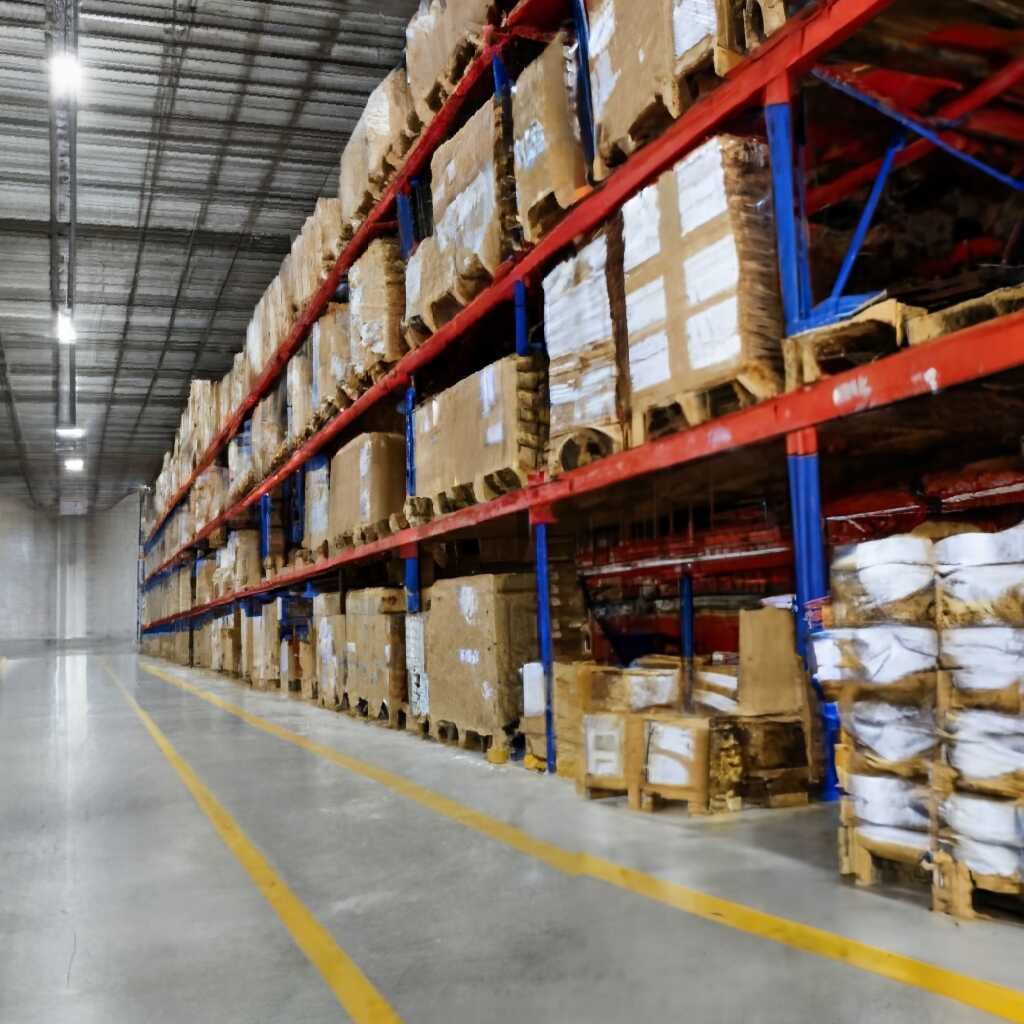} & \multicolumn{2}{c}{\multirow{2}{*}[\lastcoloffset]{\includegraphics[width=\imwidthlast\linewidth]{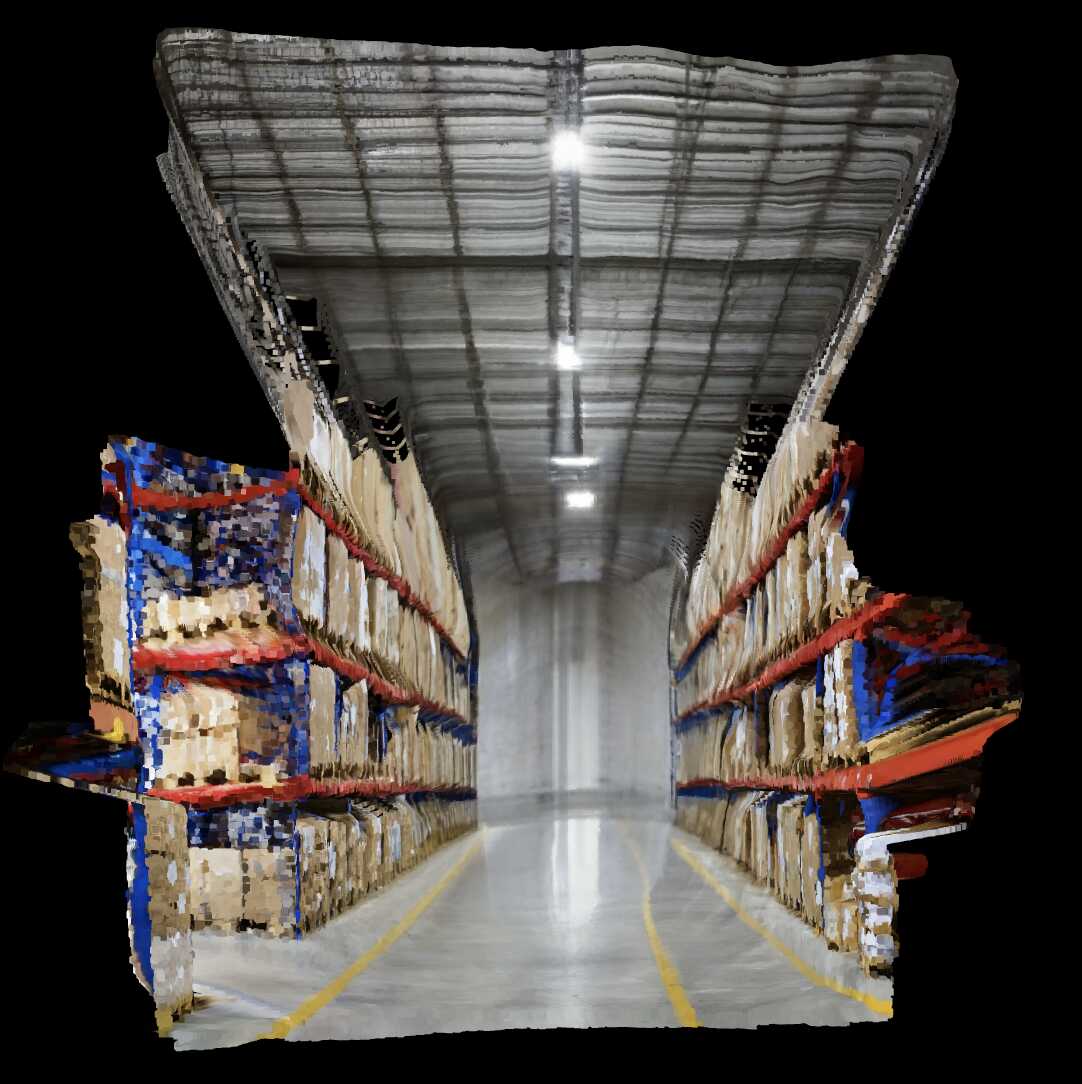}}} \\
    \includegraphics[width=\imwidth\linewidth]{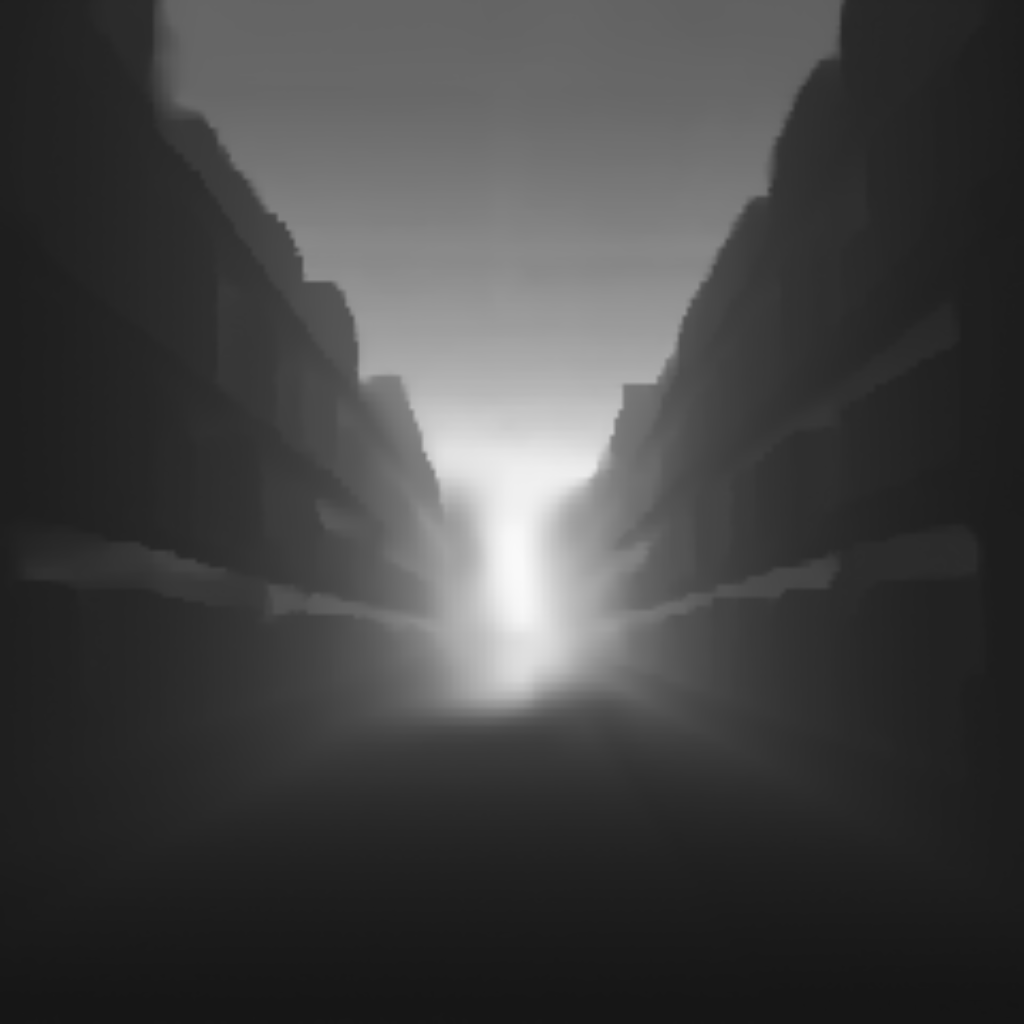} & \includegraphics[width=\imwidth\linewidth]{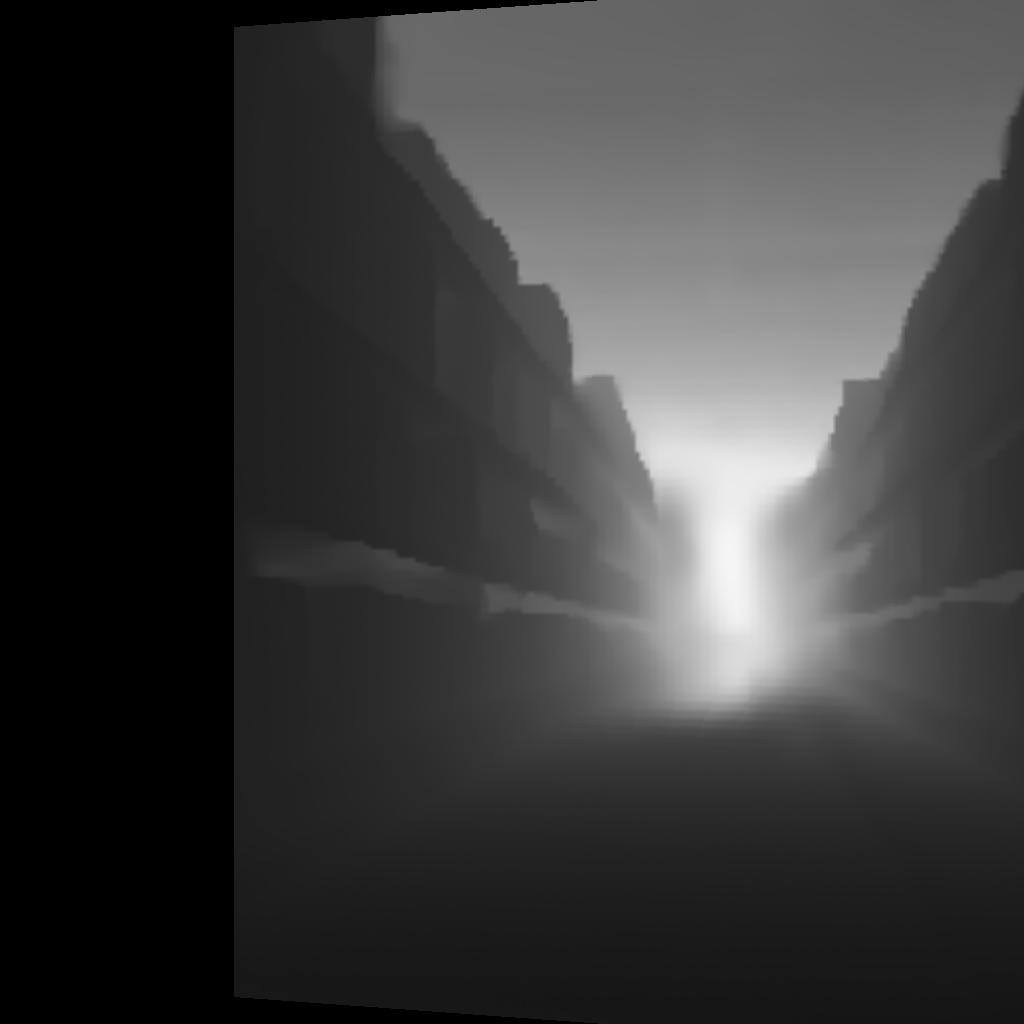} &\includegraphics[width=\imwidth\linewidth]{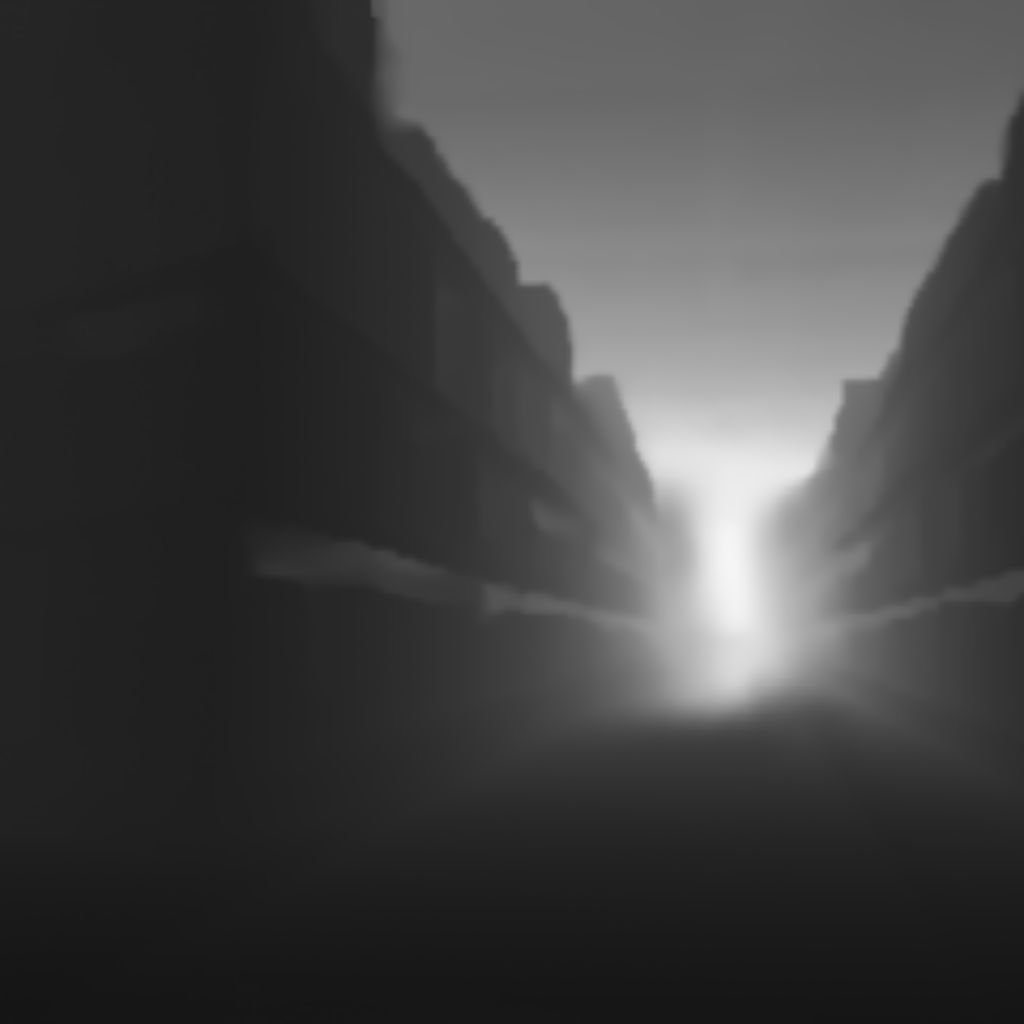} &  \includegraphics[width=\imwidth\linewidth]{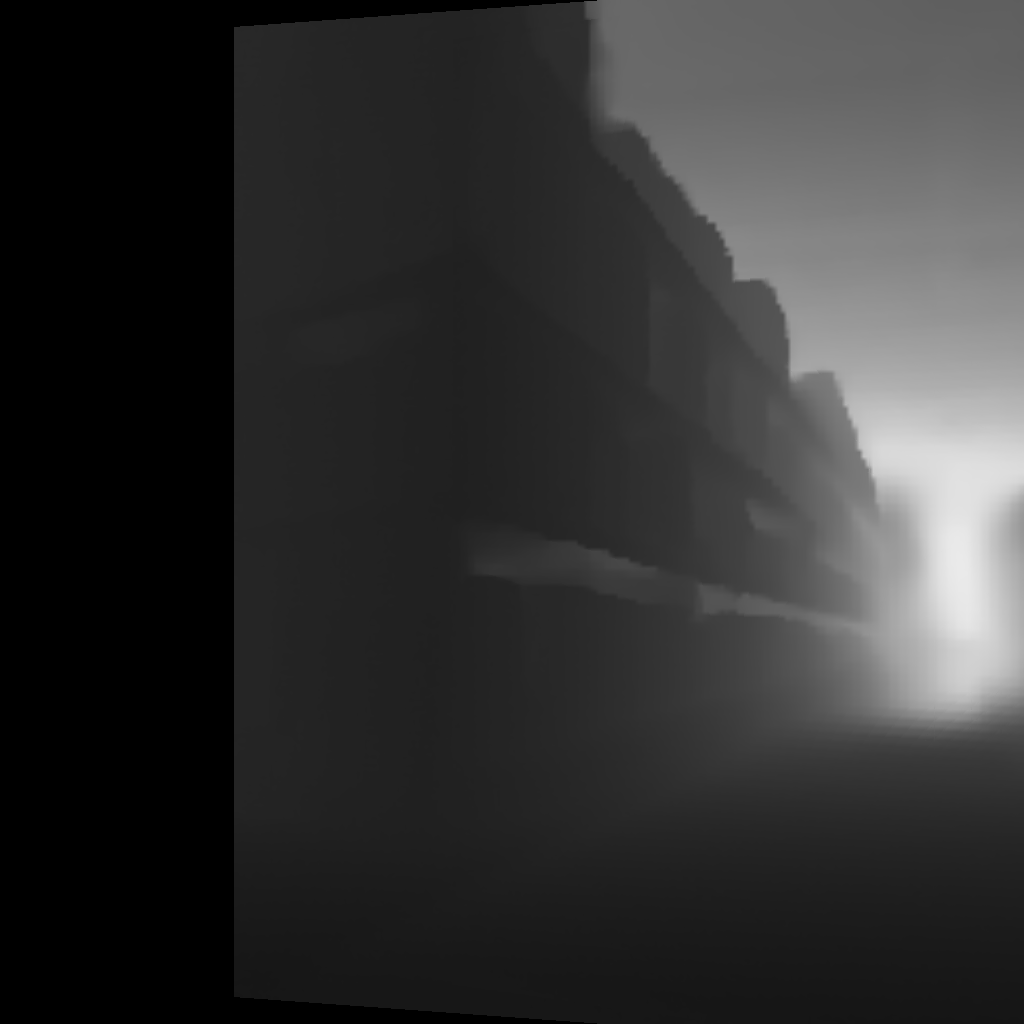} &\includegraphics[width=\imwidth\linewidth]{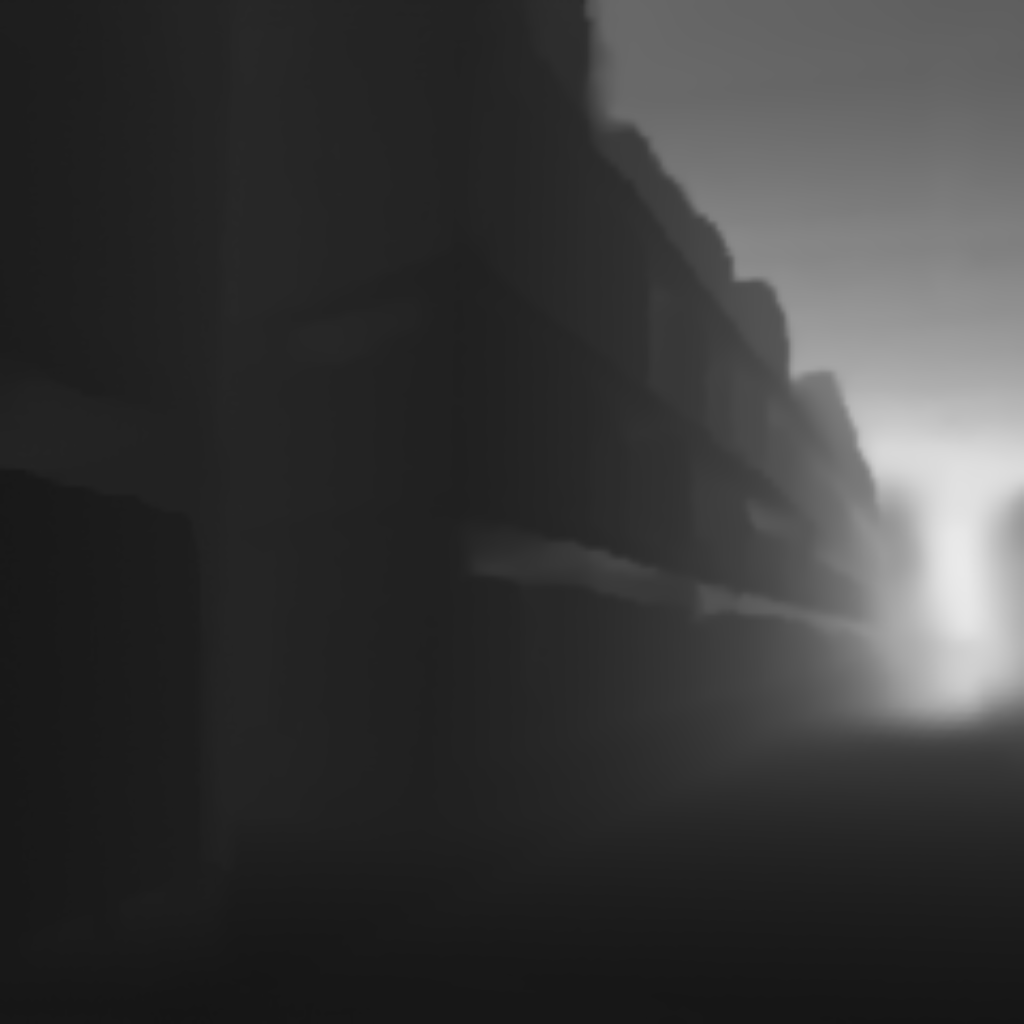} &\includegraphics[width=\imwidth\linewidth]{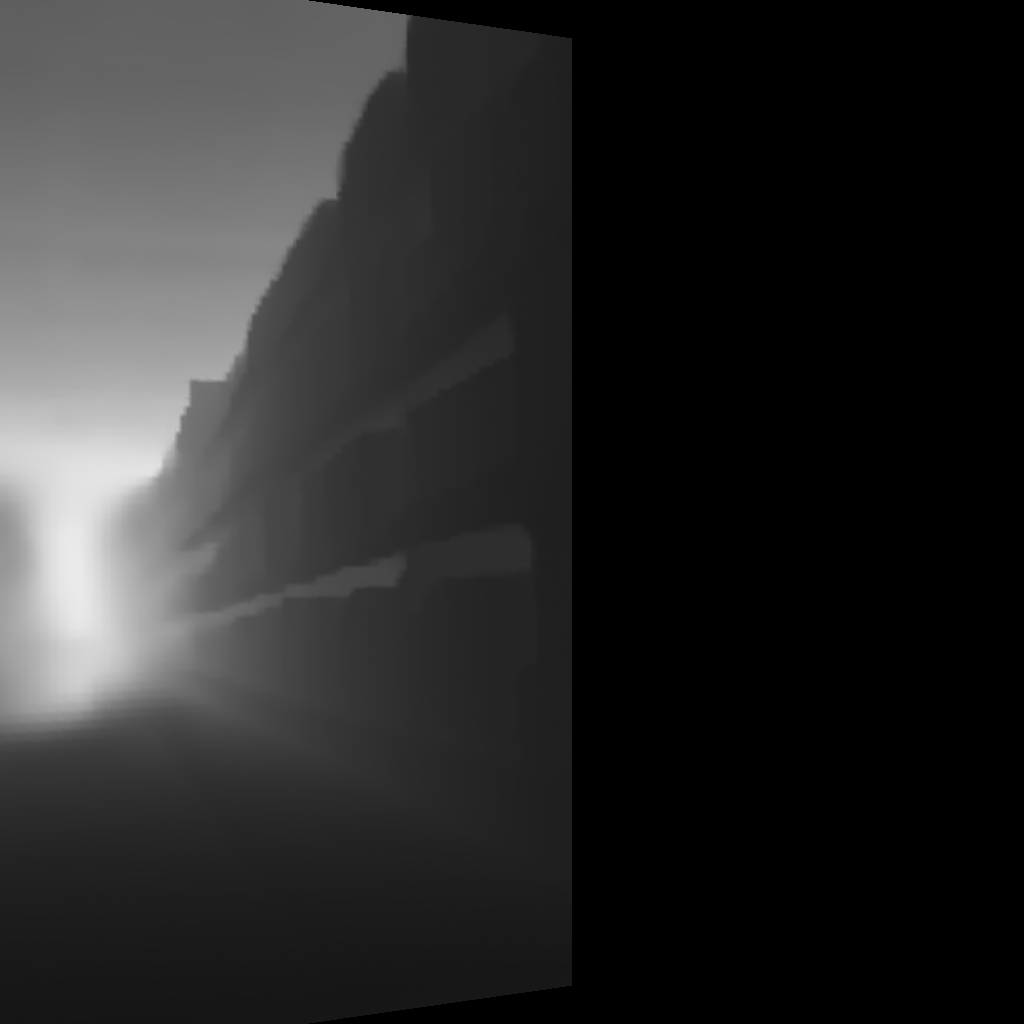} &\includegraphics[width=\imwidth\linewidth]{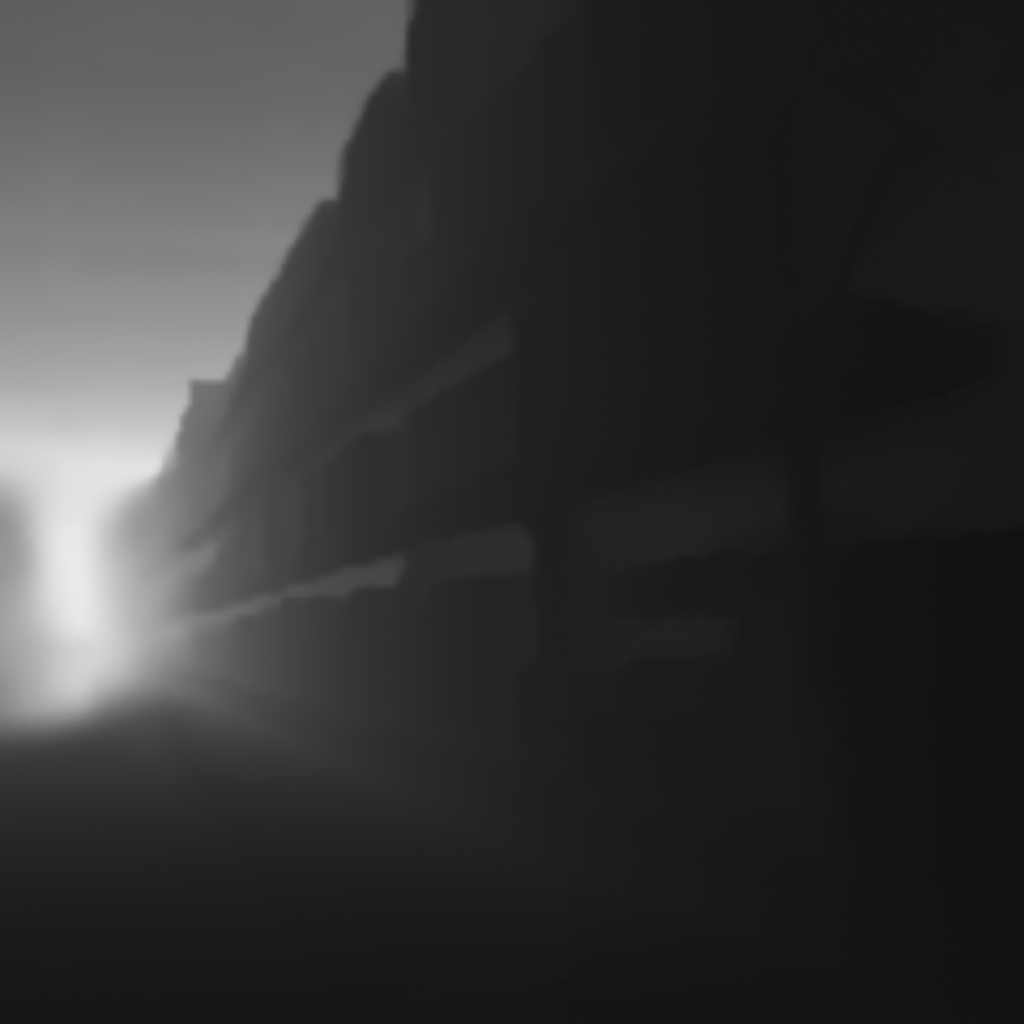} & \multicolumn{2}{c}{} \\
    \\[\textpadding]\multicolumn{\ncols}{c}{Prompt: A warehouse.}\\\\[\textpadding]

    \includegraphics[width=\imwidth\linewidth]{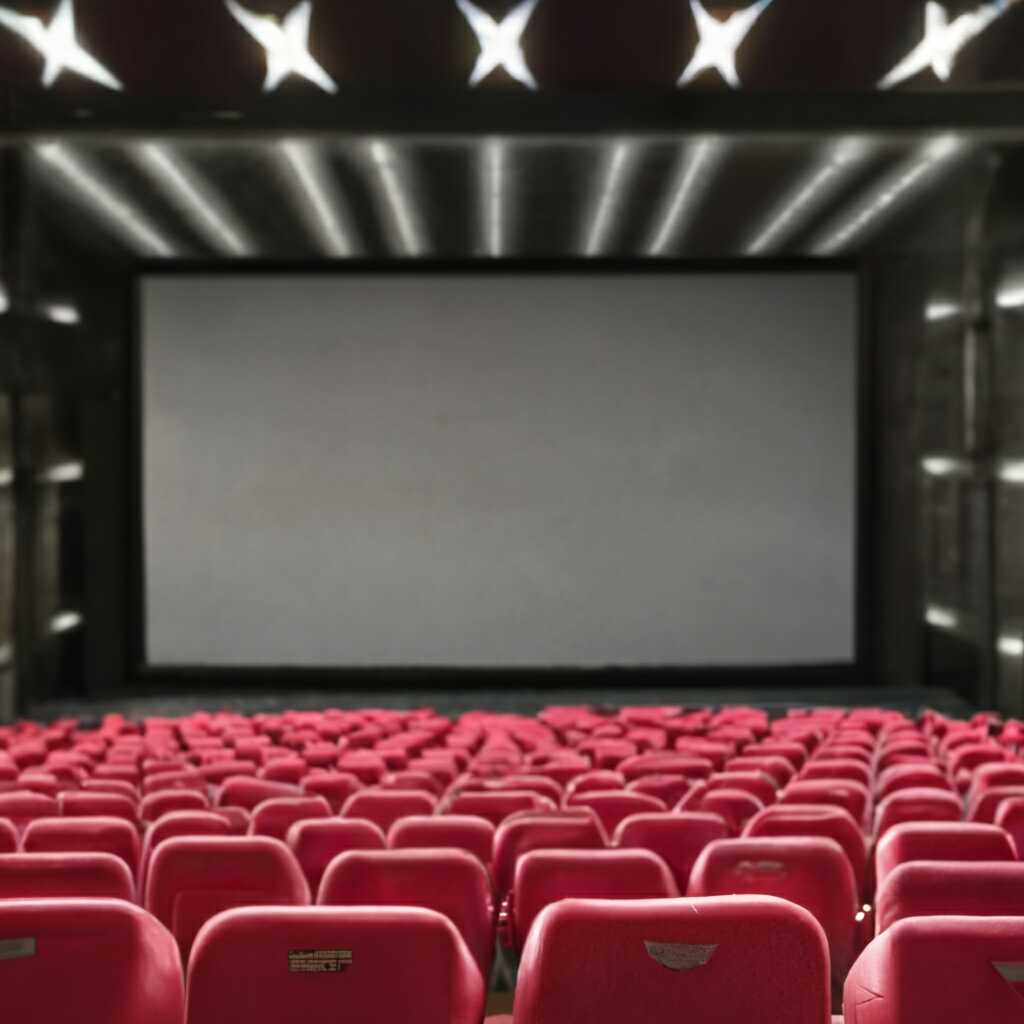} & \includegraphics[width=\imwidth\linewidth]{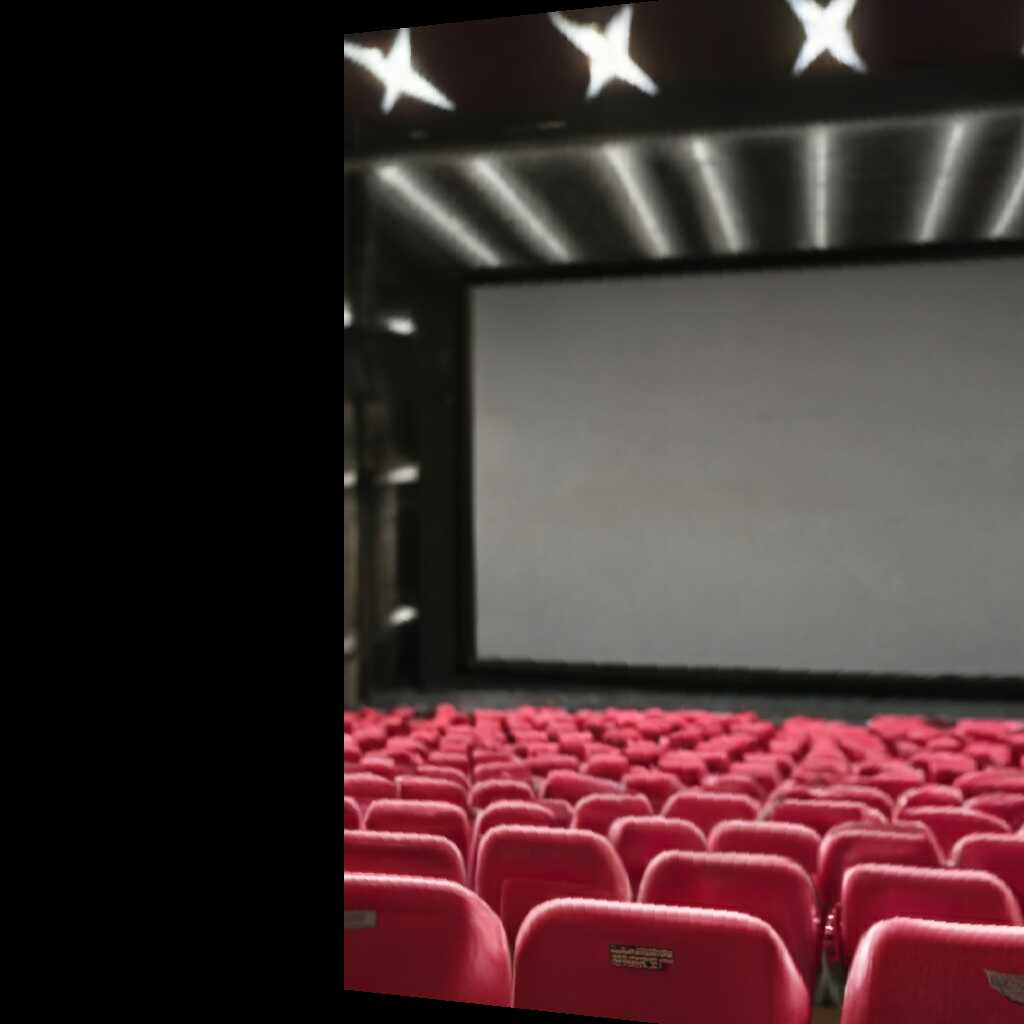} &\includegraphics[width=\imwidth\linewidth]{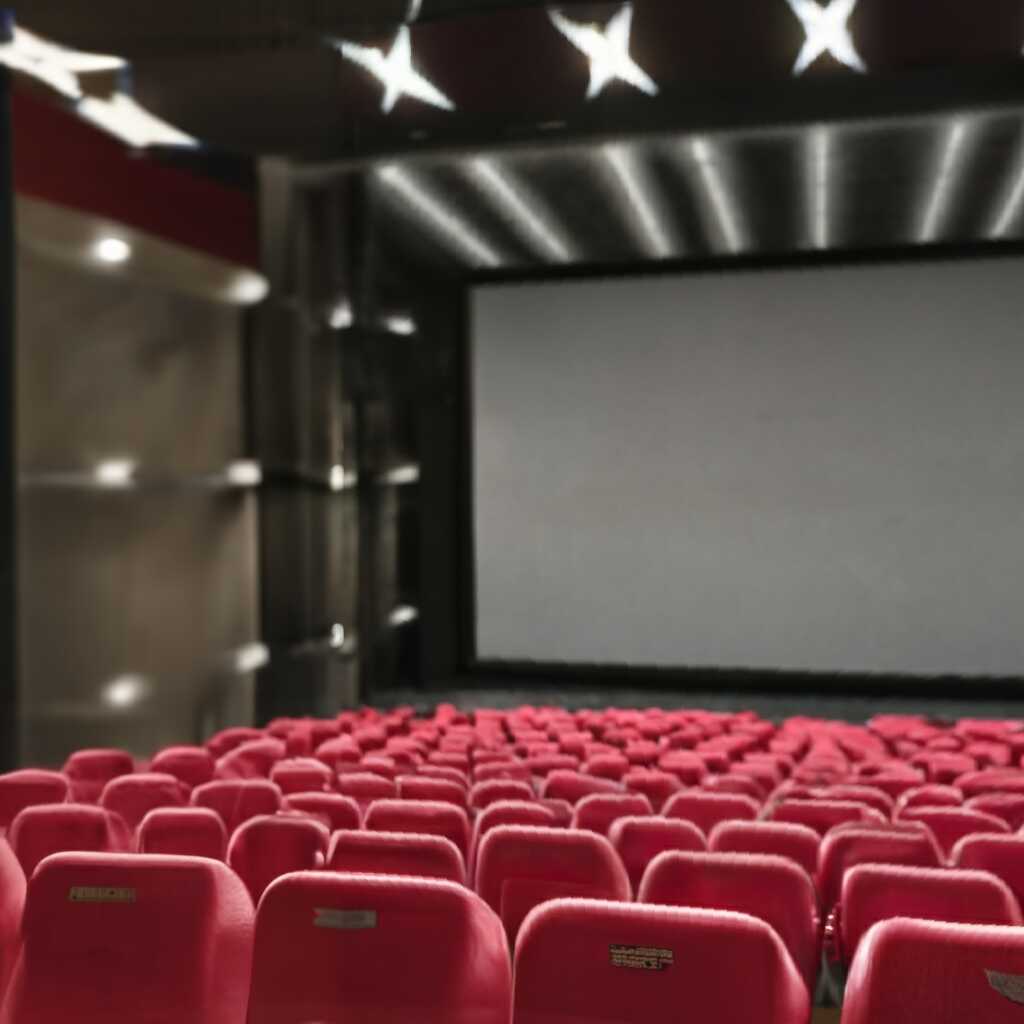} &\includegraphics[width=\imwidth\linewidth]{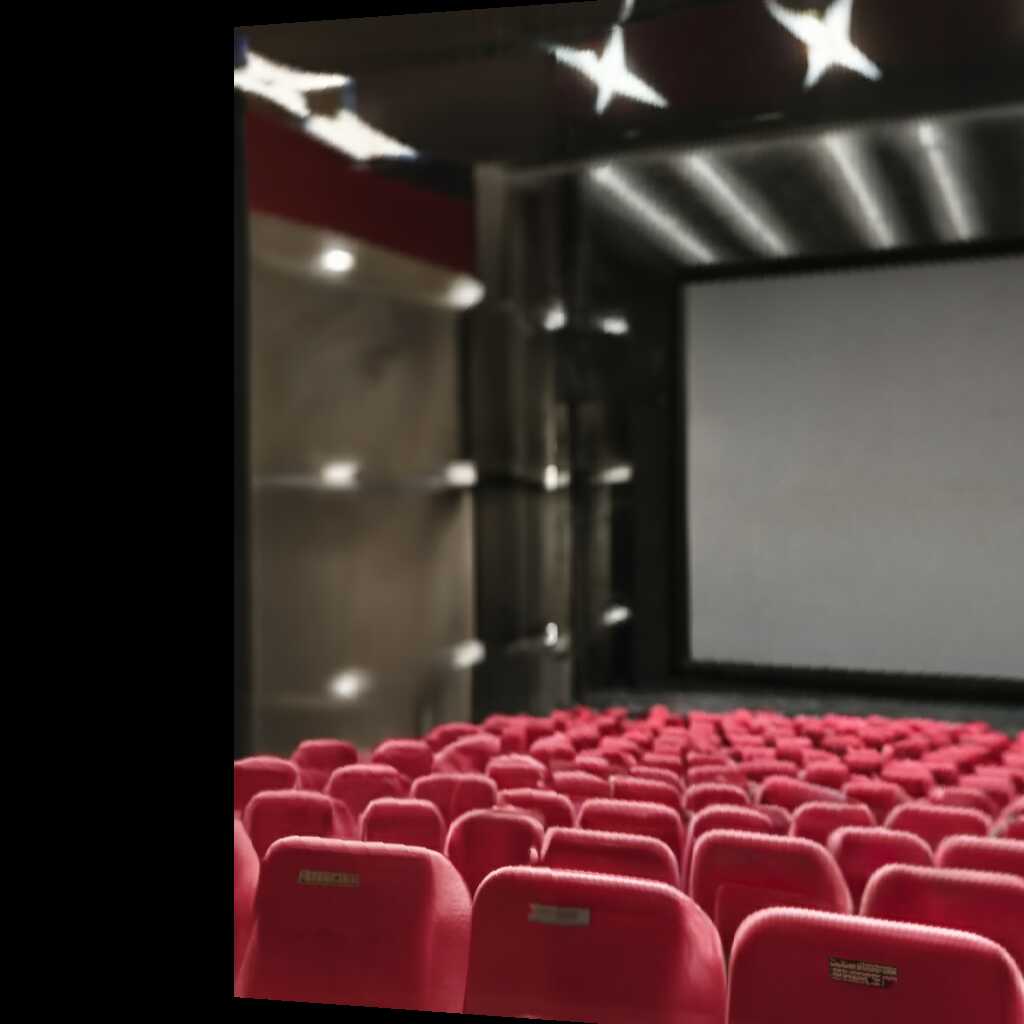} &\includegraphics[width=\imwidth\linewidth]{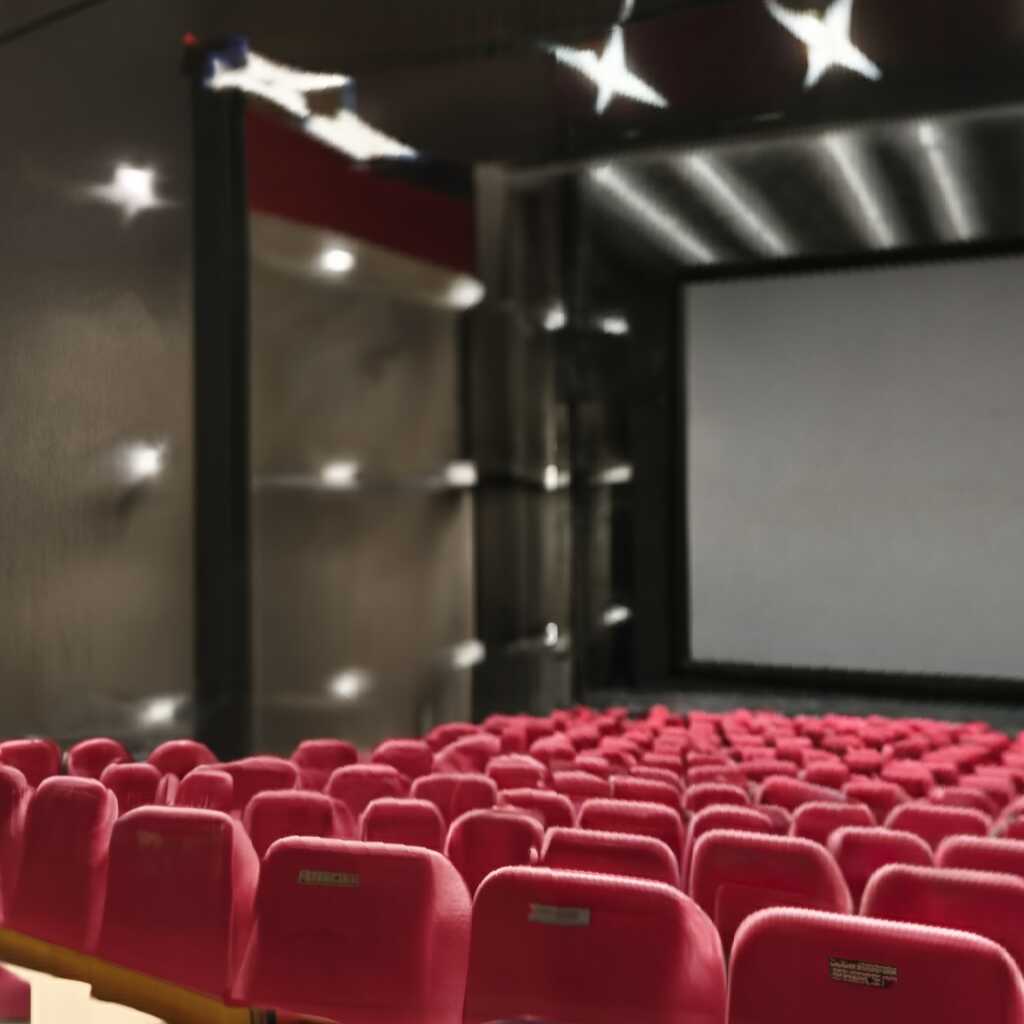} & \includegraphics[width=\imwidth\linewidth]{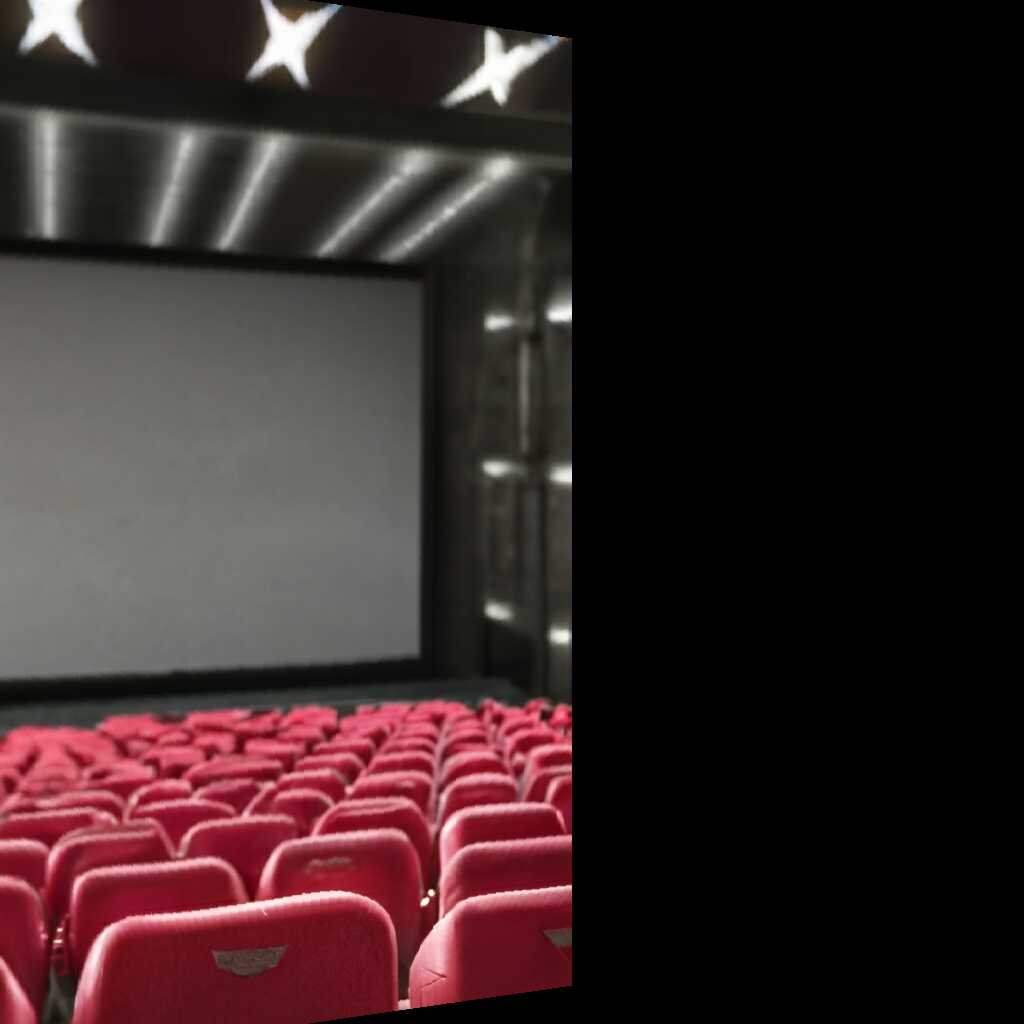} &\includegraphics[width=\imwidth\linewidth]{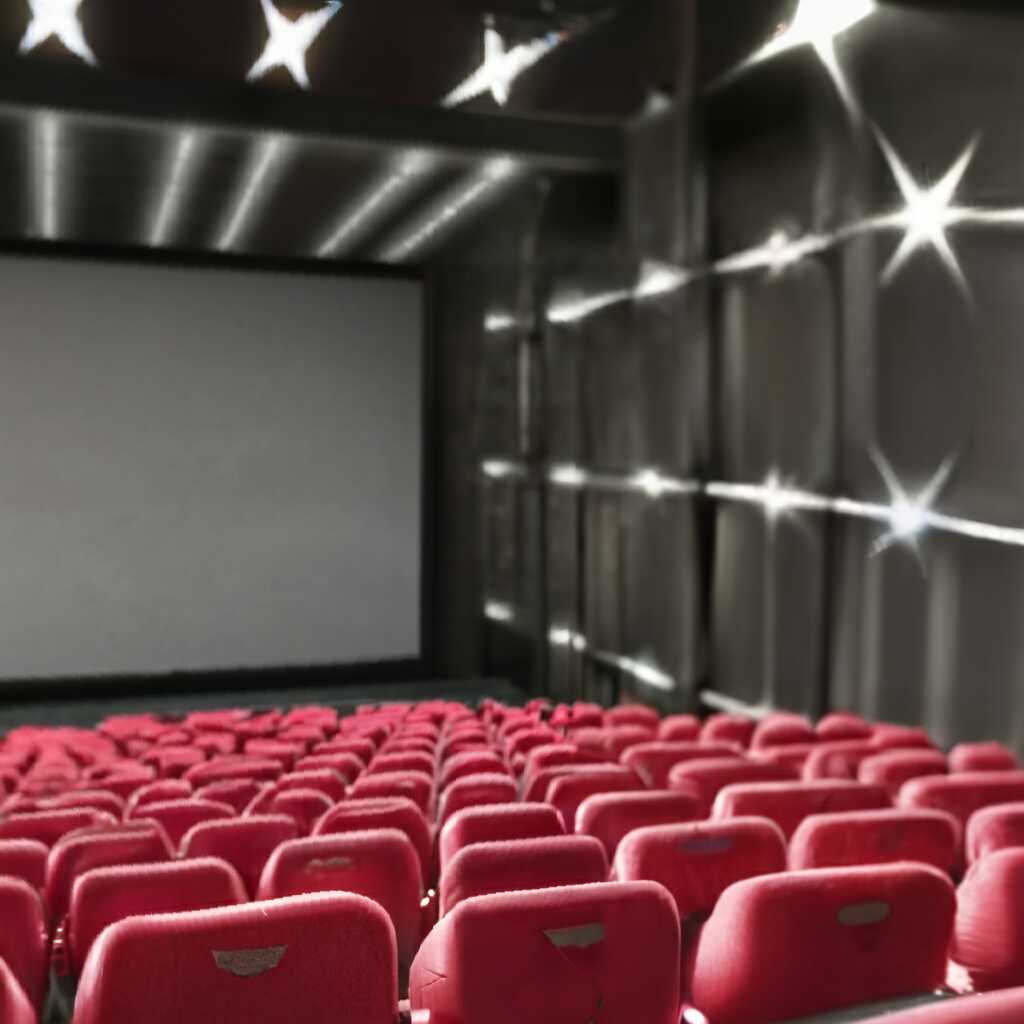} & \multicolumn{2}{c}{\multirow{2}{*}[\lastcoloffset]{\includegraphics[width=\imwidthlast\linewidth]{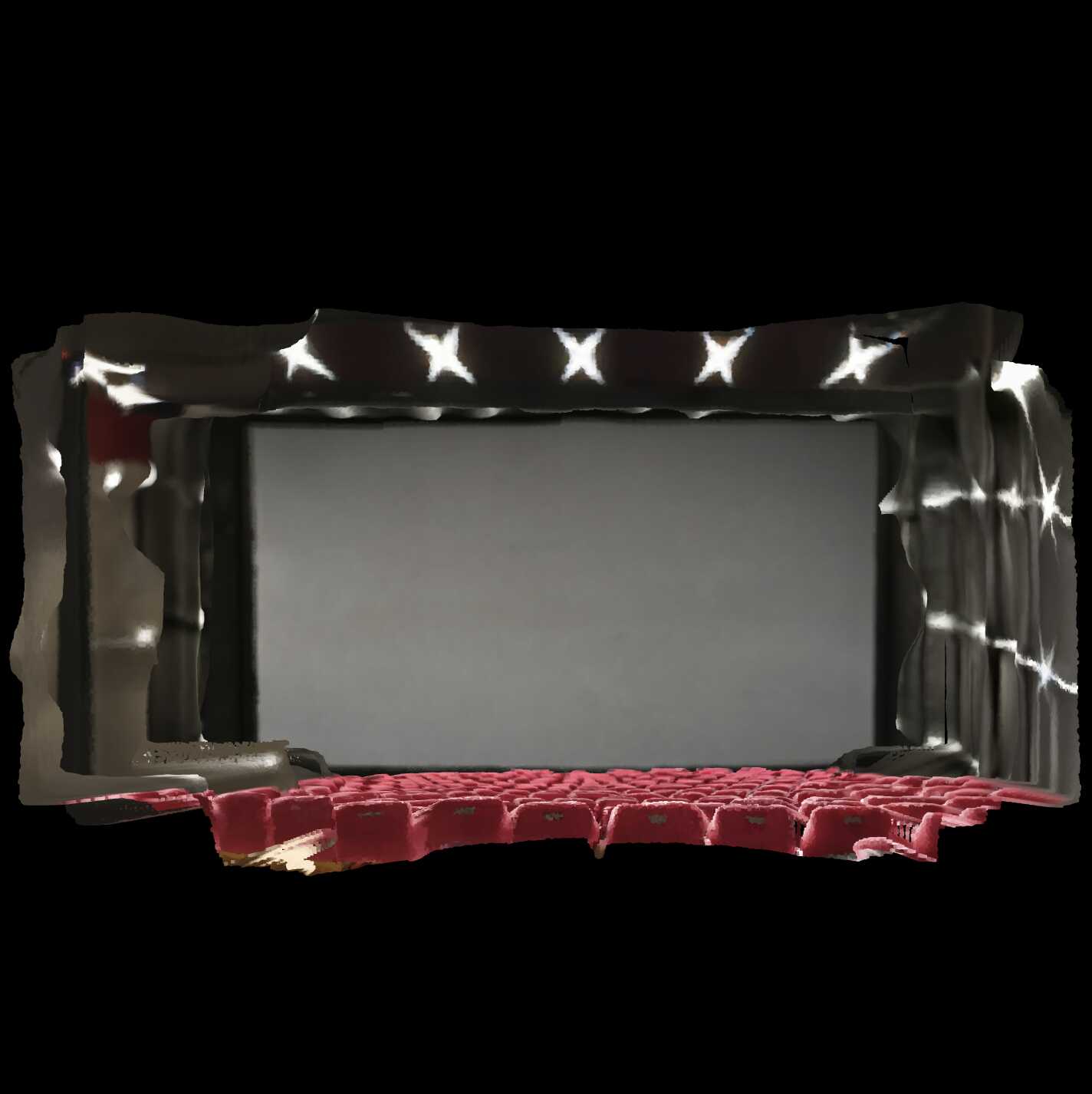}}} \\
    \includegraphics[width=\imwidth\linewidth]{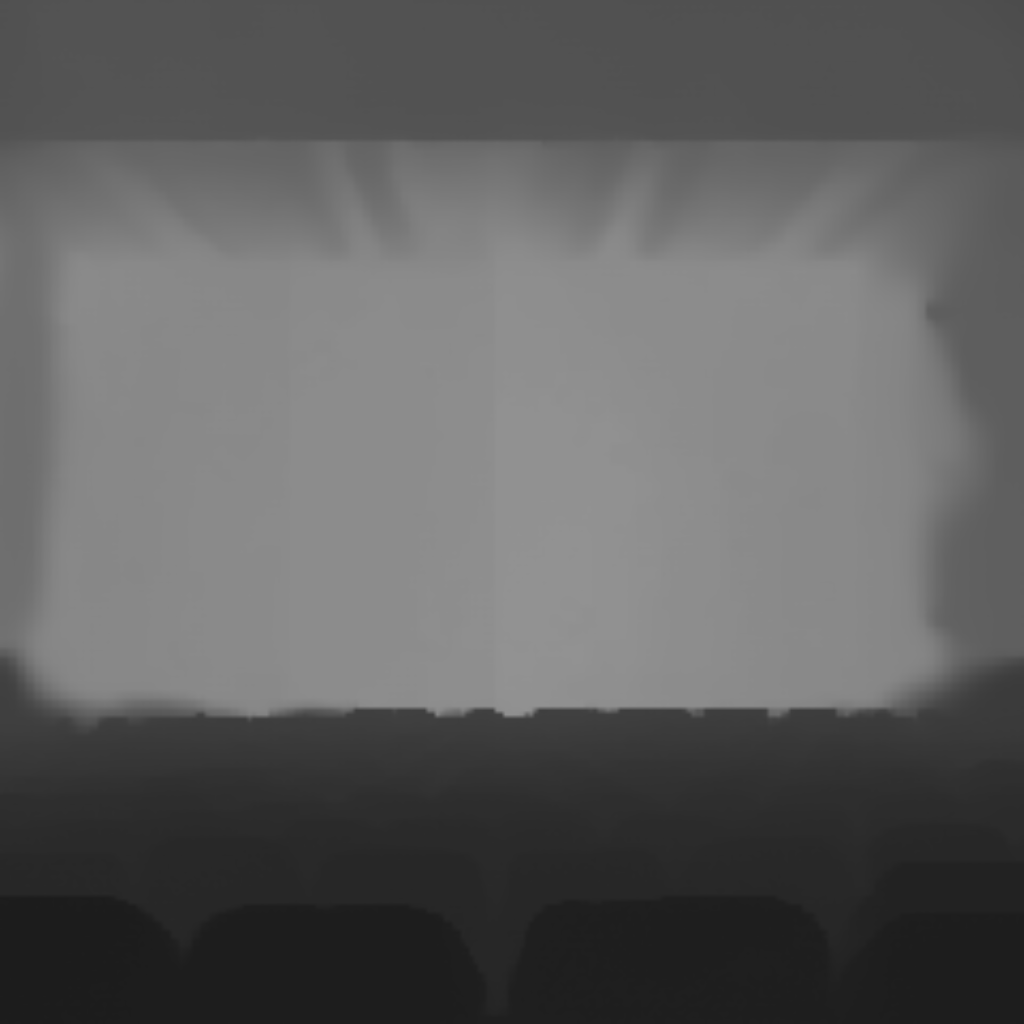} & \includegraphics[width=\imwidth\linewidth]{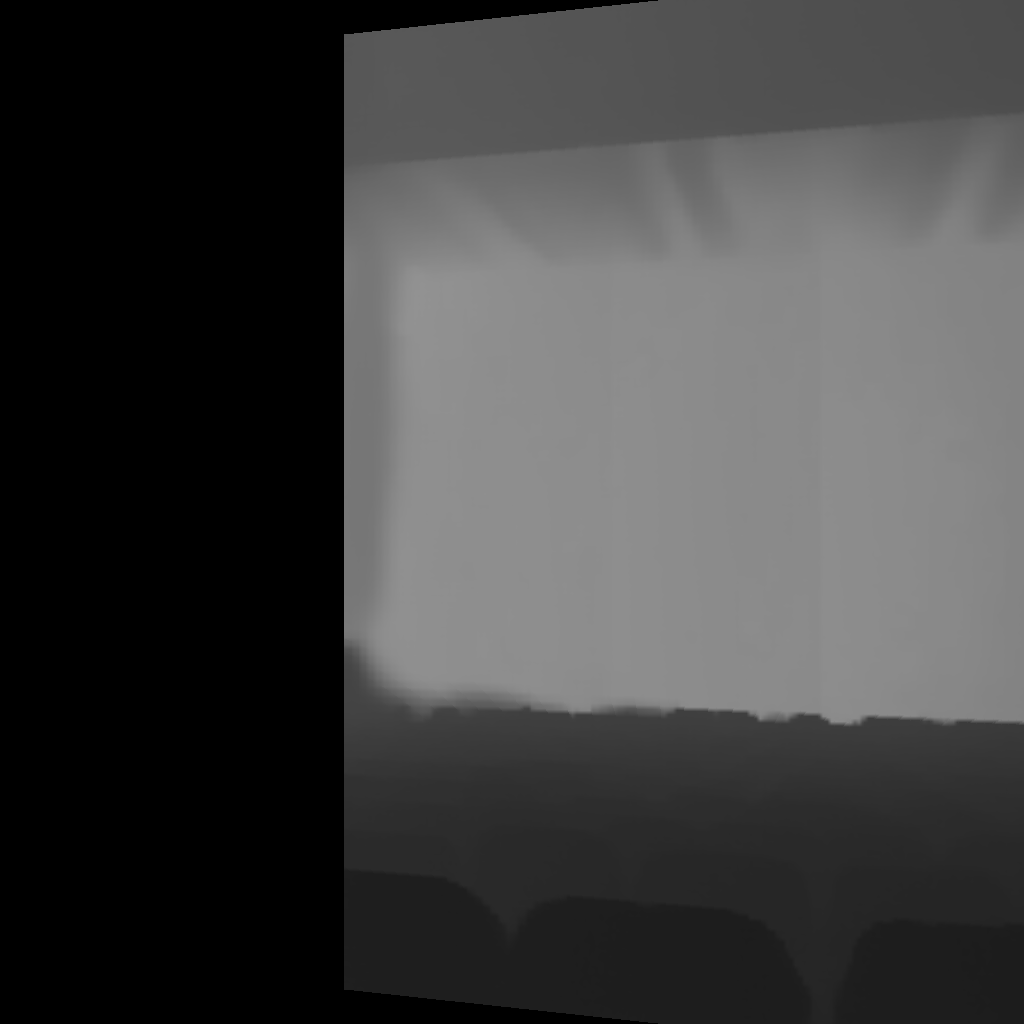} &\includegraphics[width=\imwidth\linewidth]{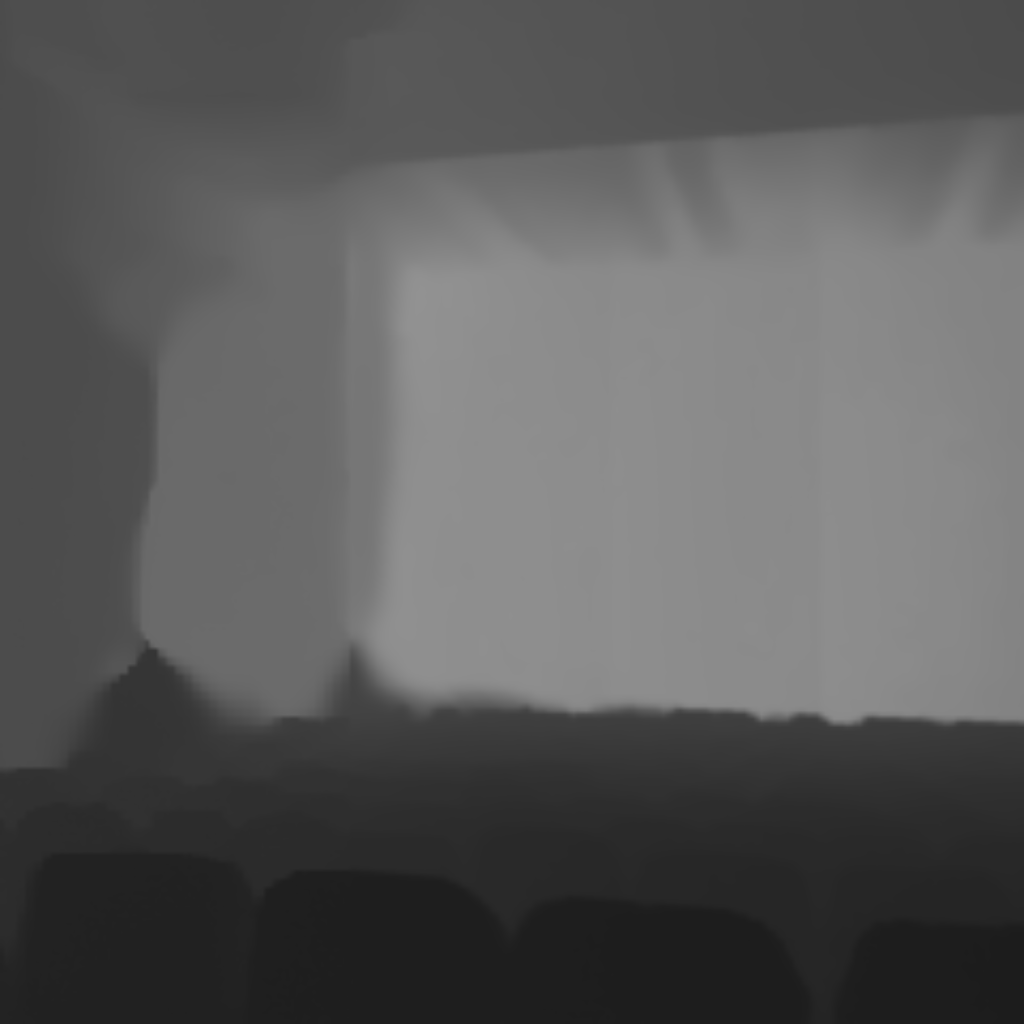} &  \includegraphics[width=\imwidth\linewidth]{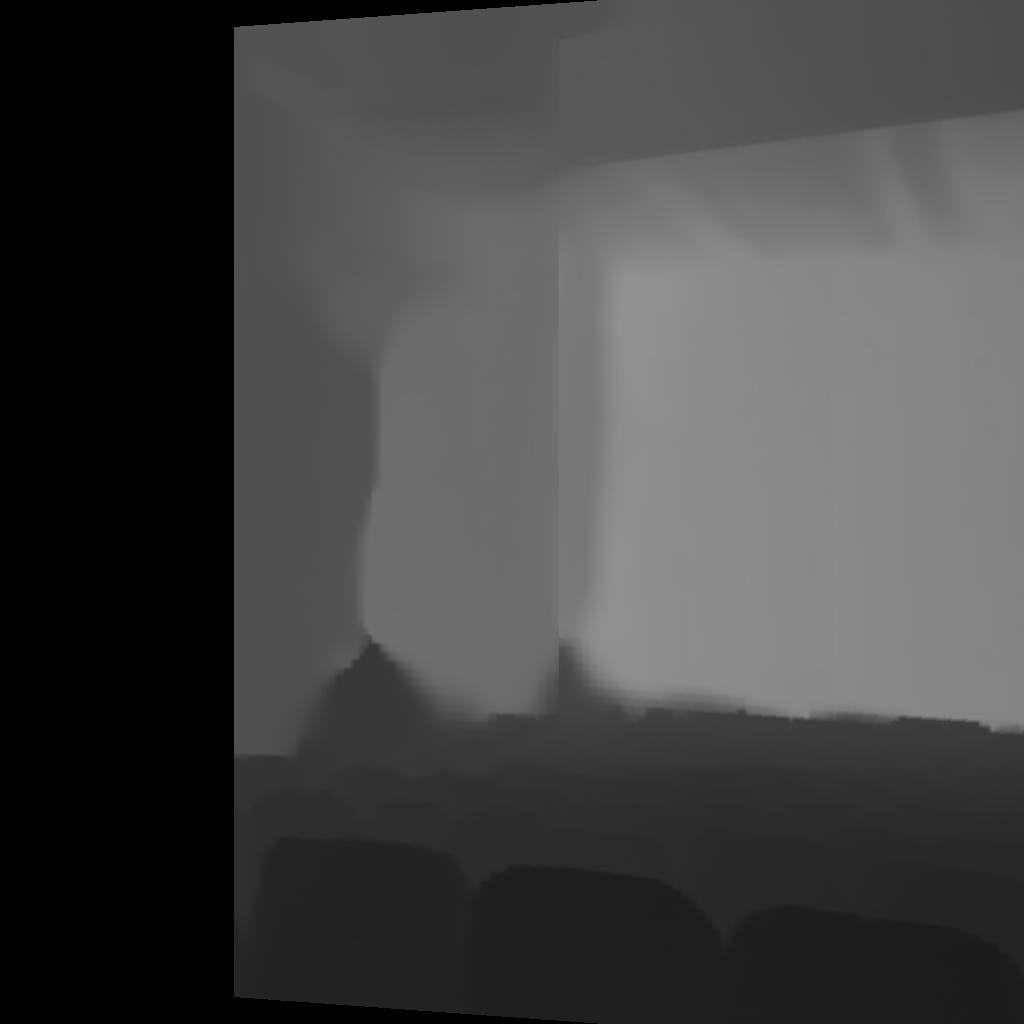} &\includegraphics[width=\imwidth\linewidth]{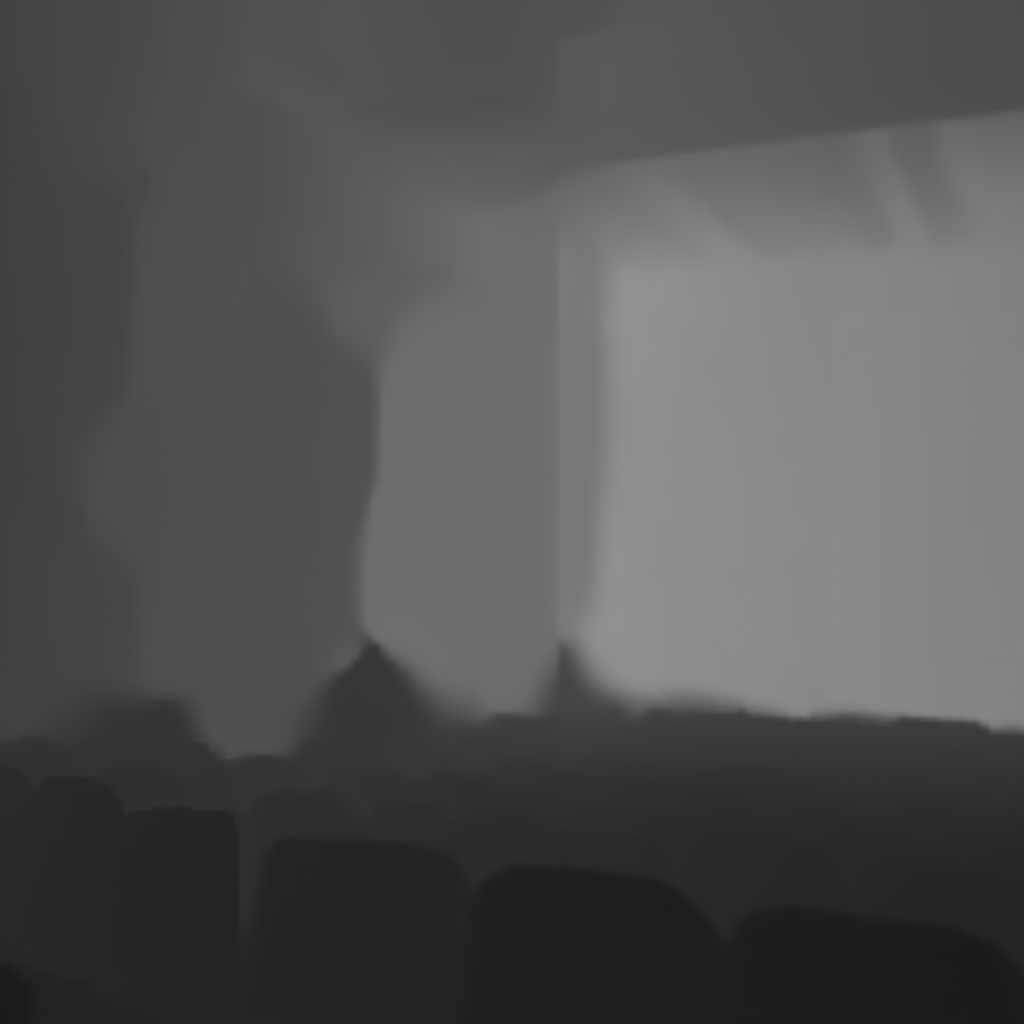} &\includegraphics[width=\imwidth\linewidth]{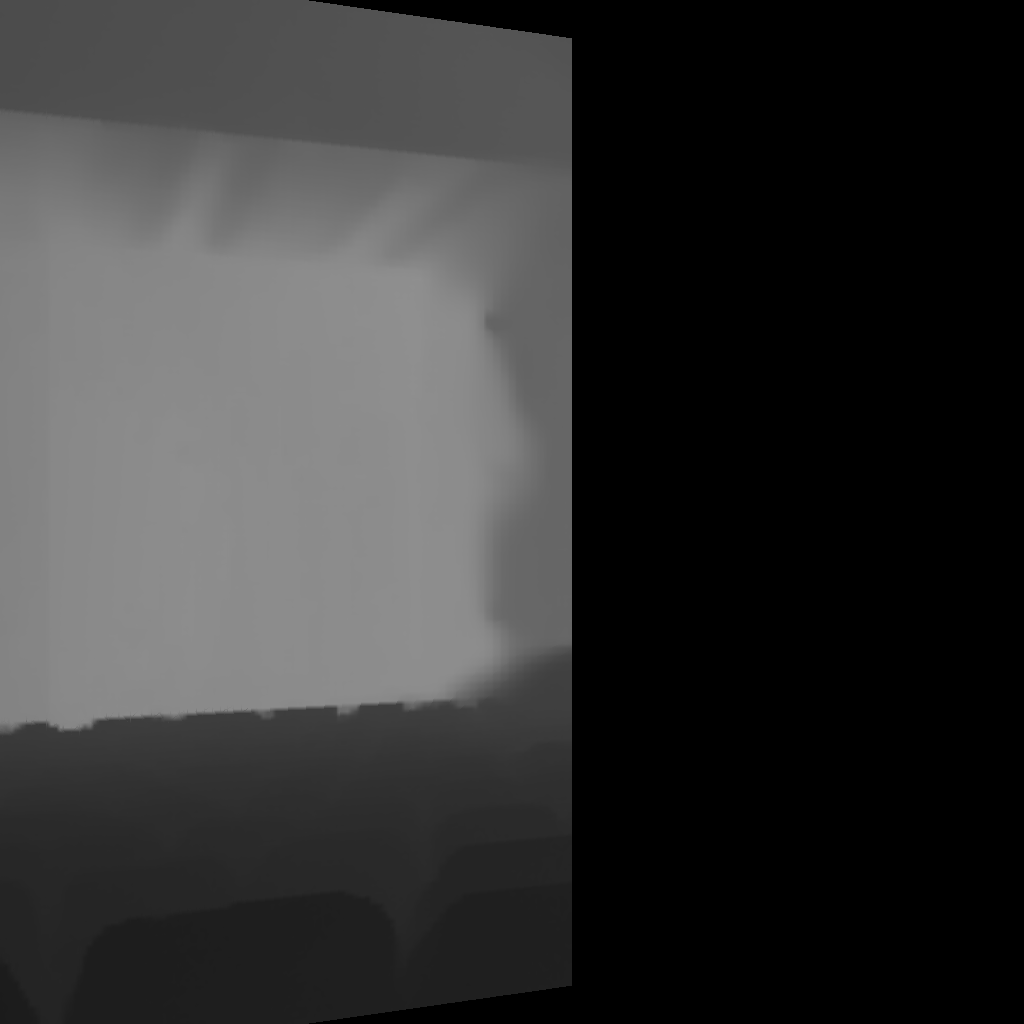} &\includegraphics[width=\imwidth\linewidth]{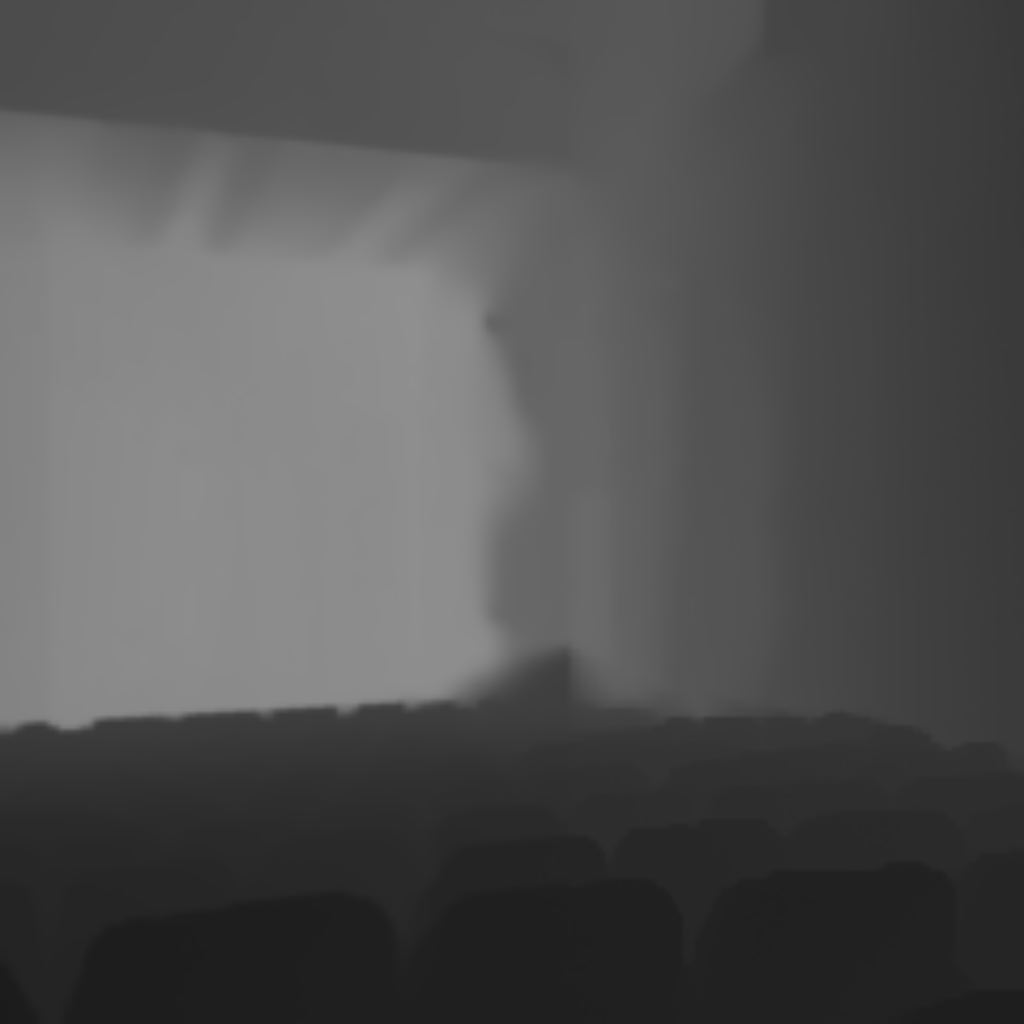} & \multicolumn{2}{c}{} \\
    \\[\textpadding]\multicolumn{\ncols}{c}{Prompt: A movie theatre.}\\\\[\textpadding]
\end{tabular}
\end{figure*}

\begin{figure*}
\centering
\renewcommand{\arraystretch}{0.2}
\addtolength{\tabcolsep}{-5pt}

\begin{tabular}{ccccccccc}
    \includegraphics[width=\imwidth\linewidth]{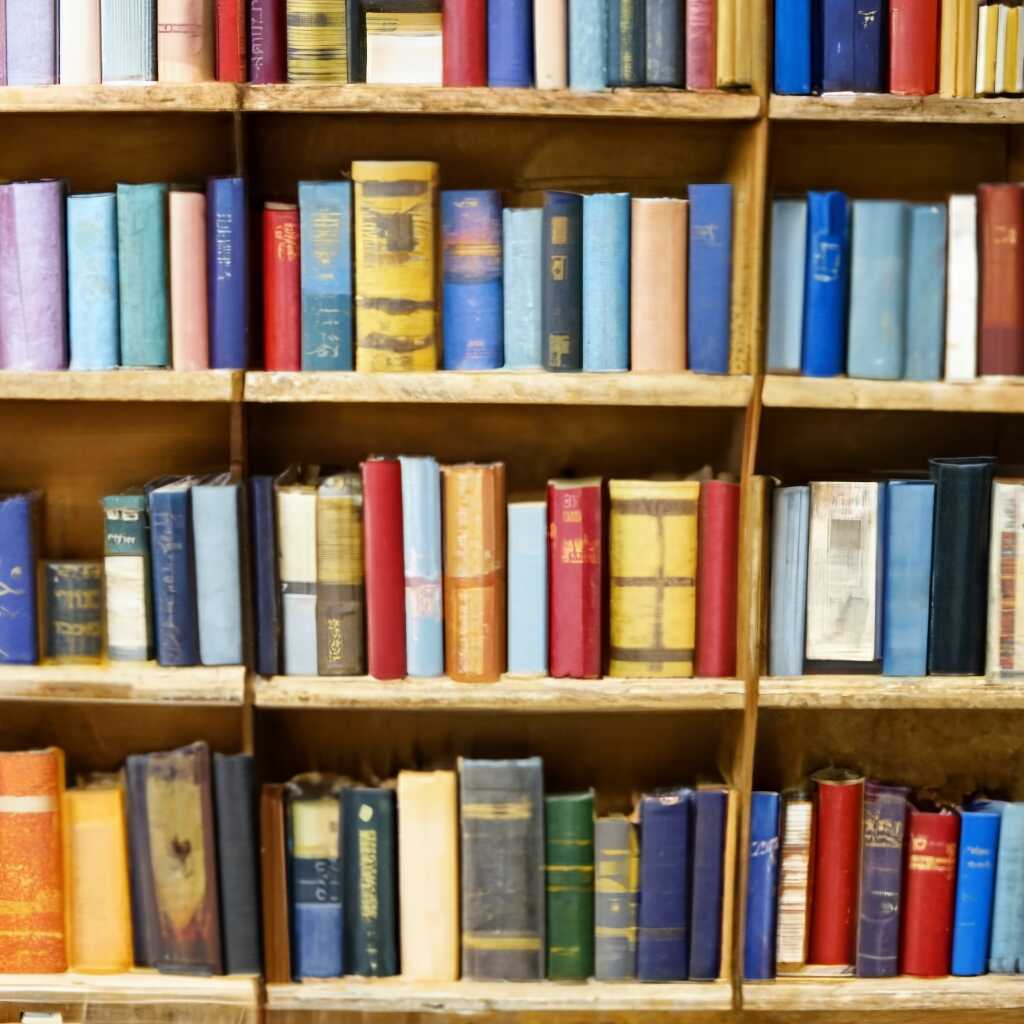} & \includegraphics[width=\imwidth\linewidth]{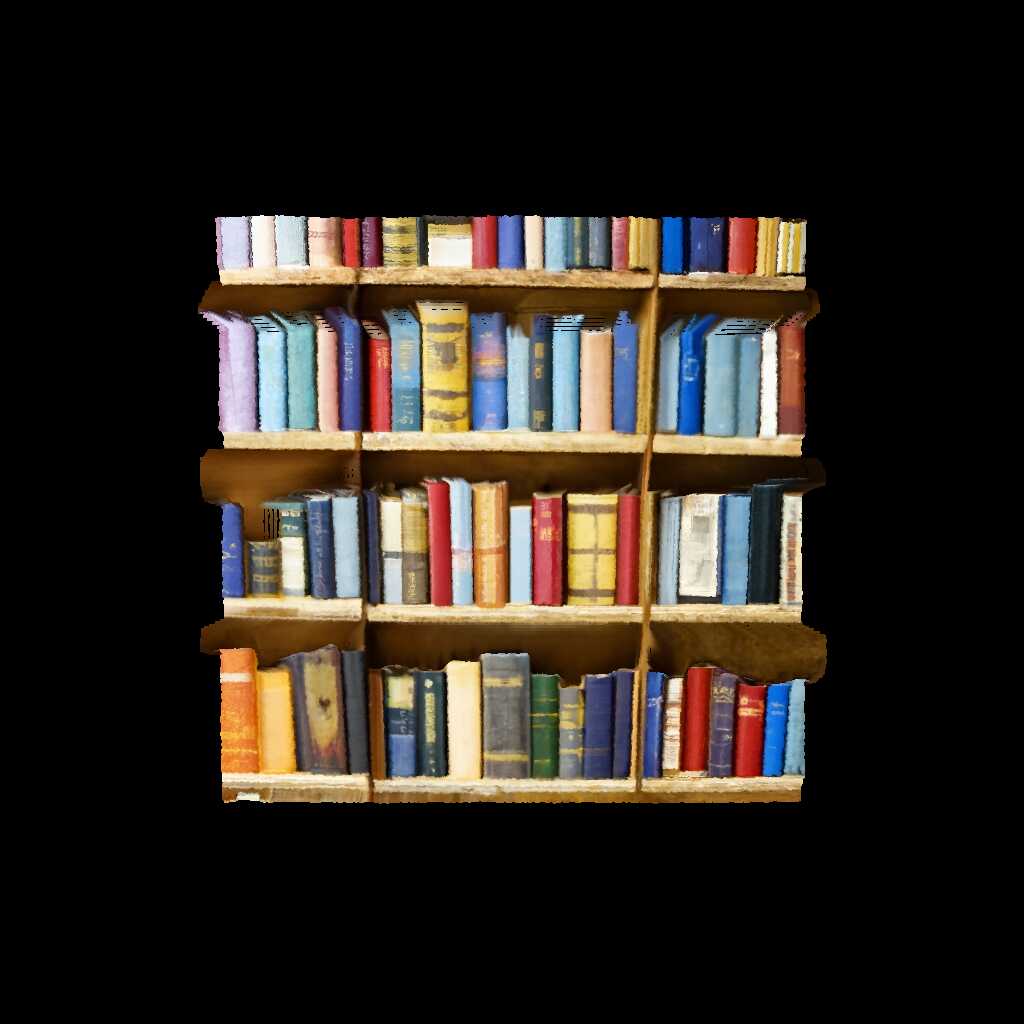} &\includegraphics[width=\imwidth\linewidth]{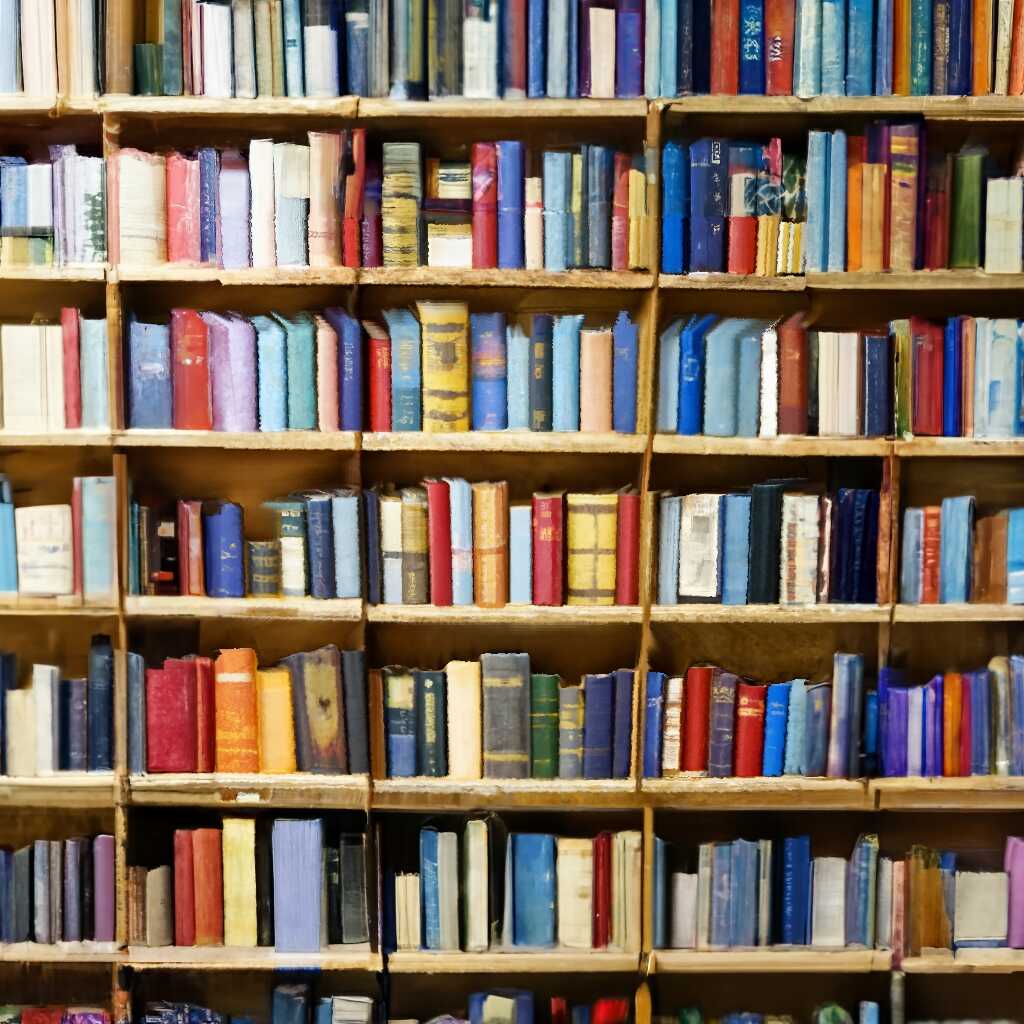} &\includegraphics[width=\imwidth\linewidth]{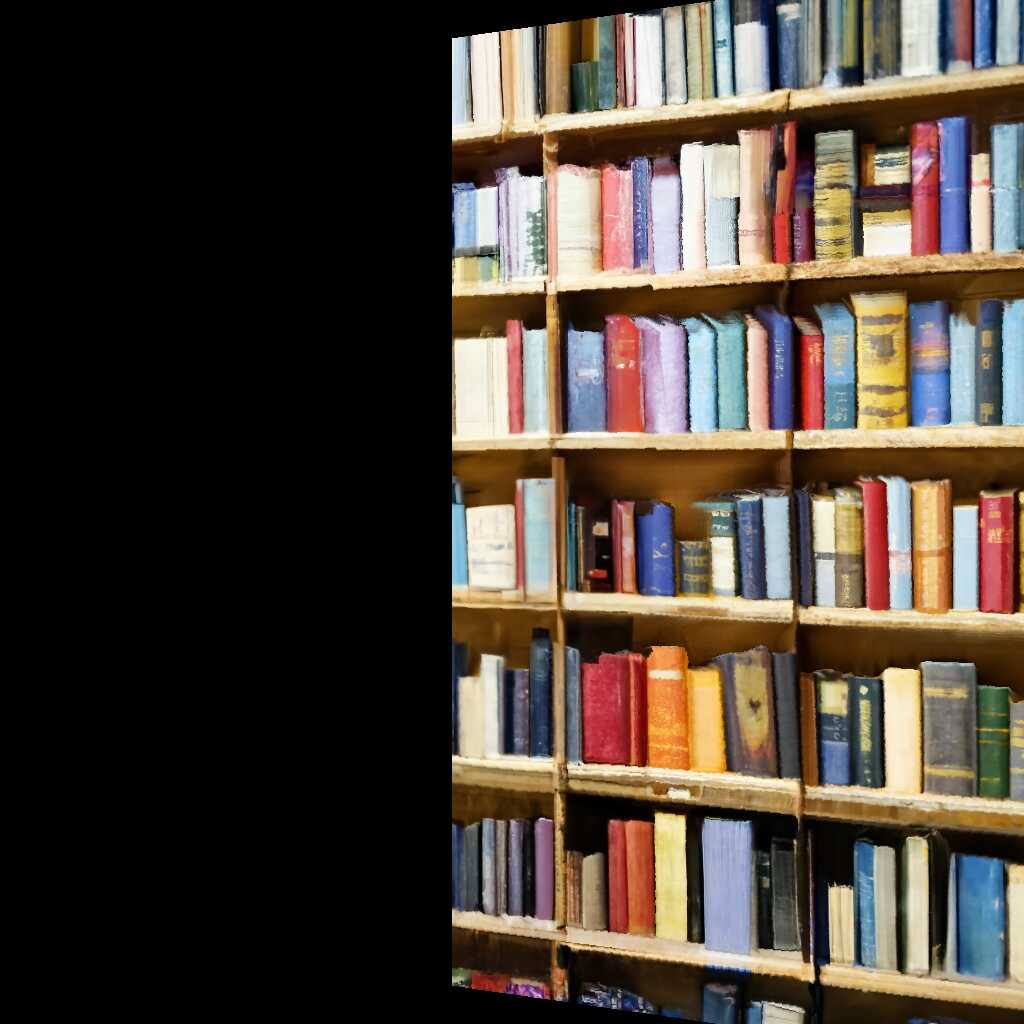} &\includegraphics[width=\imwidth\linewidth]{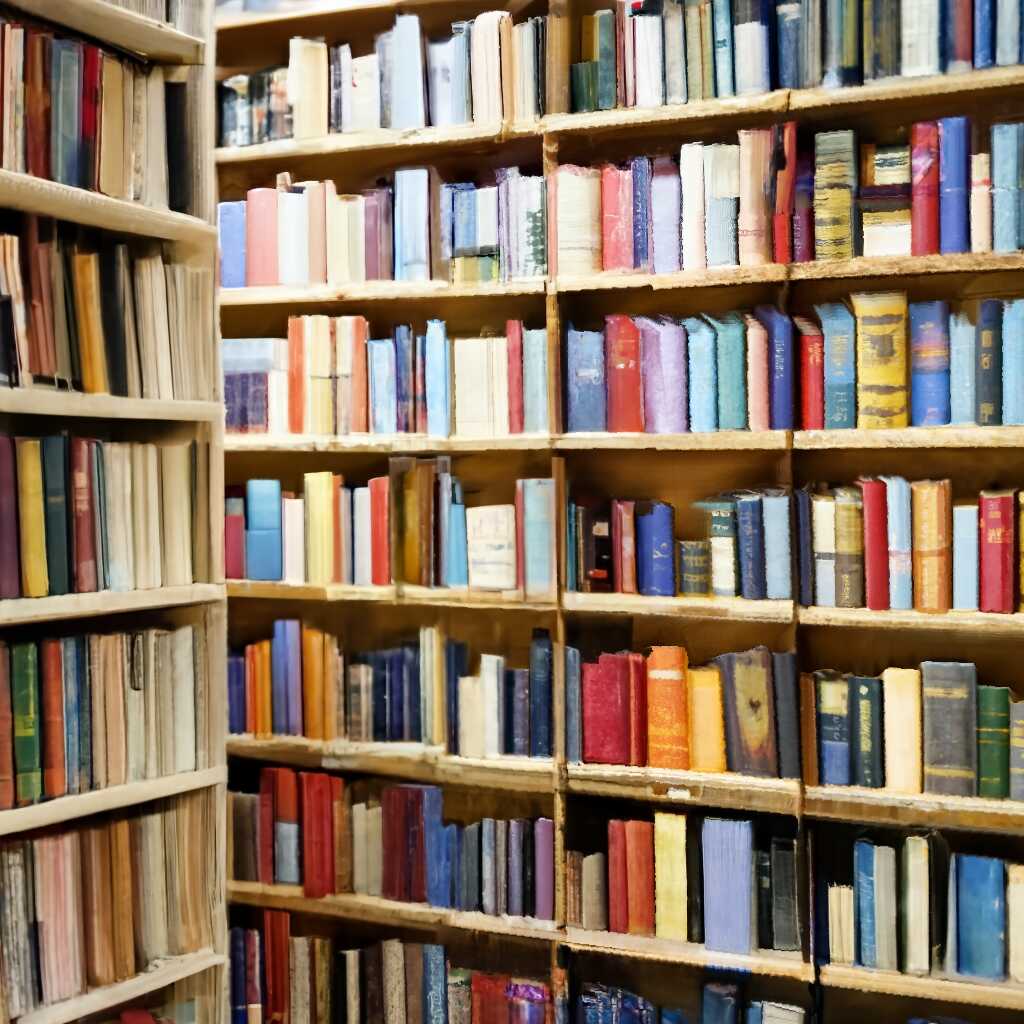} & \includegraphics[width=\imwidth\linewidth]{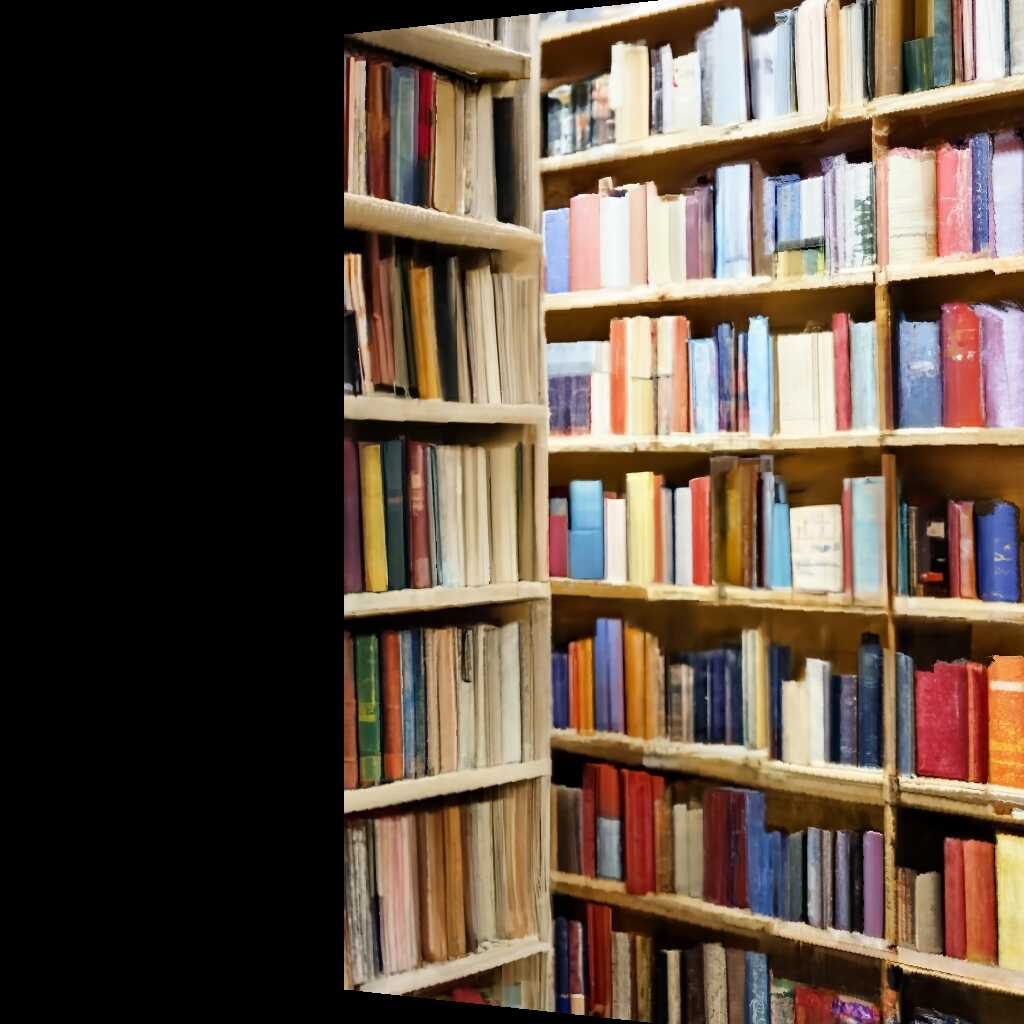} &\includegraphics[width=\imwidth\linewidth]{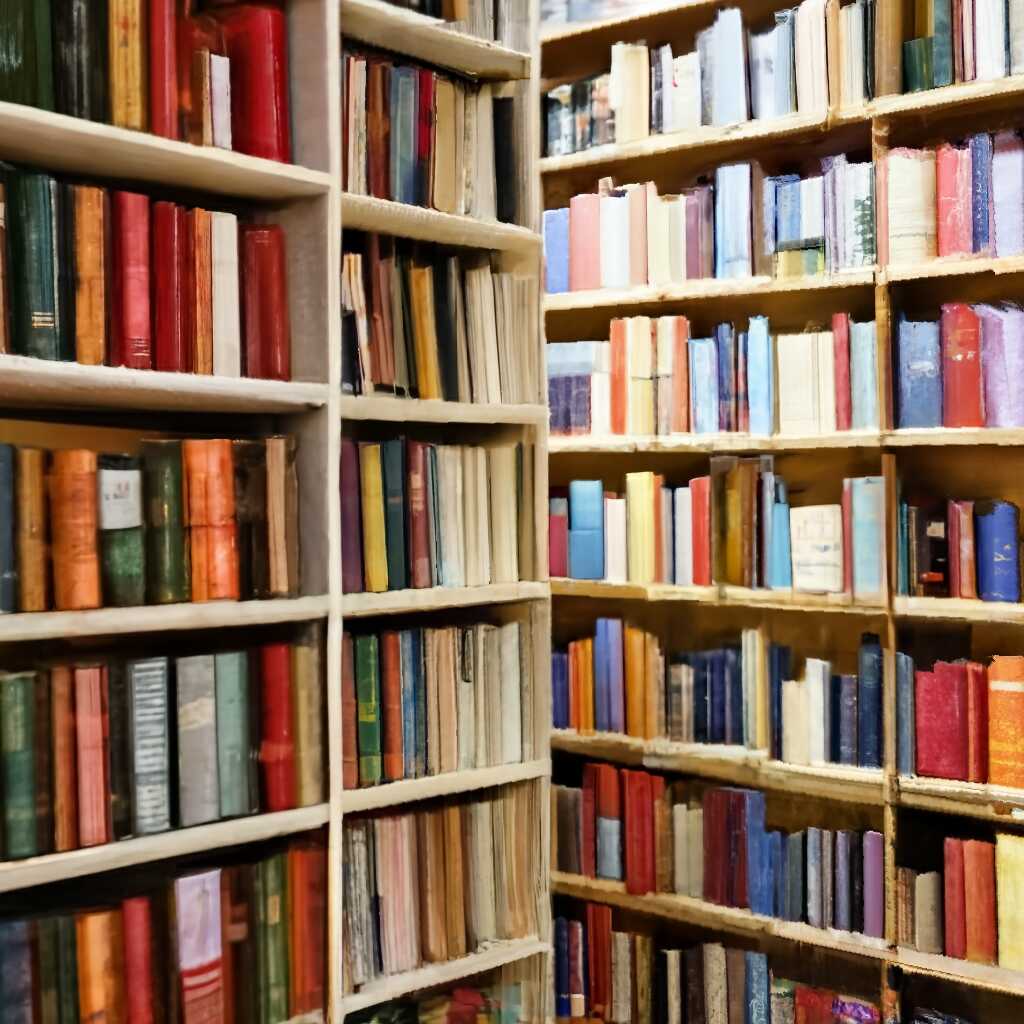} & \multicolumn{2}{c}{\multirow{2}{*}[\lastcoloffset]{\includegraphics[width=\imwidthlast\linewidth]{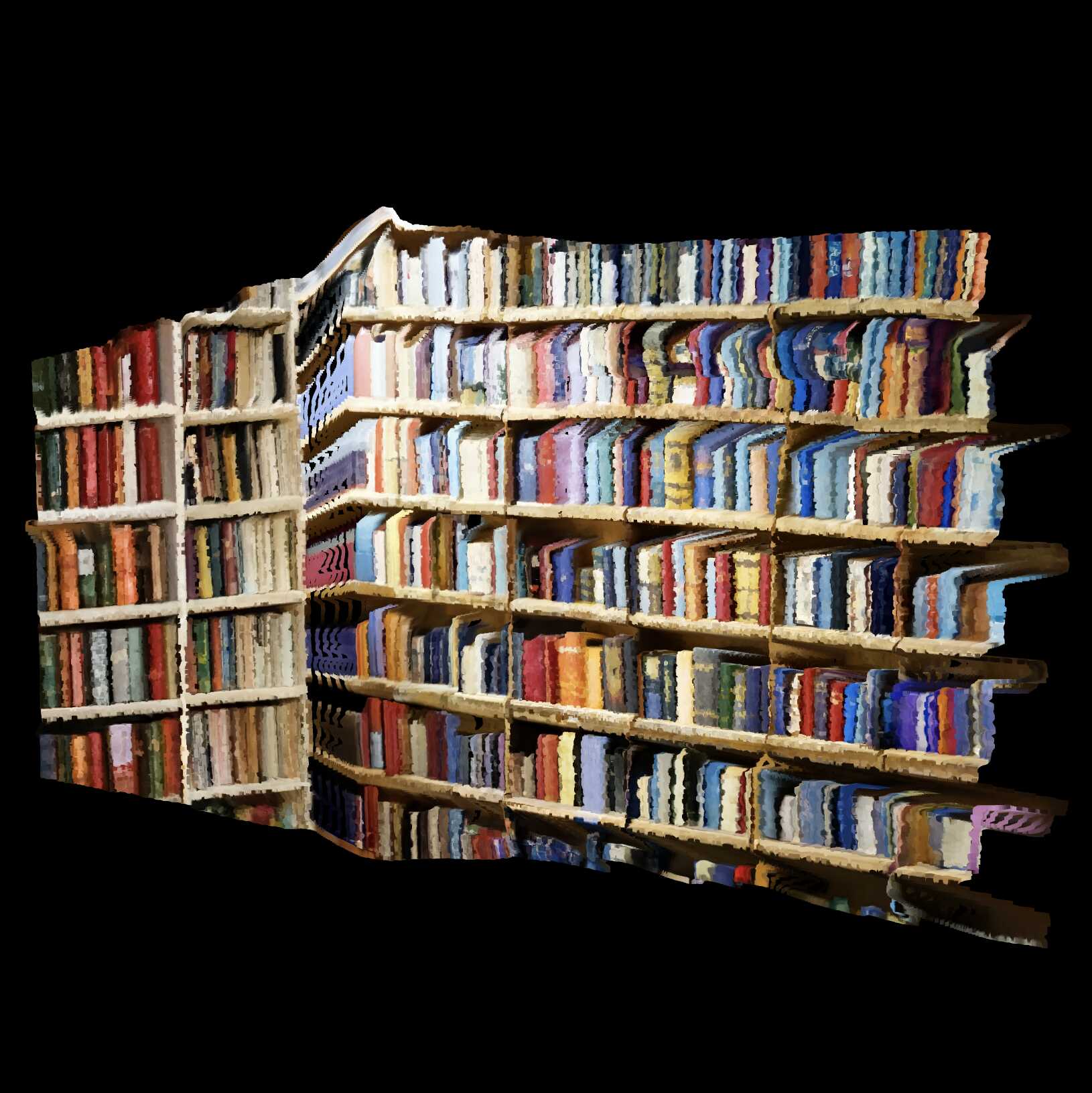}}} \\
    \includegraphics[width=\imwidth\linewidth]{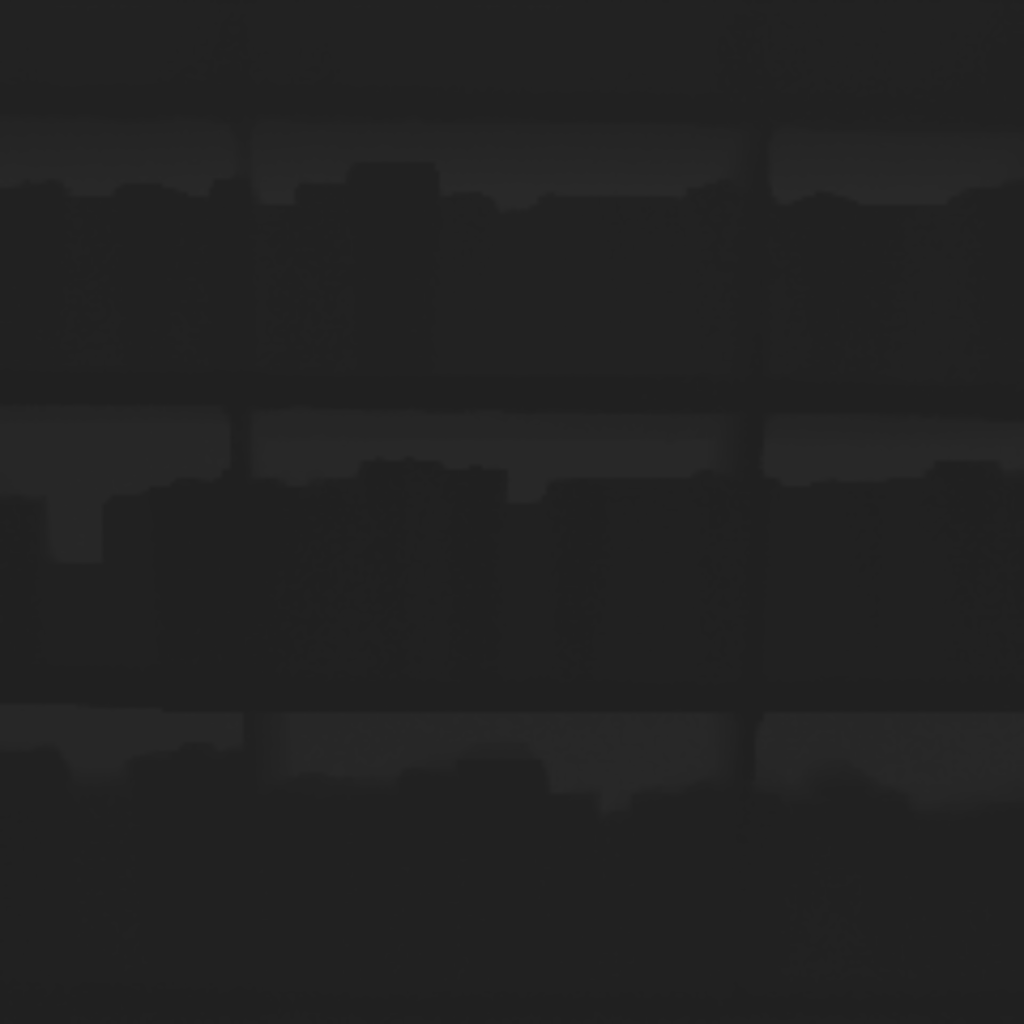} & \includegraphics[width=\imwidth\linewidth]{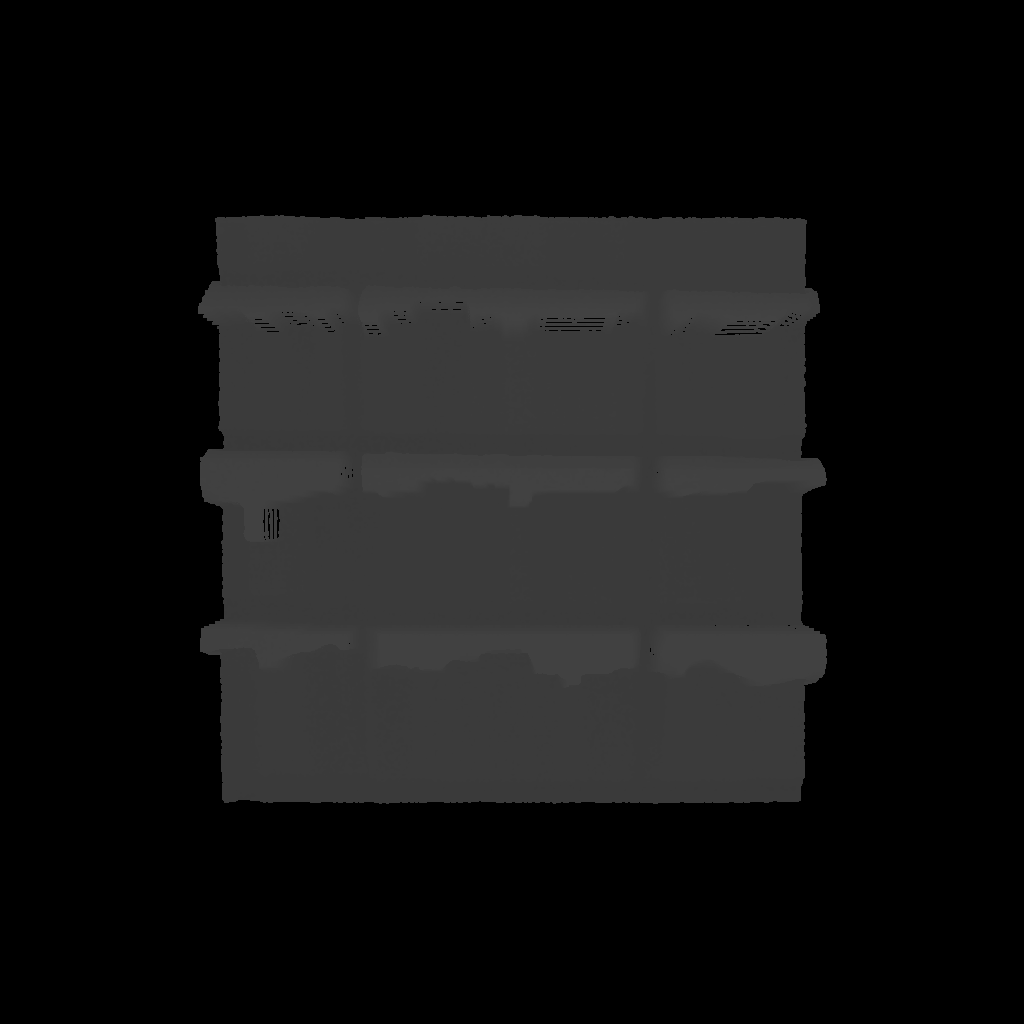} &\includegraphics[width=\imwidth\linewidth]{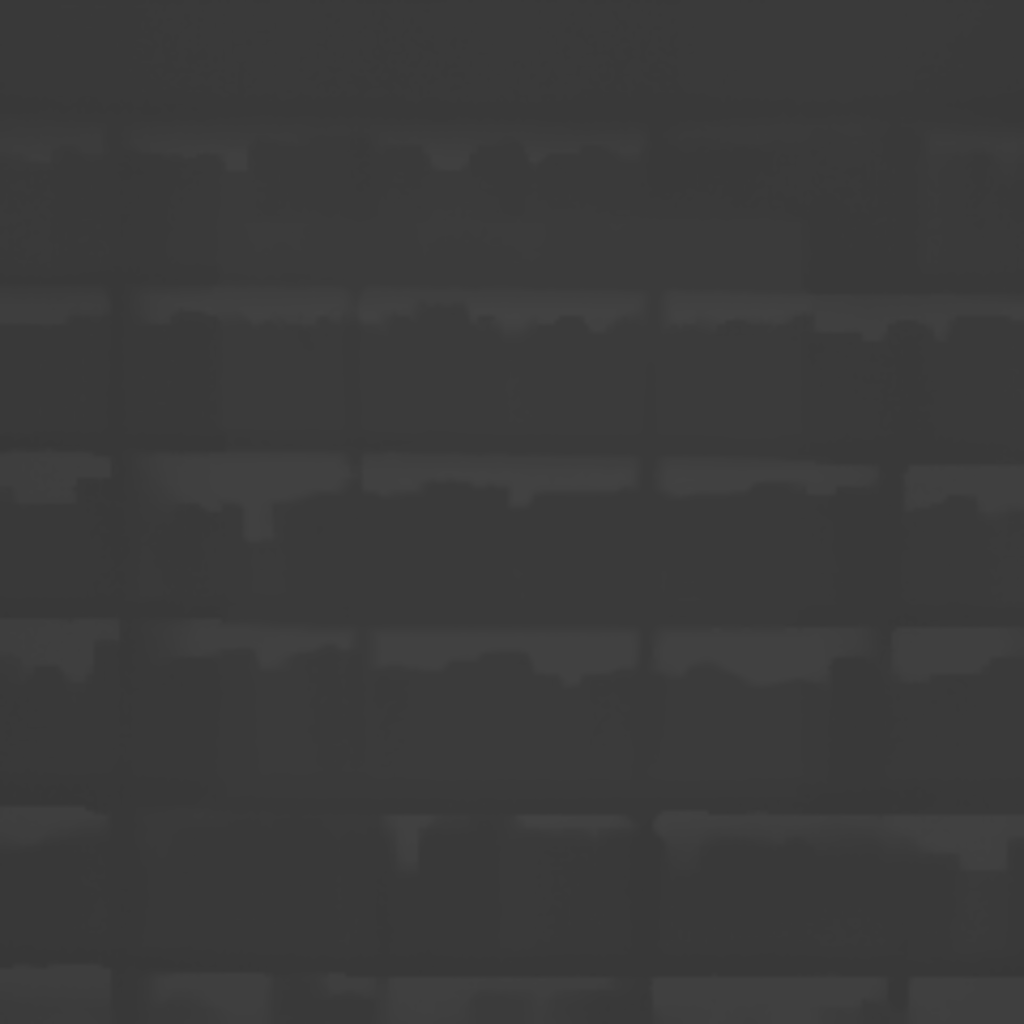} &  \includegraphics[width=\imwidth\linewidth]{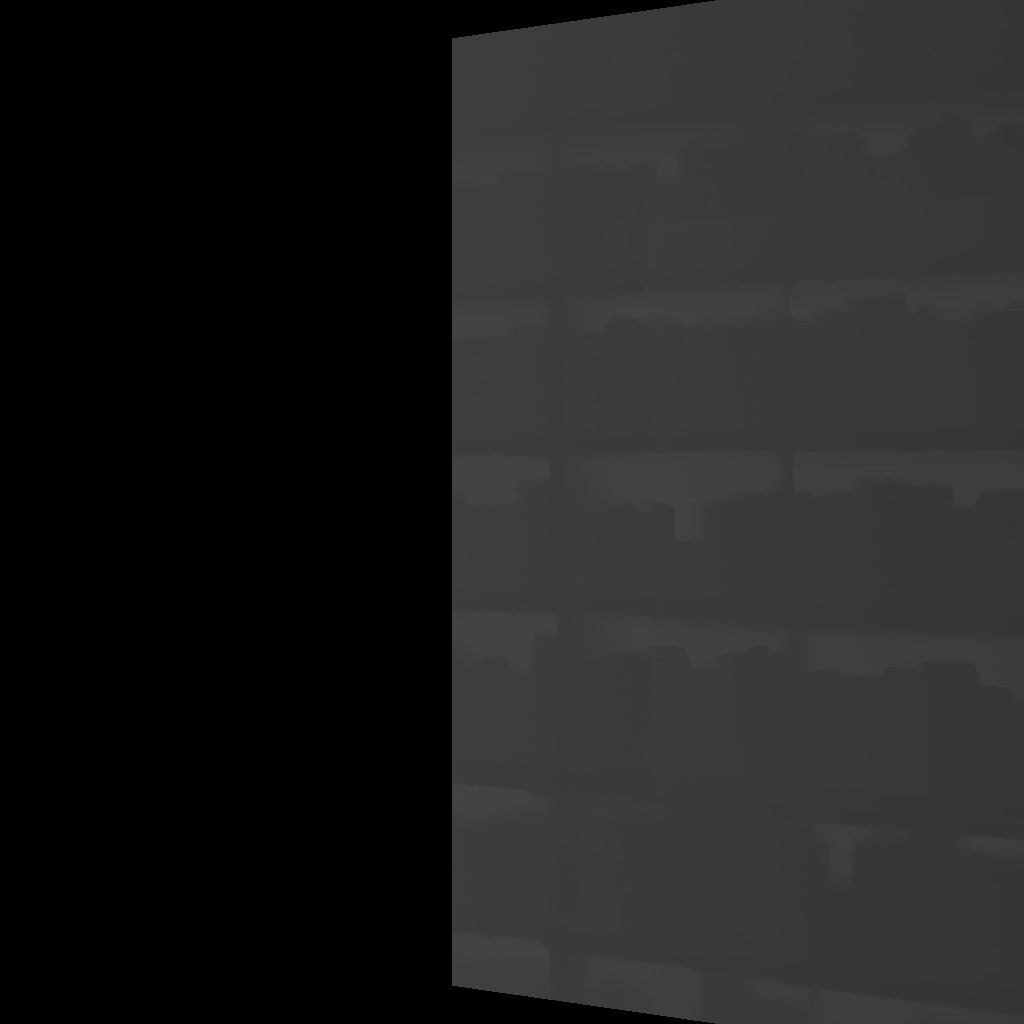} &\includegraphics[width=\imwidth\linewidth]{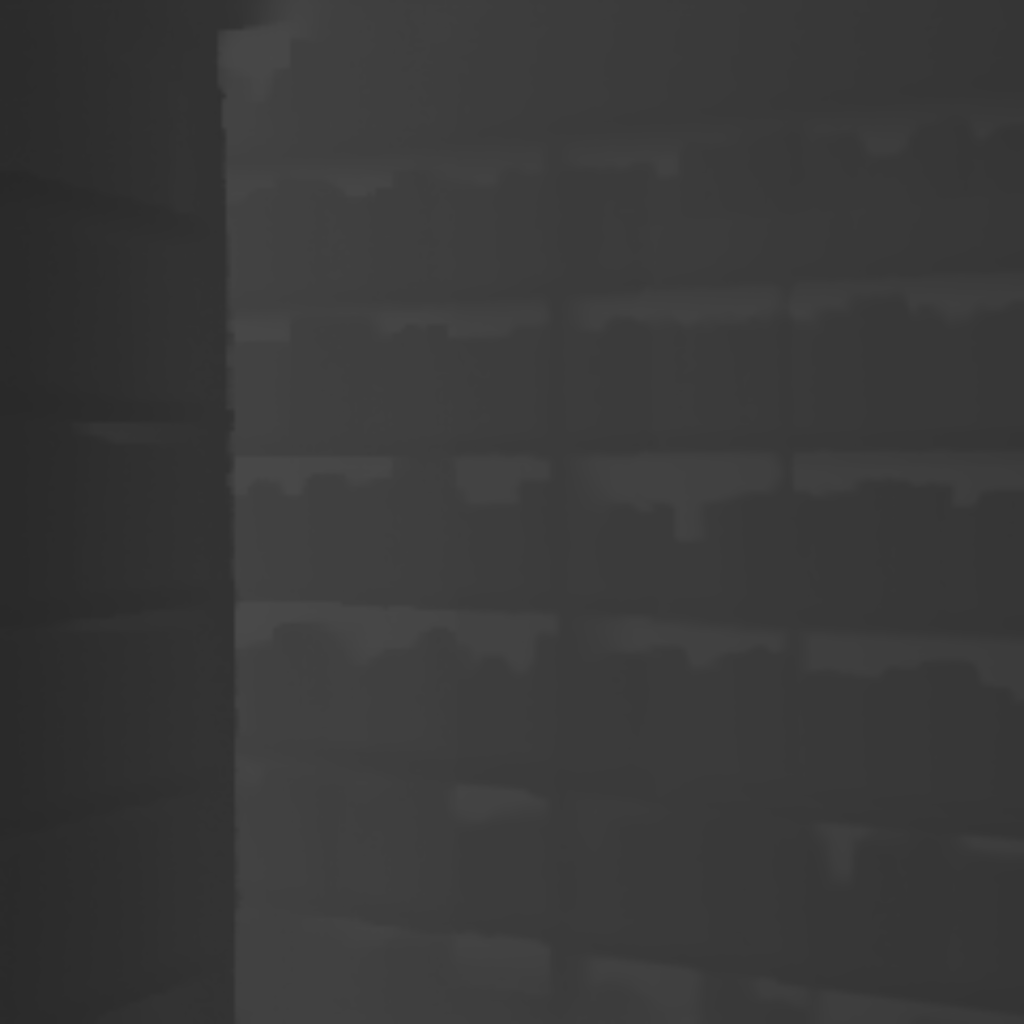} &\includegraphics[width=\imwidth\linewidth]{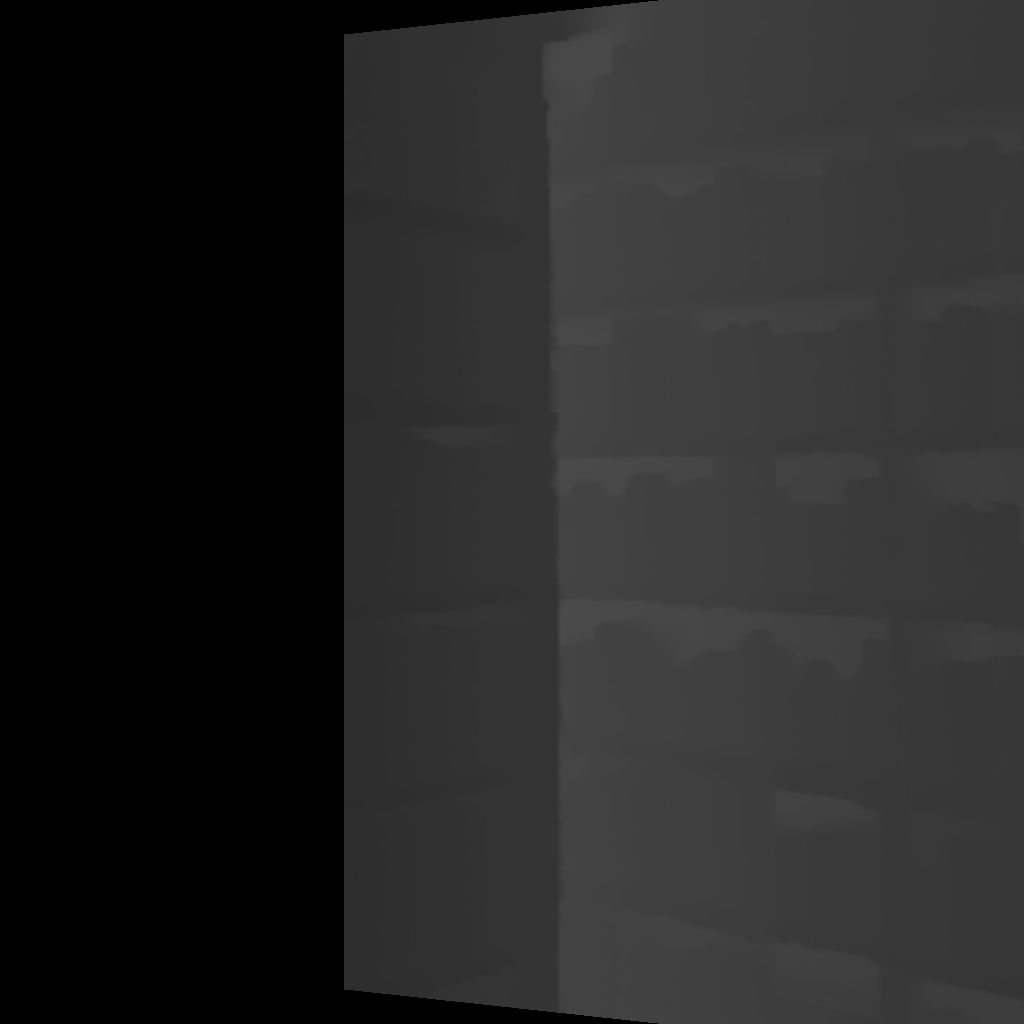} &\includegraphics[width=\imwidth\linewidth]{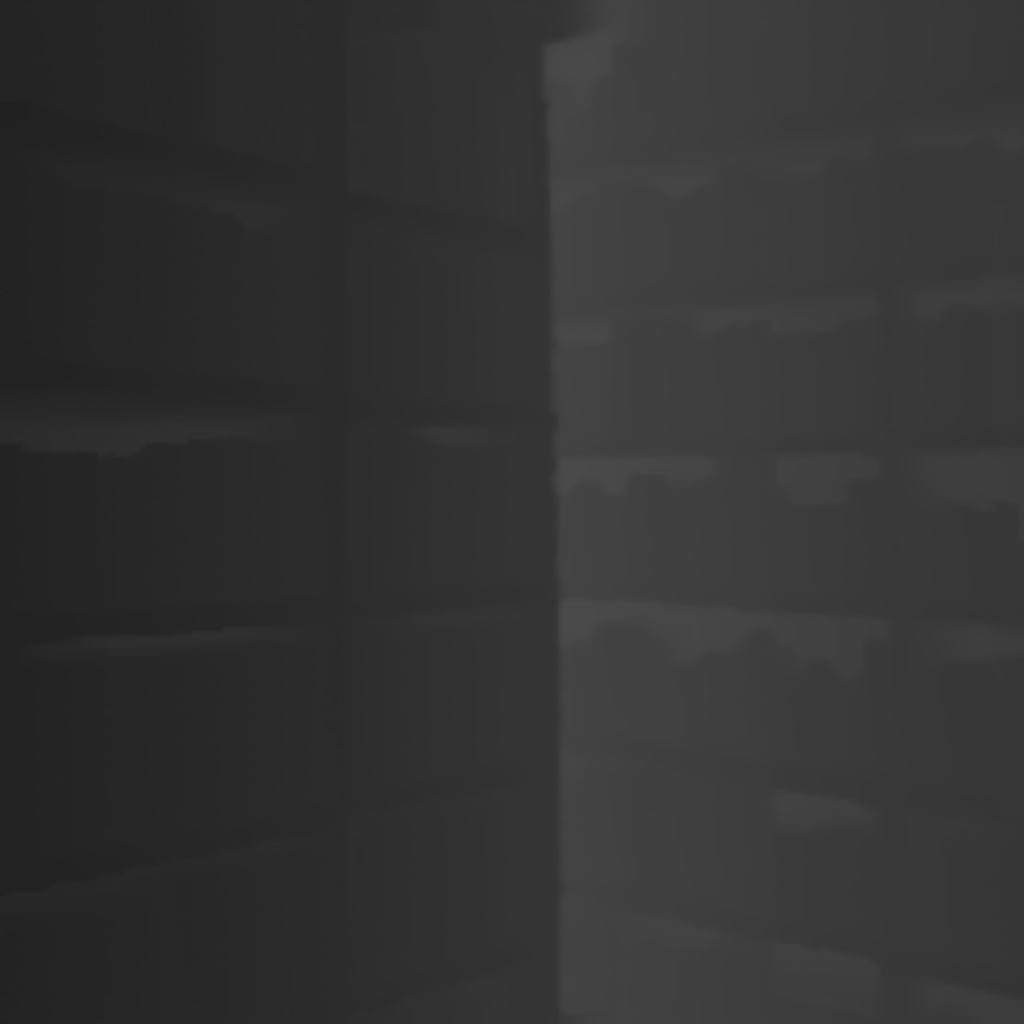} & \multicolumn{2}{c}{} \\
    \\[\textpadding]\multicolumn{\ncols}{c}{Prompt: A bookstore.}\\\\[\textpadding]
\end{tabular}

\caption{Text to 3D samples. Given a text prompt, an image is first generated using Imagen (first row of first column), after which depth is estimated (second row of first column). Subsequently the camera is moved to reveal new parts of the scene which are infilled using an image completion model and \ours~(which conditions on both the incomplete depth map and the filled image). At each step, newly generated RGBD points are added to a global point cloud which is visualized in the rightmost column.}
\label{tab:text_to_3d_samples_extras}
\end{figure*}
\begin{figure*}
\begin{tabular}{ccc}
    Image &  Prediction & Groundtruth \\
    \includegraphics[width=0.32\linewidth]{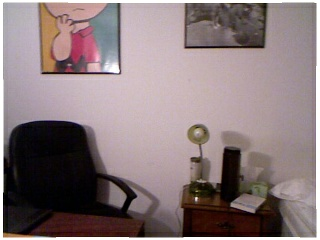} & \includegraphics[width=0.32\linewidth]{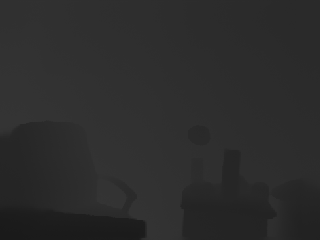} & \includegraphics[width=0.32\linewidth]{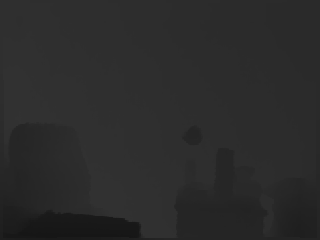} \\
    \includegraphics[width=0.32\linewidth]{} & \includegraphics[width=0.32\linewidth]{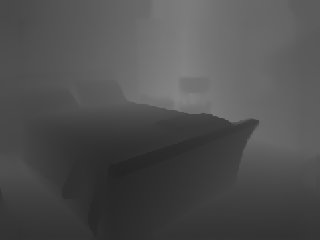} & \includegraphics[width=0.32\linewidth]{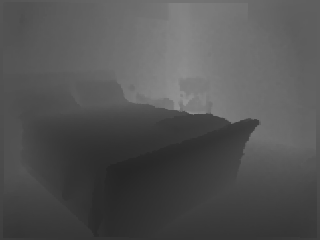} \\
    \includegraphics[width=0.32\linewidth]{} & \includegraphics[width=0.32\linewidth]{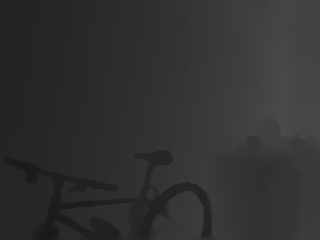} & \includegraphics[width=0.32\linewidth]{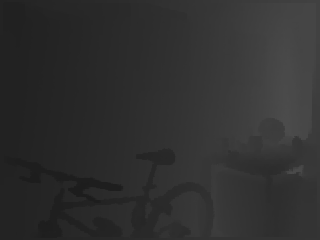} \\
\end{tabular}
\caption{Predictions on the NYU Depth V2 val dataset.}
\label{fig:NYU-examples}
\end{figure*}

\begin{figure*}
\begin{tabular}{ccc}
    Image &  Prediction & Groundtruth \\

    \includegraphics[width=0.32\linewidth]{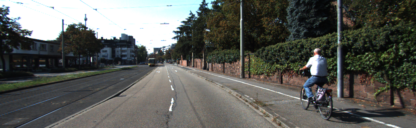} & \includegraphics[width=0.32\linewidth]{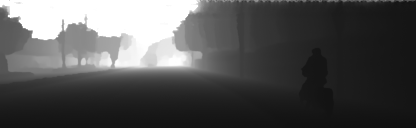} & \includegraphics[width=0.32\linewidth]{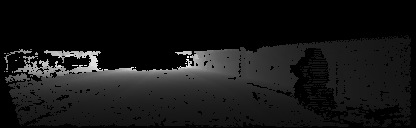} \\
    \includegraphics[width=0.32\linewidth]{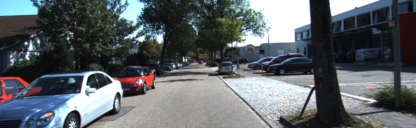} & \includegraphics[width=0.32\linewidth]{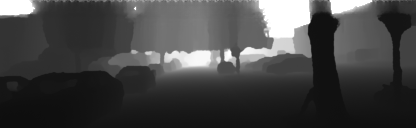} & \includegraphics[width=0.32\linewidth]{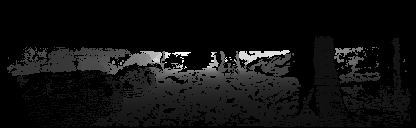} \\
    \includegraphics[width=0.32\linewidth]{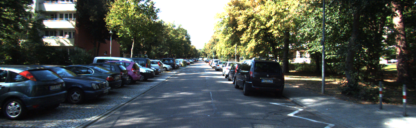} & \includegraphics[width=0.32\linewidth]{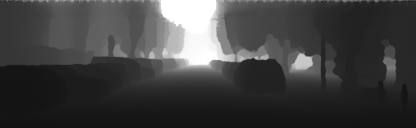} & \includegraphics[width=0.32\linewidth]{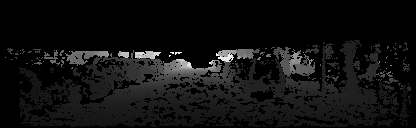} \\
\end{tabular}
\caption{Predictions on the KITTI val dataset. Note that we extract the sky region using a sky segmenter and fix the sky depth to the max depth during inference using replacement guidance.
}
\label{fig:KITTI-examples}
\end{figure*}
\begin{figure*}
\begin{tabular}{cc|cc}
    Image &  Depth & Image & Depth \\

    \includegraphics[width=0.24\linewidth]{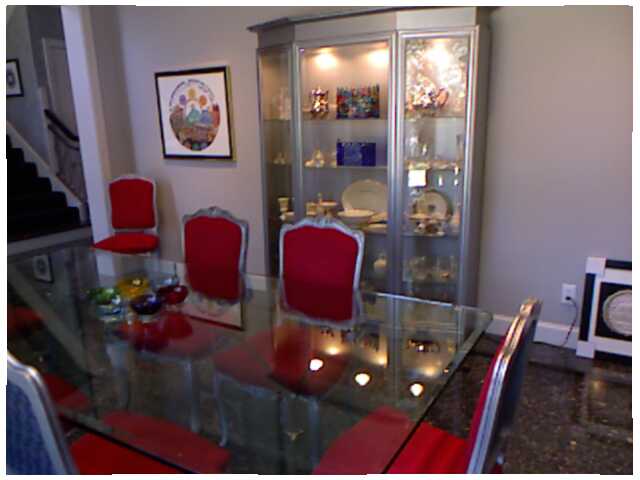} & \includegraphics[width=0.24\linewidth]{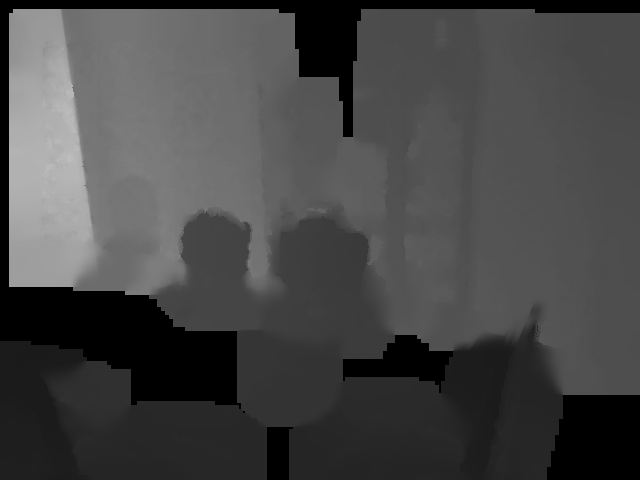} & \includegraphics[width=0.24\linewidth]{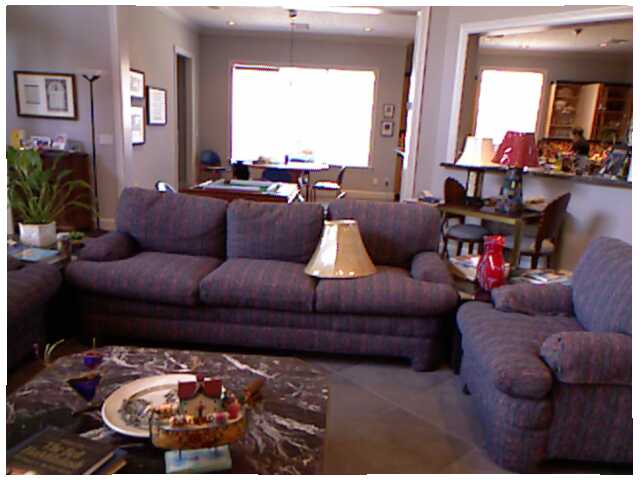} & \includegraphics[width=0.24\linewidth]{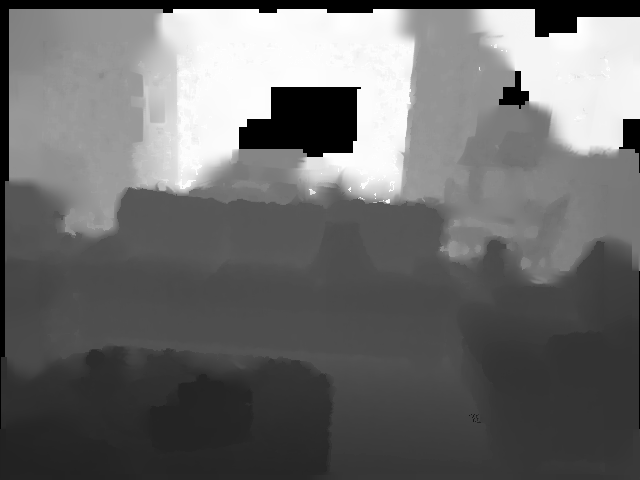} \\
    \includegraphics[width=0.24\linewidth]{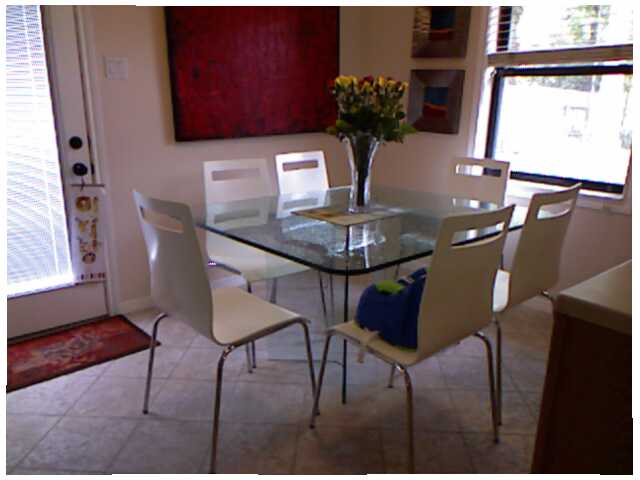} & \includegraphics[width=0.24\linewidth]{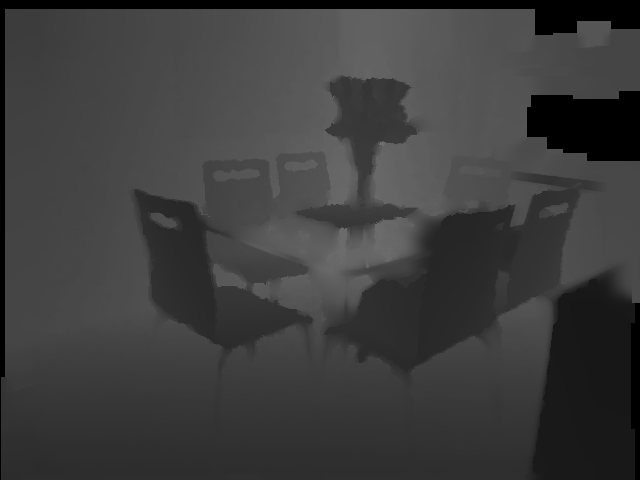} & \includegraphics[width=0.24\linewidth]{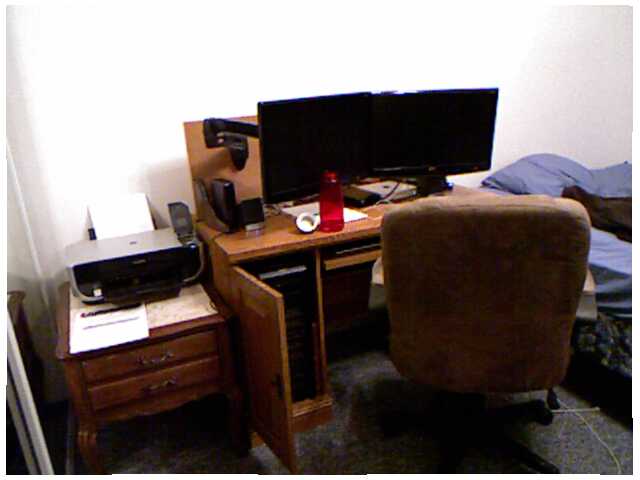} & \includegraphics[width=0.24\linewidth]{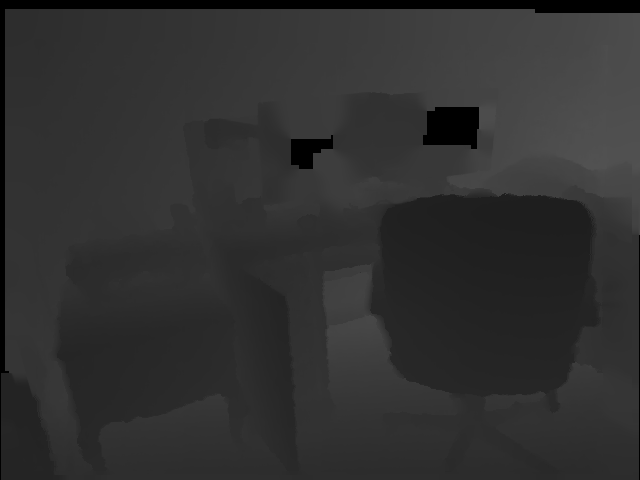} \\
    \includegraphics[width=0.24\linewidth]{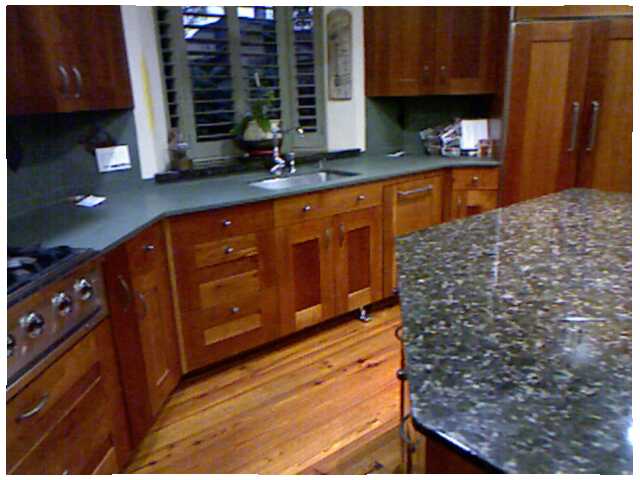} & \includegraphics[width=0.24\linewidth]{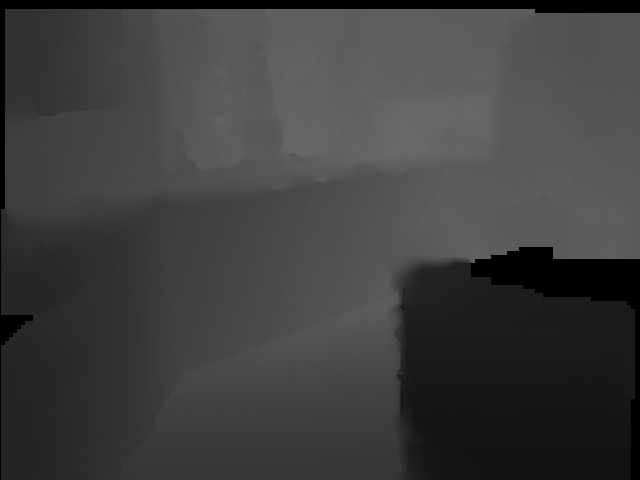} & \includegraphics[width=0.24\linewidth]{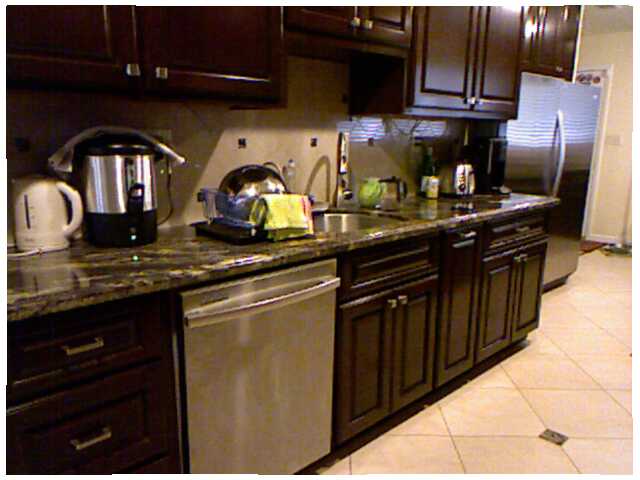} & \includegraphics[width=0.24\linewidth]{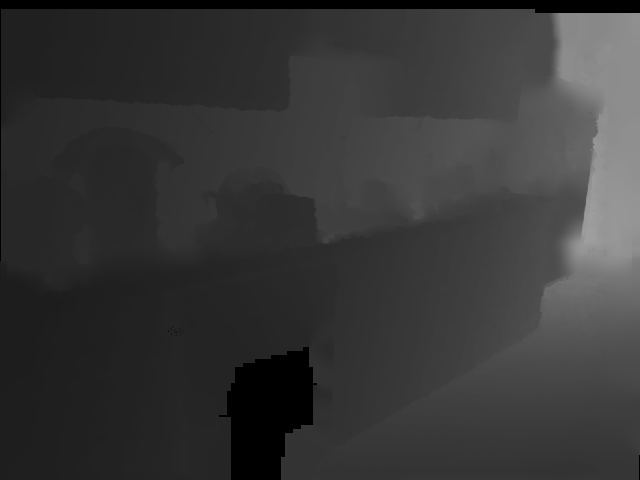} \\

\end{tabular}
\caption{Examples of RGB images and their corresponding depth maps from the NYU depth v2 training set containing missing depth.
}
\label{fig:nyu-train-missing-depth}
\end{figure*}



\end{document}